\documentclass[lettersize,journal]{IEEEtran}
\usepackage{amsmath,amsfonts}
\usepackage{algorithmic}
\usepackage{algorithm}
\usepackage{array}
\usepackage[caption=false,font=normalsize,labelfont=sf,textfont=sf]{subfig}
\usepackage{textcomp}
\usepackage{stfloats}
\usepackage{url}
\usepackage{verbatim}
\usepackage{multirow}
\usepackage{mathtools}
\usepackage{graphicx}
\usepackage{titlesec}
\titlespacing\section{0pt}{12pt plus 4pt minus 2pt}{0pt plus 2pt minus 2pt}
\titlespacing\subsection{0pt}{12pt plus 4pt minus 2pt}{0pt plus 2pt minus 2pt}
\titlespacing\subsubsection{0pt}{12pt plus 4pt minus 2pt}{0pt plus 2pt minus 2pt}
\DeclareGraphicsExtensions{.pdf,.jpeg,.png}
\usepackage{cite}
\ifCLASSINFOpdf
\else
\fi
\begin{document}

\title{Insights into the Lottery Ticket Hypothesis and Iterative Magnitude Pruning}

%
%
%

\author{Tausifa Jan Saleem, Ramanjit Ahuja, Surendra Prasad, Brejesh Lall 
       \thanks{Tausifa Jan Saleem (email: tausifa.cstaff@iitd.ac.in) is affiliated with the Bharti School of Telecommunication Technology and Management, Indian Institute of Technology Delhi.}          
\thanks{ Ramanjit Ahuja (email: ramanjit.ahuja@gmail.com) was affiliated with the Foundation for Innovation and Technology Transfer, Indian Institute of Technology Delhi as a Research Associate at the time of accomplishing this work. He is currently affiliated with  On Semiconductors (India)  as a Member of Technical Staff.}
      \thanks{ Surendra Prasad (email: sprasad@ee.iitd.ac.in) and Brejesh Lall (email: brejesh@ee.iitd.ac.in) are affiliated with the Department of Electrical Engg and the Bharti School of Telecommunication Technology and Management at Indian Institute of Technology Delhi.  }

\thanks{Acknowledgement: This work was supported by funding from IIT Delhi, Cadence Design Systems and the Indian National Science Academy (INSA). We would like to thank the authors of the papers \cite{frankle2018lottery} and \cite{li2018visualizing} for making their code available on GitHub.}}

\maketitle
\thispagestyle{empty}

\begin{abstract}
Lottery ticket hypothesis \cite{frankle2018lottery} for deep neural networks emphasizes the importance of initialization used to re-train the sparser networks obtained using the iterative magnitude pruning process. An explanation for why the specific initialization proposed by the lottery ticket hypothesis tends to work better in terms of generalization (and training) performance has been lacking. Moreover, the underlying principles in iterative magnitude pruning, like the pruning of smaller magnitude weights and the role of the iterative process, lack full understanding and explanation. In this work, we attempt to provide insights into these phenomena by empirically studying the volume/geometry and loss landscape characteristics of the solutions obtained at various stages of the iterative magnitude pruning process.
\end{abstract}

\begin{IEEEkeywords}
Lottery Ticket Hypothesis, Iterative Magnitude Pruning, Loss Landscape. 
\end{IEEEkeywords}

\IEEEpeerreviewmaketitle

\section{Introduction}

Neural network pruning is the process of removing unnecessary weights from a neural network \cite{blalock2020state}. This reduces the model size and the energy consumed by the neural network model, which makes inference efficient. It has been observed that the pruned models do not perform well without re-training, and re-training the sparser networks from the start (with random initialization) is difficult \cite{han2015learning, li2016pruning}. However, Frankle and Carbin \cite{frankle2018lottery} demonstrated that there exists a subnetwork, which if trained from the start, reaches the accuracy of the original network. More formally, their hypothesis says that a randomly initialized dense neural network contains a subnetwork that is initialized such that—when trained from the start—it can match the test accuracy of the original network after training for at most the same number of iterations. This hypothesis has been named the \textit{lottery ticket hypothesis}. Empirical corroboration of the lottery ticket hypothesis consists of a procedure called \textit{Iterative Magnitude Pruning (IMP)} \cite{frankle2018lottery, frankle2019stabilizing}. IMP consists of the following steps: 1) randomly initialize a dense neural network and pre-train it for some number of iterations. The weights of this pre-trained network is known as the rewind point; 2) train the network to convergence; 3) prune a fraction of the smallest magnitude weights from the trained network; 4) rewind the unpruned weights to their values at the rewind point; 5) repeat steps 2-4 until a sufficiently pruned network is obtained; 6) train the final network.

The final network is a \textit{winning ticket} if it is \textit{matching}, which means it trains to the same accuracy as the dense network. IMP has been successful in producing highly-sparse networks that are matching. But the underlying principles like the role of specific initialization proposed by the lottery ticket hypothesis, pruning of smaller magnitude weights, and the role of iterative process are not fully understood. \par 
Many research works related to the lottery ticket hypothesis have been published since its advent to demystify the mechanisms and principles governing it. Frankle et al. \cite{frankle2019dissecting} studied the relationship between the percentage of weights remaining and interpretability\footnote{Ability to understand and explain how the network arrived at its decisions.} of a network in case of magnitude pruning. They demonstrated that pruning does not harm interpretability until very few parameters remain in the network and there is an accuracy drop. The authors demonstrated that the network parameters considered superfluous for accuracy by the pruning algorithm are also superfluous for interpretability. However, pruning does not make the network more interpretable either. The network dissection technique given by Bau et al. \cite{bau2017network} was used to measure the interpretability of the pruned network. In another work, Frankle et al. \cite{frankle2020pruning} demonstrated that the pruning methods that prune at initialization (prior to training) perform better than random pruning but their accuracy remains smaller than the magnitude pruning after training. They also demonstrated that randomly shuffling the weights pruned by these techniques within each layer preserves or improves the accuracy. Frankle et al. \cite{frankle2020linear} performed instability analysis towards Stochastic Gradient Descent (SGD) noise to explain the success and failure of IMP. They demonstrated that IMP finds matching sub-networks only when they are stable to SGD noise (not affected by different data orders), and this happens at initialization in simple tasks but only after a few training iterations in complicated tasks. That is why pre-training the dense network for a few iterations (to obtain the rewind point) and rewinding the unpruned weights of the pruned network to their values at the rewind point is required in IMP. However, rewinding alone does not account for the success of IMP since randomly pruning the network and then rewinding to the rewind point does not produce matching sub-networks \cite{frankle2020linear}.
Hence, there is some information encoded in the associated pruning procedure, in addition to the information conveyed by the specific initialization proposed by the lottery ticket hypothesis, that leads to the success of IMP.  Larsen et al. \cite{larsen2021many} demonstrated that the number of parameters required for training a deep neural network decreases as the initial loss decreases. This explains why rewinding to a later point in training leads to better accuracy. Zhang et al. \cite{zhang2021lottery} presented a theoretical analysis of the geometrical structure of the loss function 
 in the case of pruned networks. They demonstrated that pruning enlarges the convex region near the optimal solution, and this geometric advantage makes pruned networks generalize better.  \par
Rosenfeld et al. \cite{rosenfeld2021predictability} formulated a scaling law that estimates the test error when pruning with IMP. They demonstrated that the test error depends on the size of the training dataset, depth of the network, width of the network, and pruning level. Movva et al. \cite{movva2021studying} analyzed the effect of combining pruning information across multiple training runs (different data orders) on accuracy-sparsity trade-off in the case of magnitude pruning. They demonstrated that the pruning overlap between different copies is small (not more than chance). However, with pre-training, the overlap increased significantly. Combining the information using intersection or union of masks performed similar to the one-shot magnitude pruning baseline. Paul et al. \cite{paul2022lottery} demonstrated that in the pre-training phase, only a subset of data is needed to get a matching initialization (initialization that produces a matching sub-network), and the length of this phase can be reduced if training is carried out using easy-to-learn examples. Jin et al. \cite{jin2022pruning} demonstrated that two factors essentially contribute to better generalization in the pruned models (obtained using IMP): extended model training time and model size reduction. Extended model training time improves training, and model size reduction adds regularization, which in turn improves generalization. \par The closest work to ours is the recently reported work of Paul et al. \cite{paul2022unmasking}. They studied the geometry of the error (test error) landscape to answer several questions about the lottery ticket hypothesis and IMP. Their contributions are: 1) they concluded that the pairs of IMP solutions at the successive iterations are connected linearly with no error barrier between them if and only if they are matching. This shows that in every IMP iteration, the mask produced by the pruning procedure conveys information about the axial subspace that intersects a desired linearly connected mode of a matching sublevel set; 2) they concluded that the reason of one-shot pruning not working well compared to iterative pruning is that pruning to higher sparsities using one-shot is prohibited by the sharpness of the error landscape; 3) they concluded that retraining reequilibriates the weights of the network, i.e., finds networks with new small weights ready for further pruning.

Our independently conducted studies have been in the domain of the loss (training loss) landscape rather than the error (test error) landscape. We were interested in the question: are the IMP solutions at the successive iterations (levels) linearly connected in the loss landscape? If so, then SGD should converge to the sparser solution by itself rather than converging to a less sparser solution. Consider two IMP solutions: IMP solution at level $(L-1)$ and IMP solution at level $L$. If they are linearly connected to each other in the loss landscape, SGD at level $(L-1)$ should have directly converged to level $L$ solution instead of converging to $(L-1)$ solution, but obviously, this does not happen. This is a fundamental question that we address in our work.\par In this work, we attempt to provide insights about the lottery ticket hypothesis and IMP by studying the loss landscape (not the error landscape) characteristics and volume/geometry of the IMP solutions at different iterations. Although training loss and test error are correlated with each other, studying the loss landscape characteristics instead of the error landscape makes more sense while developing an understanding of the working of neural network models because SGD navigates through the loss landscape and not the error landscape.  \\
\textbf{Contributions.} We perform extensive experimentation on a widely used network, ResNet-20, on a benchmark dataset, CIFAR-10.
The contributions are as follows:
\begin{enumerate}
    \item We demonstrate that there exist special type of solutions in the loss landscape, which generalize well but have a very small volume in the original space, and the IMP procedure exposes such solutions, which otherwise remain hidden.
    \item We provide an insight into the role played by specific initialization proposed by the lottery ticket hypothesis.
    \item We demonstrate the role played by the iterative process in IMP, which answers the question of why one-shot pruning does not work well (comparatively).
    \item We demonstrate that there exists a barrier between the  IMP solutions at successive levels in the loss landscape, implying that they are not strictly linearly connected in this scenario.
    \item We demonstrate that IMP solutions obtained using rewinding lie within the same loss sublevel set (defined in the sequel).
    \item We provide new insight into magnitude-based pruning, which answers the question of why pruning smaller weights (weights with smaller magnitudes) is beneficial and not the larger ones.
    \item We provide insight into why fine-tuning does not work at par with rewinding.
\end{enumerate}

\section{Background Information and Problem Formulation}

\subsection{Background Information}
This subsection discusses the following preliminaries: lottery ticket hypothesis, fine-tuning and rewinding, iterative pruning versus one-shot pruning, loss landscape of a neural network, error landscape of a neural network, loss sublevel set, and the role of volume in generalization performance. \\
\textbf{Lottery ticket hypothesis.} Lottery ticket hypothesis postulates that neural networks contain subnetworks that when trained from scratch reach the accuracy of the original network in a commensurate number of epochs. This hypothesis was proposed by Frankle and Carbin in 2018 \cite{frankle2018lottery}. Their assertion is supported by the observation that IMP consistently discovers such subnetworks on small vision tasks by rewinding the weights of the subnetwork to the $0^{th}$ iteration of the original dense network. However, IMP fails on deeper networks. In follow-up work, Frankle et al. \cite{frankle2019stabilizing}  demonstrated that in the case of deeper networks, such subnetworks could be obtained by rewinding to  $k^{th}$ iteration instead of rewinding to $0^{th}$ iteration after pruning, for a suitably chosen value of $k$.\\
\textbf{Fine-tuning and rewinding.} IMP consists of a number of pruning and re-training cycles. Fine-tuning and rewinding are the two retraining strategies in IMP. Rewinding was introduced by the lottery ticket hypothesis, while conventionally, IMP operated with fine-tuning or rewinding with random initialization. Fine-tuning trains the pruned network with a fixed (small) learning rate, and the starting point for retraining with fine-tuning is the final values of unpruned weights \cite{han2015learning, liu2018rethinking}. Rewinding is of two types; weight rewinding and learning rate rewinding. In the case of weight rewinding, the unpruned weights are rewound to their values at the rewind point, and apart from rewinding the weights, it also rewinds the learning rate schedule. Learning rate rewinding has the same starting point for retraining as that of fine-tuning and it follows the same learning rate schedule as followed by the weight rewinding procedure. So, the difference between fine-tuning and learning rate rewinding is the difference in their learning rates. And the difference between learning rate rewinding and weight rewinding is the difference in their starting points. Renda et al., \cite{renda2020comparing} demonstrated that fine-tuning does not perform as well as weight rewinding, however, learning rate rewinding performs at par with weight rewinding or even outperforms it in some scenarios.\\
\textbf{Iterative pruning versus one-shot pruning.} One of the crucial facets of IMP is that the weights are pruned iteratively with intervals of retraining between them. In contrast to this, one-shot pruning prunes all the weights in one go to attain the desired sparsity and then retrains. It has been demonstrated in the literature that iterative pruning always outperforms one-shot pruning \cite{frankle2018lottery, renda2020comparing}. \\
\textbf{Loss landscape of a neural network.} Neural networks are trained using feature vectors {$x_i$} and their corresponding labels {$y_i$}. In the training procedure, loss function $Loss(W)$ is minimized: 
    \begin{equation}
     Loss(W)= \frac{1}{n} \Sigma_{i=1}^n l(x_i,y_i,W)
\end{equation}
where $W$ represents the network parameters, $n$ represents the number of input data samples, and $l(x_i,y_i,W)$ is a function that measures the difference between the predicted label and the actual label. The number of parameters in neural networks is very large; hence, the neural network loss functions reside in extremely high-dimensional spaces. The plot of training loss ($Loss(W)$) with respect to the network parameters ($W$) is referred to as the loss landscape. \par
Studying the loss landscape of neural networks is important for understanding their behavior. The loss landscape of under-parameterized models has multiple isolated local minima \cite{liu2020toward}. The set of solutions of over-parameterized models, on the other hand, is generically a manifold of dimension $m-rn$ \cite{cooper2018loss}, where $m$ is the number of parameters, $n$ is the number of input data samples, and $r$ is the number of output classes. This means that the density of minima in the loss landscape of over-parameterized models is very high, and the optimizer converges to one of them irrespective of where it starts at.\\
\textbf{Error landscape of a neural network.} The variation in the test error of a neural network with respect to the network parameters is referred to as the error landscape of a neural network. It provides insights into how well the network is performing and how sensitive it is to the changes in the network parameters.\\
\textbf{Loss sublevel set.} Loss sublevel set $S(\epsilon)$ is the set of all points in the weight space for which loss $Loss(W)$ is less than or equal to some desired value $\epsilon$ \cite{larsen2021many} :
    \begin{equation}
        S(\epsilon):={W \in R^D:Loss(W) \leq \epsilon}
    \end{equation} \\
\textbf{Role of volume\footnote{For our discussion here, volume refers to the volume of minimum, not the volume of loss sublevel set.} in generalization performance}. Flatness is an indicator of network performance sensitivity to parameter perturbations. The minimum is flat if small changes to the parameters do not cause misclassifications. On the contrary, the minimum is sharp if small changes to the parameters cause a number of misclassifications, thereby increasing the value of the loss function. A number of studies have focused on establishing the relationship between the flatness/sharpness of minima and their generalization ability. The following presents a summary of those studies; \par 
    Hochreiter and Schmidhuber \cite{hochreiter1997flat} explained the relationship between the flatness of minima and their generalization ability using the Minimum Description Length (MDL) theory. They defined a flat minimum as a region in the weight space where the error remains approximately constant. Such a region requires less information for representation because of its lower complexity than a region where the error changes drastically (sharp minimum). According to MDL theory, lower complexity models have higher generalization ability. Chaudhari et al. \cite{chaudhari2019entropy} have shown that the local minima discovered by the optimizers have a flat geometry for a range of deep neural network architectures irrespective of their structures, training strategies, and the input data. These flat regions are robust to perturbations (both data perturbations as well as parameter perturbations) and noise in the activations, which makes them generalize well. Keskar et al. \cite{keskar2016large} have demonstrated that the large-batch methods are attracted towards sharp minima. They have shown that these minima have large positive eigen values of the Hessian (Hessian of the loss function) and do not have good generalization ability. In contrast, small-batch methods converge to flat minima, which have a large number of small eigen values of the Hessian and generalize well. Dinh et al. \cite{dinh2017sharp} argued that the notions of flatness cannot be directly related to the generalization performance without taking certain precautions. Their argument is based on the following grounds: the loss function of a neural network with weights much larger than one may seem to be flat because parameter perturbations by one unit will have a very small consequence on the network performance. On the contrary, in a neural network with smaller weights than one, the same perturbation will drastically affect the network performance, making the loss function appear sharp. Knowing that neural nets are scale-invariant, the large-parameter network and the small-parameter network are the same in that the large-parameter network is a rescaled version of the small-parameter network. Thus, any discrepancies in the loss function plots are simply the result of the difference in the scales of the networks. Hence, it is crucial to apply perturbations in accordance with the scale of network parameters to have a correct notion of the flatness/sharpness of minima \cite{li2018visualizing}. \par Huang et al. \cite{huang2020understanding} demonstrated that two types of minima exist in a neural network loss landscape. These are referred to as the so-called good minima and bad minima. Good minima exhibit a small training loss and a small generalization error. Bad minima, too have a small training loss but exhibit a high generalization error. They also studied the qualitative difference in the loss landscape around these minima and observed that the decision boundaries of good minima have wide margins\footnote{Distance between the class boundary and the data.} while the decision boundaries of bad minima have very narrow margins. Huang et al. \cite{huang2020understanding} also illustrated that the good minima reside in wide basins\footnote{Set of points in the neighborhood of a minimum whose loss value is smaller than some cutoff value.} that exhibit a large volume in the parameter space, while the bad minima reside in narrow basins that exhibit a much smaller volume. A larger volume also implies a higher probability of hitting the minima by SGD. Volumes of the minima are, therefore, good indicators of their robustness and can provide useful insights \cite{huang2020understanding}.\par Calculation of the volume of these basins is, however, computationally intractable because the loss function lies in an extremely high-dimensional space. Huang et al. \cite{huang2020understanding} used Monte-Carlo integration to approximate the volume of these basins. Using their method, the $n$-dimensional volume of the basin ($V$) is calculated as:
    \begin{equation}
        V ={\omega_n E_\phi[r^n(\phi)]}
    \end{equation} 
    where $r(\phi)$ represents the radius of the basin in the direction of the unit vector $\phi$ and $E$ represents the expected value. This expected value is estimated by calculating $r(\phi)$ along a large number of random directions. $\omega_n$ is the volume of the unit-$n$ ball and is given as; $\omega_n=(\pi^{n/2})/\Gamma(1+n/2)$, where $\Gamma$ represents the Euler's Gamma function.
   However, this method of volume calculation is computationally expensive. Another approximation of basin volume was given by Wu et al.\cite{wu2017towards}. They demonstrated that the product of top-$k$ positive eigen values ($\lambda$) of the Hessian could be used to approximate the inverse volume of these basins. More specifically, the inverse volume of the basin ($V^{'}$) can be approximately expressed as:
      \begin{equation}
        V^{'}(k) :={\Sigma_{i=1}^k Log(\lambda_i)} 
    \end{equation}
    This is because a large basin implies that the valley around the minimum is flat, which is associated with smaller eigen values, and vice-versa. \\
\subsection{Definitions and Notations.} 
\textbf{Sparse subnetworks:} Given a dense network with weights $W$ ($W \in \mathbb{R}^{D}$), a sparse subnetwork has weights
$m \odot W$, where $m \in$ \{${0, 1}\}^D$  is a binary mask and $\odot$ is the element-wise product. The sparsity of a
mask $m$, $S(m)$ is the fraction of zeros in the mask. \\
\textbf{Notation for IMP solution at level $L$:} We represent the IMP solution (minimum) at level $L$ by $W_{(L)}^{(min\_(L))}$, weights of the dense network at initialization by $W^{(init)}$ and weights of the dense network at rewind-point by $W^{(rewind\_point)}$. Note that all these weights are $D$-dimensional.\\
\textbf{Projection of level $L$ solution on level $(L+1)$:} Let $W^{Pr{(min\_(L))}}_{(L+1)}$ represent the projection of level $L$ solution on level $(L+1)$. It is obtained as $W^{Pr{(min\_(L))}}_{(L+1)}$=$m_{(L+1)} \odot W_{(L)}^{(min\_(L))}$, where $m_{(L+1)}$ represents the pruning mask at level $(L+1)$. For example, the projection of level $0$ solution on level $1$ will be represented as $W^{Pr{(min\_(0))}}_{(1)}$. Note that $S(m_{(L+1)}) > S(m_{(L)})$. \\
\textbf{Reverse Projection of level $(L+1)$ solution on level $(L)$:} Let $W^{RPr{(min\_(L+1))}}_{(L)}$ represent the reverse projection of level $(L+1)$ solution on level $(L)$. It is obtained as; $W^{RPr{(min\_(L+1))}}_{(L)}$=$m_{(L)} \odot W_{(L+1)}^{(min\_(L+1))}$, where $m_{(L)}$ represents the pruning mask at level $(L)$. For example, the reverse projection of level $1$ solution on level $0$ will be represented as $W^{RPr{(min\_(1))}}_{(0)}$.\\
\subsection{Problem Statement and Questions of Interest.} The primary objective of this study is to develop a deeper insight into the nature of the loss landscape, especially regarding the distribution and characteristics of its minima and the behavior of the SGD and of the solutions obtained using IMP and rewinding at various levels. The specific questions that can lead to better insight are as follows:\\
    1) Why are these solutions (at different levels) not discoverable directly? \\
    2) What is the role played by specific initialization proposed by the lottery ticket hypothesis?\\
    3) What is the role played by the iterative process in IMP?\\
    4) Why is the pruning of smaller weights beneficial and not the larger ones?\\
    5) Why fine-tuning does not work at par with rewinding?\\
    Our experiments reported below have been conducted to elicit accurate answers to these questions.
\section{Methodology}
We perform experimentation on a widely used network, ResNet-20, on a benchmark dataset, CIFAR-10. We use iterative magnitude
pruning with weight rewinding (IMP-WR) and run $10$ iterations of IMP. The unpruned weights are rewound to their values at $2000^{th}$ training step of the original dense network in each iteration. For an initial and quick appreciation, a plot of training loss and test accuracy at different levels of IMP-WR is presented in Fig. 1. \par
\begin{figure}[h!]
\centering
\includegraphics[width=4cm, height=3.5cm]{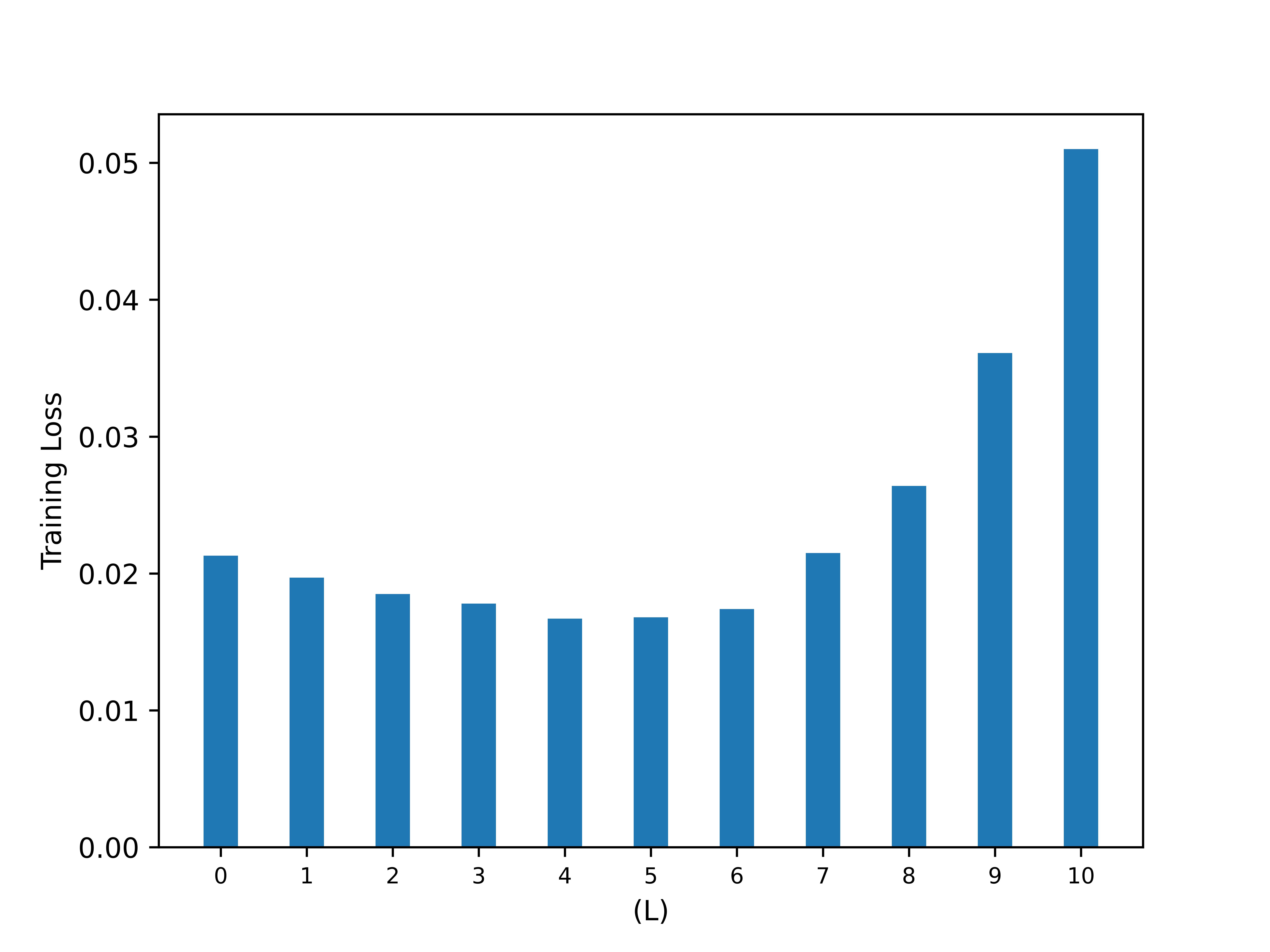}
\includegraphics[width=4cm, height=3.5cm]{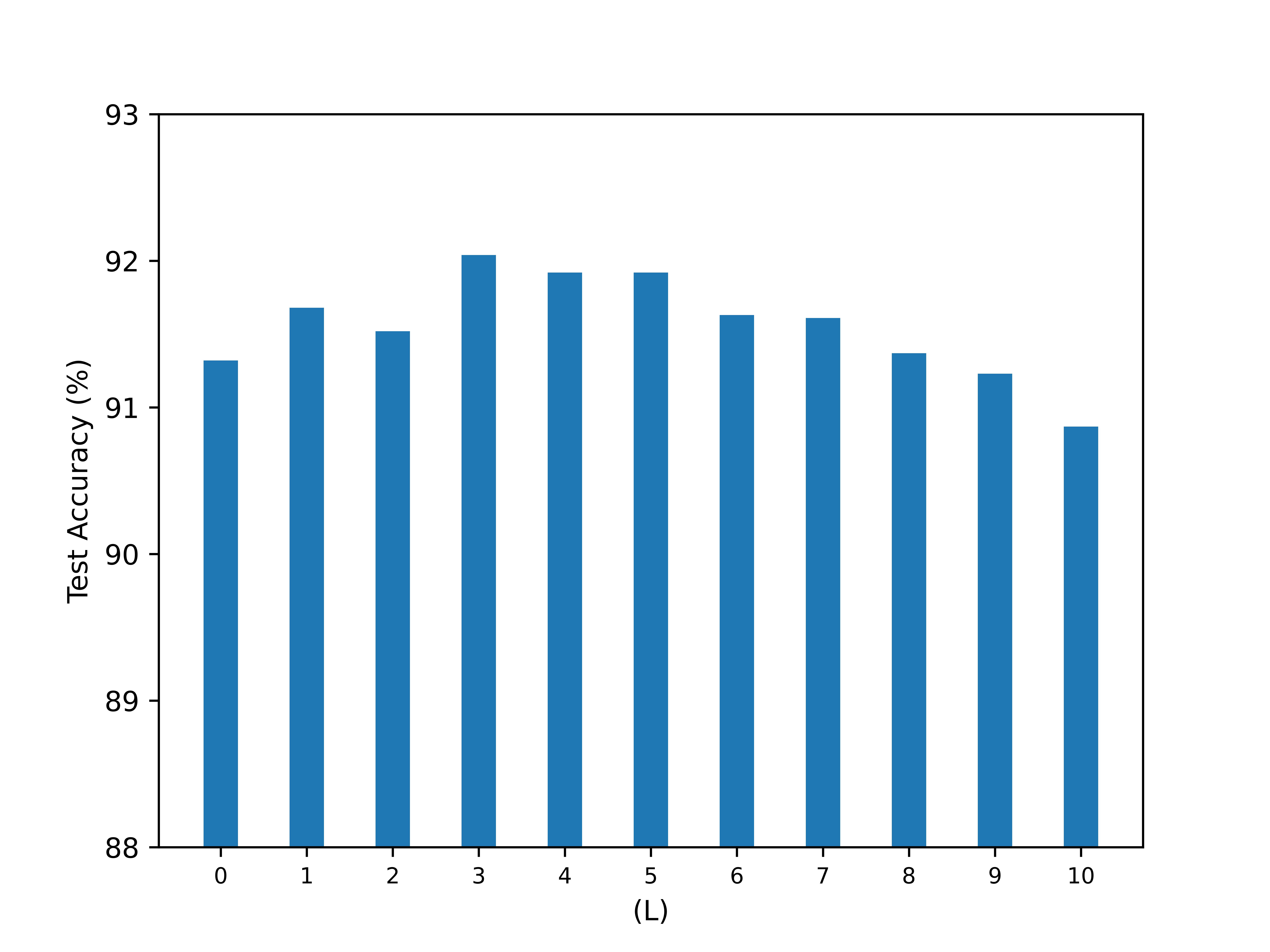}
\caption{Training loss and test accuracy at different levels of IMP-WR. \textbf{Left:} Training loss. \textbf{Right:} Test accuracy.}
\end{figure}
It is also useful to compare the performance of IMP-WR with other strategies that may either produce networks with a similar structure as that produced with IMP-WR but are initialized differently or produce networks with the same sparsity but different structures. To this end, we apply techniques like one-shot pruning, fine-tuning, random initialization of the pruned network, and random pruning on the aforementioned network. The details of these techniques are summarized below:\\ \textbf{One-shot pruned network} is obtained by pruning the weights of the trained dense network $W_{(0)}^{(min\_(0))}$ based on magnitude pruning in one go to attain the desired sparsity\footnote{Same sparsity as that of $W_{(10)}^{(min\_(10))}$.} and then rewinding the unpruned weights to their values at $W^{(rewind\_point)}$ and retraining. We represent the solution obtained using one-shot pruning by $W^{(one\_shot)}_{(10)}$. \\ \textbf{Fine-tuned network} is obtained by pruning $20\%$ smallest magnitude weights from $W_{(9)}^{(min\_(9))}$ and then re-training the unpruned weights (without rewinding) with a learning rate of $0.001$ for $40$ epochs. We represent the solution obtained using fine-tuning by $W^{(FT)}_{(10)}$. \\ \textbf{Randomly initialized pruned network} is obtained by pruning $20\%$ smallest magnitude weights from $W_{(9)}^{(min\_(9))}$ and then randomly initializing the unpruned weights and retraining. We call this network \textit{Randomly Initialized Pruned Network} and represent the solution by $W^{(RIPN)}_{(10)}$. \\ \textbf{Randomly pruned network} is obtained by randomly pruning $20\%$ weights from $W_{(9)}^{(min\_(9))}$ and then rewinding and retraining. We call this network \textit{Randomly Pruned Network} and represent the solution by $W^{(RPN\_1)}_{(10)}$. We also consider another randomly pruned network by randomly pruning the weights of $W_{(0)}^{(min\_(0))}$ in one go to attain the desired sparsity and then rewinding and re-training. We represent the solution obtained in this manner by $W^{(RPN\_2)}_{(10)}$. Note that the only difference between $W^{(RPN\_2)}_{(10)}$ and $W^{(one\_shot)}_{(10)}$ is that while obtaining $W^{(RPN\_2)}_{(10)}$, the weights of $W_{(0)}^{(min\_(0))}$ are pruned randomly whereas while obtaining $W^{(one\_shot)}_{(10)}$, the weights of $W_{(0)}^{(min\_(0))}$ are pruned using magnitude based pruning. \par 
A comparison of training loss and test accuracy between $W_{(10)}^{(min\_(10))}$, $W^{(one\_shot)}_{(10)}$, $W^{(FT)}_{(10)}$, $W^{(RIPN)}_{(10)}$, $W^{(RPN\_1)}_{(10)}$ and $W^{(RPN\_2)}_{(10)}$ presented in Fig. 2 shows that $W_{(10)}^{(min\_(10))}$ outperforms the other solutions. To explain the reason behind this, we study the loss landscape of these networks.
\begin{figure}[h!]
\centering
\includegraphics[width=4cm, height=3.5cm]{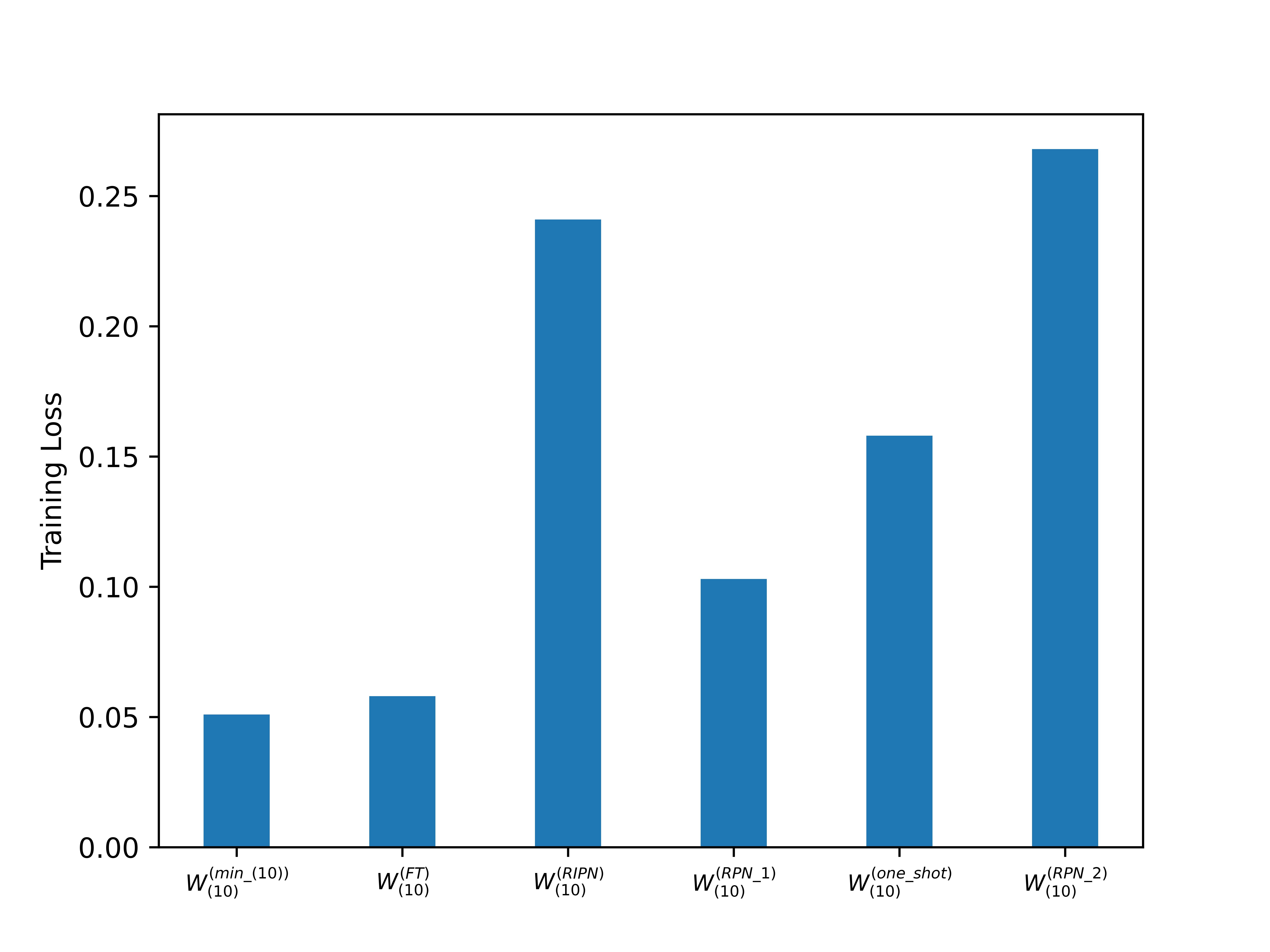}
\includegraphics[width=4cm, height=3.5cm]{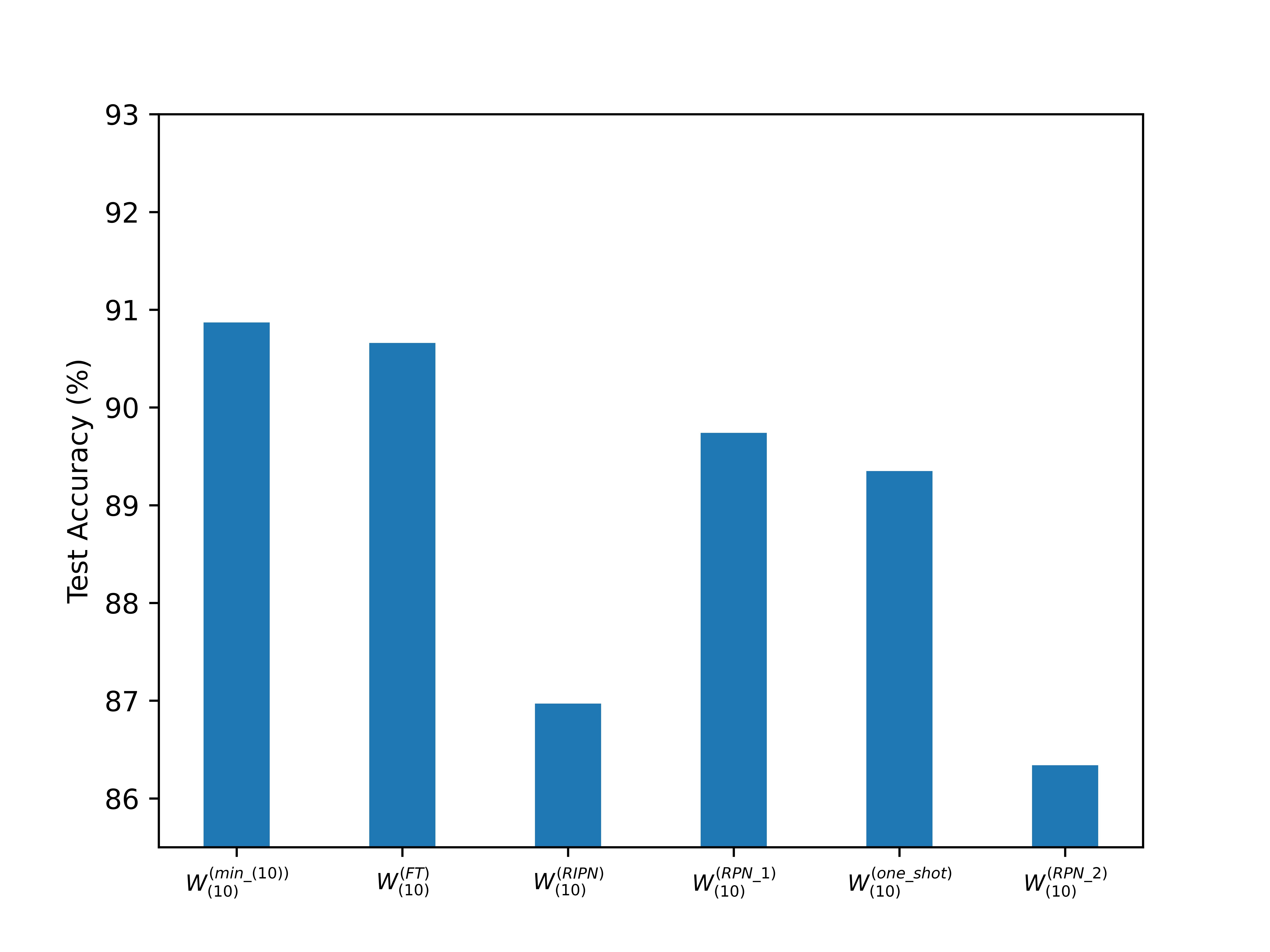}
\caption{Comparison of training loss and test accuracy between $W_{(10)}^{(min\_(10))}$, $W^{(one\_shot)}_{(10)}$, $W^{(FT)}_{(10)}$, $W^{(RIPN)}_{(10)}$, $W^{(RPN\_1)}_{(10)}$ and $W^{(RPN\_2)}_{(10)}$. \textbf{Left:} Training loss. \textbf{Right:} Test accuracy.}
\label{train_loss and test_acc}
\end{figure}
 
\section{Results and Findings}
In this section, we summarize our major findings based on extensive experimentation. We also provide a logical rationale/explanation for the observed behavior.

\subsection{Result 1: Special solutions with small volume exist.} Huang et al. \cite{huang2020understanding} had proposed that there exist two kinds of minima in a neural network loss landscape. The so-called \textit{good minima} have a low training loss and a large volume associated with their basins. These tend to have good generalization performance. Then there are the \textit{bad minima}, which too have a low training loss but have a small volume and do not exhibit good generalization performance. \par We demonstrate that there also exist another kind of solutions \footnote{We can not say for sure that these solutions are true minima, but they lie in the neighbourhood of minima because the gradient of the vast majority of parameters at these points is zero.} which have good generalization performance but have a (relatively) small volume. This implies that volume is not the only criterion for generalization performance; there is more to it. A careful experimental study leads us to hypothesize that the small volume of these solutions is due to very sharp curvature in certain dimensions, but the coefficients in these dimensions are zero. Their \textit{volume} measure tends to increase when these inferior dimensions are removed (possibly via pruning at another point) but is small when considered in the original space. \textbf{This makes these solutions almost undiscoverable by SGD in the original space but can be readily discovered in the pruned space.} This is the main result of our study. \par
Consider the two IMP solutions, that at level $(L-1)$, denoted by $W_{(L-1)}^{(min\_(L-1))}$ and that at level $(L)$, $W_{(L)}^{(min\_(L))}$. $W_{(L-1)}^{(min\_(L-1))}$ is a baseline for  $W_{(L)}^{(min\_(L))}$ because the pruning mask for level $L$ is determined by $W_{(L-1)}^{(min\_(L-1))}$. A comparison of the euclidean distance between $W^{Pr{(min\_(L-1))}}_{(L)}$ and $W^{Pr{(rewind\_point)}}_{(L)}$, and the euclidean distance between $W_{(L)}^{(min\_(L))}$ and $W^{Pr{(rewind\_point)}}_{(L)}$ given in Fig. 3 shows that for all $L$ except $2$ and $3$, $W^{Pr{(min\_(L-1))}}_{(L)}$ is closer to $W^{Pr{(rewind\_point)}}_{(L)}$ than $W_{(L)}^{(min\_(L))}$. 
\begin{figure}[h!]
\centering
\includegraphics[width=7cm, height=4cm]{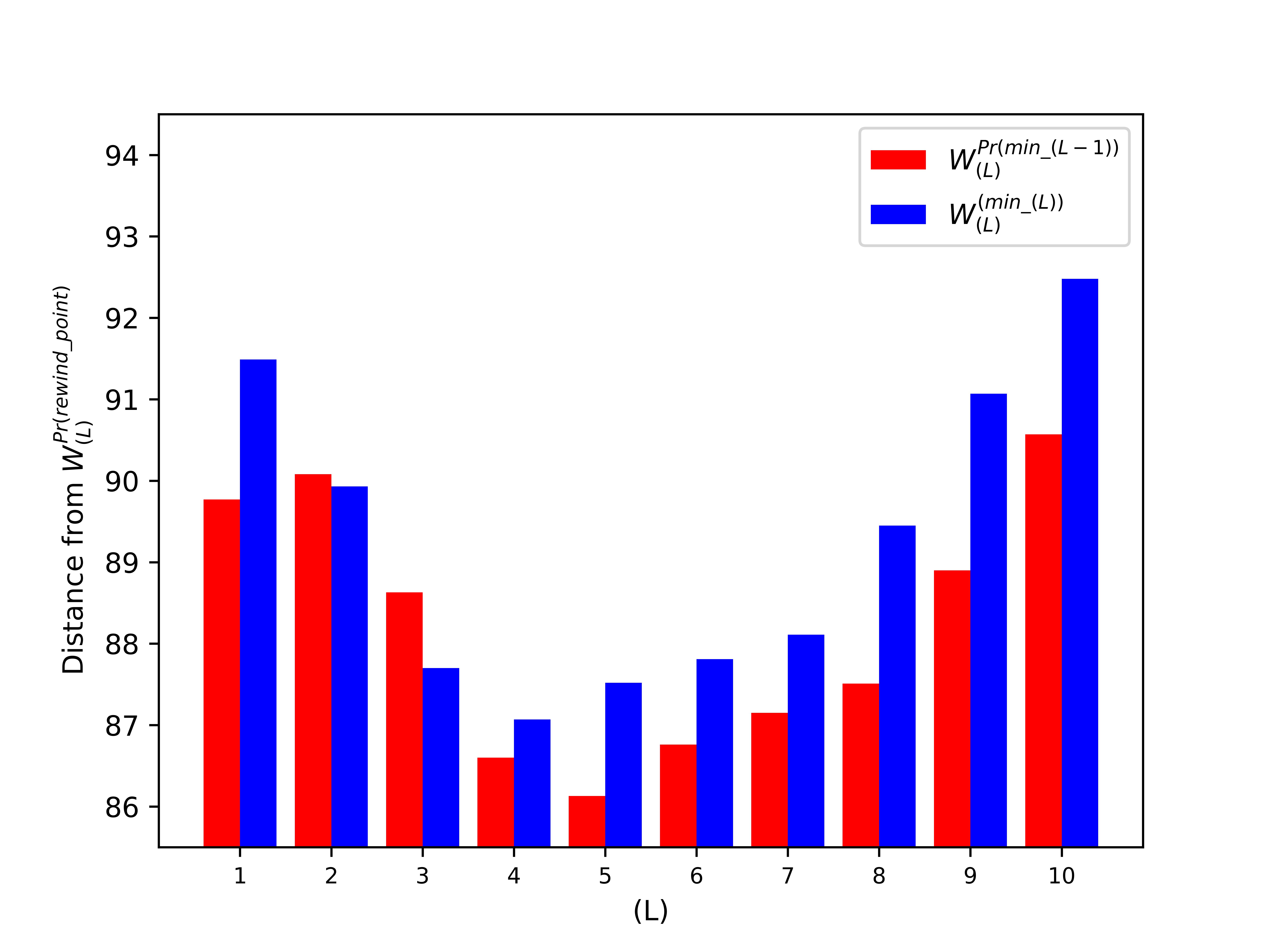}
\caption{Distance from $W^{Pr{(rewind\_point)}}_{(L)}$ to $W^{Pr{(min\_(L-1))}}_{(L)}$ and to $W_{(L)}^{(min\_(L))}$ for $L$ ranging from $1$ to $10$.}
\end{figure} 
Despite this at level $(L)$, SGD converges to $W_{(L)}^{(min\_(L))}$ and not to $W^{Pr{(min\_(L-1))}}_{(L)}$. In order to find the underlying reason, we plot the trajectory of SGD for level $L$ and that for level $(L-1)$ projected on level $L$ in Fig. 4. 
\begin{figure*}
\centering
\includegraphics[width=4cm, height=3.5cm]{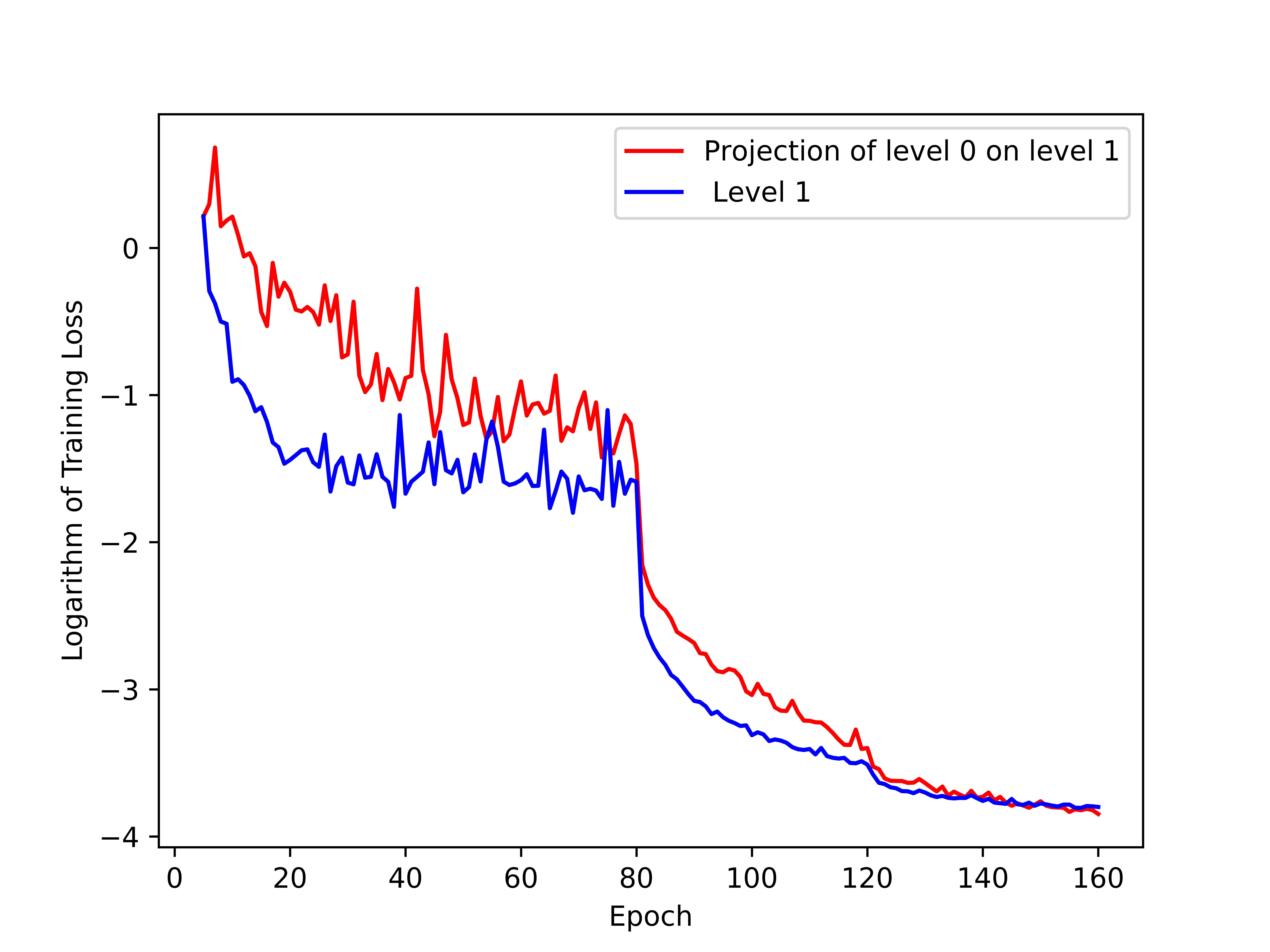}
\includegraphics[width=4cm, height=3.5cm]{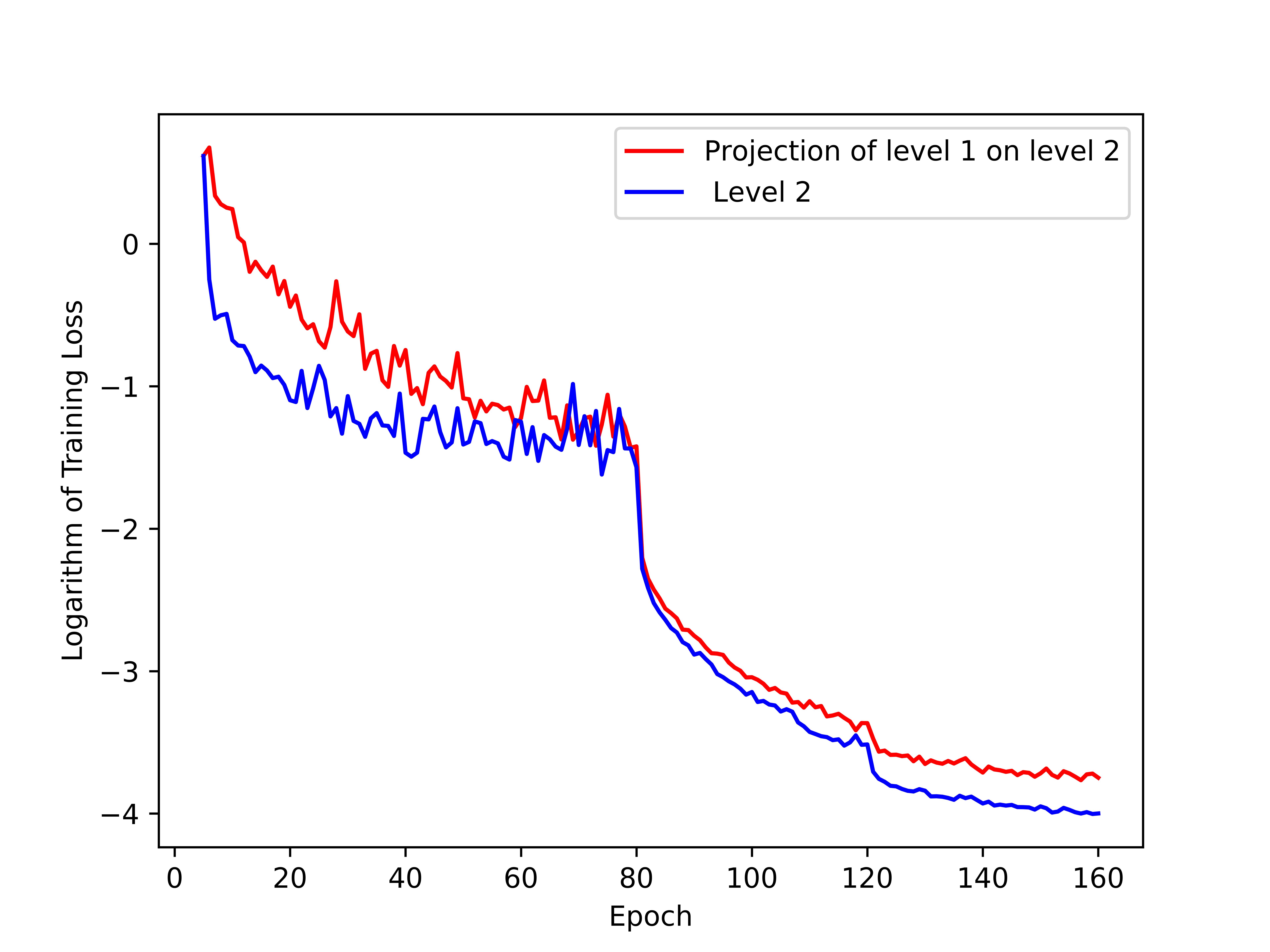}
\includegraphics[width=4cm, height=3.5cm]{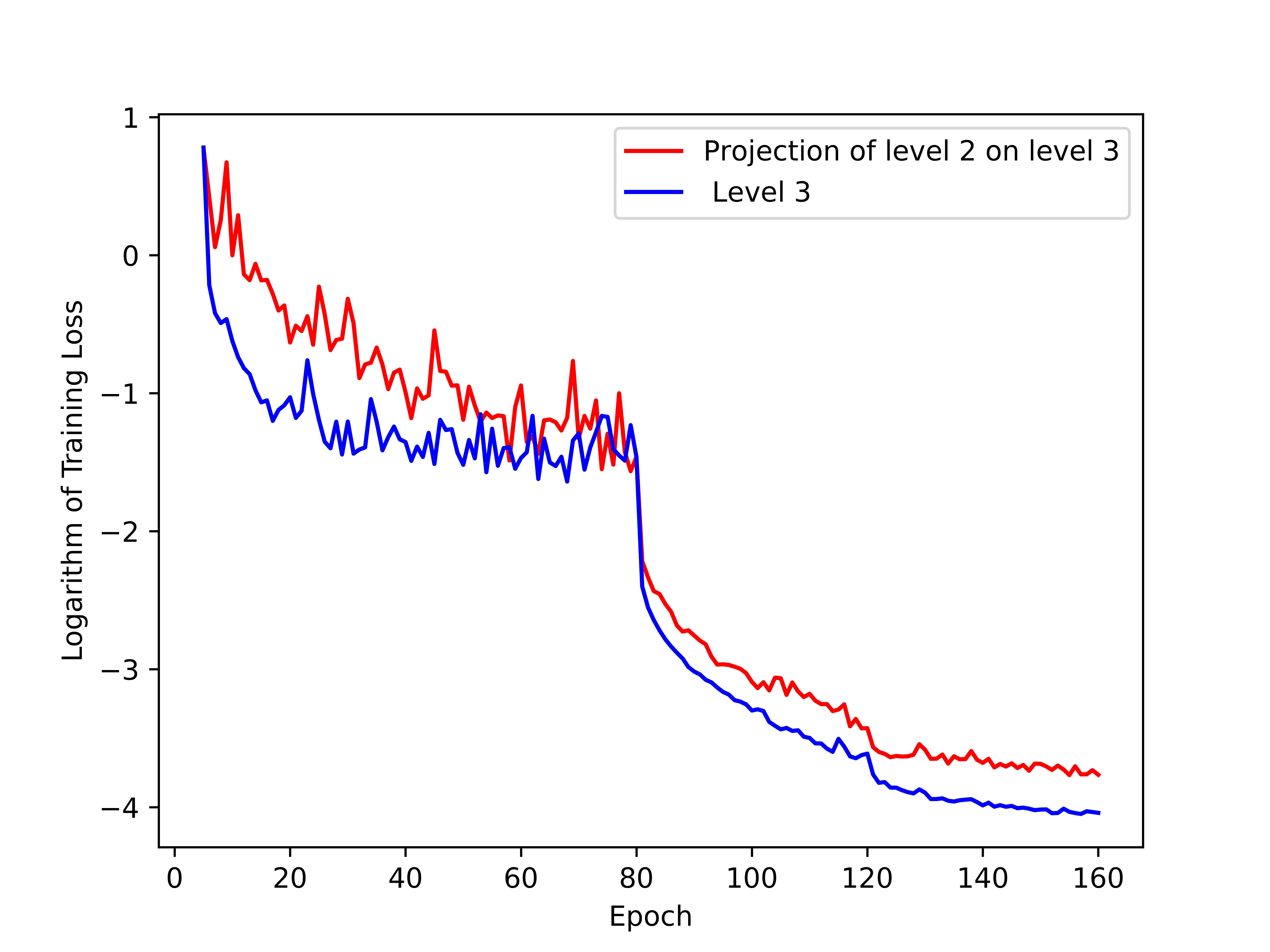}
\includegraphics[width=4cm, height=3.5cm]{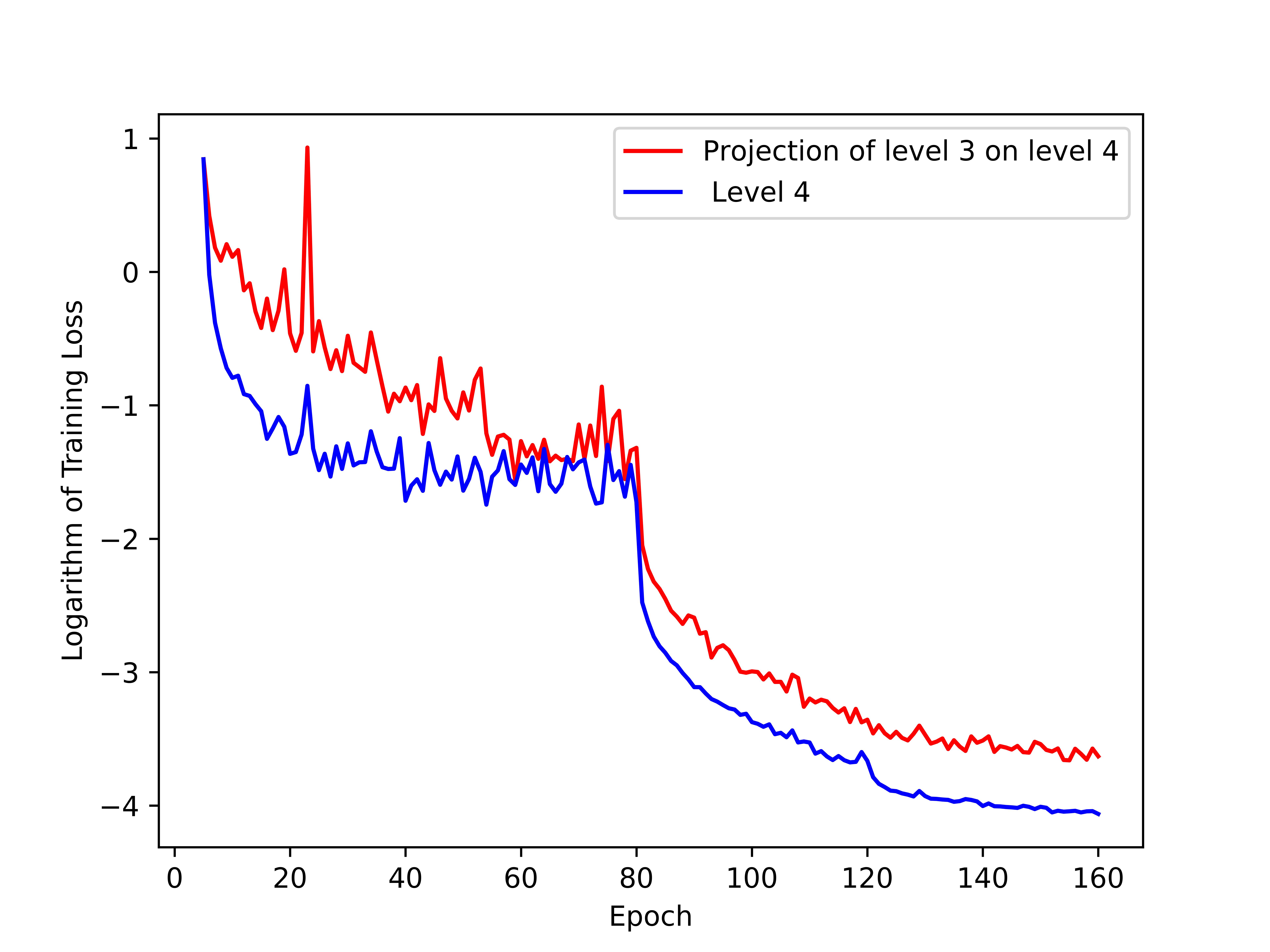} 
\includegraphics[width=4cm, height=3.5cm]{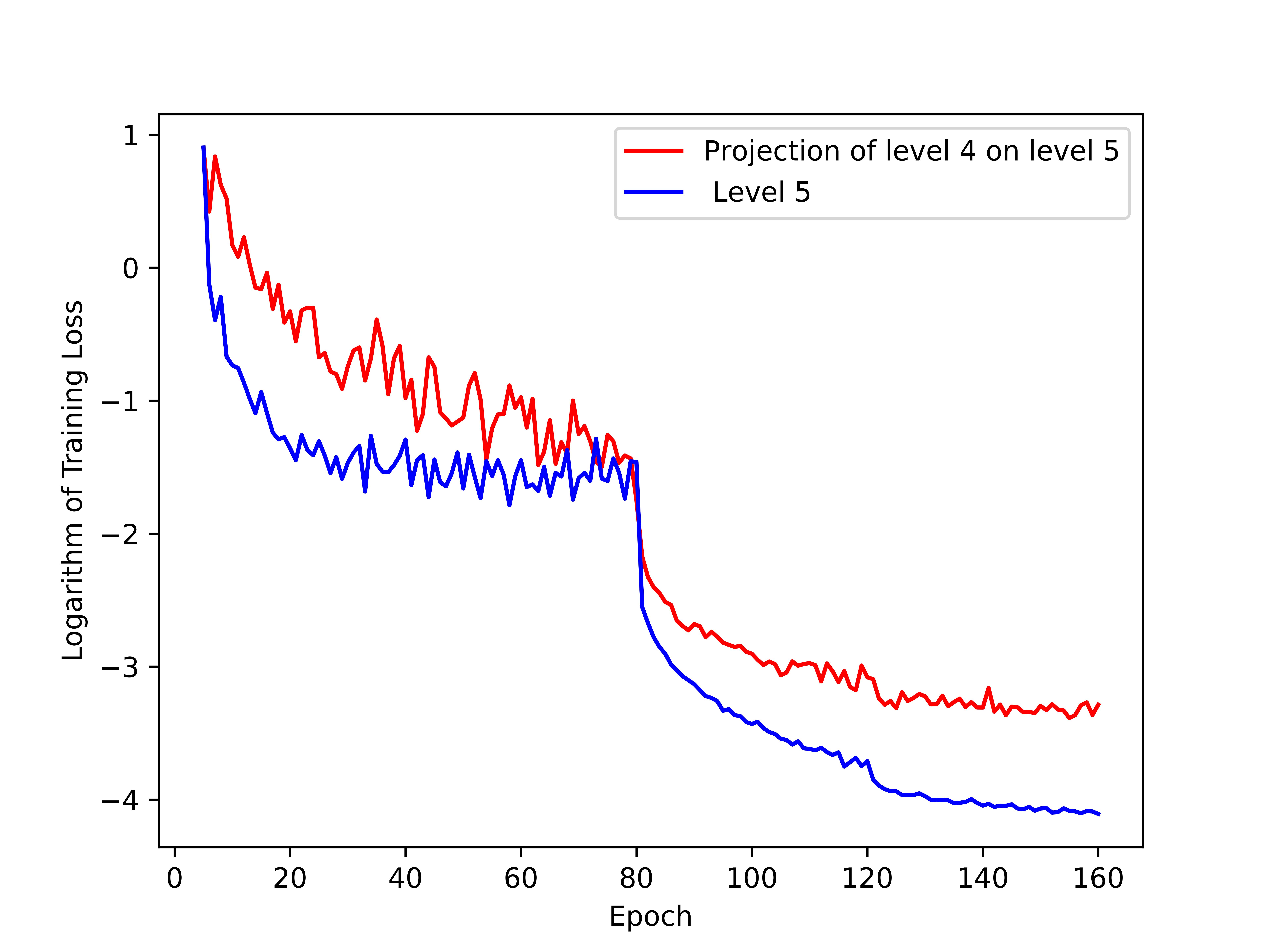}
\includegraphics[width=4cm, height=3.5cm]{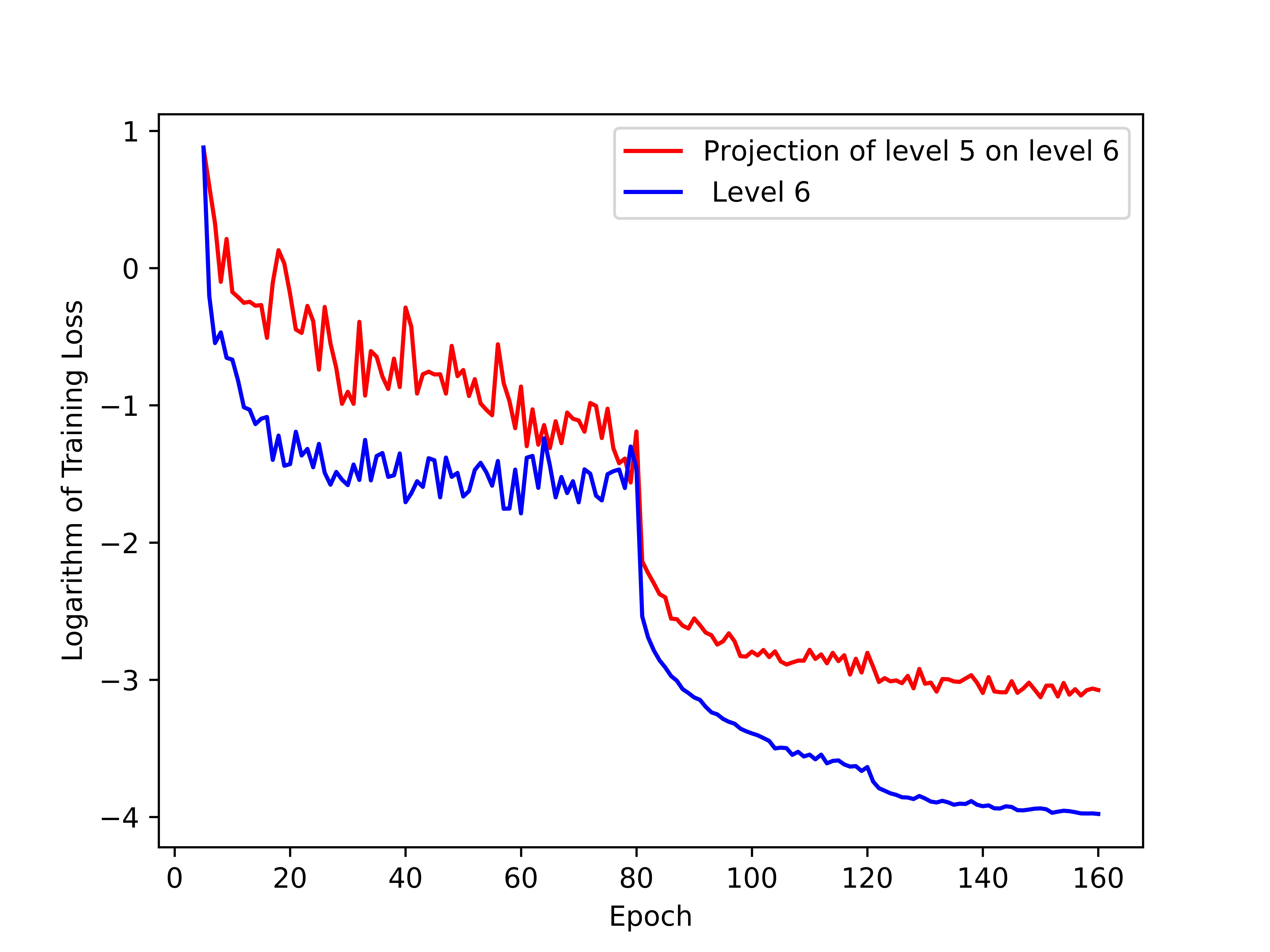}
\includegraphics[width=4cm, height=3.5cm]{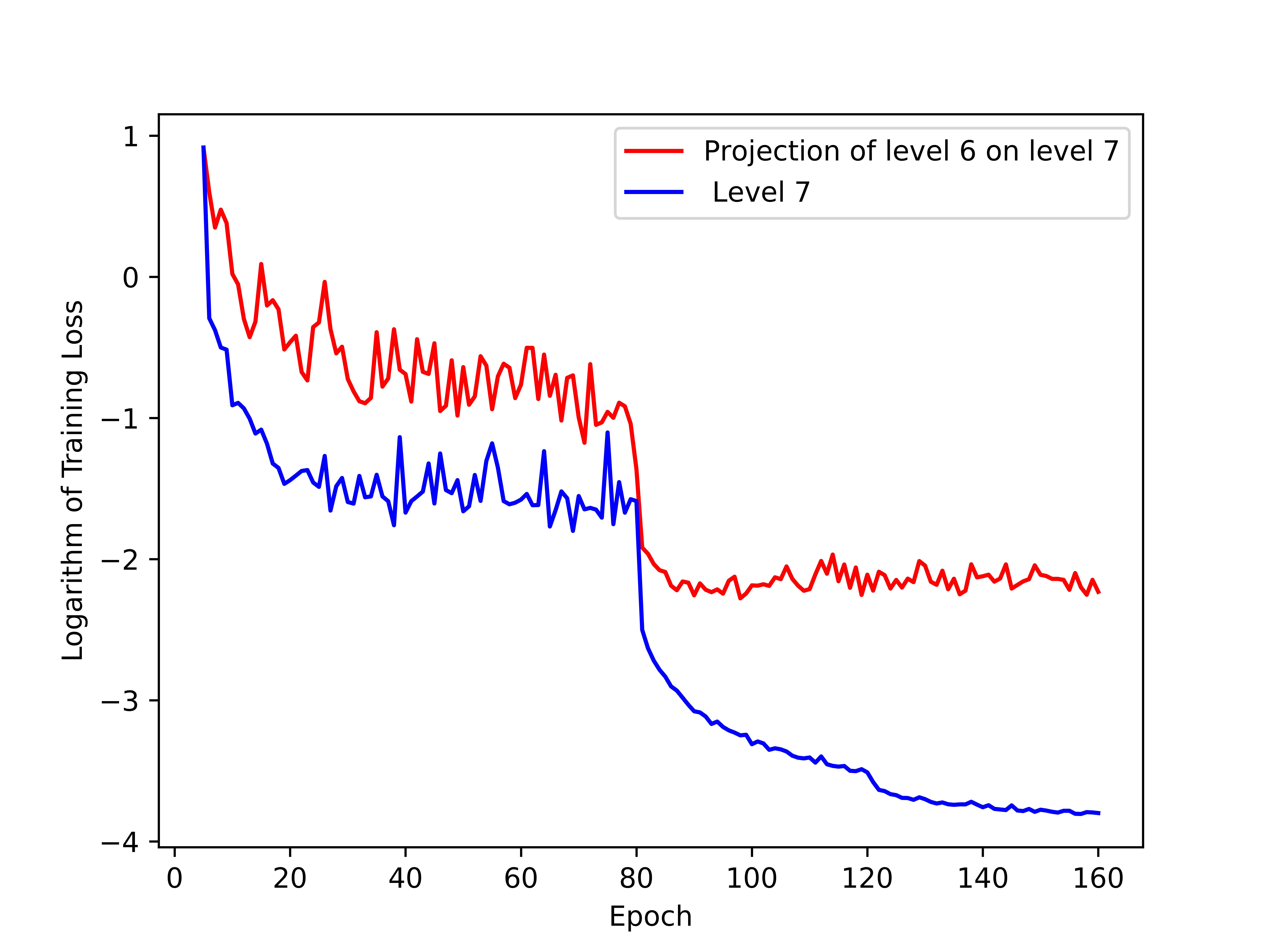} 
\includegraphics[width=4cm, height=3.5cm]{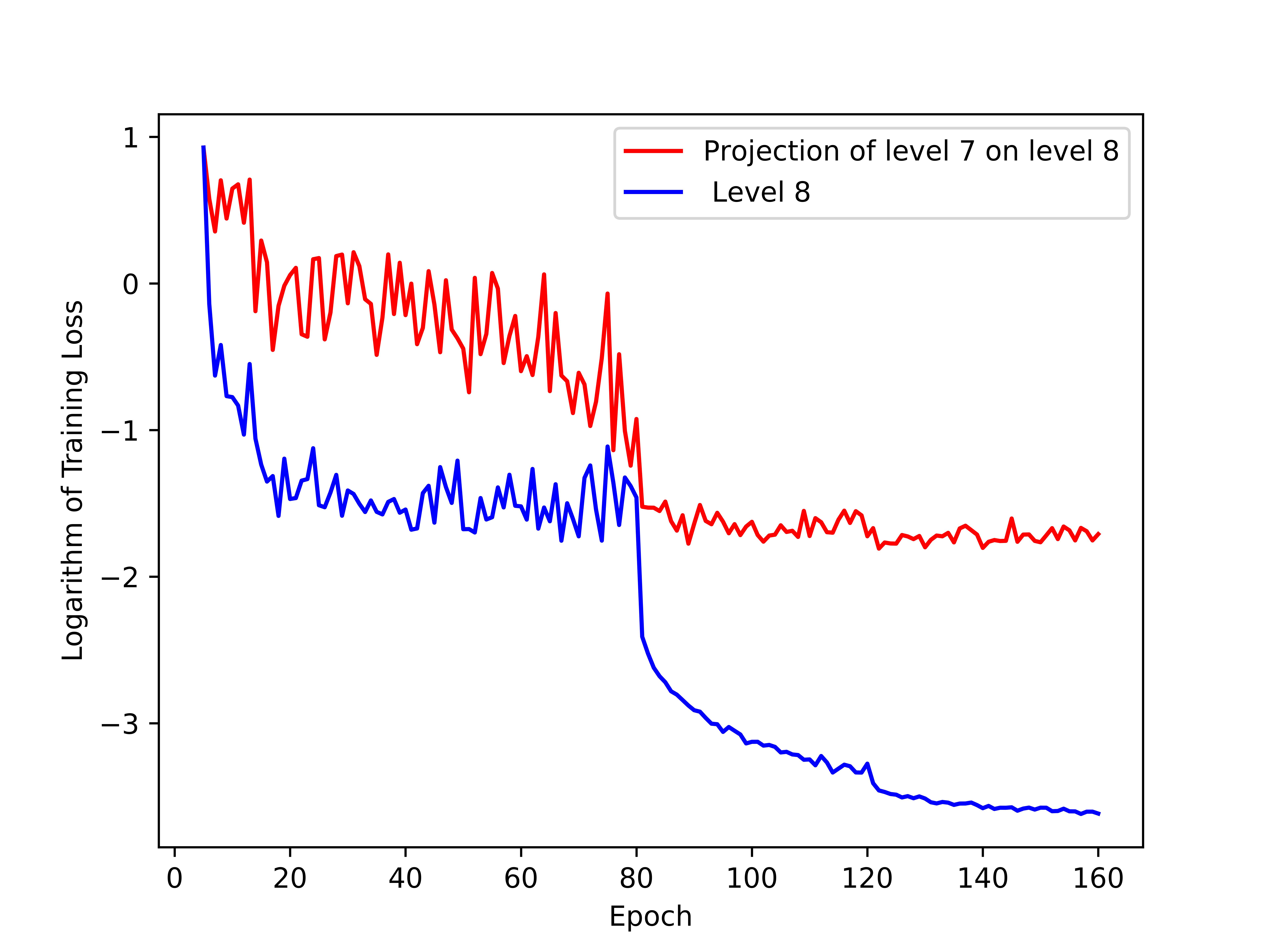}
\includegraphics[width=4cm, height=3.5cm]{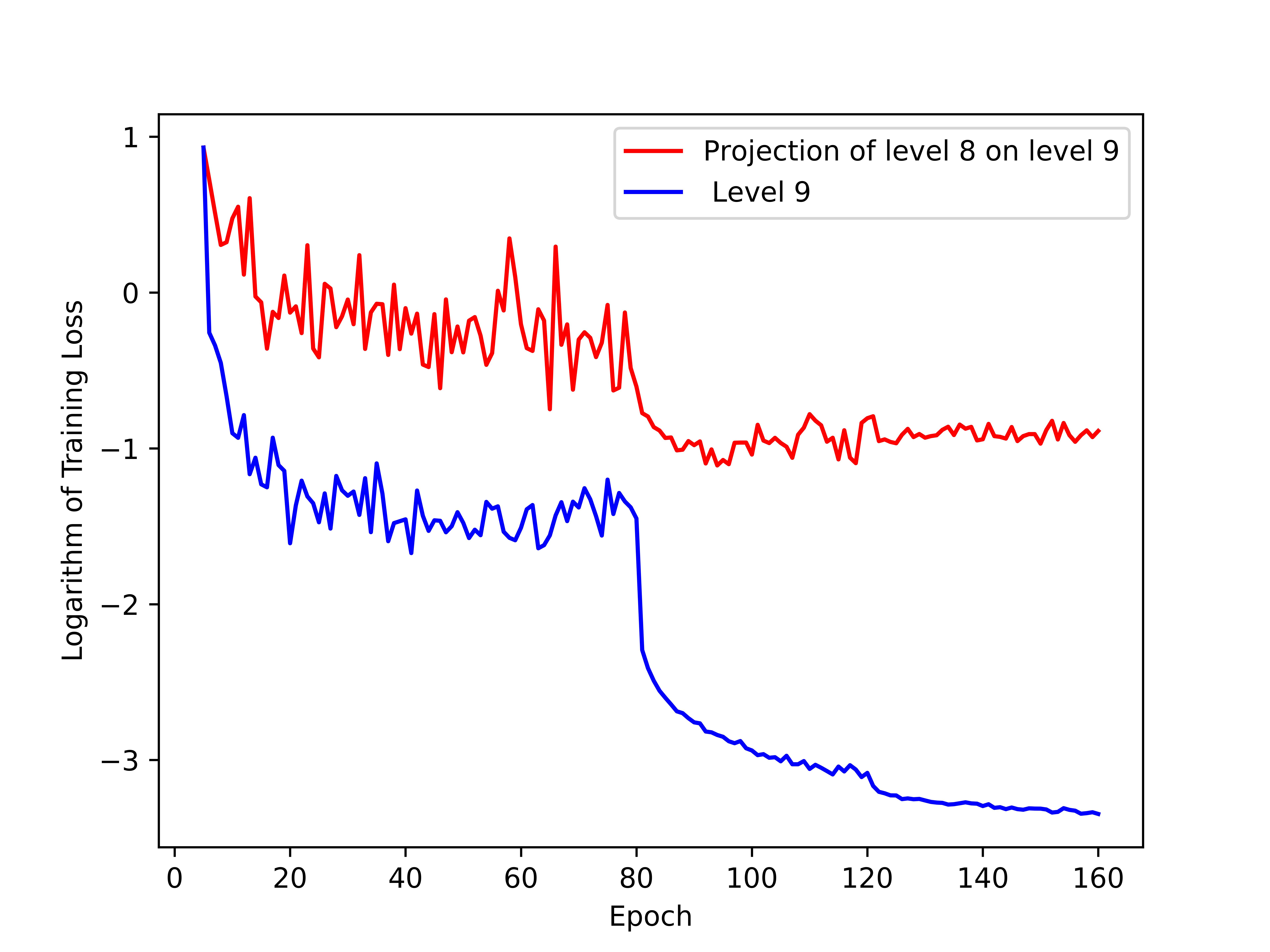}
\includegraphics[width=4cm, height=3.5cm]{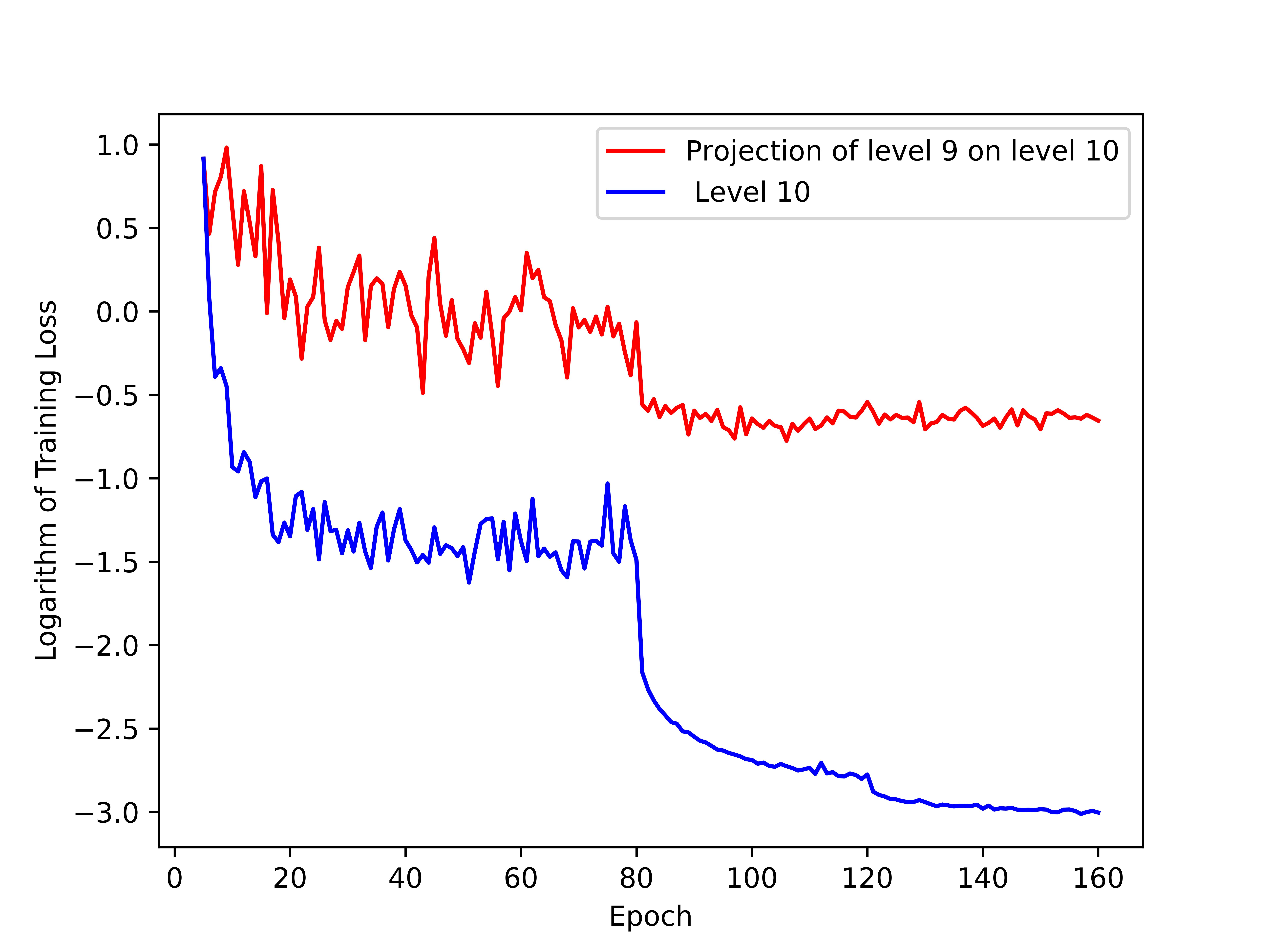} 
\caption{Comparison of logarithm of training loss versus epoch between level $(L)$ and level $(L-1)$ projected on level $(L)$ for $L$ ranging from $1$ to $10$.}
\end{figure*}
It is evident from the figure that the trajectory for level $(L)$ is steeper than for level $(L-1)$ projected on level $L$. This makes SGD converge to  $W_{(L)}^{(min\_(L))}$ and not to $W^{Pr{(min\_(L-1))}}_{(L)}$. \par Further, if we compare the volume of basins around $W_{(L)}^{(min\_(L))}$ and $W^{Pr{(min\_(L-1))}}_{(L)}$ by calculating the product of top-$100$ positive eigen values of the Hessian (of the loss function), the volume of basin around $W_{(L)}^{(min\_(L))}$ is seen to be larger than the volume of the basin around $W^{Pr{(min\_(L-1))}}_{(L)}$. This is demonstrated in Fig. 5 and Table I.
\begin{figure*}
\centering
\includegraphics[width=4cm, height=3.5cm]{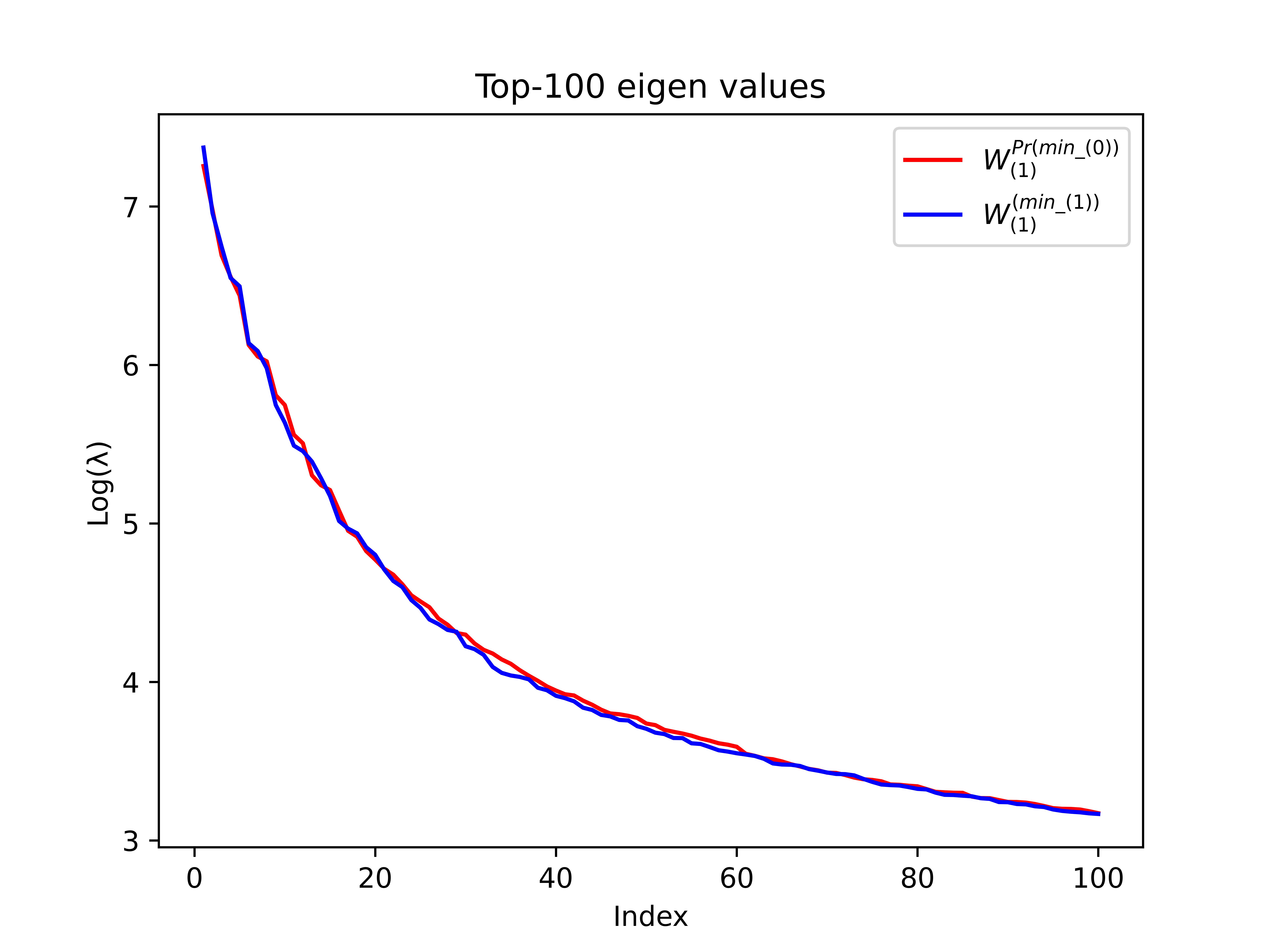}
\includegraphics[width=4cm, height=3.5cm]{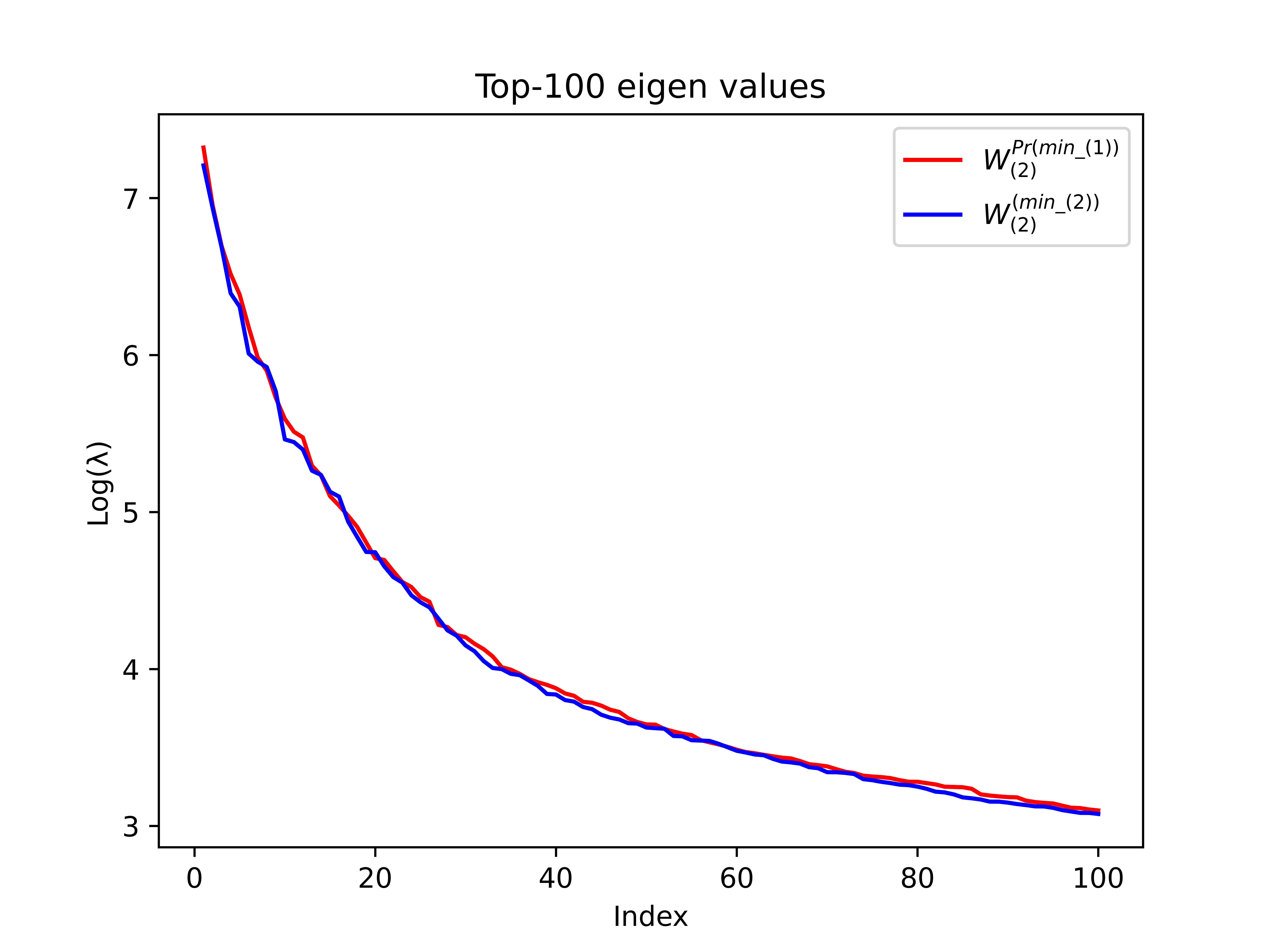}
\includegraphics[width=4cm, height=3.5cm]{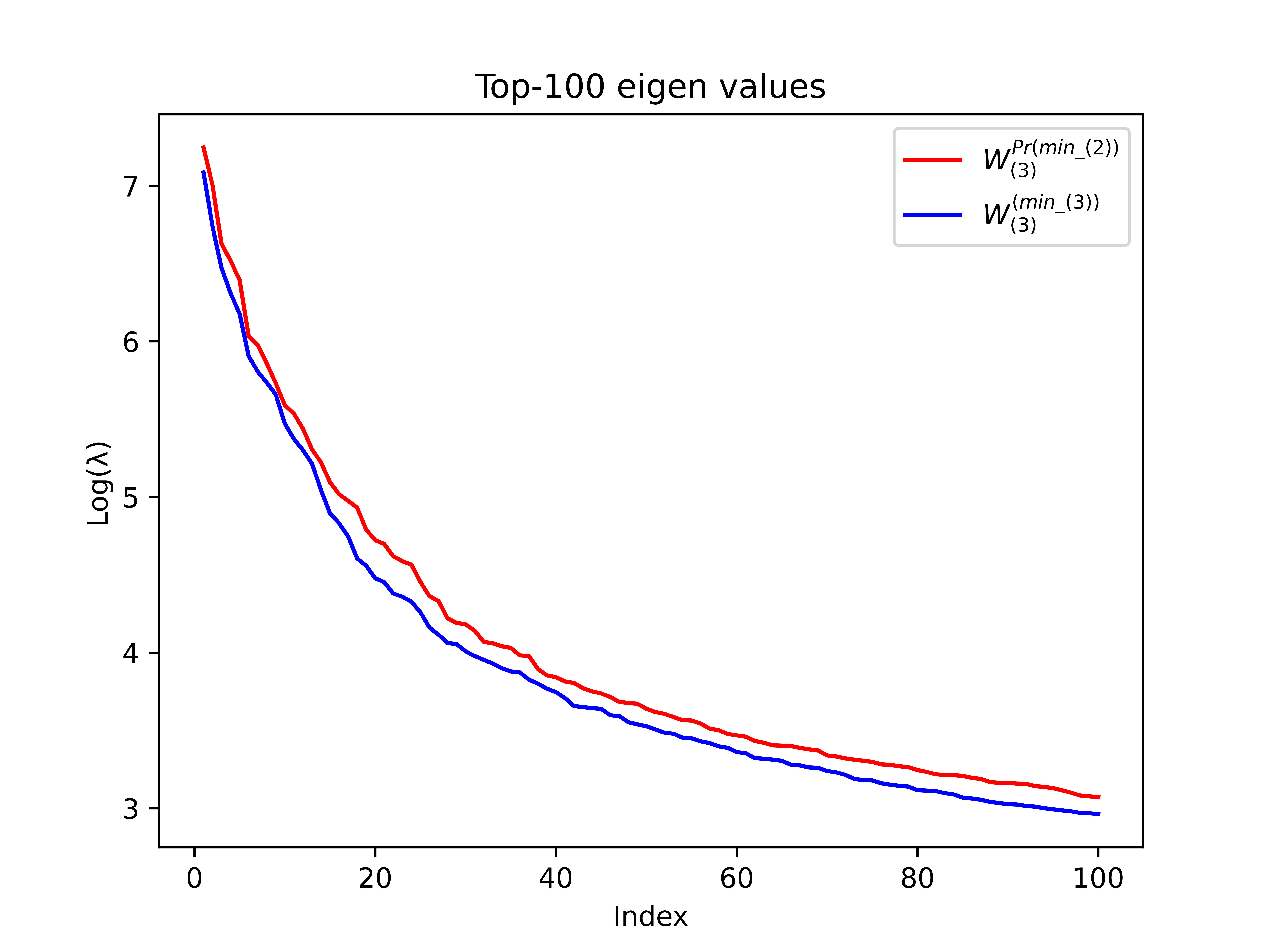}
\includegraphics[width=4cm, height=3.5cm]{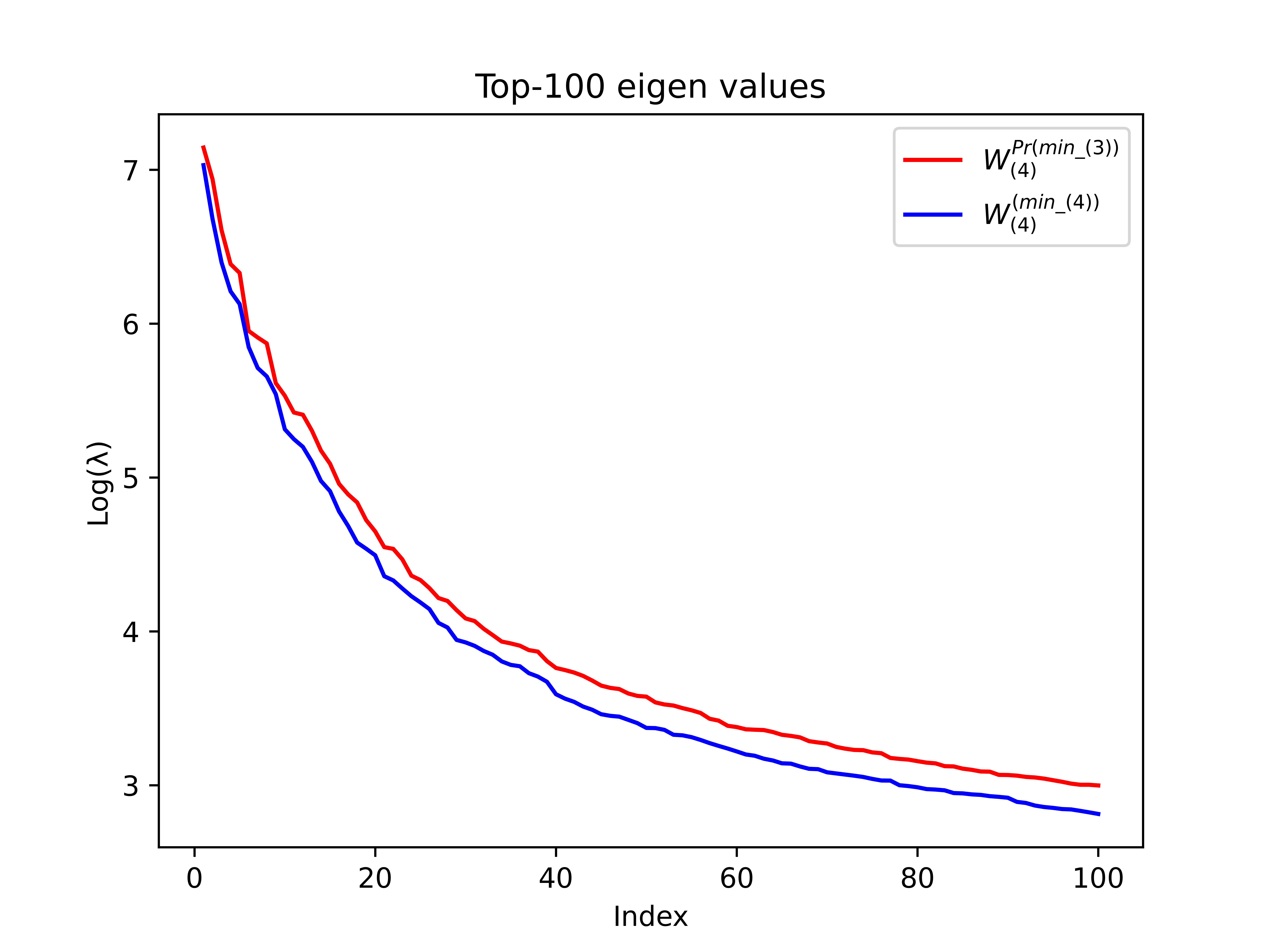} 
\includegraphics[width=4cm, height=3.5cm]{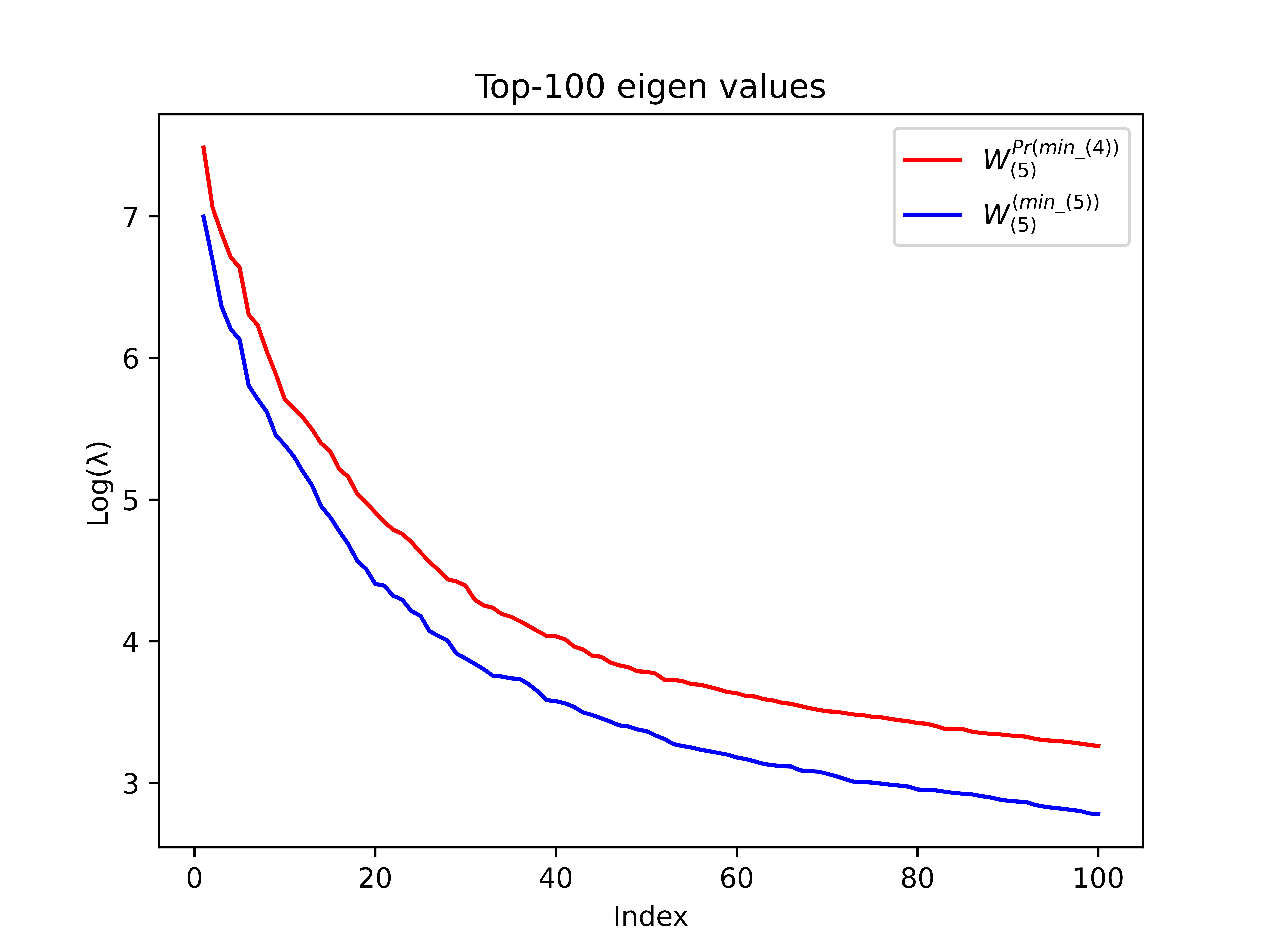}
\includegraphics[width=4cm, height=3.5cm]{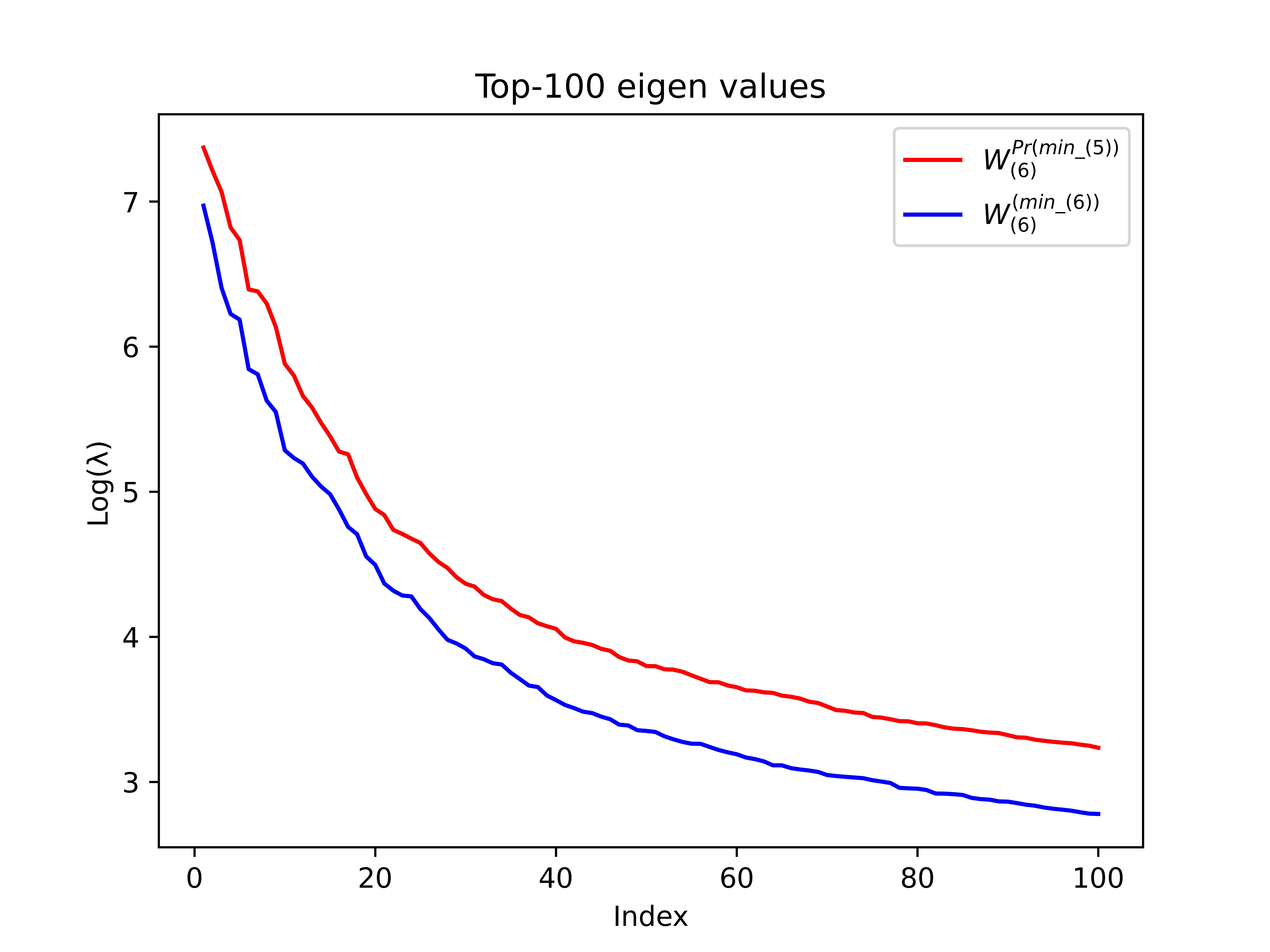}
\includegraphics[width=4cm, height=3.5cm]{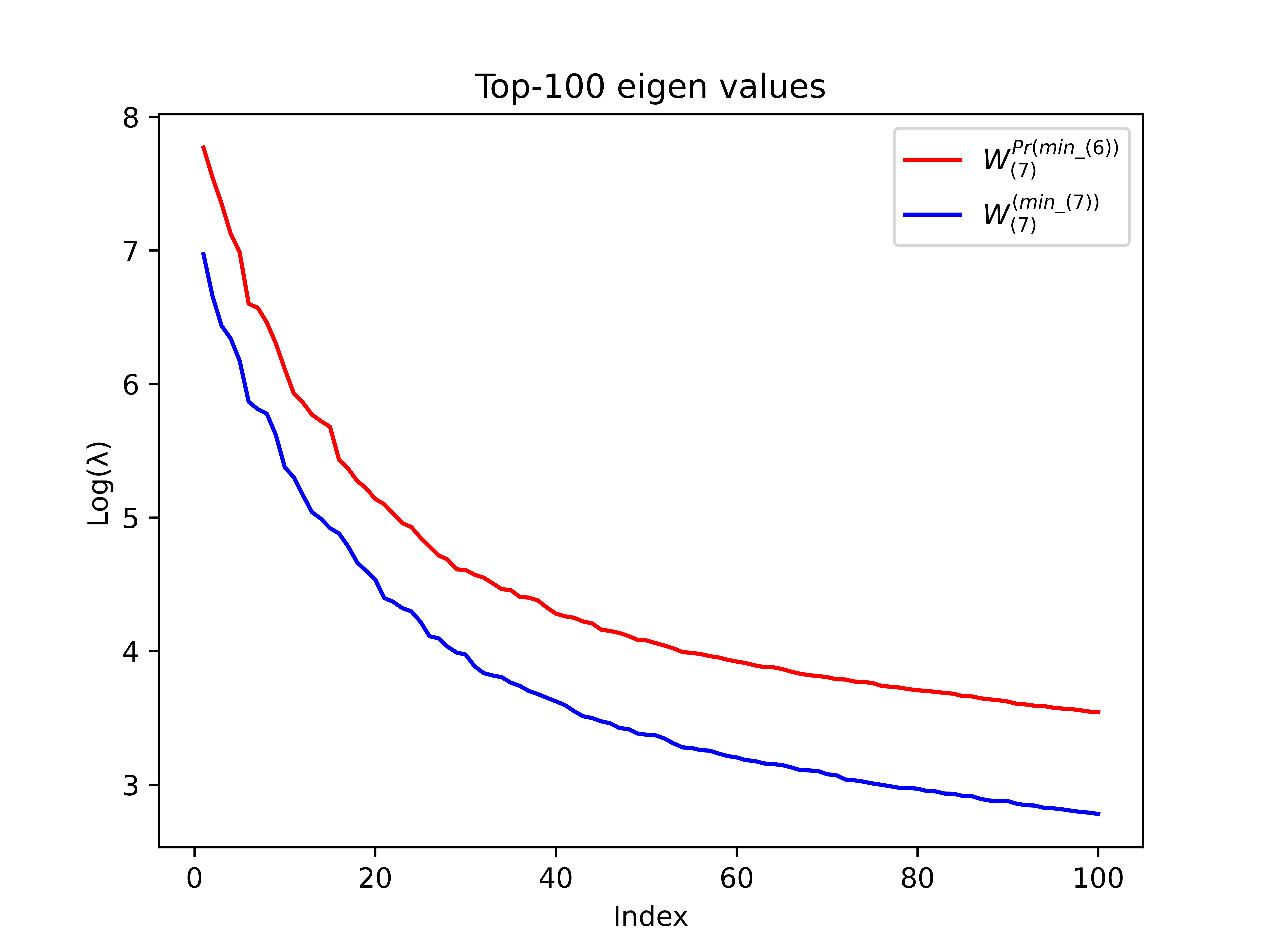} 
\includegraphics[width=4cm, height=3.5cm]{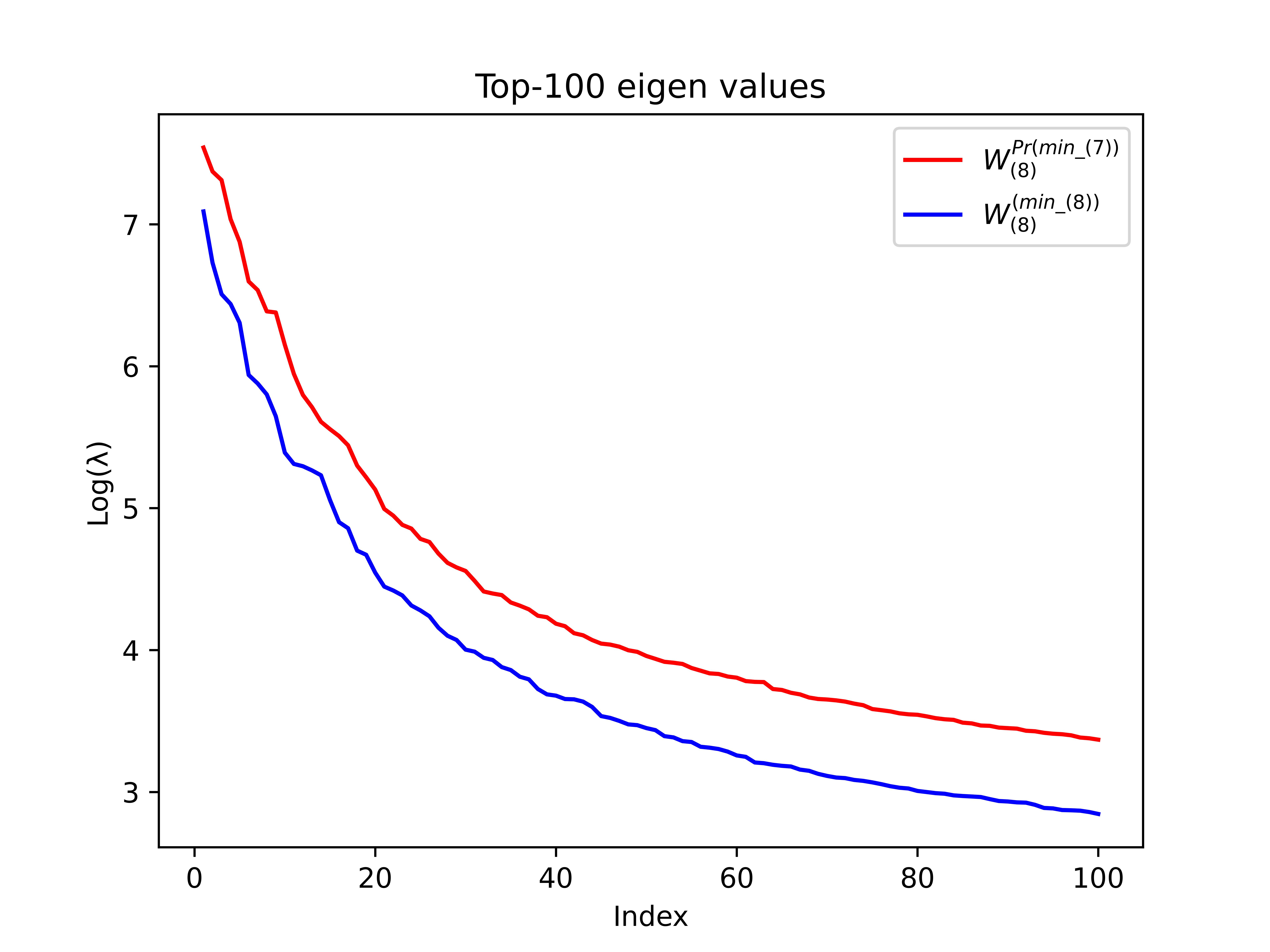}
\includegraphics[width=4cm, height=3.5cm]{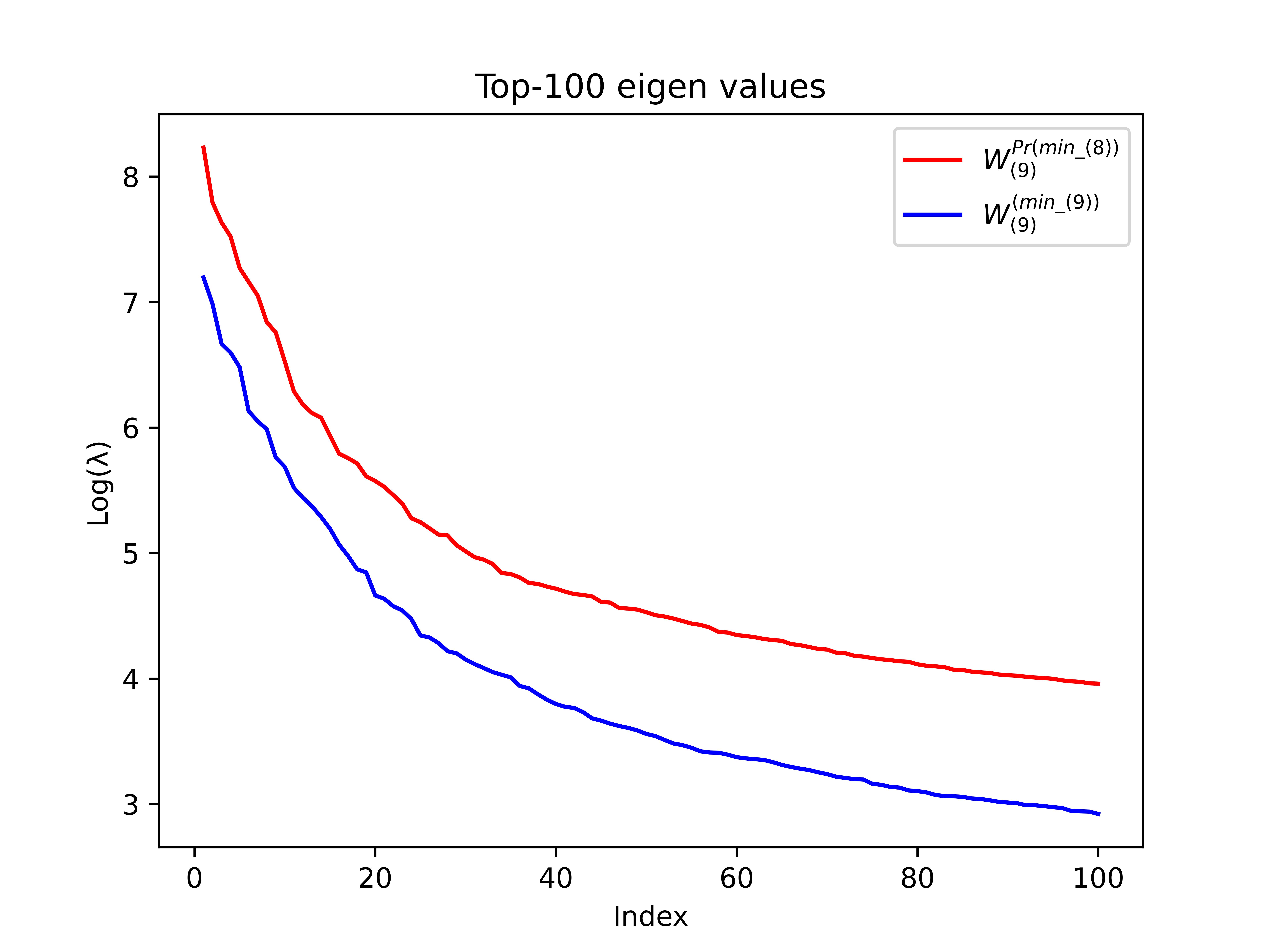}
\includegraphics[width=4cm, height=3.5cm]{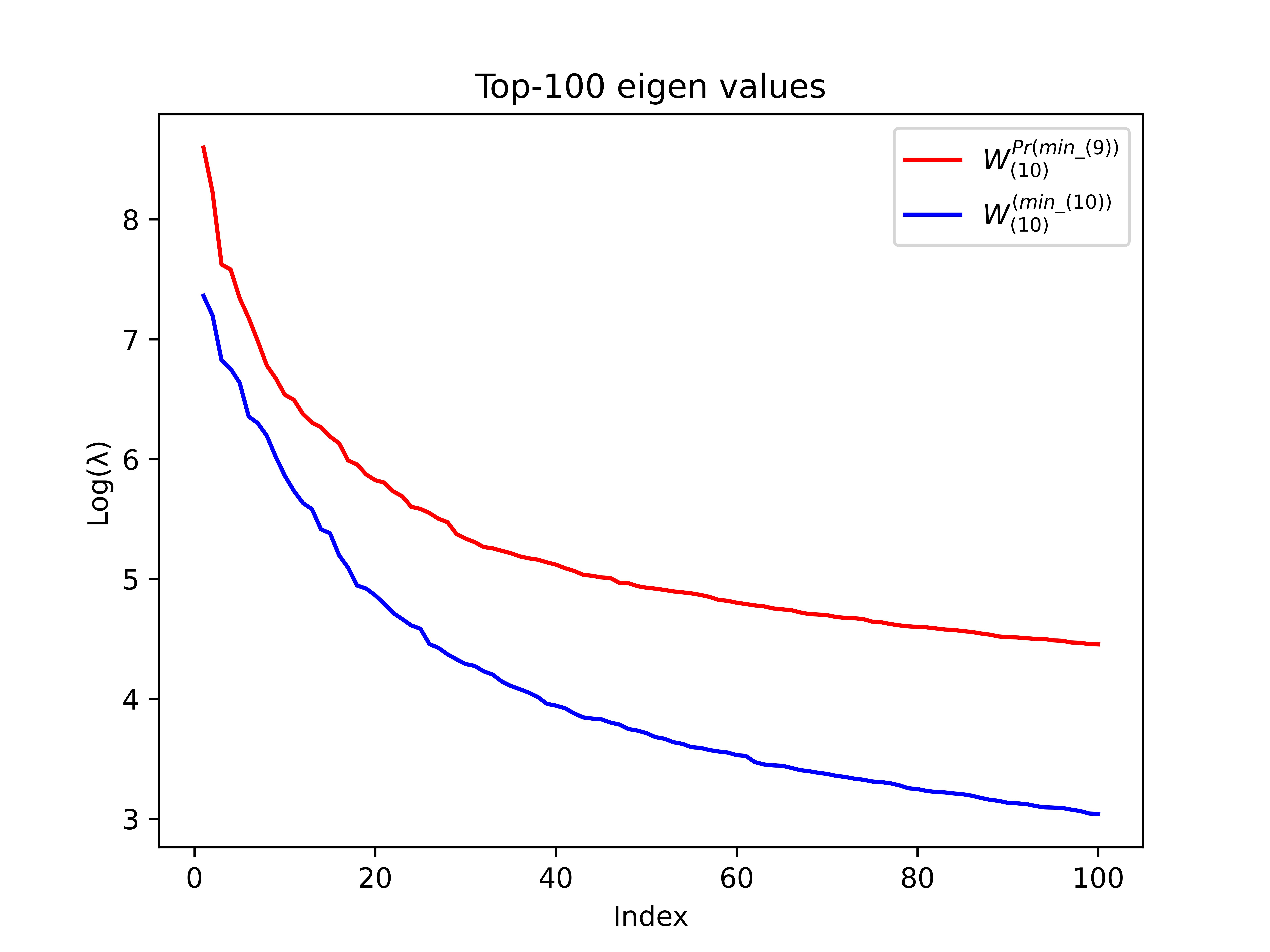} 
\caption{Comparison of top-100 positive eigen values of the Hessian at $W_{(L)}^{(min\_(L))}$ and $W^{Pr{(min\_(L-1))}}_{(L)}$ for $L$ ranging from $1$ to $10$. The figure shows that the eigen values of the Hessian at $W_{(L)}^{(min\_(L))}$ are smaller than that at $W^{Pr{(min\_(L-1))}}_{(L)}$. And smaller the eigen values, the smaller their product will be, and the larger would be the volume of the basin around the minimum.} 
\end{figure*}

\begin{table}[h!]
\caption{Comparison of inverse volume of basin, $V^{'} (100)$ at $W^{Pr{(min\_(L-1))}}_{(L)}$ and $W_{(L)}^{(min\_(L))}$.}
\begin{center}
\begin{tabular}{|p{0.03\textwidth} | p{0.14\textwidth}  | p{0.10\textwidth}|} 
  \hline
 $L$ &Solution&  $V^{'} (100)$  \\ 
  
  \hline
  \multirow{2}{4em}{1} 
&$W^{Pr{(min\_(0))}}_{(1)}$
 &409.242\\ \cline{2-3}
 &$W_{(1)}^{(min\_(1))}$
 & 407.417\\
  \hline
  \multirow{2}{4em}{2}
  &$W^{Pr{(min\_(1))}}_{(2)}$  
 & 402.678\\\cline{2-3}
 &$W_{(2)}^{(min\_(2))}$
 & 399.680\\ 
  \hline 
 \multirow{2}{4em}{3}
  &$W^{Pr{(min\_(2))}}_{(3)}$ 
 & 400.855\\ \cline{2-3}
  &$W_{(3)}^{(min\_(3))}$
 & 386.948\\ 
  \hline
\multirow{2}{4em}{4}
    &$W^{Pr{(min\_(3))}}_{(4)}$ 
 & 392.829\\ \cline{2-3}
  &$W_{(4)}^{(min\_(4))}$
  & 375.590\\
  \hline
    \multirow{2}{4em}{5}
 &$W^{Pr{(min\_(4))}}_{(5)}$ 
 &417.984\\ \cline{2-3}
 &$W_{(5)}^{(min\_(5))}$ 
 & 372.753\\ 
  \hline
  \multirow{2}{4em}{6}
 &$W^{Pr{(min\_(5))}}_{(6)}$
 & 420.431\\  \cline{2-3}
 &$W_{(6)}^{(min\_(6))}$
 & 373.472\\ 
  \hline 
 \multirow{2}{4em}{7}
   &$W^{Pr{(min\_(6))}}_{(7)}$ 
 & 446.758\\  \cline{2-3}
  &$W_{(7)}^{(min\_(7))}$
 & 375.401\\   
  \hline
\multirow{2}{4em}{8}
    &$W^{Pr{(min\_(7))}}_{(8)}$
 & 432.505\\   \cline{2-3}
  &$W_{(8)}^{(min\_(8))}$
  & 382.051\\
  \hline
   \multirow{2}{4em}{9}
   &$W^{Pr{(min\_(8))}}_{(9)}$
  &488.004\\  \cline{2-3}
  &$W_{(9)}^{(min\_(9))}$
 & 394.166\\  
  \hline
\multirow{2}{4em}{10}
 &$W^{Pr{(min\_(9))}}_{(10)}$
 & 525.621\\    \cline{2-3}
  &$W_{(10)}^{(min\_(10))}$  & 408.810\\

  \hline
\end{tabular}
\label{product}
\end{center}
\end{table}
We also confirmed this by using the method given by Huang et al. \cite{huang2020understanding} to calculate the radii of the basin. We calculated the radii of the basin along $500$ different random directions around $W_{(10)}^{(min\_(10))}$ and $W^{Pr{(min\_(9))}}_{(10)}$. The radius of the basin in each direction has been calculated by choosing a suitable cut-off value in proximity to the loss values at the minima:
\begin{equation}
cut-off=2 \times max\{Loss(W_{(10)}^{(min\_(10))}), Loss(W^{Pr{(min\_(9))}}_{(10)})\}
\end{equation}
This normalized cut-off value turns out to be $1.0$ in this case using equation $5$. A comparison of radii of the basin, around $W_{(10)}^{(min\_(10))}$ and $W^{Pr{(min\_(9))}}_{(10)}$ is given in Fig. 6 and the average radii of basins around $W_{(10)}^{(min\_(10))}$ and $W^{Pr{(min\_(9))}}_{(10)}$ are given in Table II. It is clear from the figure and the table that the average radius of the basin around $W_{(10)}^{(min\_(10))}$ is larger than that around $W^{Pr{(min\_(9))}}_{(10)}$. Since the neural network loss functions lie in high-dimensional spaces, even a small difference in the radii of basins around minima translates to exponentially large disparities in the volume of their surrounding basins.
\begin{figure}[h!]
\centering
\includegraphics[width=4cm, height=3.5cm]{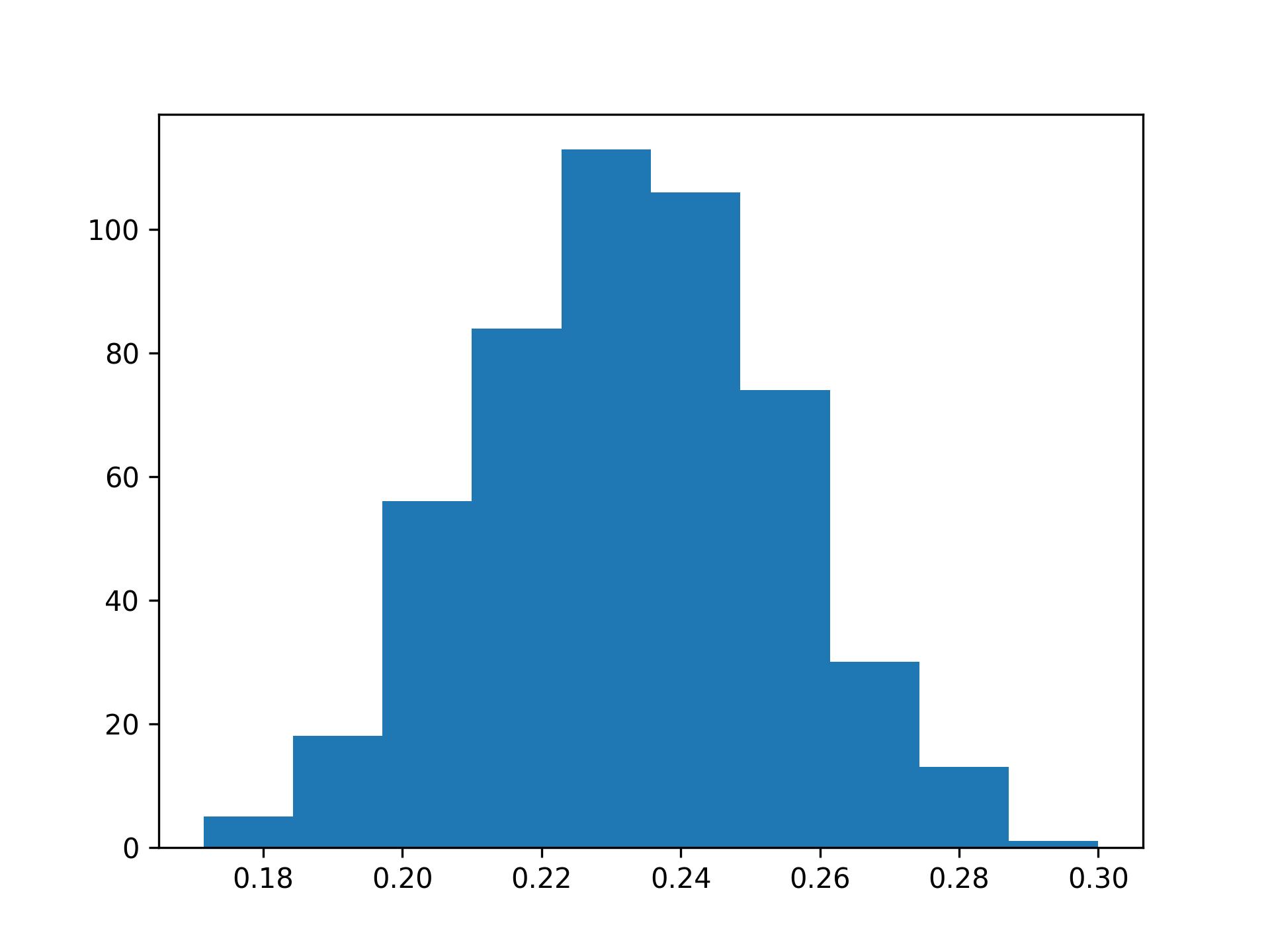}
\includegraphics[width=4cm, height=3.5cm]{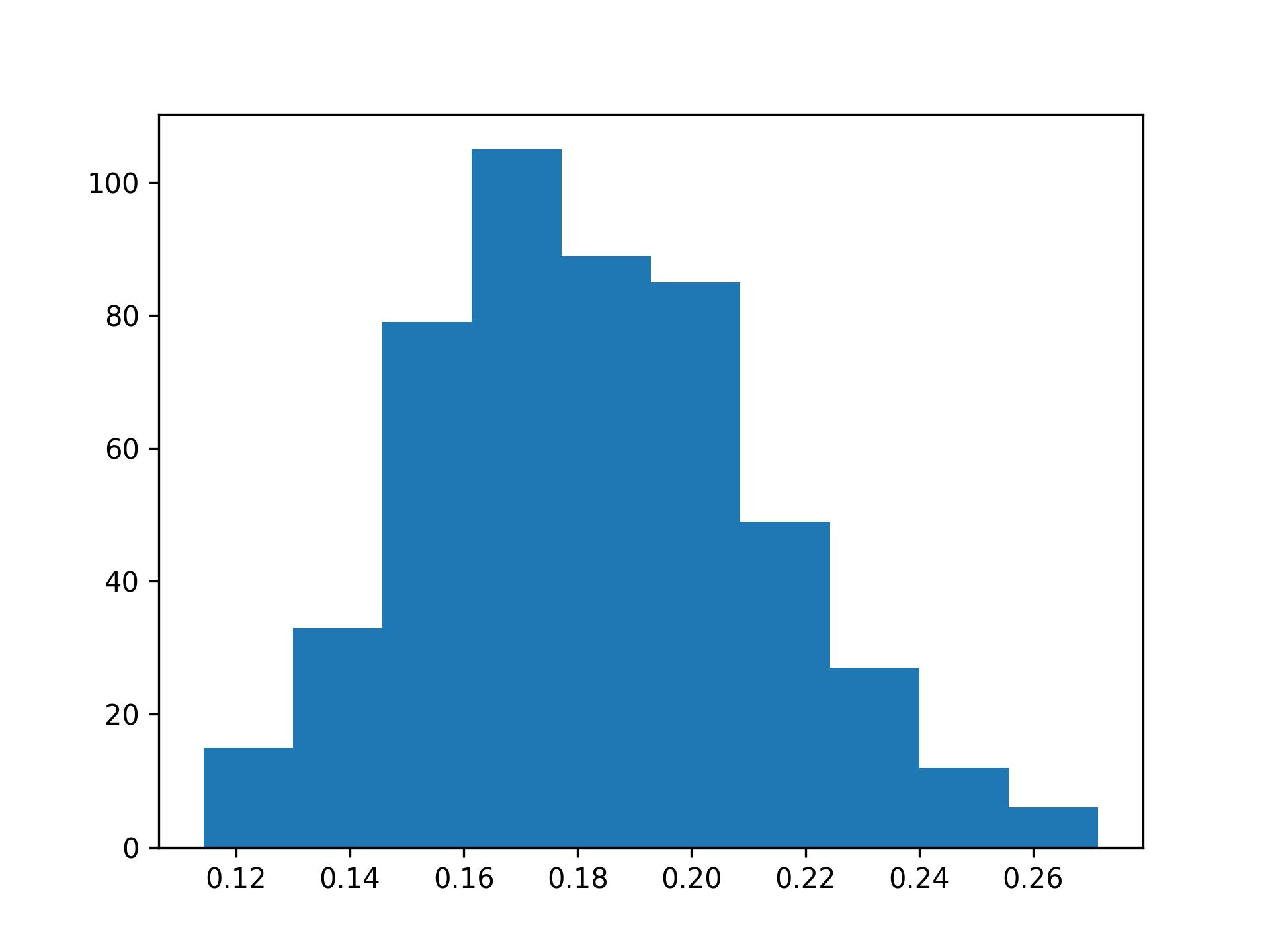}   
\caption{Comparison of  radii of basin along $500$ random directions around $W_{(L)}^{(min\_(L))}$ and $W^{Pr{(min\_(L-1))}}_{(L)}$ for $L=10$. The x-axis in the above histograms represents the radius of the basin, and the y-axis represents the number of random directions. \textbf{Left:} Radii of basin around $W_{(10)}^{(min\_(10))}$. \textbf{Right:} Radii of basin around $W^{Pr{(min\_(9))}}_{(10)}$.} 
\end{figure}

\begin{table}[h!]
\caption{Comparison of Average Radius of Basin around $W_{(L)}^{(min\_(L))}$ and $W^{Pr{(min\_(L-1))}}_{(L)}$ for $L=10$.  }
\begin{center}

\begin{tabular}{ | p{0.1\textwidth} | p{0.2\textwidth}|} 
  \hline
  Solution& Average Radius of Basin   \\ 
 
  \hline
$W_{(10)}^{(min\_(10))}$
 &0.232314
  \\ 
  \hline
  $W^{Pr{(min\_(9))}}_{(10)}$
 &0.183428 
 \\ 
  \hline
\end{tabular}
\end{center}
\end {table}
 Next, if we consider the reverse projection of level $(L)$ solution on level $(L-1)$, $W^{RPr{(min\_(L))}}_{(L-1)}$ and the level $(L-1)$ solution, $W_{(L-1)}^{(min\_(L-1))}$, the volume of basin around $W_{(L-1)}^{(min\_(L-1))}$ is seen to be larger than the volume of basin around $W^{RPr{(min\_(L))}}_{(L-1)}$. This is demonstrated in Fig. 7 and Table III.
\begin{figure*}
\centering
\includegraphics[width=4cm, height=3.5cm]{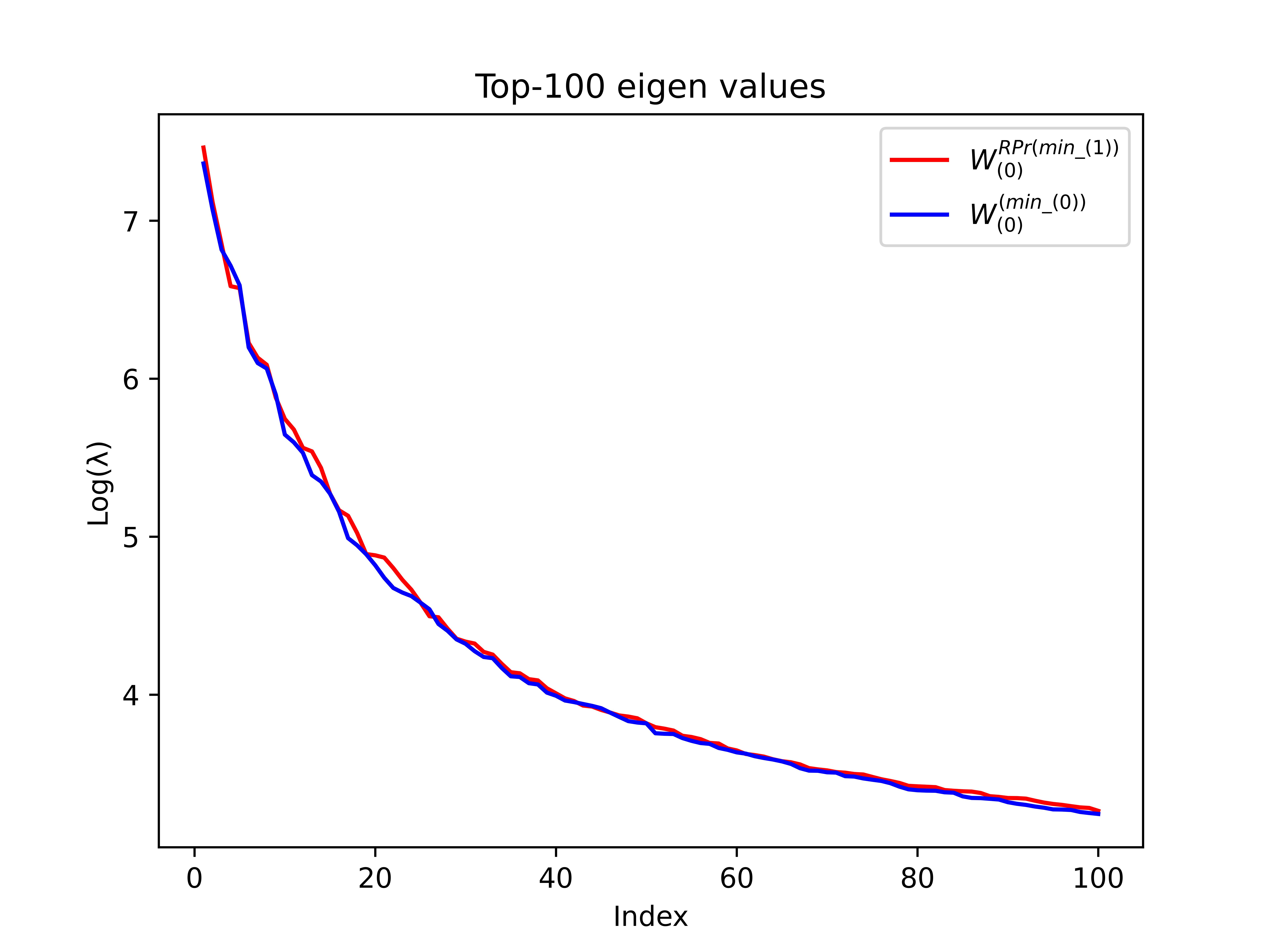} 
\includegraphics[width=4cm, height=3.5cm]{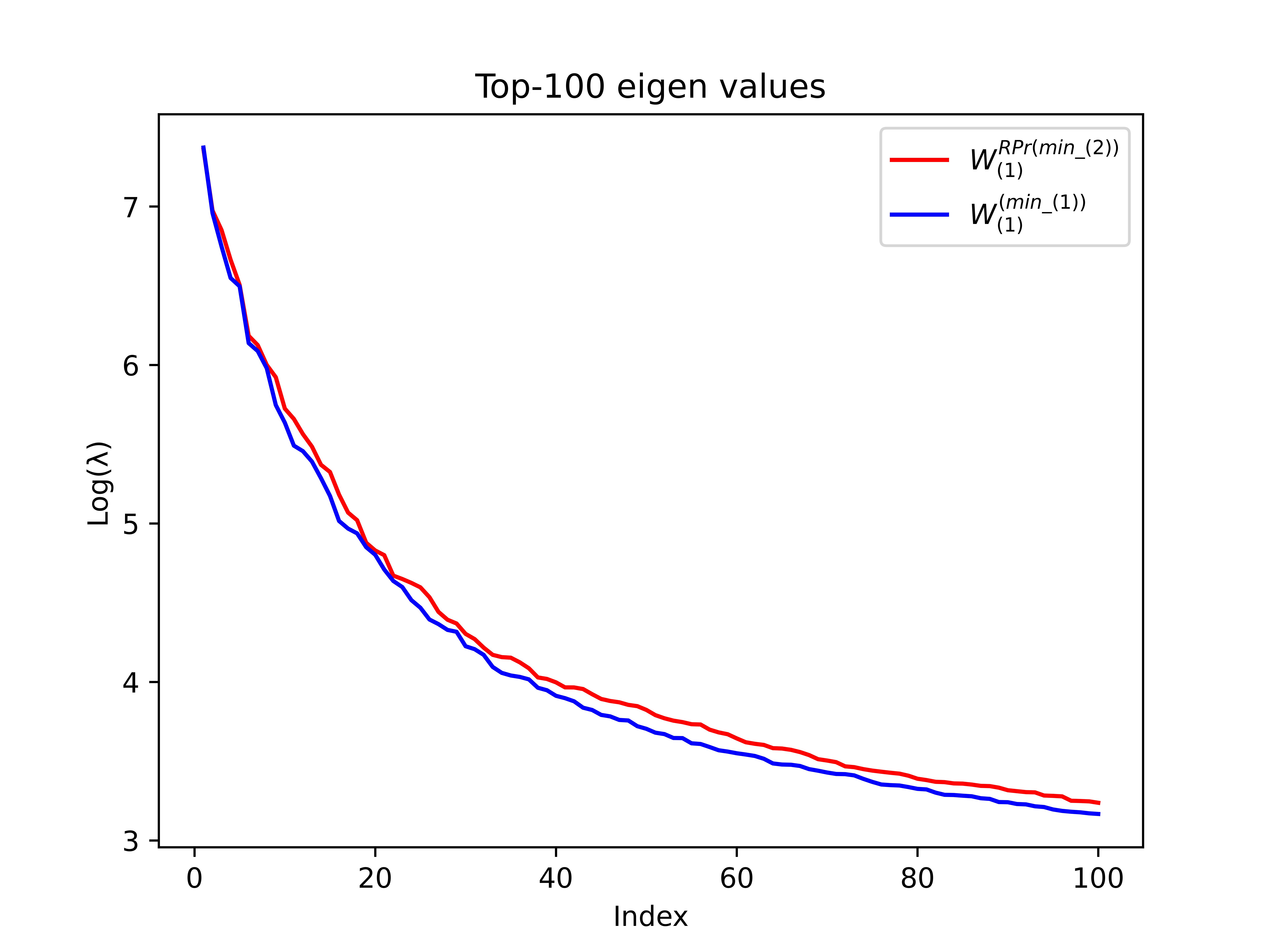} 
\includegraphics[width=4cm, height=3.5cm]{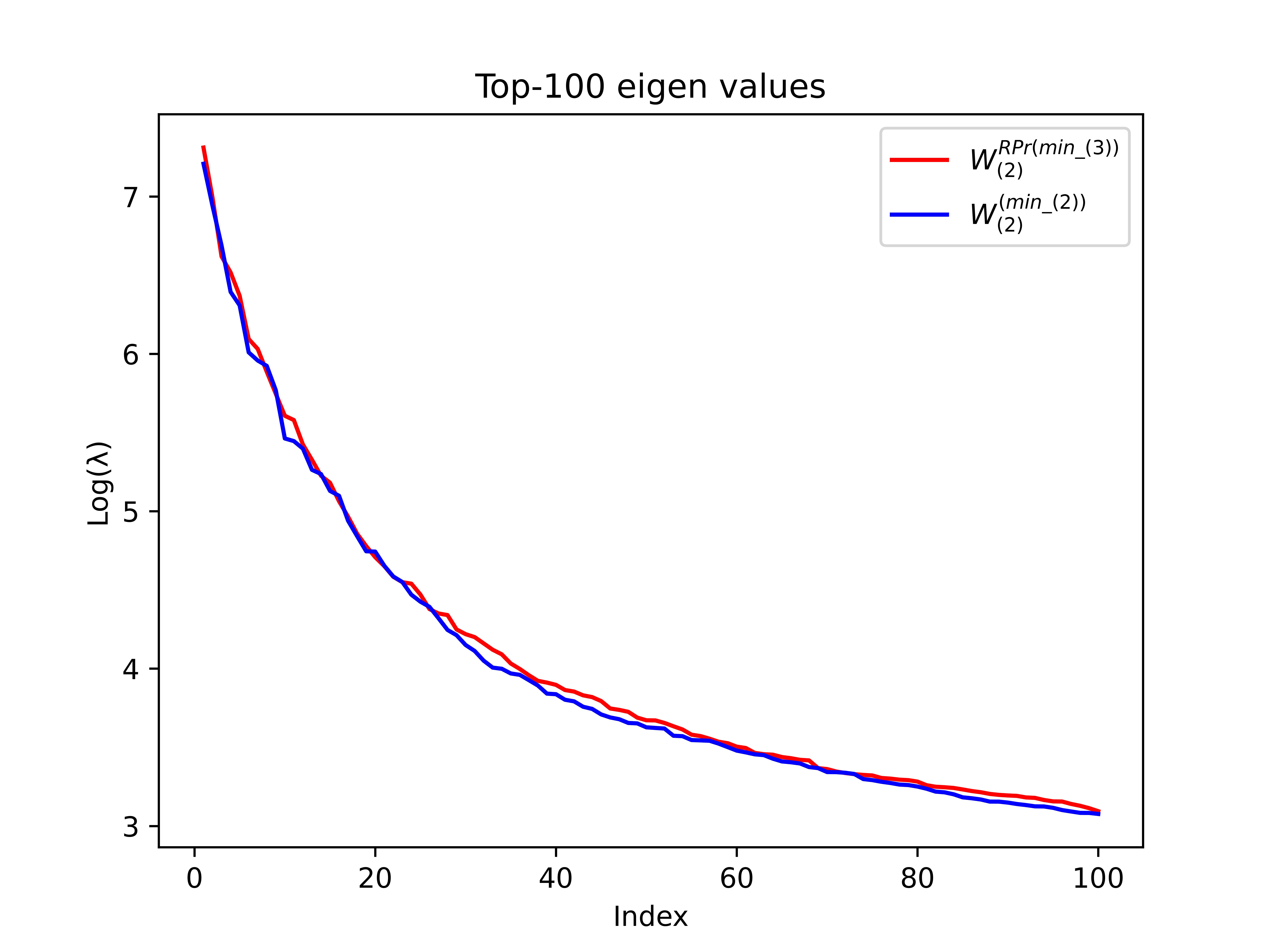} 
\includegraphics[width=4cm, height=3.5cm]{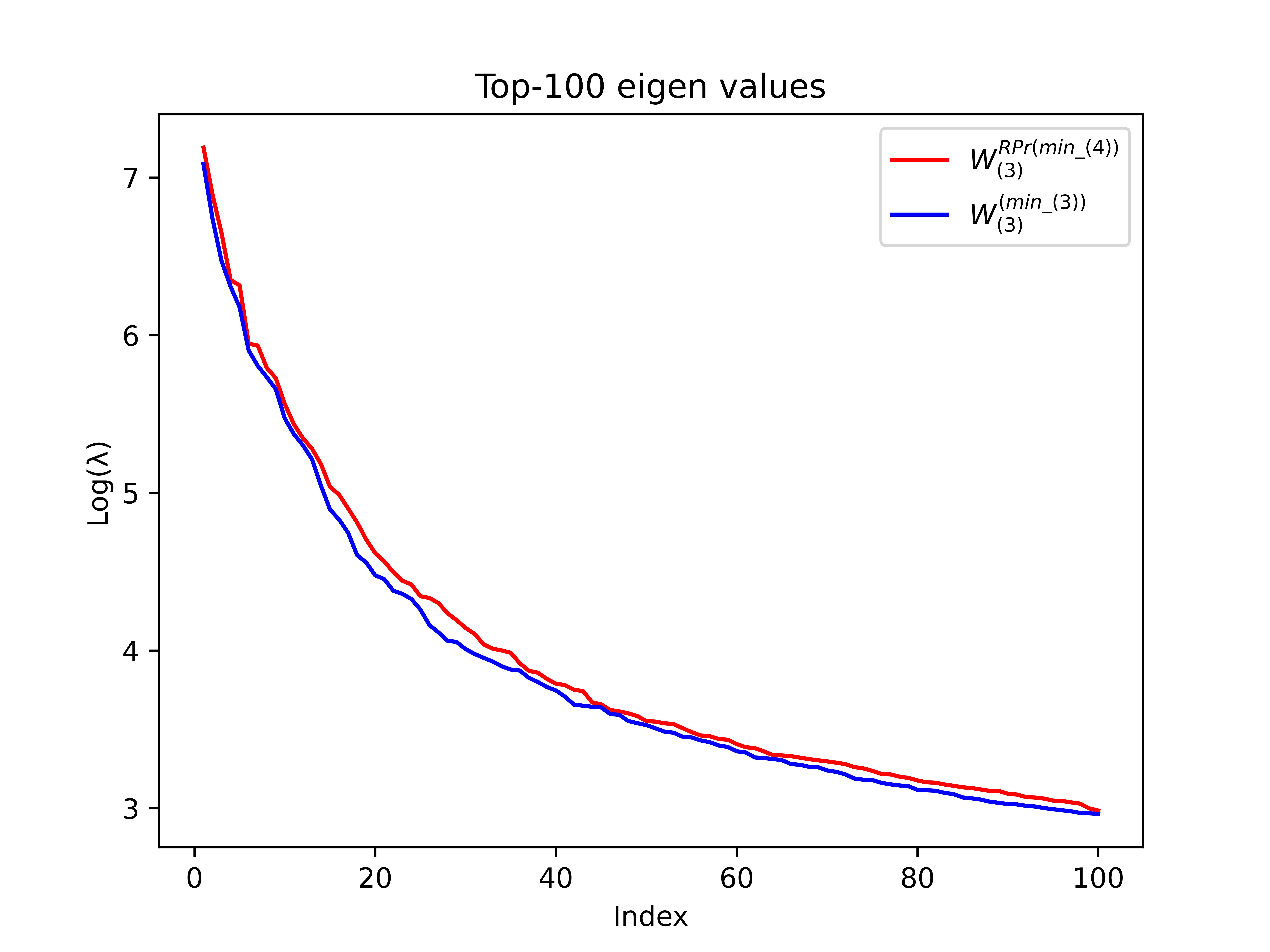} 
\includegraphics[width=4cm, height=3.5cm]{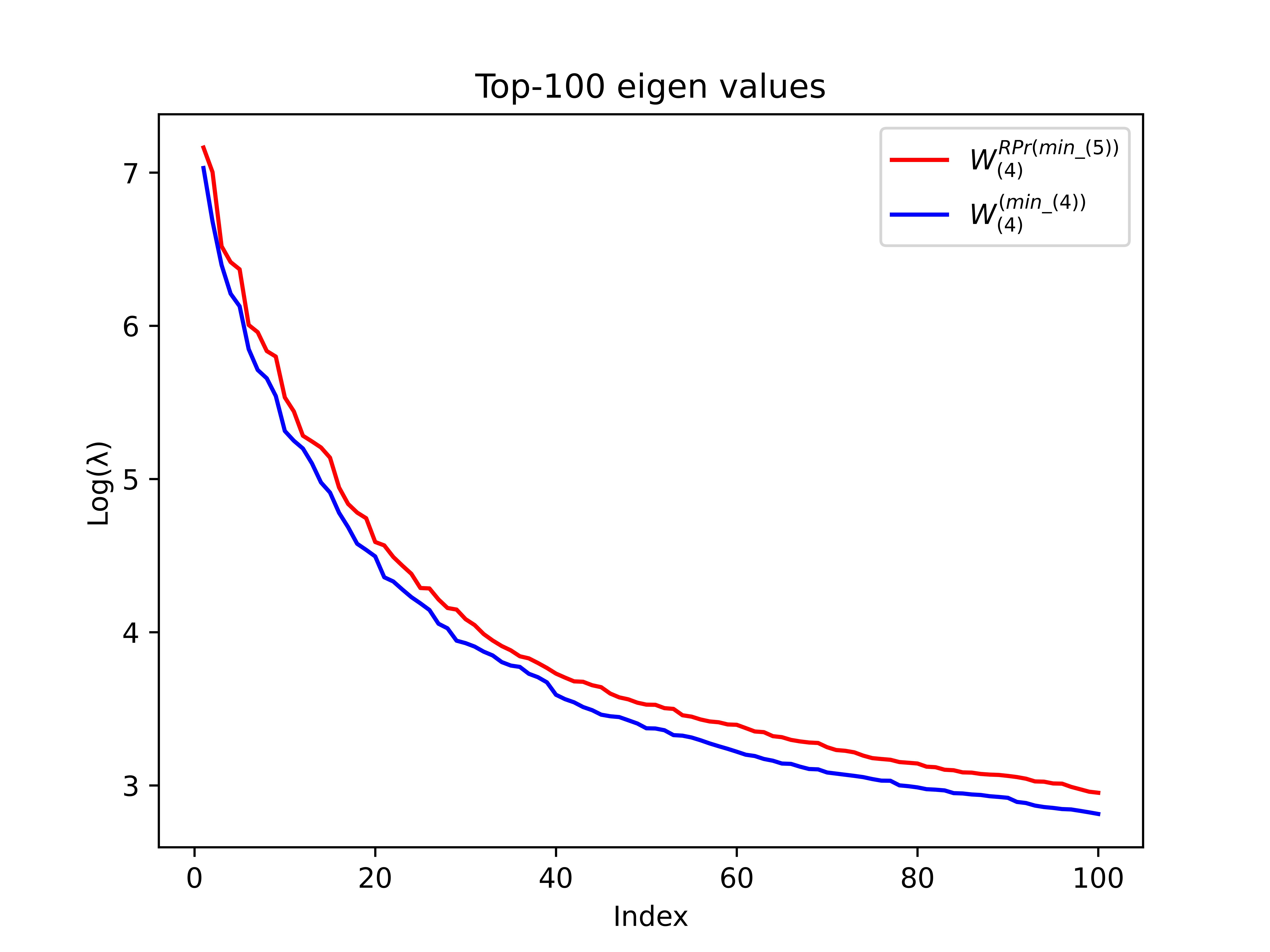} 
\includegraphics[width=4cm, height=3.5cm]{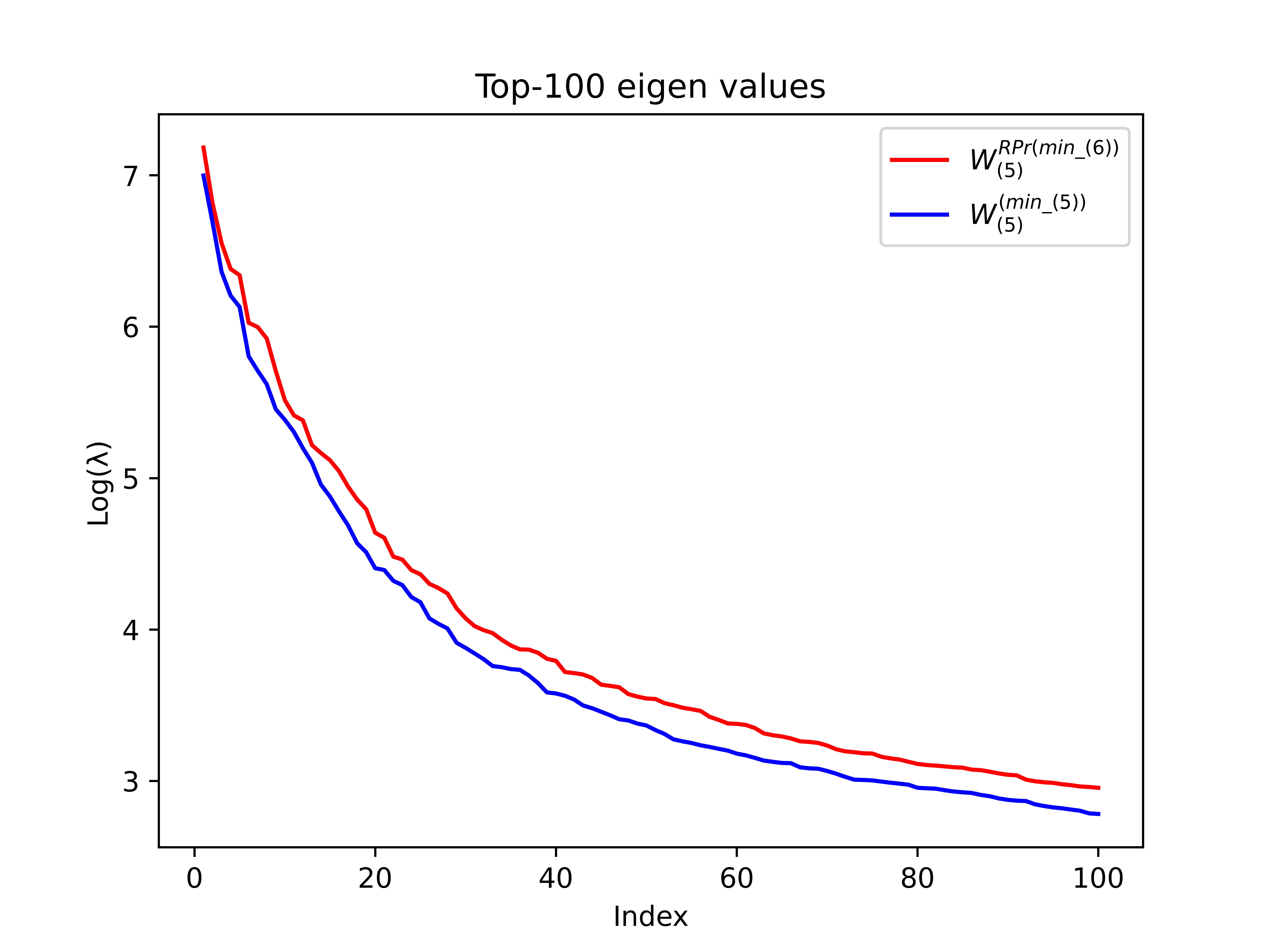} 
\includegraphics[width=4cm, height=3.5cm]{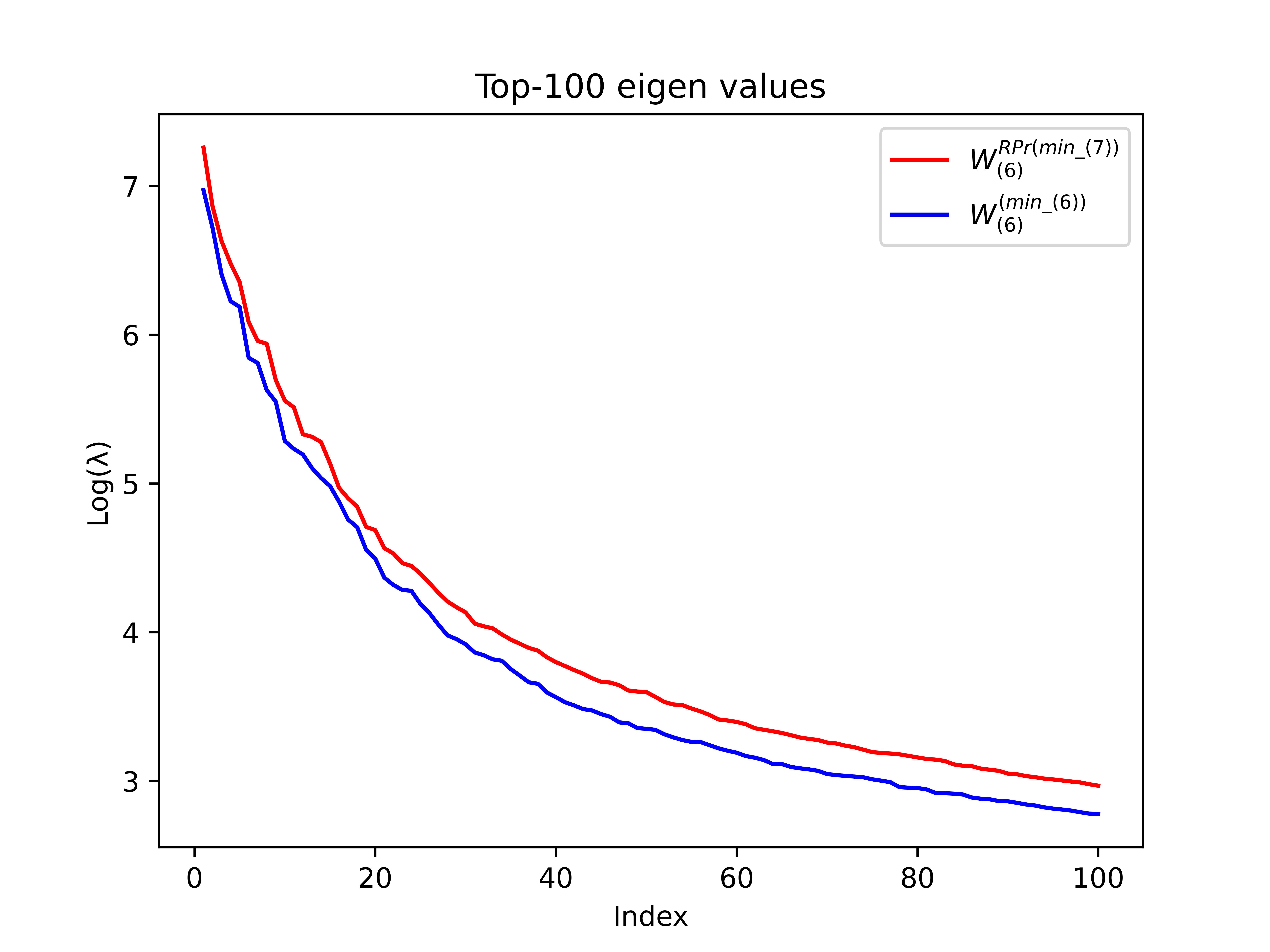}
\includegraphics[width=4cm, height=3.5cm]{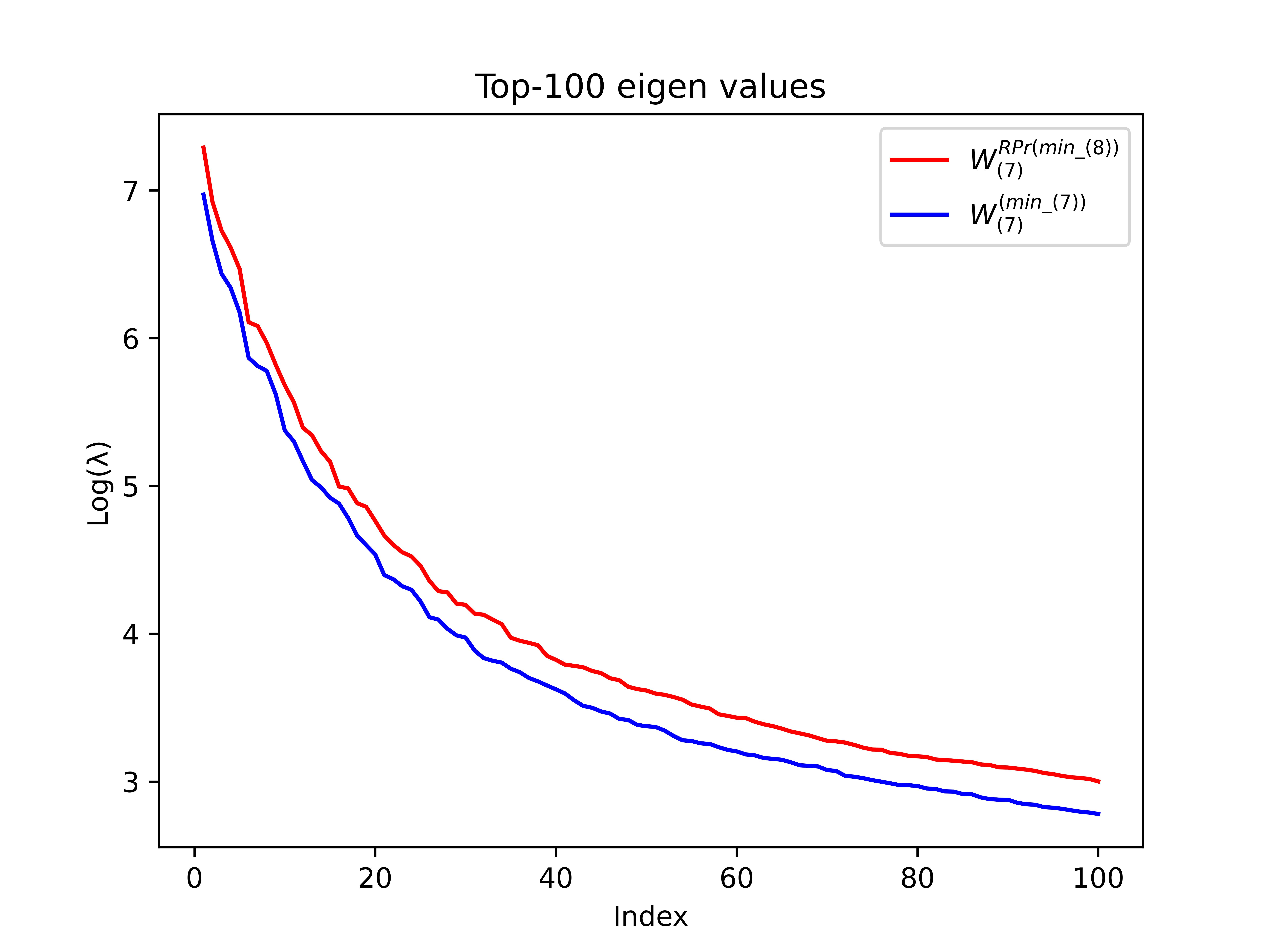} 
\includegraphics[width=4cm, height=3.5cm]{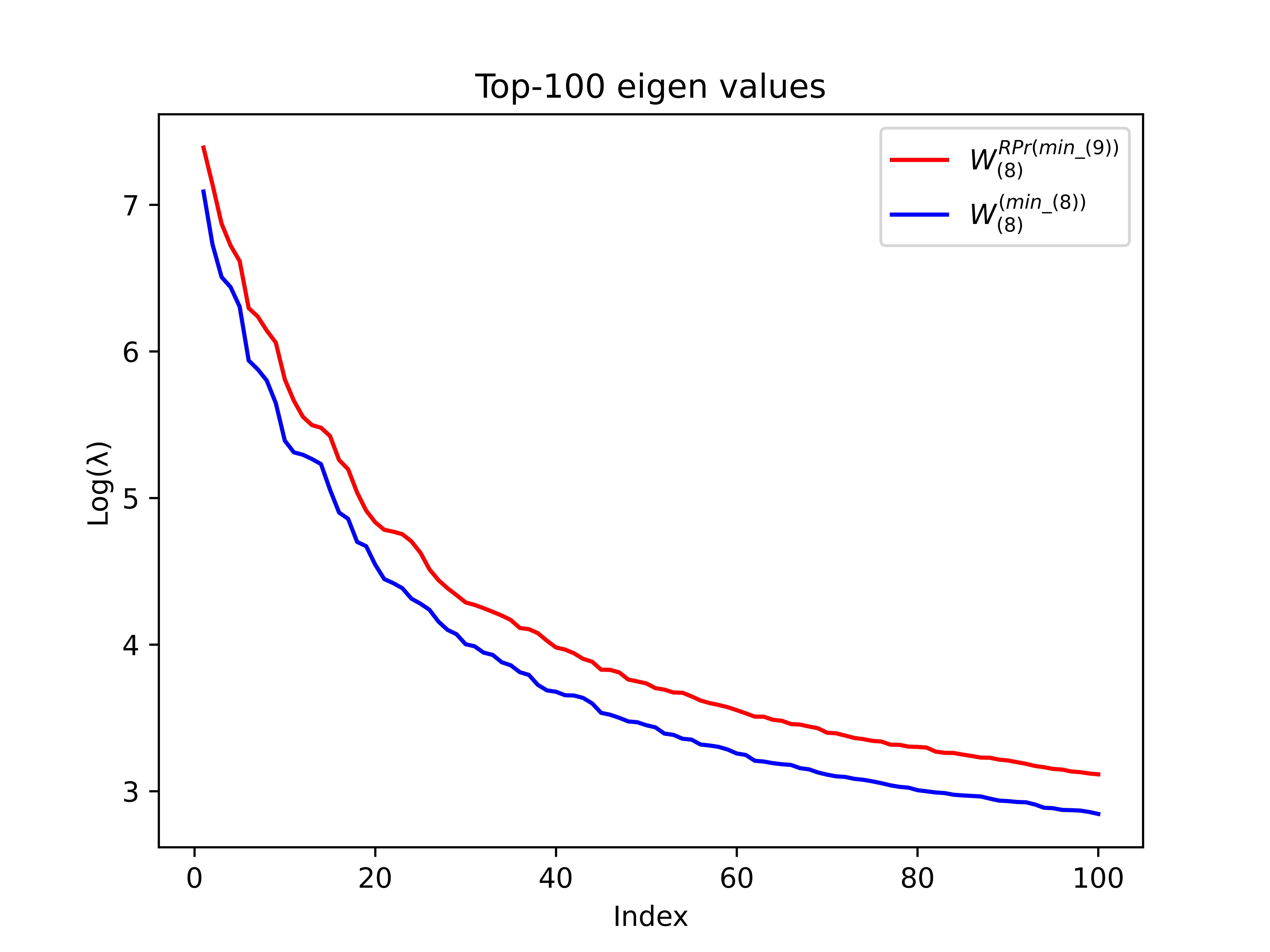} 
\includegraphics[width=4cm, height=3.5cm]{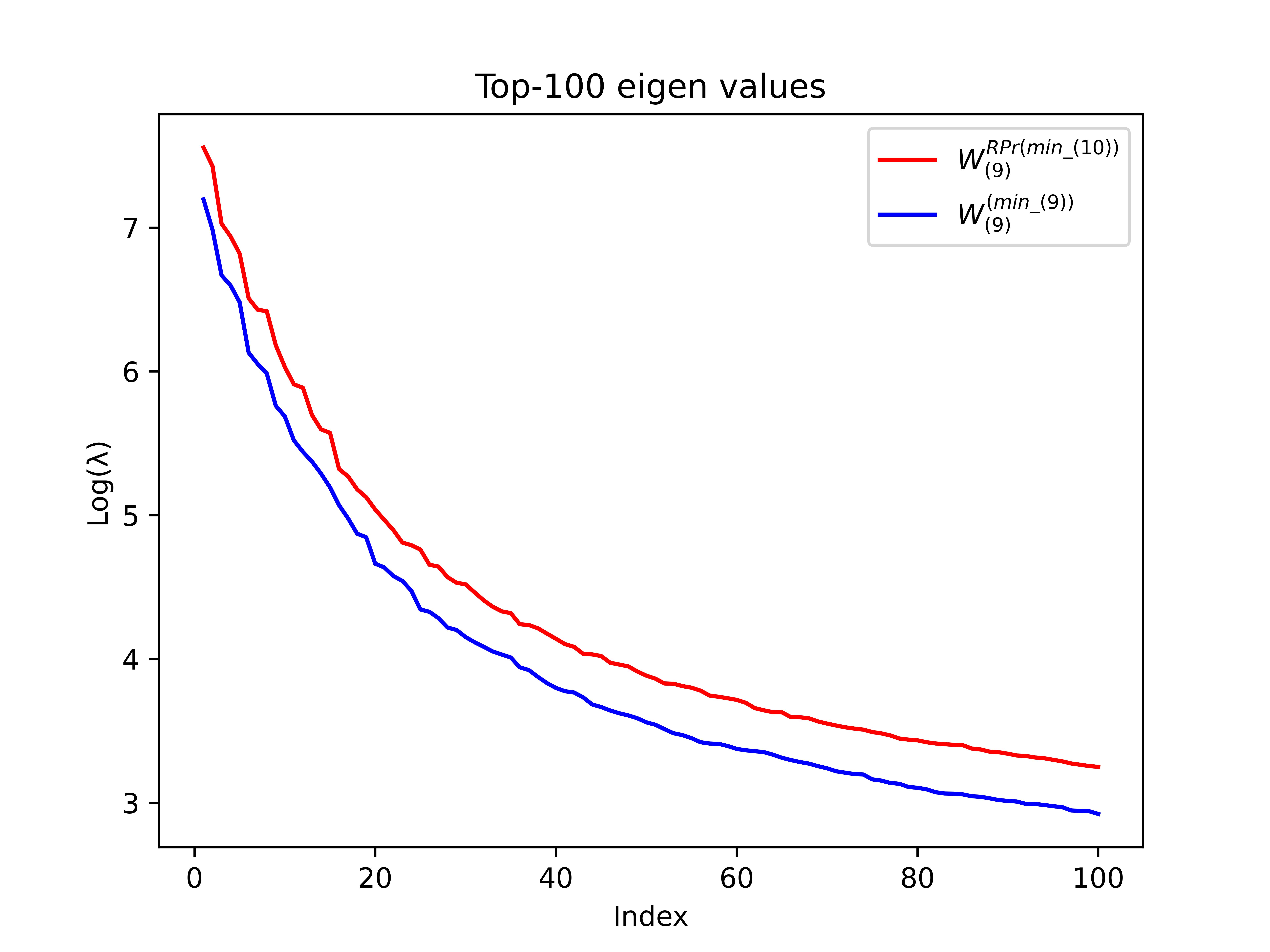} 
\label{eigen}
{\caption{Comparison of top-100 positive eigen values of the Hessian at $W_{(L-1)}^{(min\_(L-1))}$ and $W^{RPr{(min\_(L))}}_{(L-1)}$ for $L$ ranging from $1$ to $10$. The figure shows that the eigen values of the Hessian at $W_{(L-1)}^{(min\_(L-1))}$ are smaller than that at $W^{RPr{(min\_(L))}}_{(L-1)}$.} }
\end{figure*}
And this explains why SGD does not converge to $W^{RPr{(min\_(L))}}_{(L-1)}$ at level $(L-1)$. \par
\begin{table}[h!]
\caption{Comparison of $V^{'} (100)$ at $W^{RPr{(min\_(L))}}_{(L-1)}$ and $W_{(L-1)}^{(min\_(L-1))}$.}
\begin{center}
\begin{tabular}{|p{0.03\textwidth} | p{0.14\textwidth}  | p{0.10\textwidth}|}
  \hline
 $L$ &Solution & $V^{'} (100)$  \\   
  \hline
  \multirow{2}{4em}{1}
 &$W^{RPr{(min\_(1))}}_{(0)}$
 &418.034\\  \cline{2-3}
 &$W_{(0)}^{(min\_(0))}$
 & 415.42\\ 
  \hline
  \multirow{2}{4em}{2}
  &$W^{RPr{(min\_(2))}}_{(1)}$ 
 & 415.826\\  \cline{2-3}
 &$W_{(1)}^{(min\_(1))}$
 & 407.417\\ 
  \hline 
 \multirow{2}{4em}{3}
  &$W^{RPr{(min\_(3))}}_{(2)}$
 & 403.674\\  \cline{2-3}
  &$W_{(2)}^{(min\_(2))}$
 & 399.68\\  
  \hline
\multirow{2}{4em}{4}
   &$W^{RPr{(min\_(4))}}_{(3)}$
 & 394.347\\   \cline{2-3}
  &$W_{(3)}^{(min\_(3))}$  & 386.948\\
  \hline
    \multirow{2}{4em}{5}
 &$W^{RPr{(min\_(5))}}_{(4)}$
 &391.093\\  \cline{2-3}
 &$W_{(4)}^{(min\_(4))}$
 & 375.59\\ 
  \hline
  \multirow{2}{4em}{6}
  &$W^{RPr{(min\_(6))}}_{(5)}$
 & 391.471\\   \cline{2-3}
 &$W_{(5)}^{(min\_(5))}$
 & 372.753\\ 
  \hline 
 \multirow{2}{4em}{7}
&$W^{RPr{(min\_(7))}}_{(6)}$
 & 394.086\\   \cline{2-3}
 &$W_{(6)}^{(min\_(6))}$
 & 373.472\\ 
  \hline
\multirow{2}{4em}{8}
    &$W^{RPr{(min\_(8))}}_{(7)}$
 & 398.559\\  \cline{2-3}
  &$W_{(7)}^{(min\_(7))}$  & 375.401\\ 
  \hline
   \multirow{2}{4em}{9}
    &$W^{RPr{(min\_(9))}}_{(8)}$ 
 & 411.997\\   \cline{2-3}
  &$W_{(8)}^{(min\_(8))}$
 & 382.051\\  
  \hline
\multirow{2}{4em}{10}
  &$W^{RPr{(min\_(10))}}_{(9)}$
 & 427.475\\   \cline{2-3}
  &$W_{(9)}^{(min\_(9))}$  & 394.166\\
  \hline
\end{tabular}
\label{product}
\end{center}
\end{table}

This is confirmed by a comparison of radii of the basin around $W_{(9)}^{(min\_(9))}$ and $W^{RPr{(min\_(10))}}_{(9)}$ using the method given by Huang et al. \cite{huang2020understanding}. Here, the normalized cut-off value for the radii of the basins turns out to be $0.1$ using equation $5$. A comparison of radii of the basin around $W_{(9)}^{(min\_(9))}$ and $W^{RPr{(min\_(10))}}_{(9)}$ is given in Fig. 8, and the average radii of the two basins are given in Table IV. It is clear from the figure and the table that the average radius of the basin for $W_{(9)}^{(min\_(9))}$ is larger than that for $W^{RPr{(min\_(10))}}_{(9)}$. 

\begin{figure}[h!]
\centering  
\includegraphics[width=4cm, height=3.5cm]{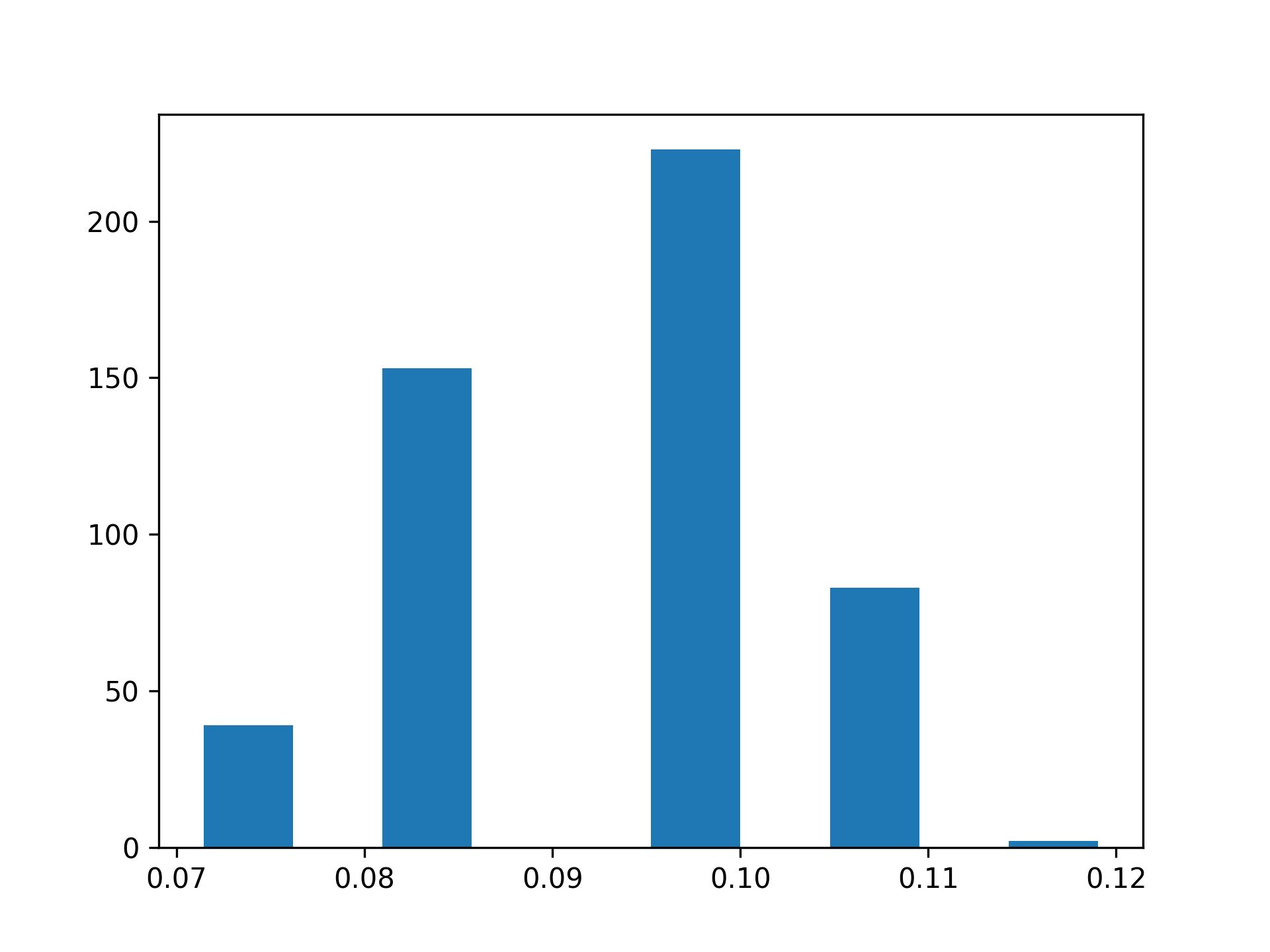}  
\includegraphics[width=4cm, height=3.5cm]{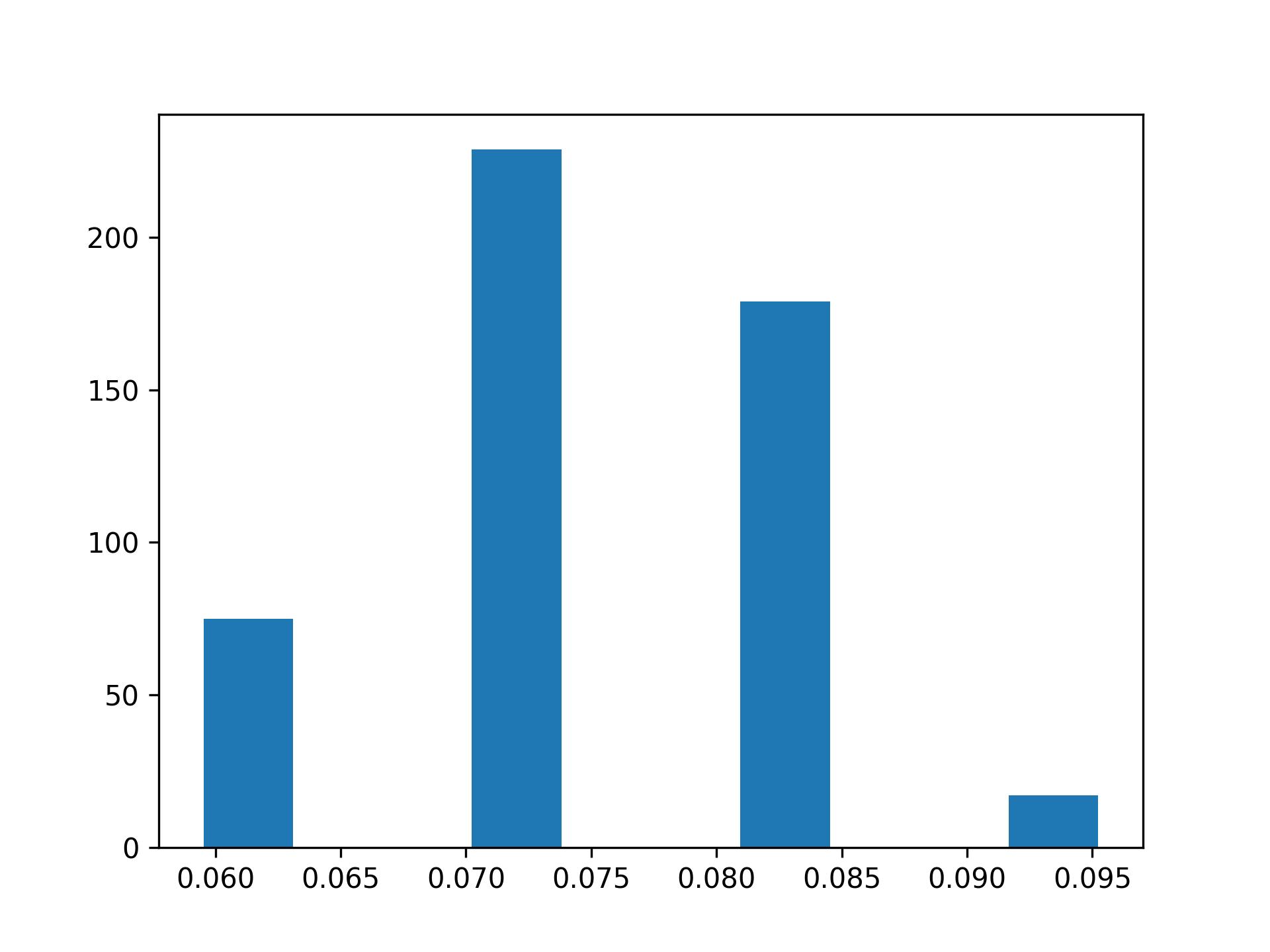}  
\caption{Comparison of  radii of basin along $500$ random directions around $W_{(L-1)}^{(min\_(L-1))}$ and $W^{RPr{(min\_(L))}}_{(L-1)}$ for $L=10$. The x-axis in the above histograms represents the radius of the basin, and the y-axis represents the number of random directions. \textbf{Left:} Radii of basin around $W_{(9)}^{(min\_(9))}$. \textbf{Right:} Radii of basin around $W^{RPr{(min\_(10))}}_{(9)}$.} 
\end{figure}

\begin{table}[h!]
\caption{Comparison of Average Radii of Basin around $W_{(L-1)}^{(min\_(L-1))}$ and $W^{RPr{(min\_(L))}}_{(L-1)}$ for $L=10$.}
\begin{center}
\begin{tabular}{ | p{0.11\textwidth} | p{0.2\textwidth}|} 
  \hline
  Solution& Average Radius of Basin   \\ 
 
  \hline
   $W_{(9)}^{(min\_(9))}$ &0.091809 \\ 
  \hline
  $W^{RPr{(min\_(10))}}_{(9)}$
 &0.074714 
  \\
  \hline
\end{tabular}
\label{product}
\end{center}
\end{table}

To sum up the discussion above, SGD converges to $W_{(L)}^{(min\_(L))}$ at level $(L)$ instead of $W^{Pr{(min\_(L-1))}}_{(L)}$ because the volume of the basin around $W_{(L)}^{(min\_(L))}$ is much larger than the volume of the basin around $W^{Pr{(min\_(L-1))}}_{(L)}$, and the path to $W_{(L)}^{(min\_(L))}$ is steeper than the path to $W^{Pr{(min\_(L-1))}}_{(L)}$. However, in the $(L-1)$ space, the volume of the basin around $W^{RPr{(min\_(L))}}_{(L-1)}$ is much smaller than the volume of basin around $W_{(L-1)}^{(min\_(L-1))}$. This means that the volume comparison gets flipped in the two spaces, $(L-1)$ space and $(L)$ space. To make this more clear, let us consider two spaces, for example, level $0$ space and level $1$ space. In level $0$ space, the volume of the basin surrounding level $0$ solution is larger than the volume of the basin surrounding level $1$ solution; however, in level $1$ space, the volume of the basin surrounding level $0$ solution is smaller than the volume of the basin surrounding level $1$ solution. This is shown in Table V. 

\begin{table}[h!]
\caption{Comparison of $V^{'} (100)$ between $W_{(0)}^{(min\_(0))}$ and $W_{(1)}^{(min\_(1))}$ in level $0$ and level $1$ space.}
\begin{center}
\begin{tabular}{|p{0.05\textwidth} |  p{0.25\textwidth} | p{0.07\textwidth}|}
  \hline
Space &Solution  & $V^{'} (100)$  \\   
  \hline
  \multirow{2}{4em}{Level $0$}
   &$W_{(0)}^{(min\_(0))}$ (Level $0$ solution) 
 & 415.42\\ \cline{2-3}
 &$W^{RPr{(min\_(1))}}_{(0)}$ (Level $1$ solution in level $0$ space)
 &418.034\\ 
  \hline
   \multirow{2}{4em}{Level $1$} 
 &$W_{(1)}^{(min\_(1))}$ (Level $1$ solution)
 & 407.417\\ \cline{2-3}
&$W^{Pr{(min\_(0))}}_{(1)}$  (Level $0$ solution in level $1$ space)
 &409.242\\
 \hline
\end{tabular}
\end{center}
\end{table}

\subsection{Result 2: Why does the initialization proposed by the lottery ticket hypothesis work well?}

 The above results also provide a strong clue to the positive role of the initialization proposed by the lottery ticket hypothesis in finding good minima in the sparser weight space. Pruning out the smaller weights leads to a new minimum or a saddle point with a slightly larger training loss value and a smaller volume. Retraining from this point via learning rate rewinding or from a point in close proximity (via weight rewinding) leads SGD to converge to a new minimum, which has a larger volume (compared to the earlier minimum) and a lower training loss. This minimum was not discoverable earlier since it spanned some dimensions where the loss function was steeply increasing and therefore, had an overall smaller volume, and pruning out these dimensions exposed this minimum to the SGD. Hence, the starting minimum acts as a baseline, avoidance of which leads to a better minimum. Choosing any other initialization point places the SGD at a point outside the loss sublevel set and there is no guarantee of finding a better minimum. Fig. 9 presents the training loss along a straight line connecting $W^{(min\_(9))}_{(9)}$ (baseline for $W^{(RIPN)}_{(10)}$) and $W^{(RIPN)}_{(10)}$. 
\begin{figure}[h!]
\centering
\includegraphics[width=4cm, height=3.5cm]{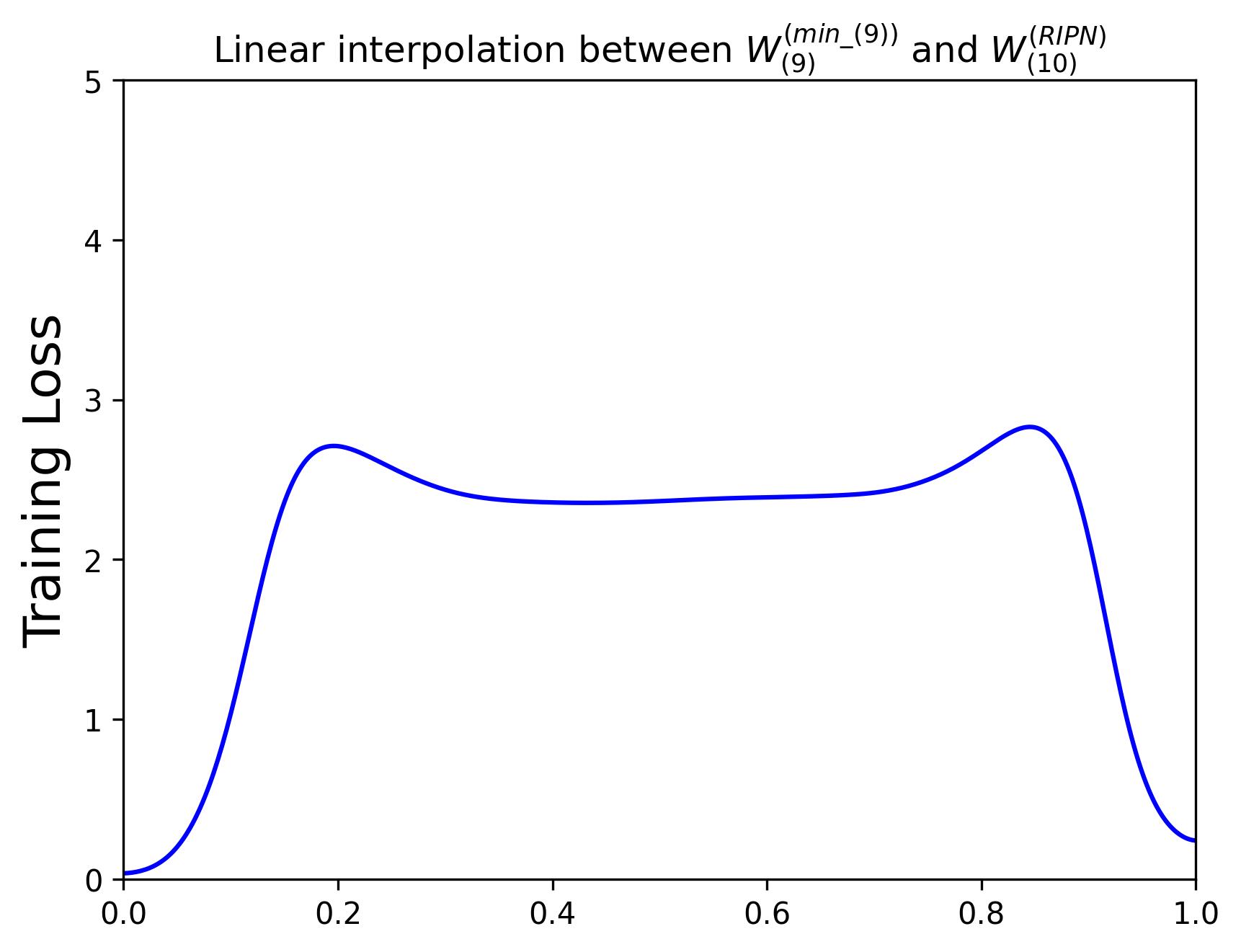}  
\caption{Training loss along a straight line between $W^{(min\_(9))}_{(9)}$ and $W^{(RIPN)}_{(10)}$. } 
\end{figure}
The figure shows a significant barrier between the two points, demonstrating clearly that these two points lie in different loss sublevel sets. \par Further, a comparison of top-100 positive eigen values of the Hessian at $W^{(RIPN)}_{(10)}$ and $W^{(min\_(10))}_{(10)}$ (Fig. 10) shows that $W^{(RIPN)}_{(10)}$ has larger eigen values than $W^{(min\_(10))}_{(10)}$, which indicates a smaller volume for the basin around $W^{(RIPN)}_{(10)}$.
\begin{figure}[h!]
\centering
\includegraphics[width=4cm, height=3.5cm]{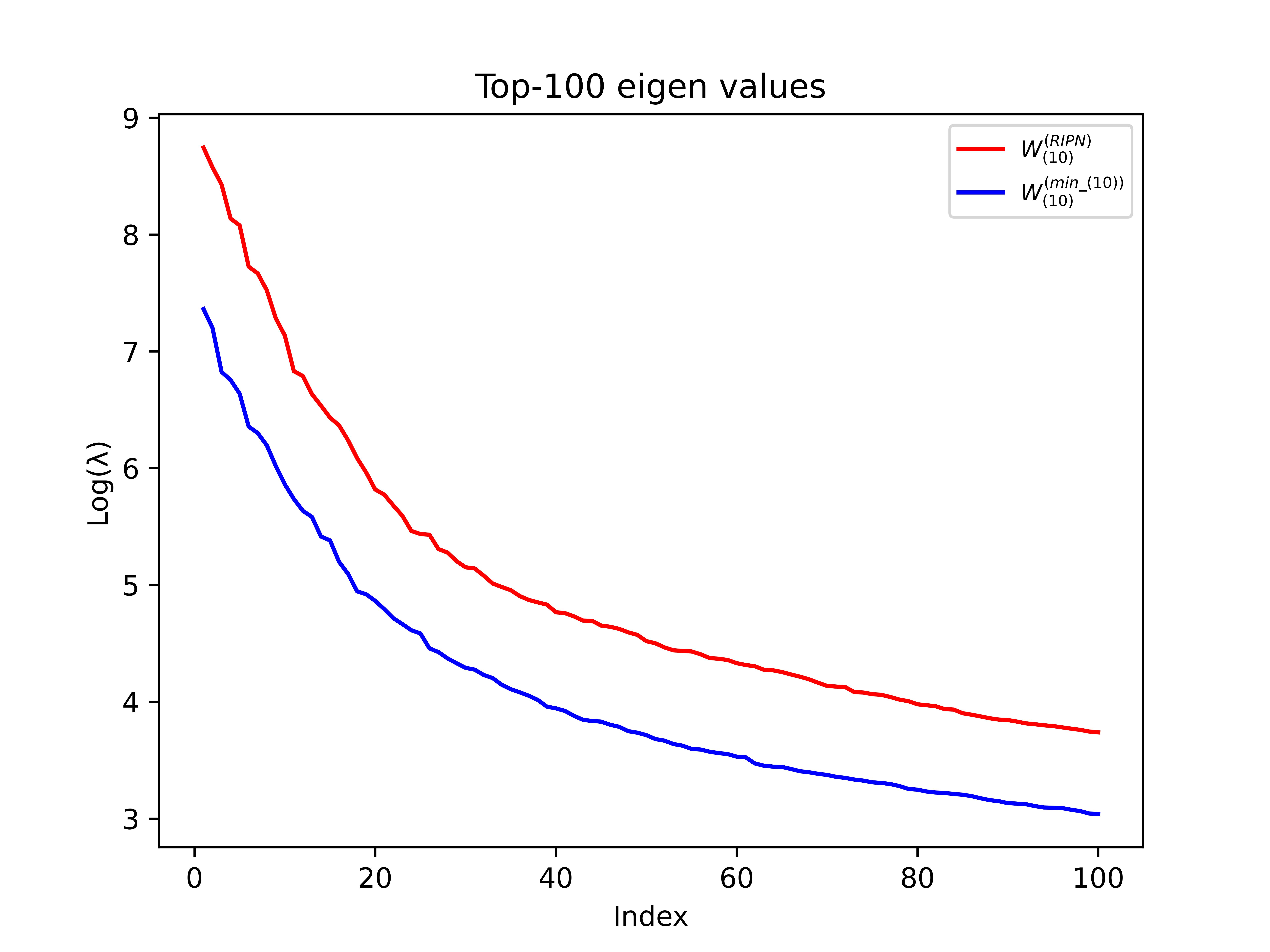}  
\caption{Comparison of top-100 positive eigen values of the Hessian at $W^{(RIPN)}_{(10)}$ and $W^{(min\_(10))}_{(10)}$. } 
\end{figure} \par 
Hence, random initialization of a pruned network takes SGD out of the sublevel set, and after retraining, converges to a minimum with a smaller volume. The importance of initialization proposed by the lottery ticket hypothesis, therefore, becomes quite apparent.
\subsection{Result 3: Why do we need the Iterative Process, and why does one-shot pruning not work as well?}
At each step in the iterative process via the magnitude pruning, we remove dimensions with smaller weights. This increases the loss of the minimum arrived at in the previous step and decreases its volume measure. We then rewind the weights and start looking for an alternate solution in the vicinity of this minimum. As explained and shown earlier, the new (potentially better) minimum that is now exposed to SGD has a larger volume and lower training loss than the earlier pruned minimum. Thus, the pruned minimum acts as a baseline for the minimum to be found in the next step. If too many weights are pruned out in a step, it leads to a bigger increase in the training loss and a bigger decrease in volume, thus lowering the baseline for the next solution. If too small a number of weights are pruned out in a step, we risk returning to the same pruned minimum, thus not offering any improvement. One-shot pruning removes too many weights and therefore, has inferior performance.\par 
A comparison of top-100 positive eigen values of the Hessian between $W^{(one\_shot)}_{(10)}$ and $W^{(min\_(10))}_{(10)}$ (Fig. 11) shows that $W^{(one\_shot)}_{(10)}$ has larger eigen values than $W^{(min\_(10))}_{(10)}$, which implies a smaller volume for the basin around $W^{(one\_shot)}_{(10)}$. This brings out and confirms the importance of the iterative process in pruning.
\begin{figure}[h!]
\centering
\includegraphics[width=4cm, height=3.5cm]{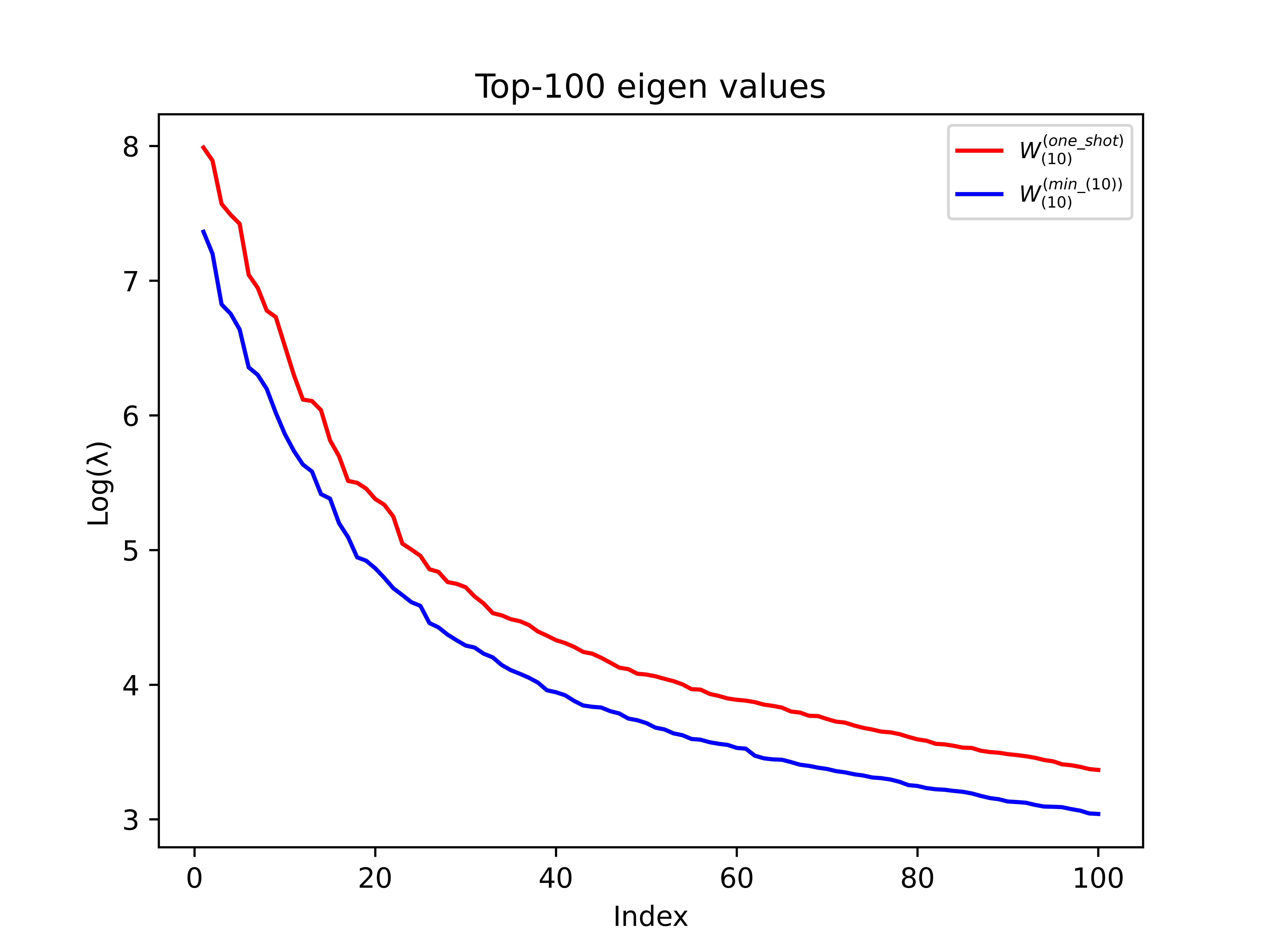}  
\caption{Comparison of top-100 positive eigen values of the Hessian at $W^{(one\_shot)}_{(10)}$ and $W^{(min\_(10))}_{(10)}$. } 
\end{figure}

\subsection{Result 4: There exists a barrier between IMP solutions at successive levels in the loss landscape.}
One interpretation of result 1 would clearly be that IMP solutions at different levels are distinct minima. In order to test this hypothesis more definitively, we examine whether these are separated by distinct barriers in the loss landscape. If this was not the case, SGD should converge to a sparser solution in the first place and would not stop at a less sparse solution. Evidence for our hypothesis is presented in Fig. 12. It gives the training loss along a straight line connecting  $W^{(min\_(L-1))}_{(L-1)}$ and $W^{(min\_(L))}_{(L)}$. The barriers between successive minima are clearly visible. Thus, at least in the loss landscape, the IMP solutions at successive levels are not linearly connected, unlike the corresponding claim by \cite{paul2022unmasking} in the test error landscape. The implication is that these loss barriers are large enough to prevent SGD from crossing over and small enough to have almost similar generalization performance all over the region between these solutions.
\begin{figure*}
\centering
\includegraphics[width=4cm, height=3.5cm]{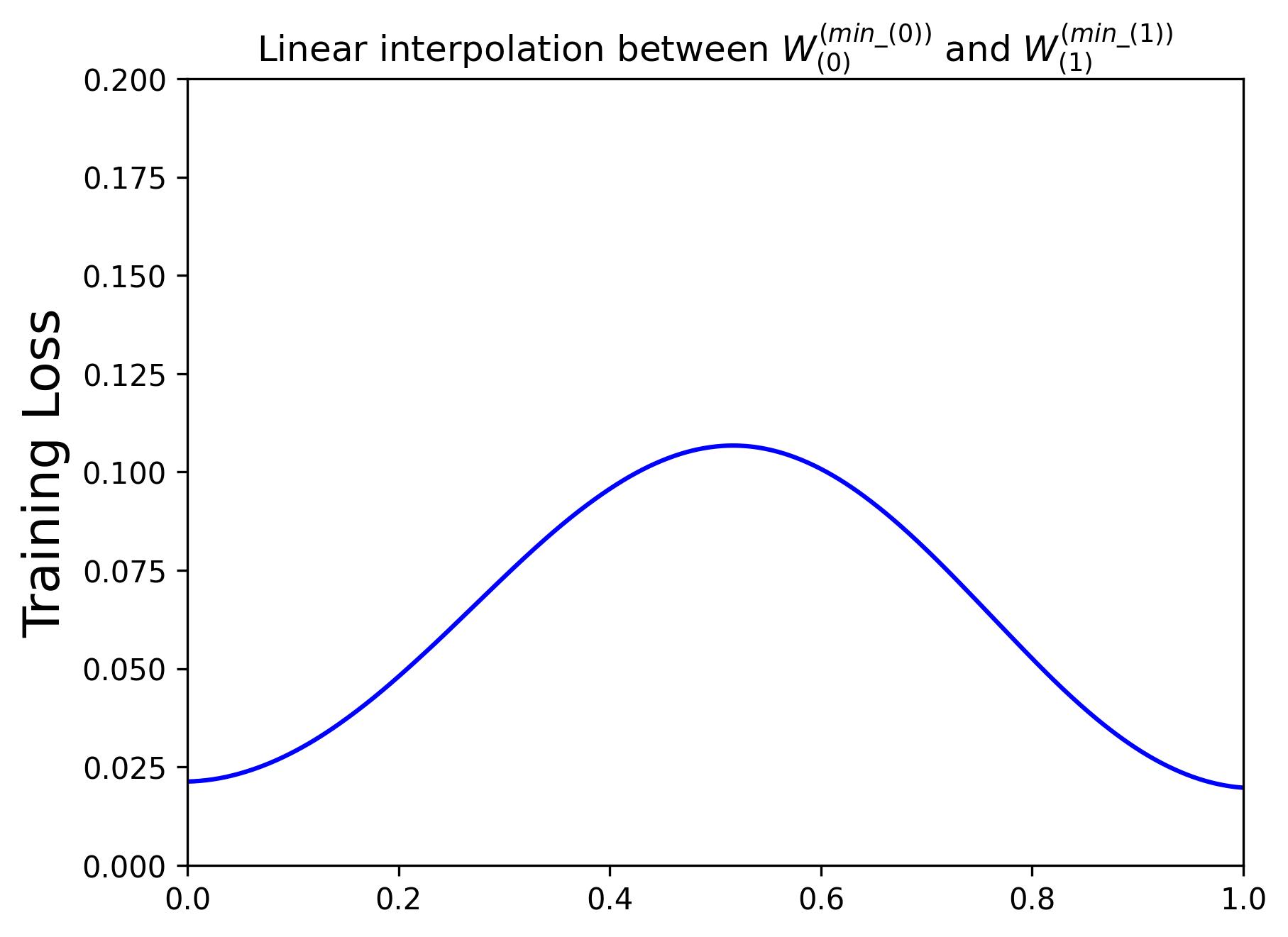}
\includegraphics[width=4cm, height=3.5cm]{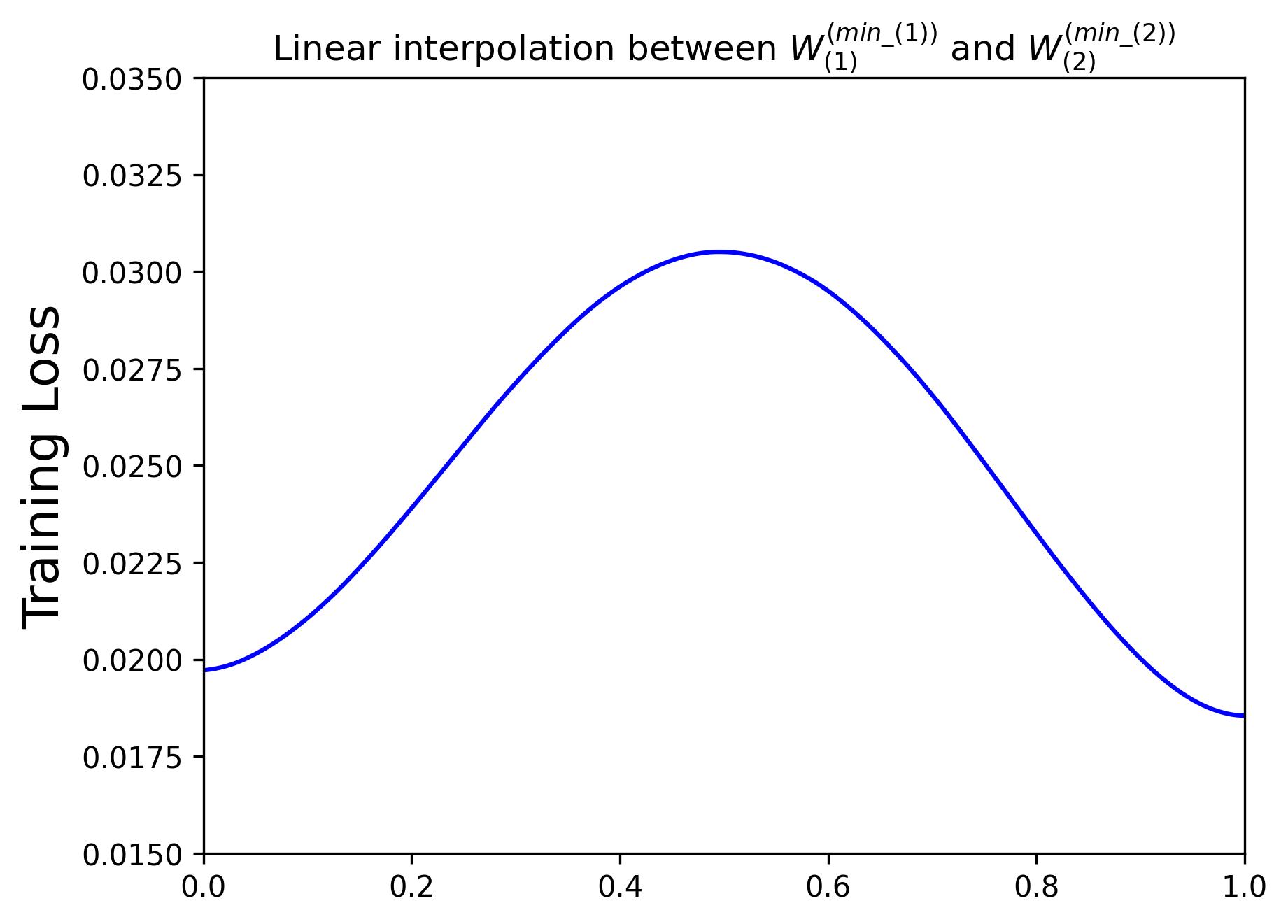}
\includegraphics[width=4cm, height=3.5cm]{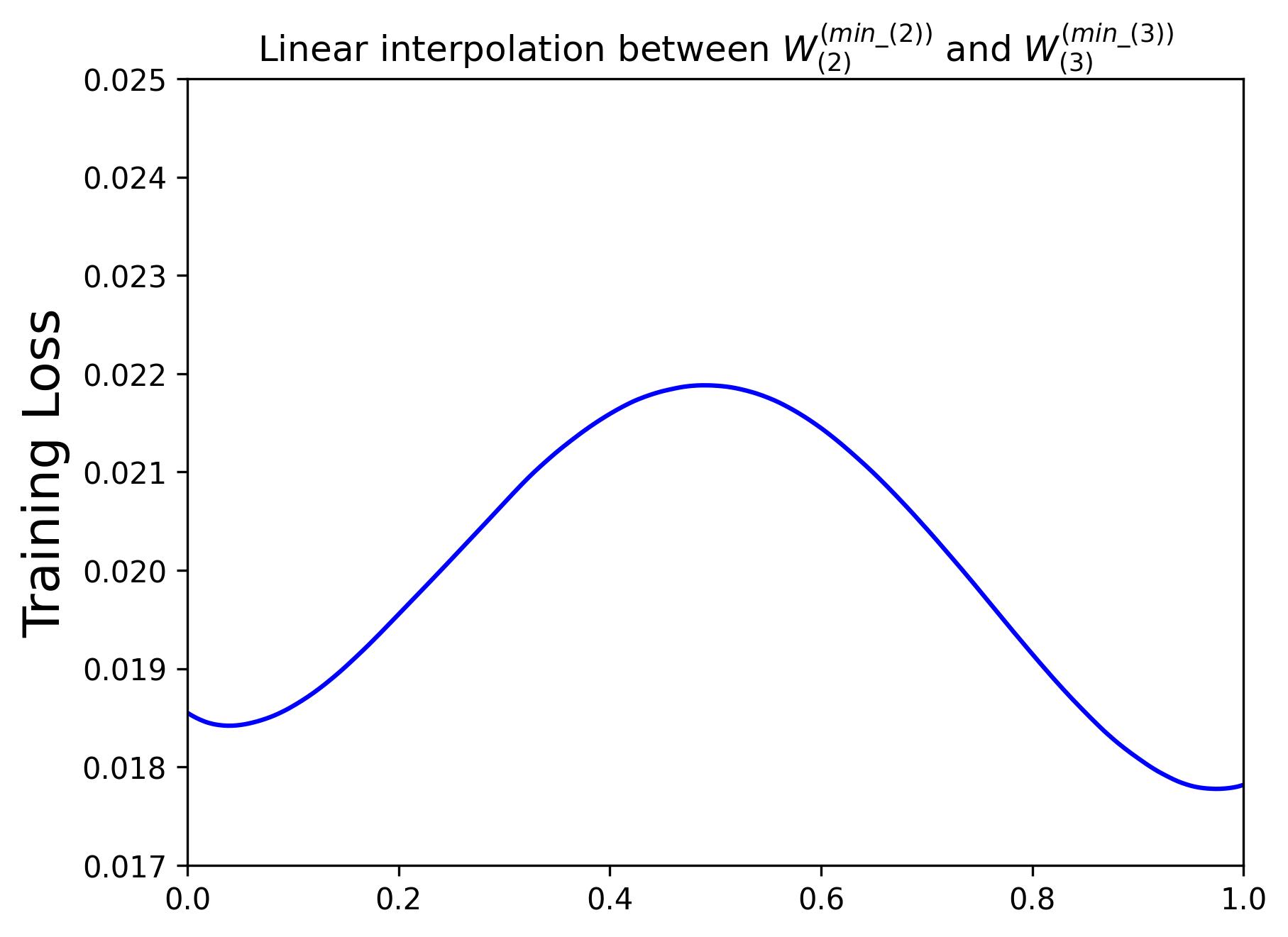}
\includegraphics[width=4cm, height=3.5cm]{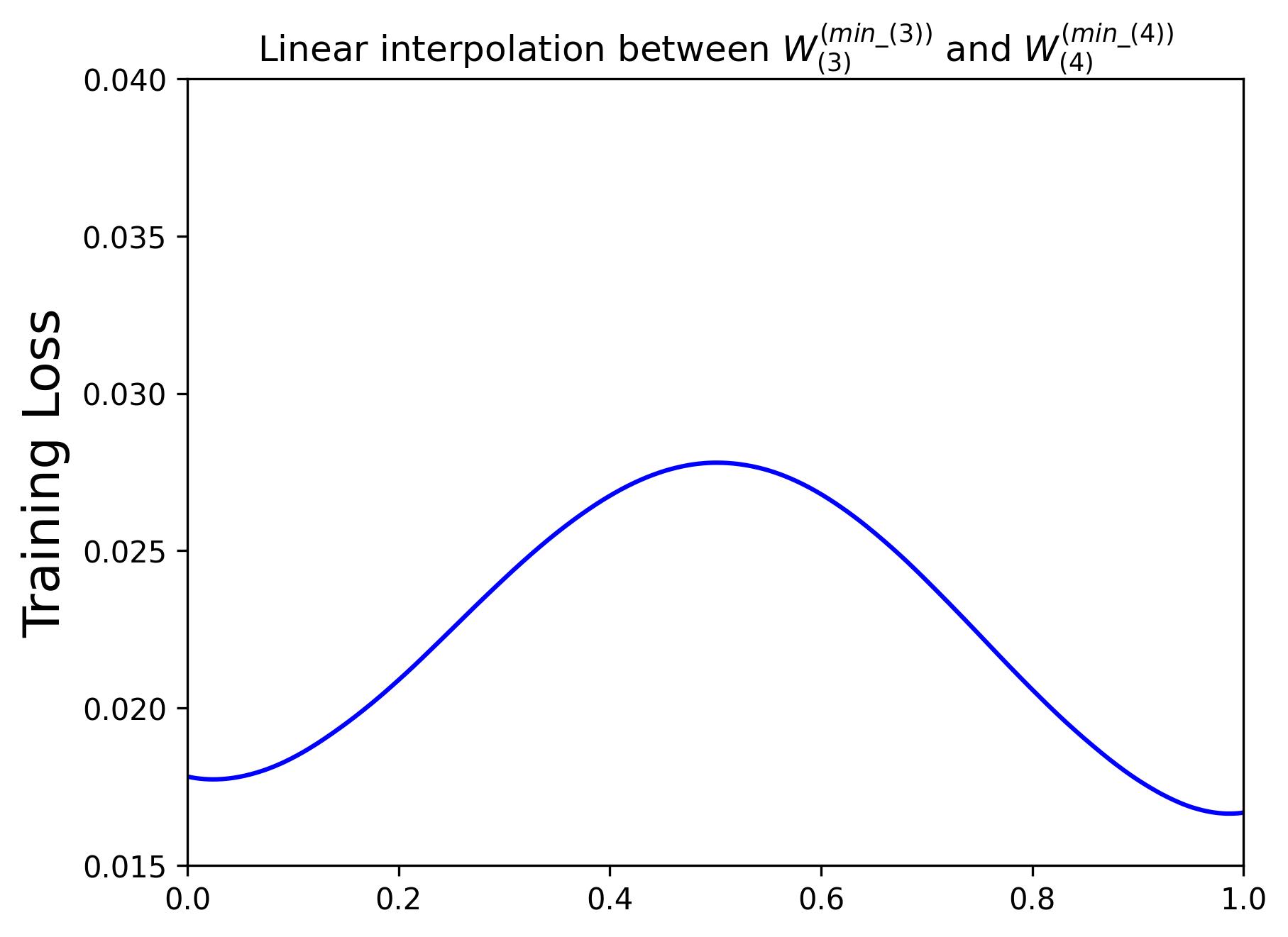} 
\includegraphics[width=4cm, height=3.5cm]{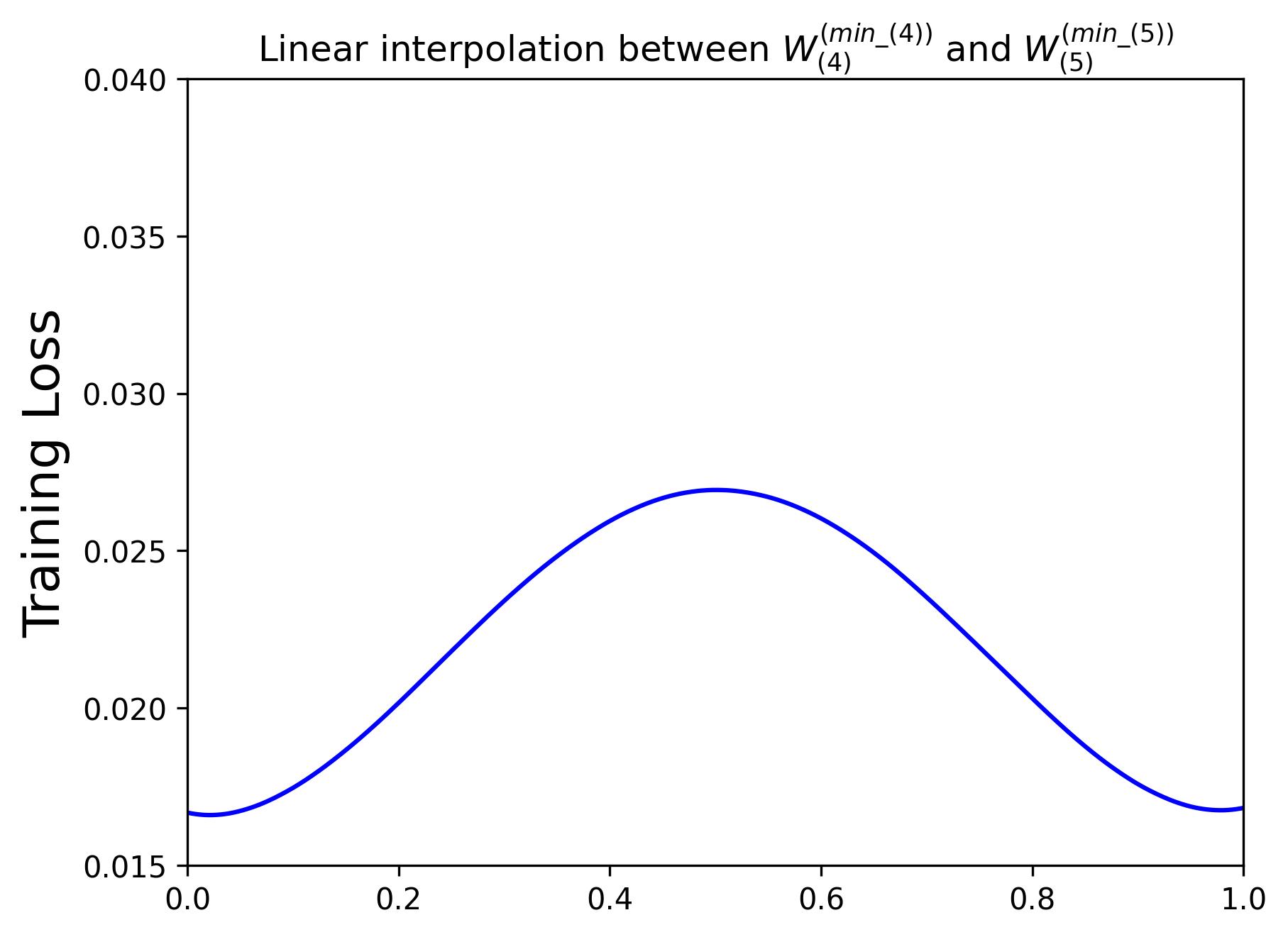}
\includegraphics[width=4cm, height=3.5cm]{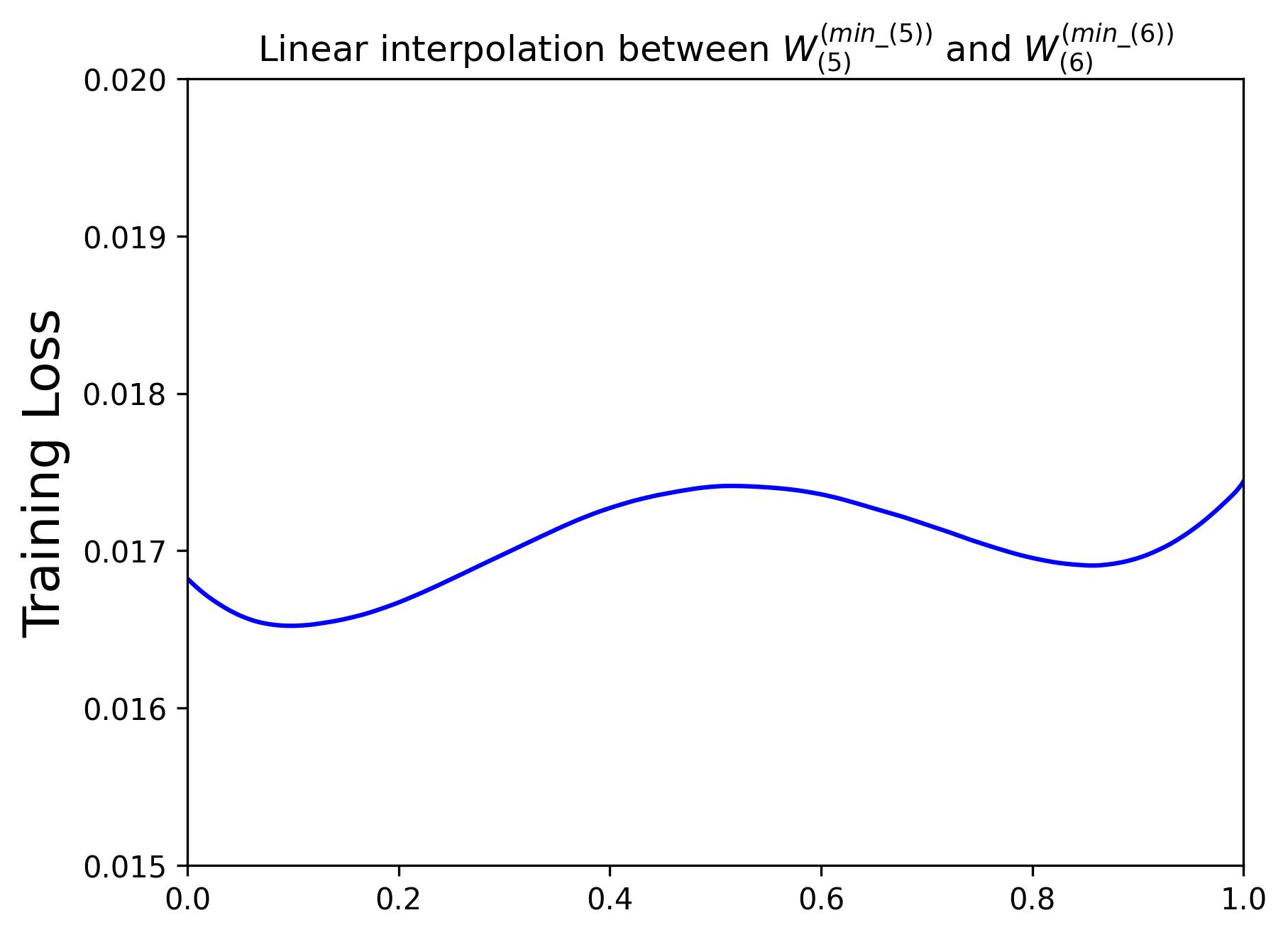}
\includegraphics[width=4cm, height=3.5cm]{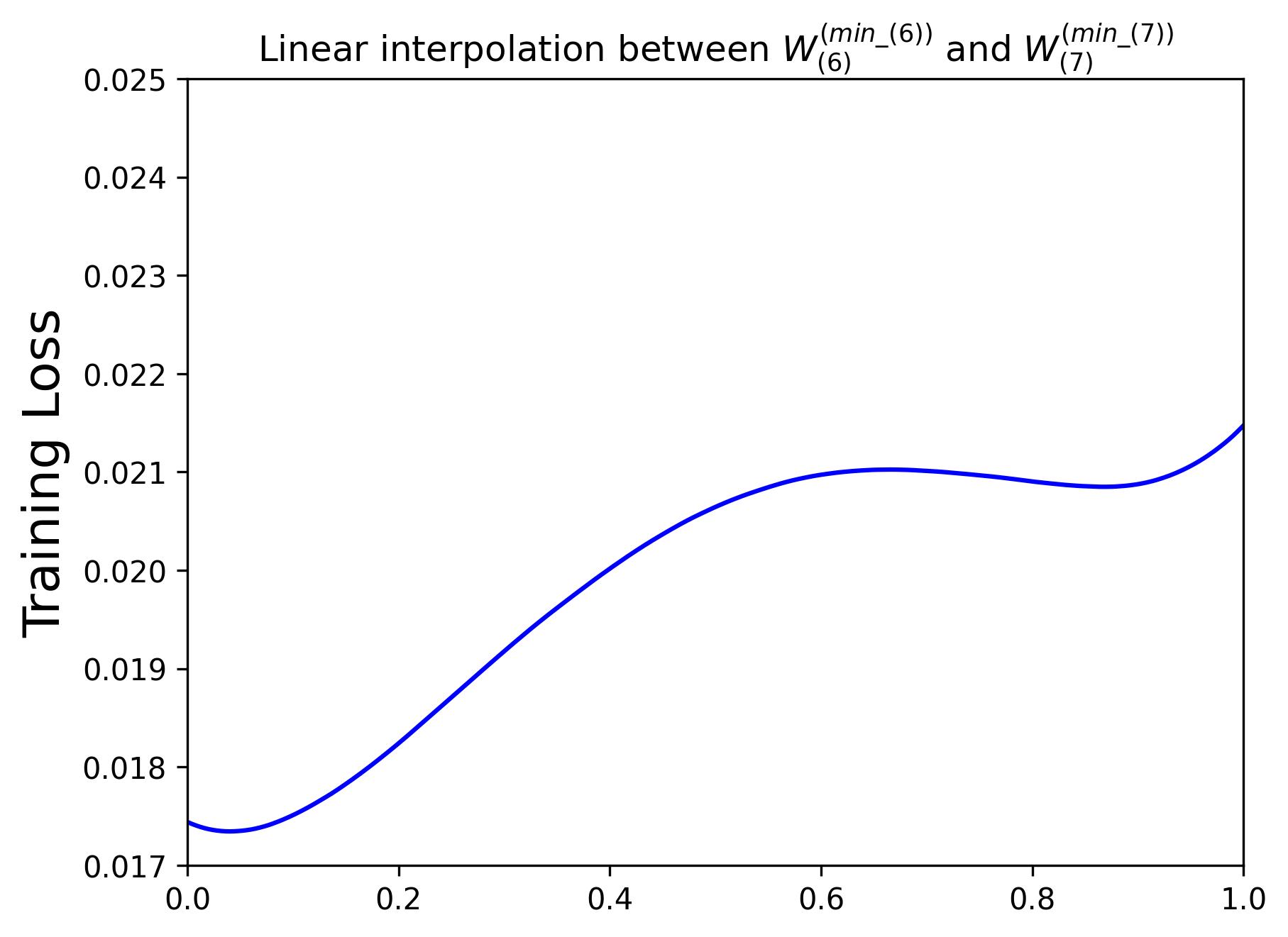} 
\includegraphics[width=4cm, height=3.5cm]{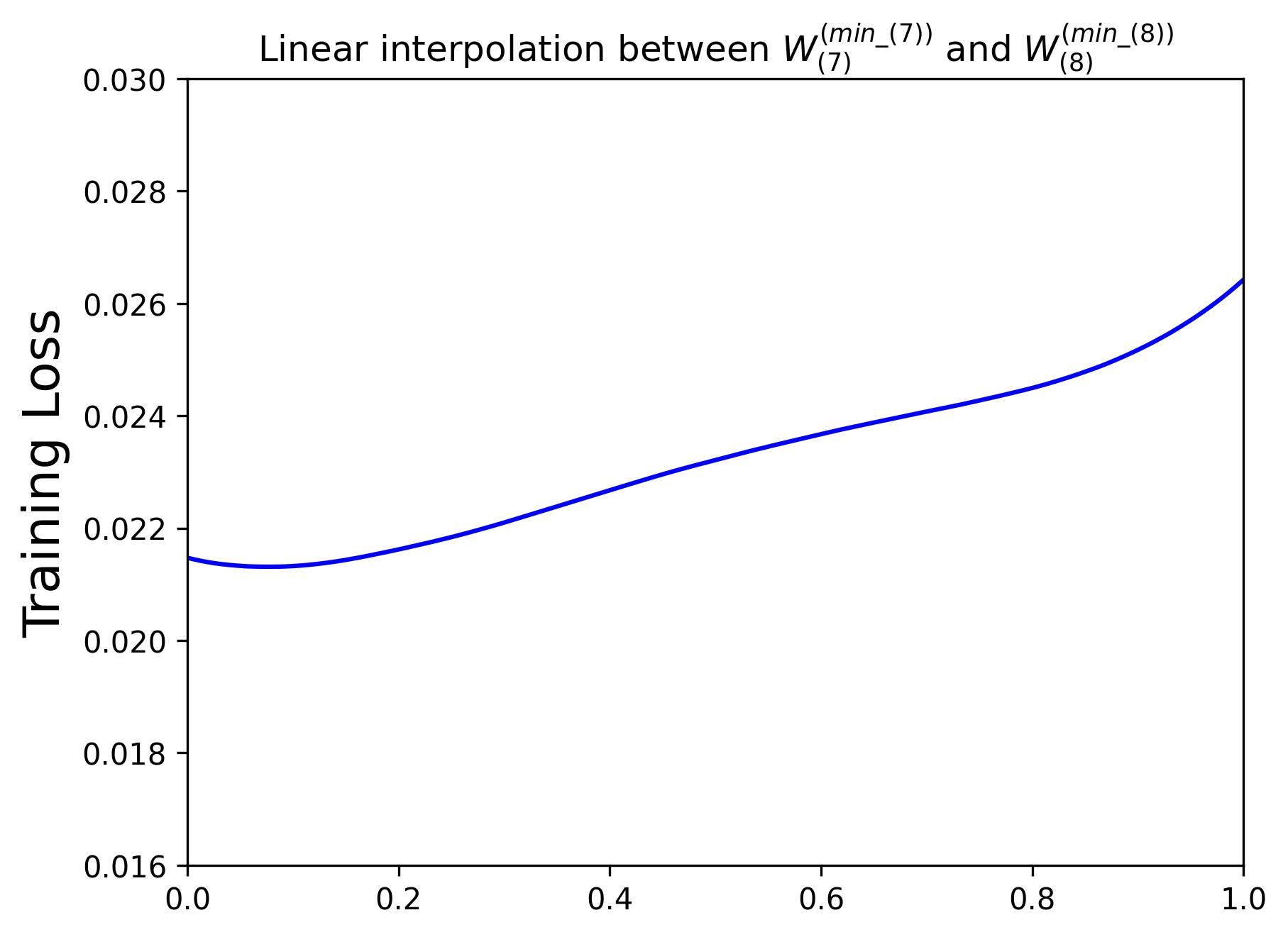}
\includegraphics[width=4cm, height=3.5cm]{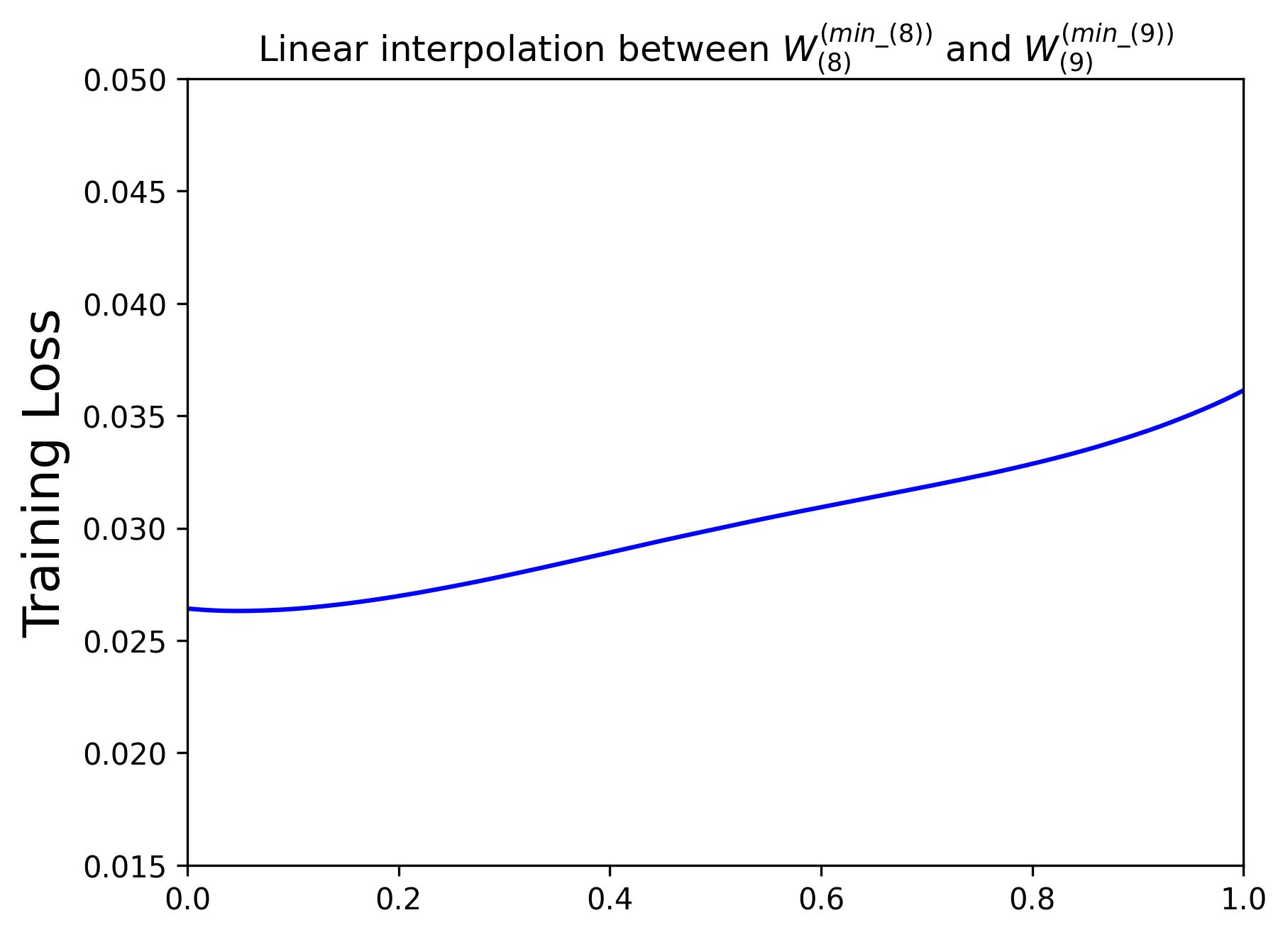}
\includegraphics[width=4cm, height=3.5cm]{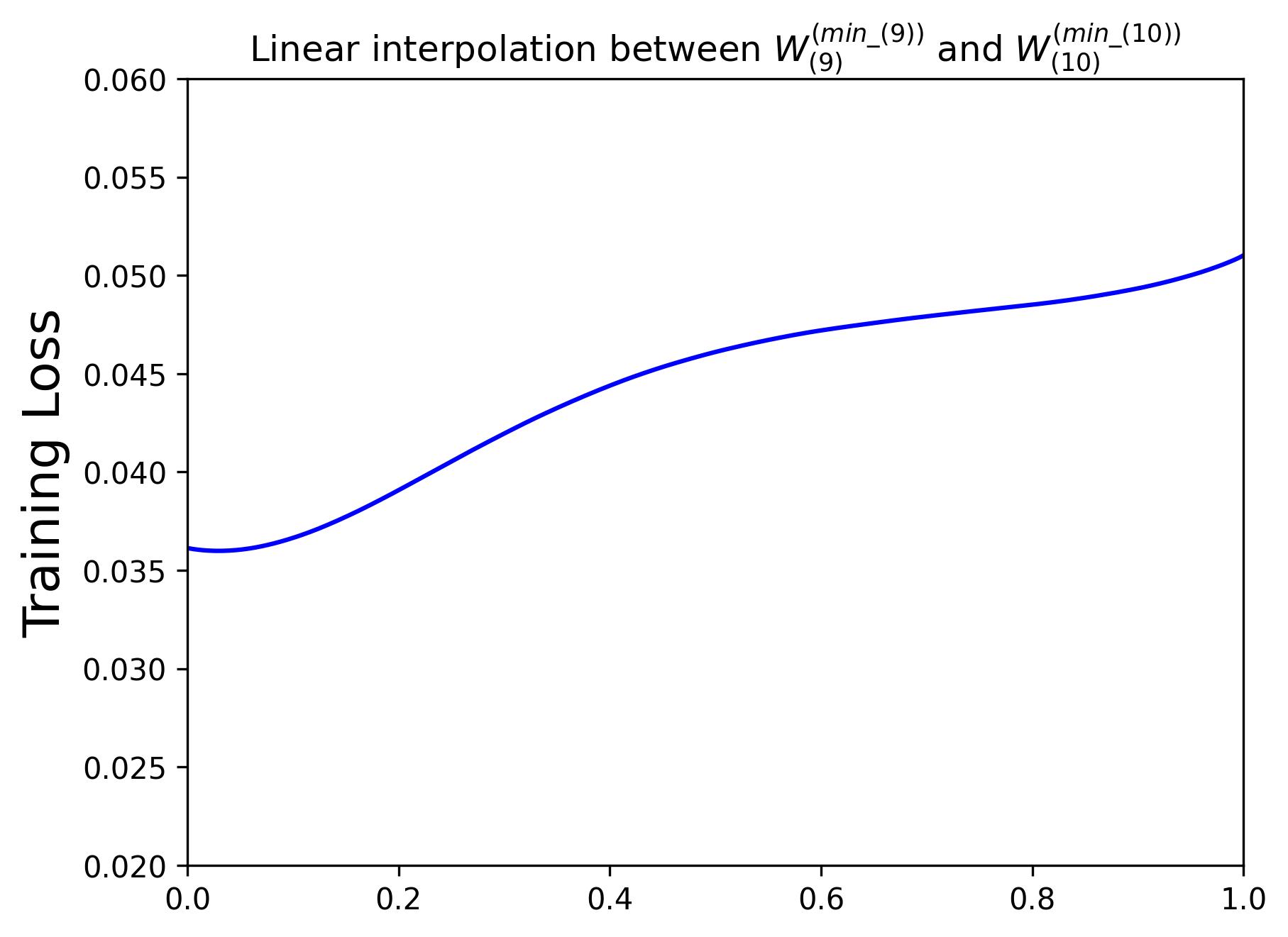} 
\caption{Training Loss along a straight line connecting $W^{(min\_(L-1))}_{(L-1)}$ and $W^{(min\_(L))}_{(L)}$ for $L$ ranging from $1$ to $10$. The x-axis represents the interpolation co-efficient $\alpha$. Each plot depicts the training loss at $501$ points between $W^{(min\_(L-1))}_{(L-1)}$ and $W^{(min\_(L))}_{(L)}$. } 
\end{figure*}
\subsection{Result 5:  IMP solutions obtained using rewinding lie within the same loss sublevel set.}
Any random initialization of a dense neural network makes SGD converge to a good minimum. This can be ascribed to the well-understood fact that the density of good minima in such scenarios is very high \cite{huang2020understanding}. However, in sparse subspaces, the density of such minima is much smaller. If we consider starting from a random point in such subspaces, the solution would not be as good as we get with rewinding, as is observed in our experiments. \par
We also hypothesize that all the IMP solutions obtained using rewinding lie within the same loss sublevel set as that of the original dense network, whereas random initialization of the pruned network takes SGD out of that sublevel set. Fig. 13 presents the evidence for our claim. The plot has been obtained by calculating training loss at $4200$ points in the high-dimensional space and then projecting the points on the two-dimensional plane spanned by $3$ points, $W_{(0)}^{(min\_(0))}$, $W_{(10)}^{(min\_(10))}$ and $W^{(RIPN)}_{(10)}$. However, the two axes\footnote{ ($W_{(10)}^{(min\_(10))} - W_{(0)}^{(min\_(0))}$) forms the x-axis and ($W^{(RIPN)}_{(10)}- W_{(0)}^{(min\_(0))}$) forms the y-axis.} formed by these points are not orthogonal. Hence, orthogonalization of these axes is carried out before projecting the points from the high-dimensional space. The figure clearly shows that all the IMP solutions ($W^{(min\_(0))}_{(0)}$ to $W^{(min\_(10))}_{(10)}$) lie within the same connected loss sublevel set (dark green region). However, $W^{(RIPN)}_{(10)}$ lies outside the sublevel set. Similarly, random pruning of a large number of weights also takes SGD outside the sublevel set. It is interesting to see that $W^{(one\_shot)}_{(10)}$ also lies in the same sublevel set as the IMP solutions; however, $W^{(RPN\_2)}_{(10)}$ lies outside the sublevel set. In other words, a randomly pruned network obtained by pruning a large number of parameters does not yield an equally good solution as that of the network obtained with magnitude based pruning, as it is not seen to lie in the same sublevel set.
\begin{figure}[h!]
\centering
\includegraphics[width=7cm, height=5.5cm]{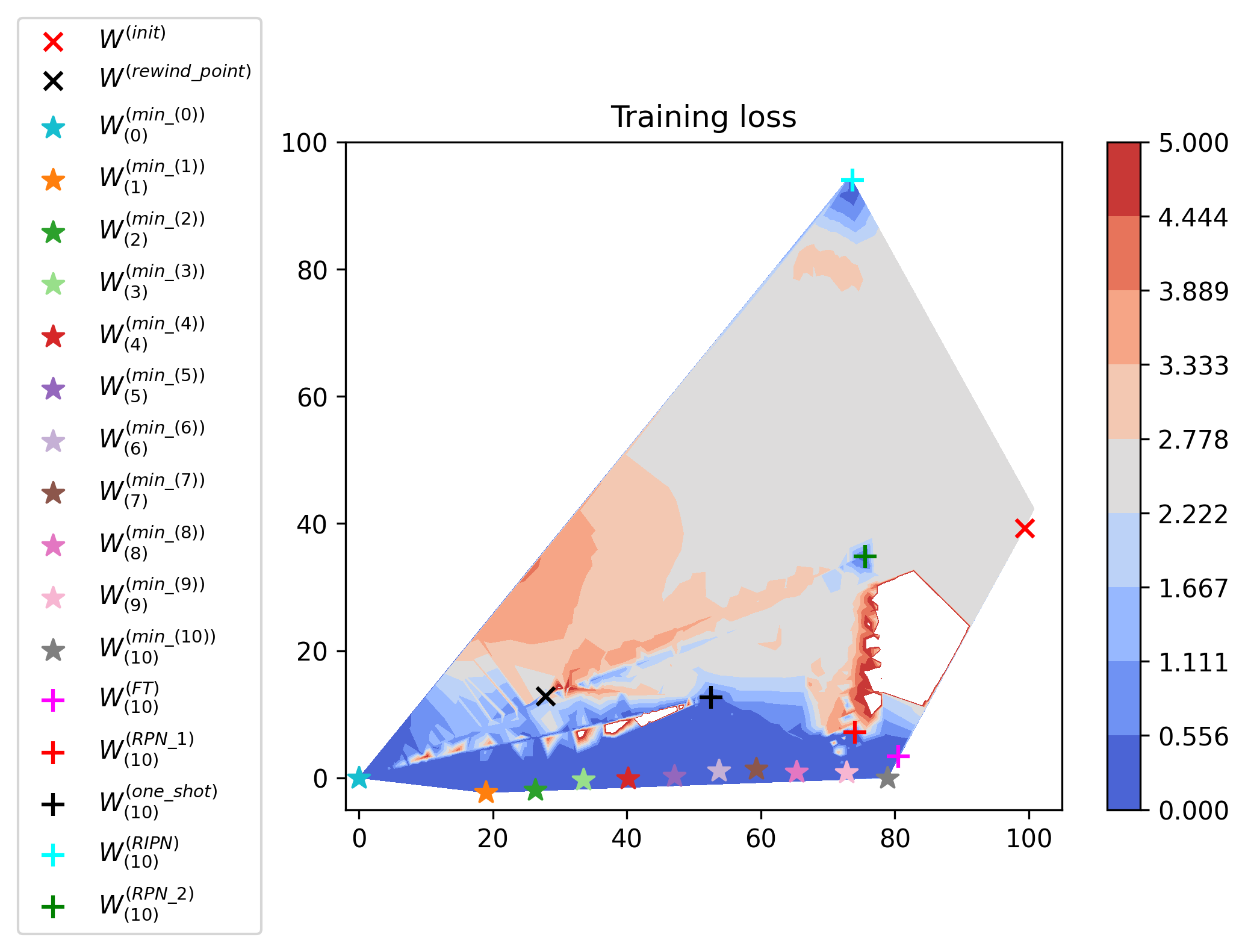}
\caption{Training loss at and around the neighborhood of the points of interest. The plot shows multiple loss sublevel sets. However, all the IMP-WR solutions lie within the same sublevel set.}
\end{figure} \\
This can be further confirmed by calculating the euclidean distances and cosine similarities (angular distances) between different points of interest. Fig. 14 shows the euclidean distance between different points of interest, and Fig. 15 gives the cosine similarity between different points of interest.
\begin{figure}[h!]
\centering
\includegraphics[width=4cm, height=3.5cm]{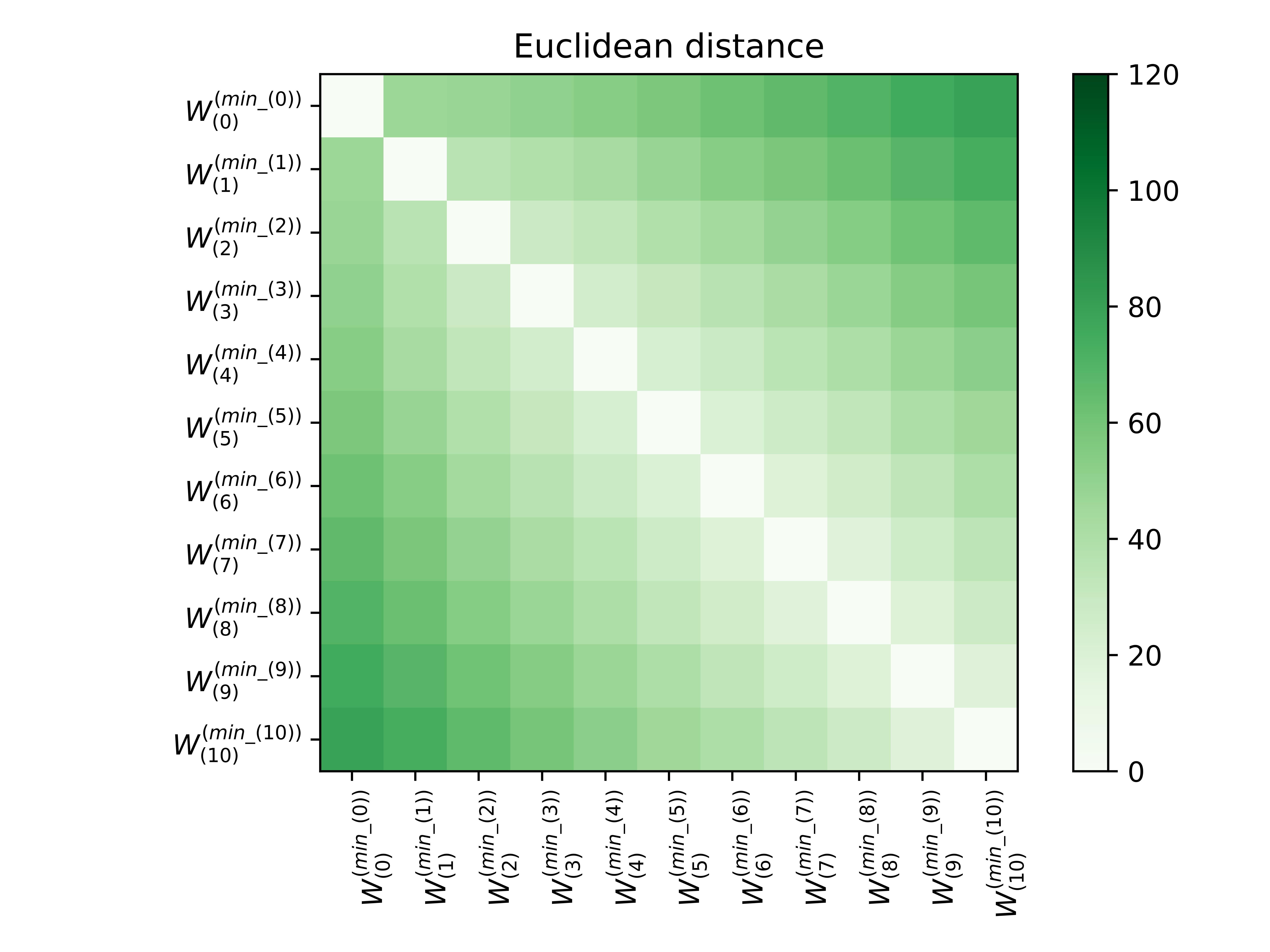}
\includegraphics[width=4cm, height=3.5cm]{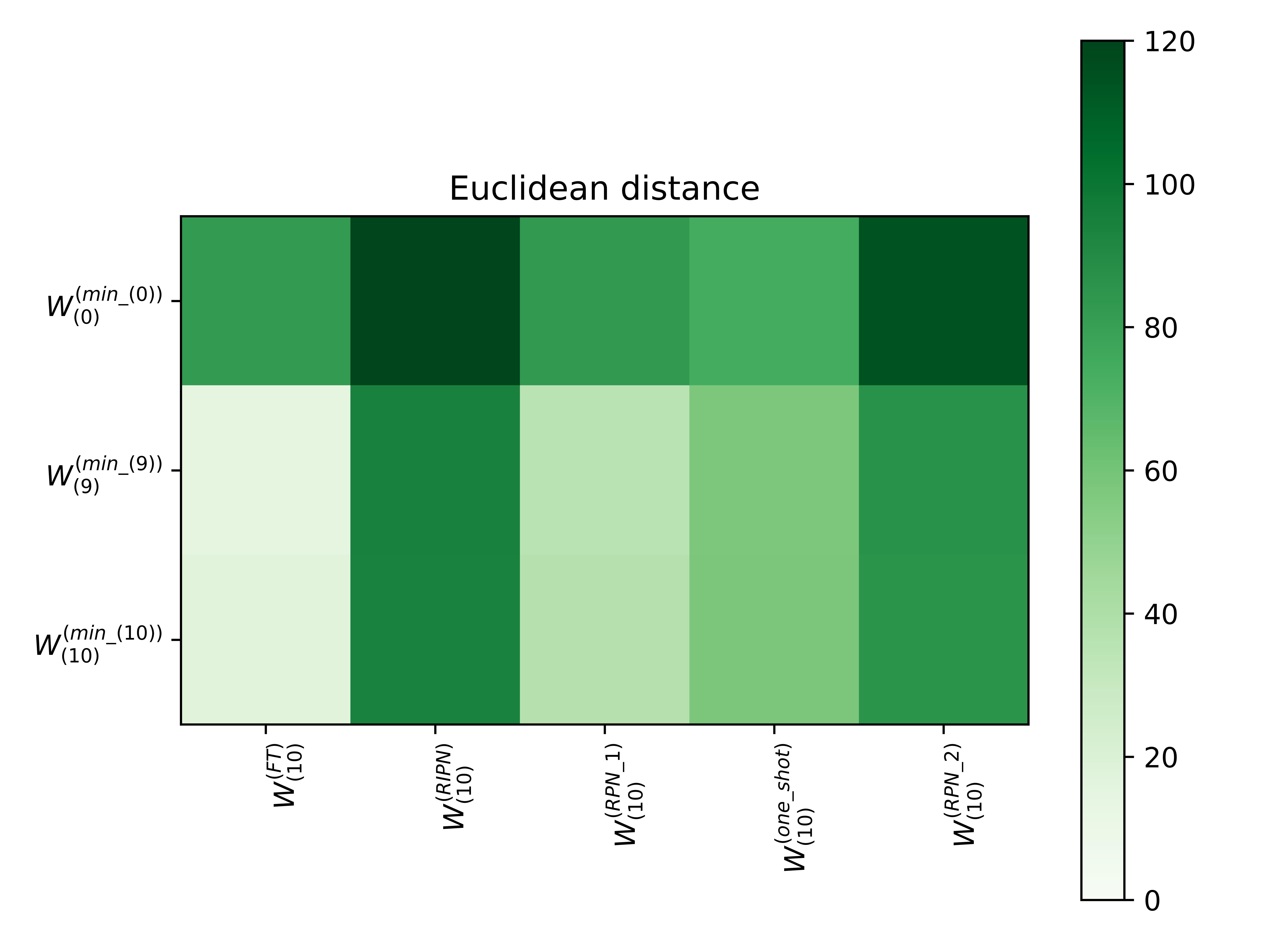}
\caption{Euclidean distance between different points of interest. \textbf{Left:} Euclidean distance between IMP solutions at different levels. \textbf{Right:} Euclidean distance of $W^{(FT)}_{(10)}$, $W^{(RIPN)}_{(10)}$, $W^{(RPN\_1)}_{(10)}$, $W^{(one\_shot)}_{(10)}$ and $W^{(RPN\_2)}_{(10)}$ with $W^{(min\_(0))}_{(0)}$, $W^{(min\_(9))}_{(9)}$ and $W^{(min\_(10))}_{(10)}$ .}
\end{figure}
\begin{figure}[h!]
\centering
\includegraphics[width=4cm, height=3.5cm]{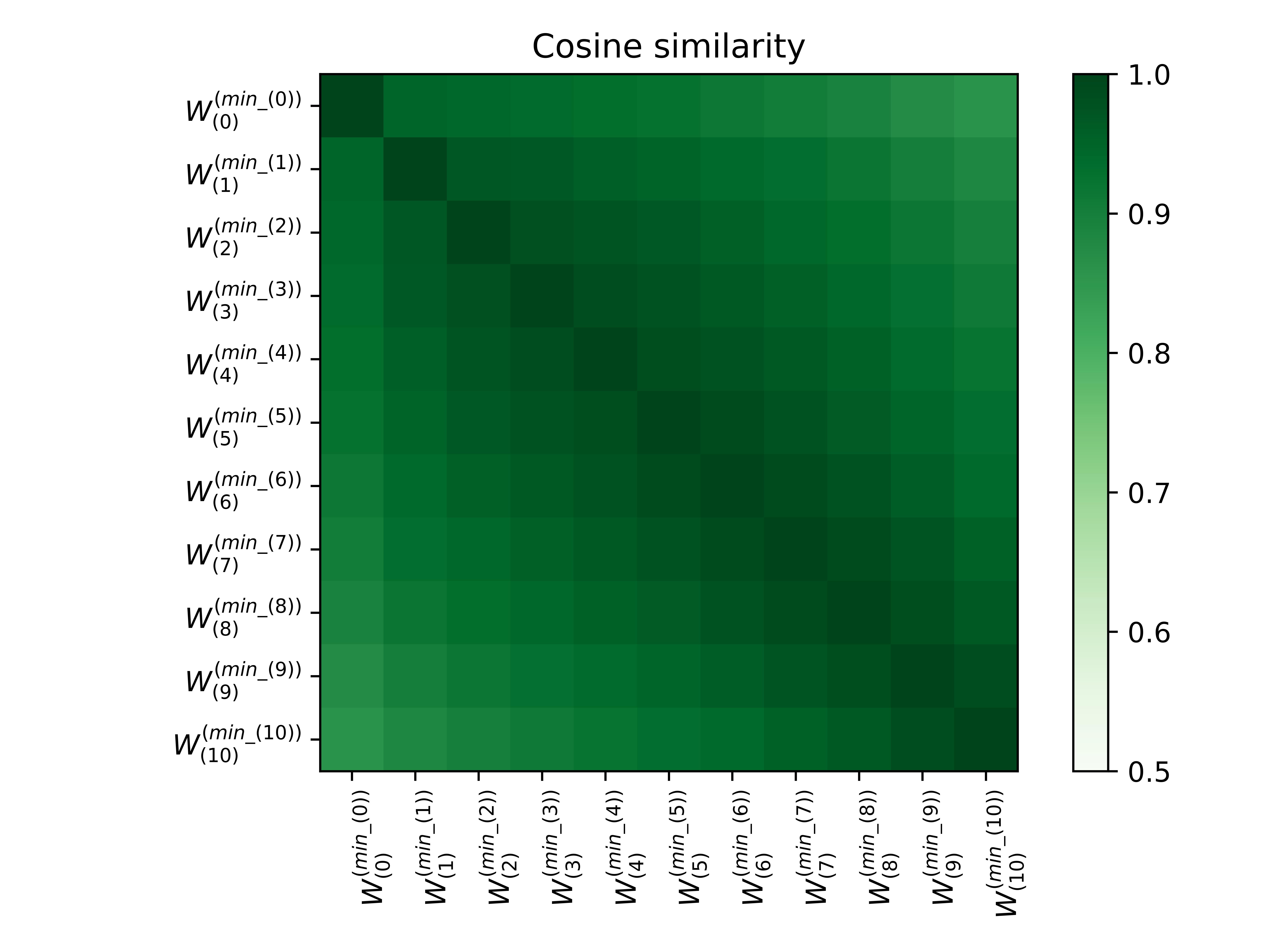}
\includegraphics[width=4cm, height=3.5cm]{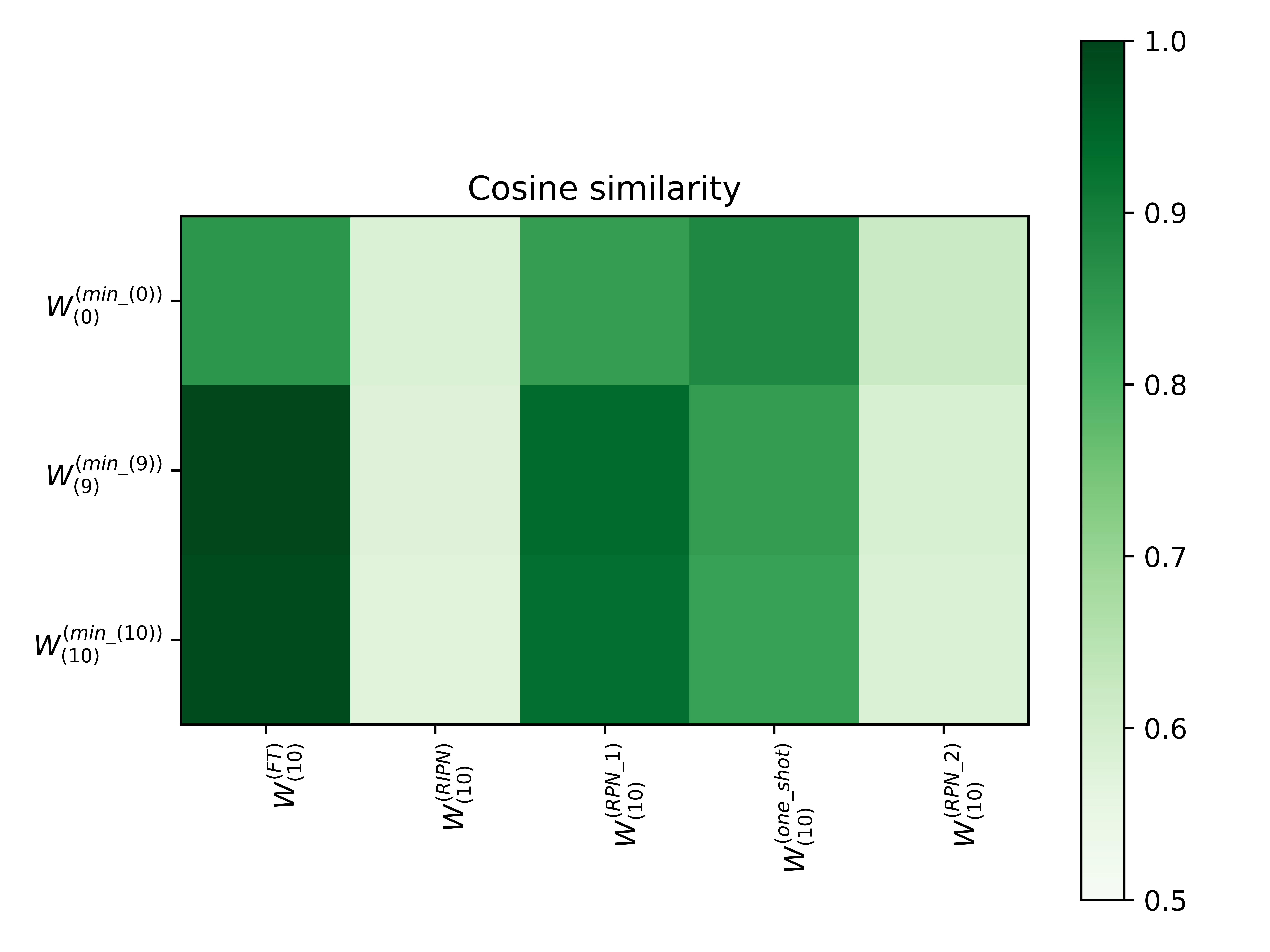}
\caption{Cosine similarity between different points of interest. \textbf{Left:} Cosine similarity between IMP solutions at different levels. \textbf{Right:} Cosine similarity of $W^{(FT)}_{(10)}$, $W^{(RIPN)}_{(10)}$, $W^{(RPN\_1)}_{(10)}$, $W^{(one\_shot)}_{(10)}$ and $W^{(RPN\_2)}_{(10)}$ with $W^{(min\_(0))}_{(0)}$, $W^{(min\_(9))}_{(9)}$ and $W^{(min\_(10))}_{(10)}$.}
\end{figure}
 It is apparent from these figures that the euclidean distance between IMP solutions is smaller than the euclidean distance of $W^{(RIPN)}_{(10)}$ with $W^{(min\_(9))}_{(9)}$  and $W^{(min\_(10))}_{(10)}$. Similarly, the cosine similarity between IMP solutions is greater than the cosine similarity of $W^{(RIPN)}_{(10)}$ with $W^{(min\_(9))}_{(9)}$ and $W^{(min\_(10))}_{(10)}$. Also the euclidean distance between $W^{(min\_(0))}_{(0)}$ and $W^{(one\_shot)}$ is smaller than the euclidean distance between $W^{(min\_(0))}_{(0)}$ and $W^{(RPN\_2)}_{(10)}$, while the cosine similarity between $W^{(min\_(0))}_{(0)}$  and $W^{(one\_shot)}_{(10)}$ is greater than the cosine similarity of $W^{(min\_(0))}_{(0)}$ with $W^{(RPN\_2)}_{(10)}$. This further buttresses the claim.

\subsection{Result 6: What happens when you prune the smaller weights?}
Finding the weights whose removal causes a minimum increase in the loss function value is a combinatorial problem, and searching for such weights in a combinatorial space is impractical due to the massive number of parameters in a neural network. But somehow, the magnitude based pruning algorithm finds such weights. Le cun et al. \cite{lecun1989optimal} proposed an explanation for this based on approximating the loss function by a $2nd$ order Taylor series expansion:
\begin{align*}
    Loss({W_{pruned}}) = Loss({W}) + {(W_{pruned}-W)^T}\frac{\partial Loss}{\partial {W}}_{{W}={W}}   \end{align*} 
\begin{align}
    + {(W_{pruned}-W)^T}\frac{\partial^2Loss}{\partial {W}^2}{(W_{pruned}-W)} 
\end{align} 
where $W$ represents the weights of a dense network and $W_{pruned}$ represents the weights of a pruned network. To simplify the explanation, it is assumed that the loss function is smooth (which may not be true in reality due to the presence of non-linearity induced by ReLU). It is also assumed that the Hessian of the loss function is a diagonal matrix. Calculating the actual Hessian is too complex, and this simplification works well in practice. \par Pruning out the smallest components in the weight vector ${W}$ results in only a small increase in the loss value. Pruning out the larger weights tends to bring about a bigger increase in the loss value since it makes the vector ${(W_{pruned}-W)}$ have a larger magnitude (norm). This can invalidate the second-order approximation given in Eq. $6$, since, in general, the loss function will not be a smooth one, and may have several points of inflection in the neighborhood. \par We hypothesize that pruning larger magnitude weights either takes SGD out of the sublevel set and/or reduces the quality of the baseline minimum so drastically that after re-training, it converges to a minimum with inferior performance. This is corroborated when we compare the pruning of smaller magnitude weights against slightly larger ones. The volume of minimum in the latter case is generally reduced by a great deal, as seen in Fig. 16. This also means that the directions with smaller weights do not have the flattest profile; otherwise, the volume reduction should have been larger in the earlier case than the latter one. \par
\begin{figure}[h!]
\centering
\includegraphics[width=4cm, height=3.5cm]{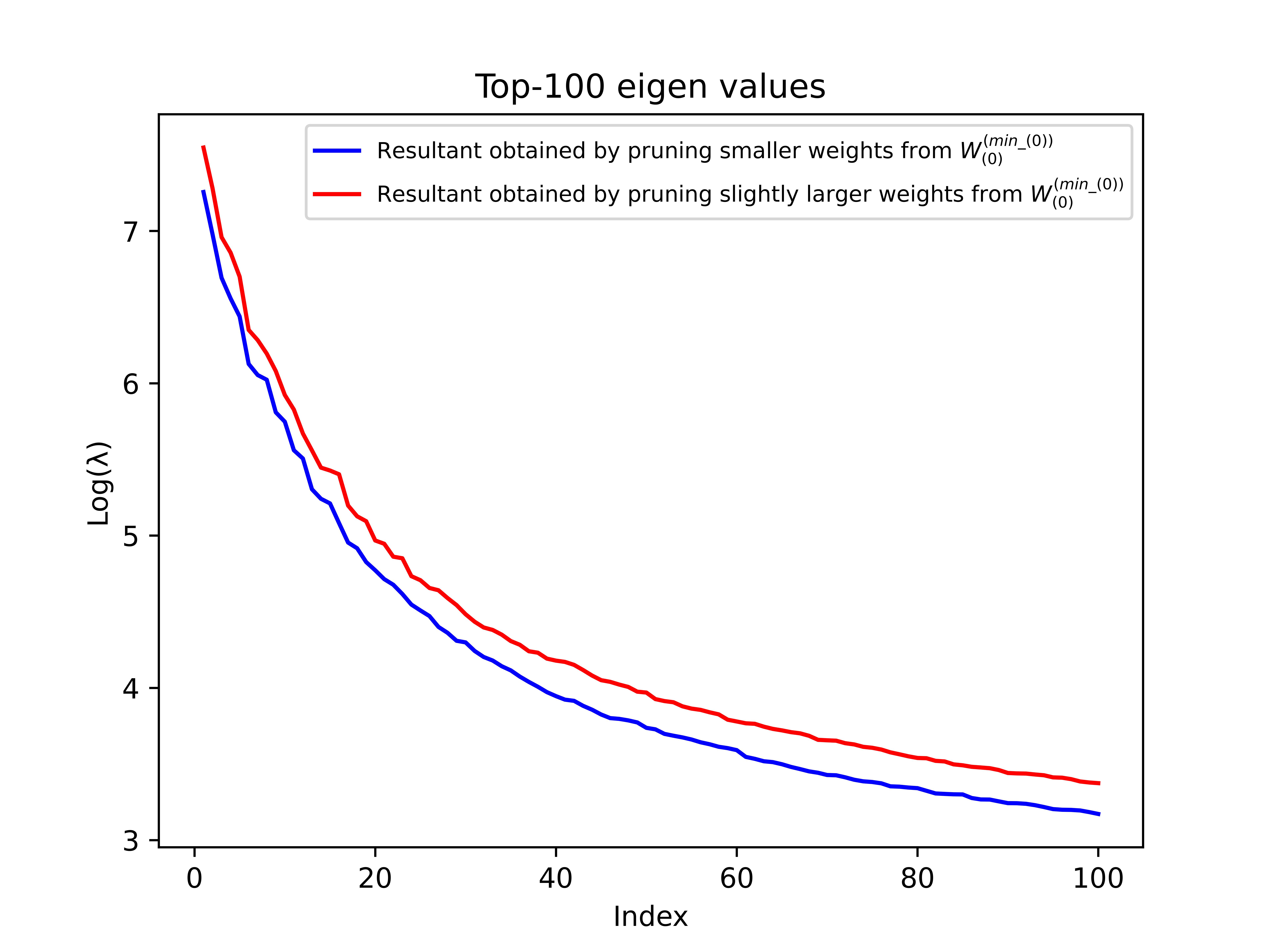}  
\includegraphics[width=4cm, height=3.5cm]{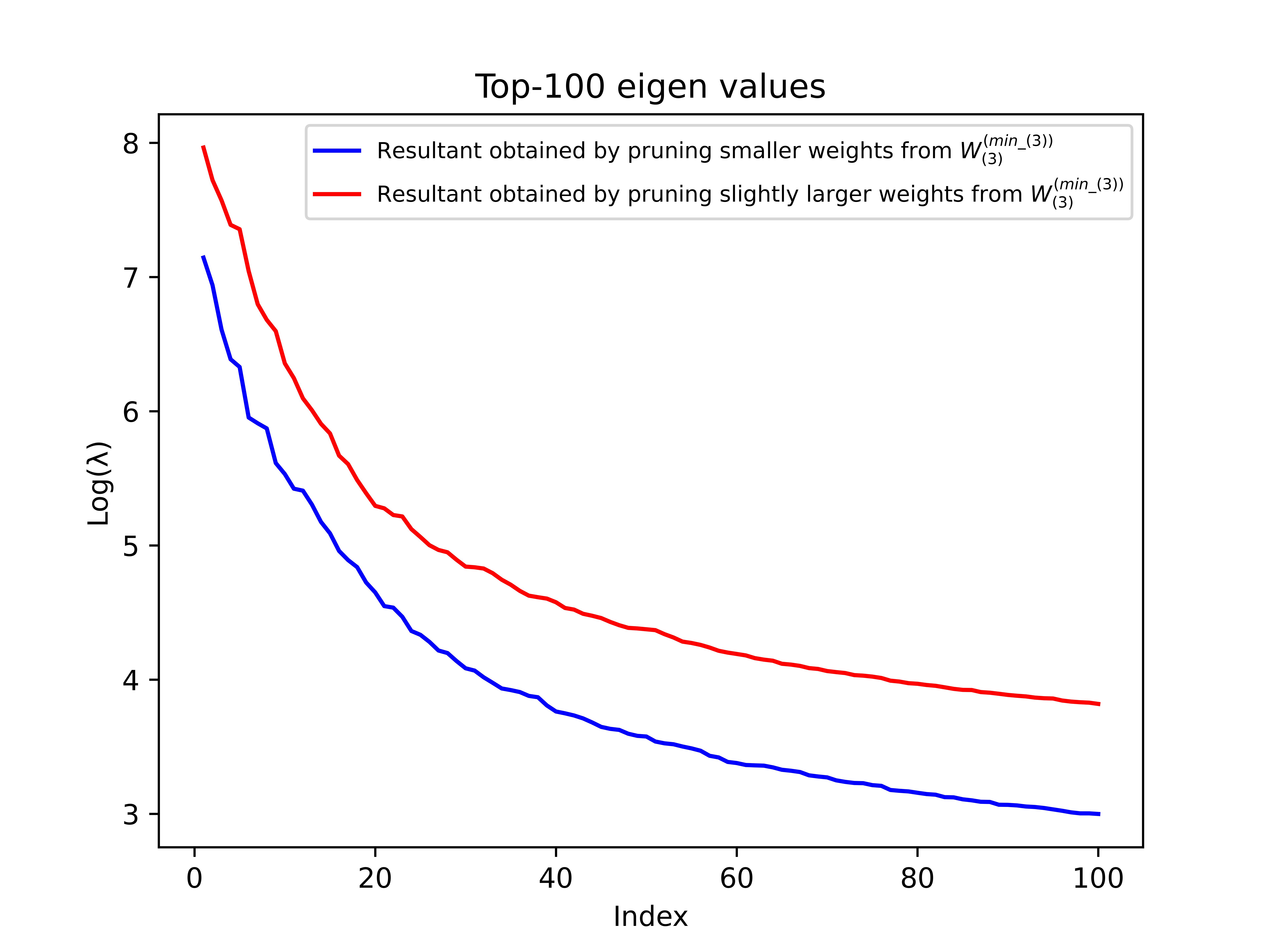} 
\caption{Comparison of top-$100$ positive eigen values of the Hessian at the minimum obtained by pruning smaller magnitude weights and at the minimum obtained by pruning slightly larger weights. \textbf{Left:} Weights are pruned from $W^{(min\_(0))}_{(0)}$.  \textbf{Right:} Weights are pruned from $W^{(min\_(3))}_{(3)}$.} 
\end{figure} \par
Randomly pruning a large number of weights also has a similar effect in that it may lead  SGD to a position outside the sublevel set region, so that even re-training with weight rewinding cannot bring the network back to the sublevel set. Fig. 17 presents the training loss along a straight line connecting $W^{(min\_(0))}_{(0)}$ and  $W^{(one\_shot)}_{(10)}$, and $W^{(min\_(0))}_{(0)}$ (baseline for $W^{(RPN\_2)}_{(10)}$) and $W^{(RPN\_2)}_{(10)}$. 
 \begin{figure}[h!]
\centering
\includegraphics[width=4cm, height=3.5cm]{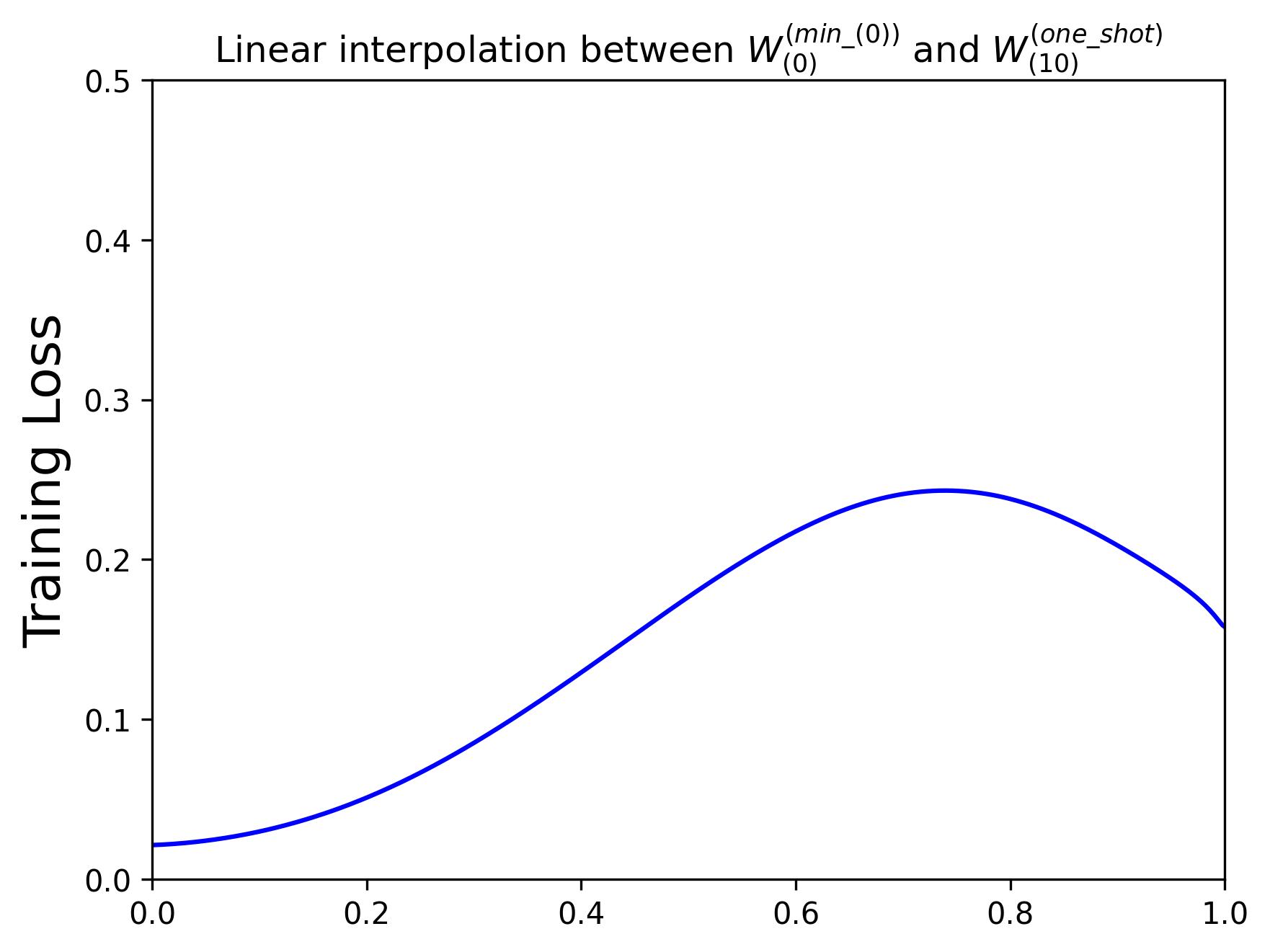}  
\includegraphics[width=4cm, height=3.5cm]{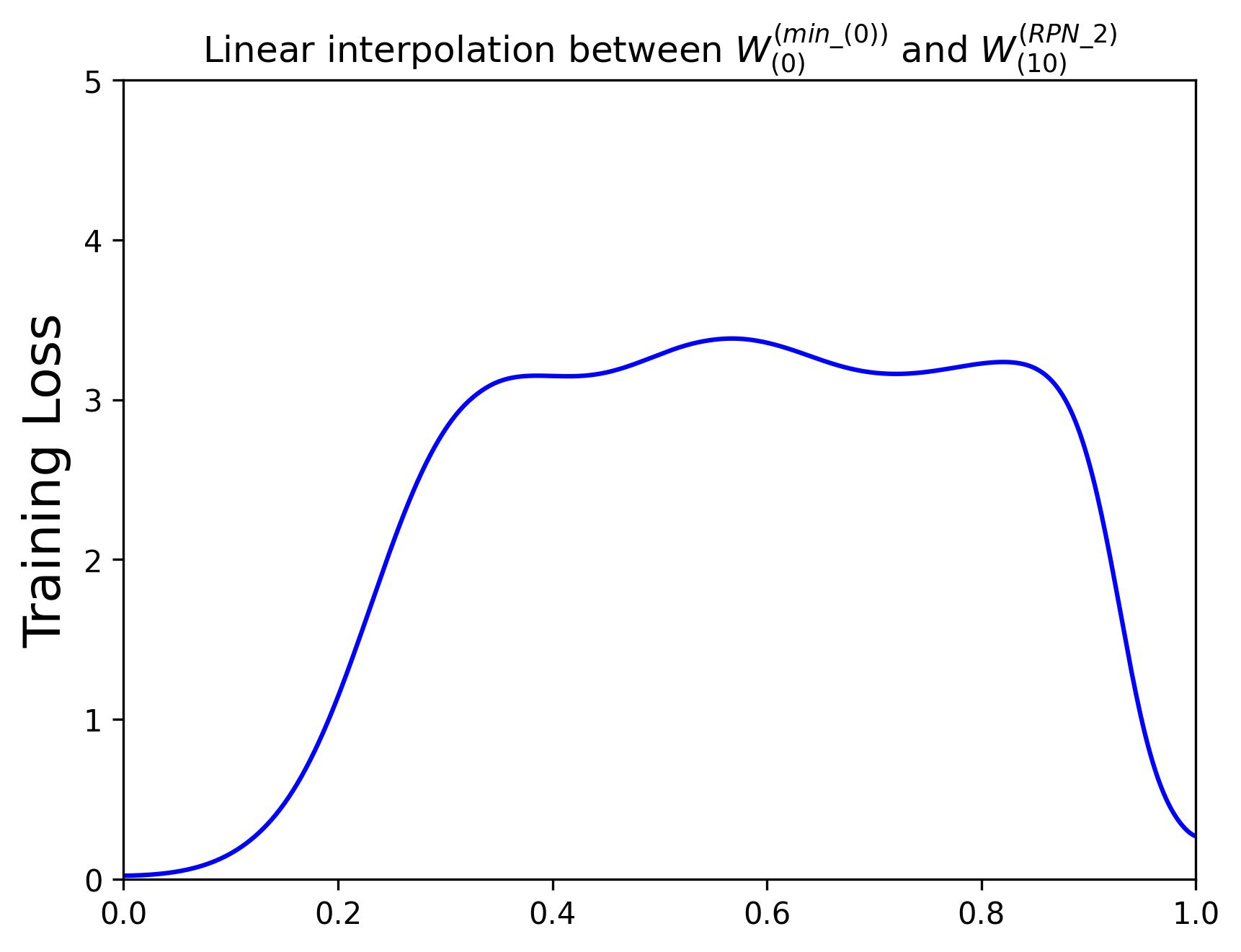} 
\caption{\textbf{Left:} Training Loss along a straight connecting $W^{(min\_(0))}_{(0)}$ and  $W^{(one\_shot)}_{(10)}$. \textbf{Right:} Training Loss along a straight connecting $W^{(min\_(0))}_{(0)}$ and $W^{(RPN\_2)}_{(10)}$.} 
\end{figure}
The figure shows a small barrier between $W^{(min\_(0))}_{(0)}$ and $W^{(one\_shot)}_{(10)}$, and a huge barrier between $W^{(min\_(0))}_{(0)}$ and $W^{(RPN\_2)}_{(10)}$, which again demonstrates that $W^{(min\_(0))}_{(0)}$ and $W^{(one\_shot)}_{(10)}$ lie in the same sublevel set, while $W^{(min\_(0))}_{(0)}$ and $W^{(RPN\_2)}_{(10)}$ lie in different sublevel sets. \par A comparison of top-100 positive eigen values of Hessians at $W^{(one\_shot)}_{(10)}$ and $W^{(RPN\_2)}_{(10)}$ is given in Fig. 18.
\begin{figure}[h!]
\centering
\includegraphics[width=4cm, height=3.5cm]{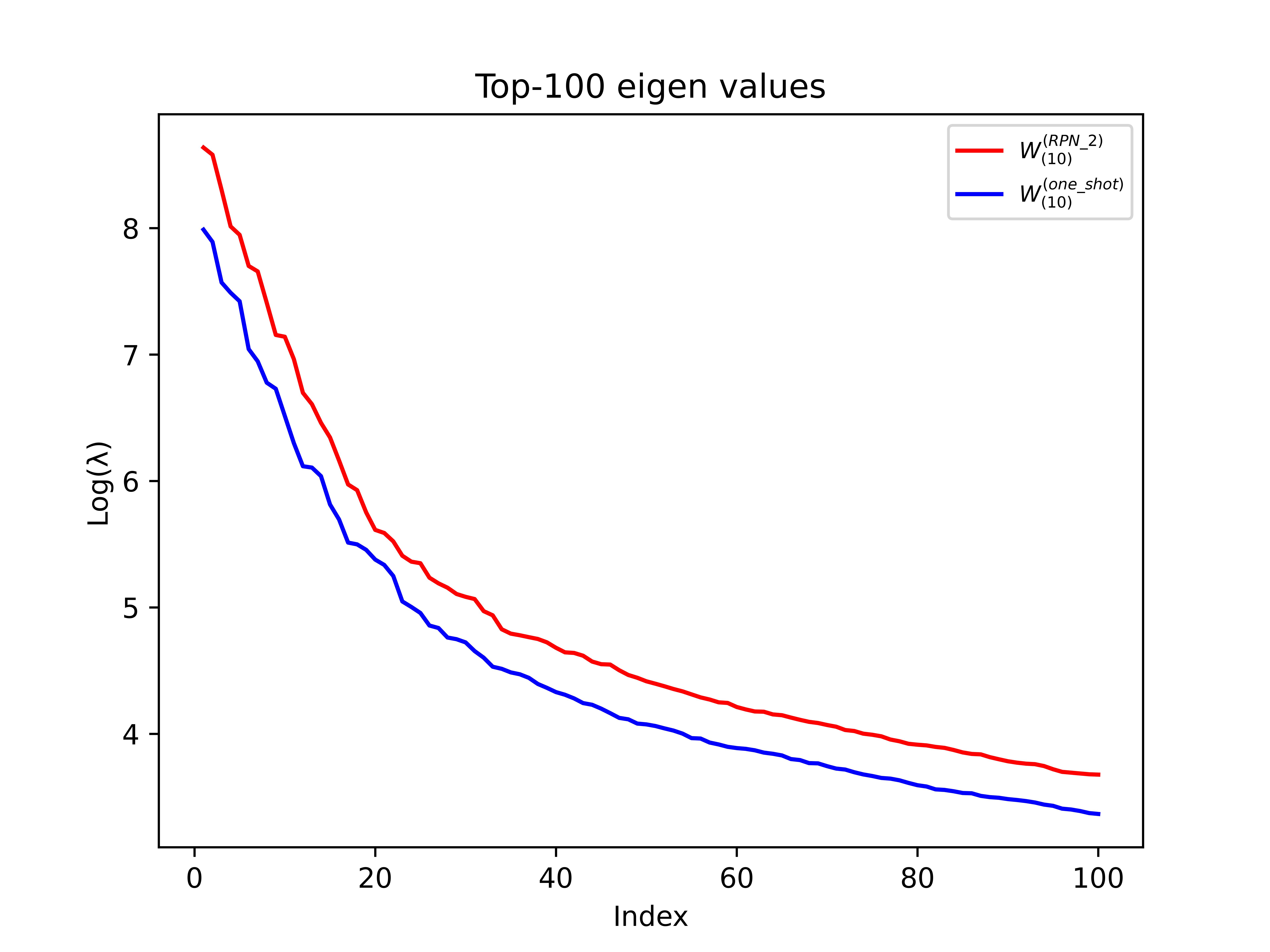}
\centering
\caption{Comparison of top-100 positive eigen values of the Hessian at $W^{(one\_shot)}_{(10)}$ and $W^{(RPN\_2)}_{(10)}$.} 
\end{figure}
It can be observed from the figure that $W^{(RPN\_2)}_{(10)}$ has larger eigen values of Hessian than that at $W^{(one\_shot)}_{(10)}$, which implies a smaller volume for the basin around $W^{(RPN\_2)}_{(10)}$. \par
This demonstrates our contention that random pruning of a large number of weights can take the SGD out of the sublevel set, and after retraining, converge to a minimum with a smaller volume. \par
Fig. 19 presents a training loss along a straight line joining $W^{(min\_(9))}_{(9)}$ and $W^{(RPN\_1)}_{(10)}$. The figure shows a small barrier between $W^{(min\_(9))}_{(9)}$ and $W^{(RPN\_1)}_{(10)}$, indicating that they lie in the same sublevel set. However, the volume of the basin around $W^{(RPN\_1)}_{(10)}$ is smaller than the volume of the basin around $W^{(min\_(10))}_{(10)}$. This is illustrated in Fig. 20.
 \begin{figure}[h!]
\centering
\includegraphics[width=4cm, height=3.5cm]{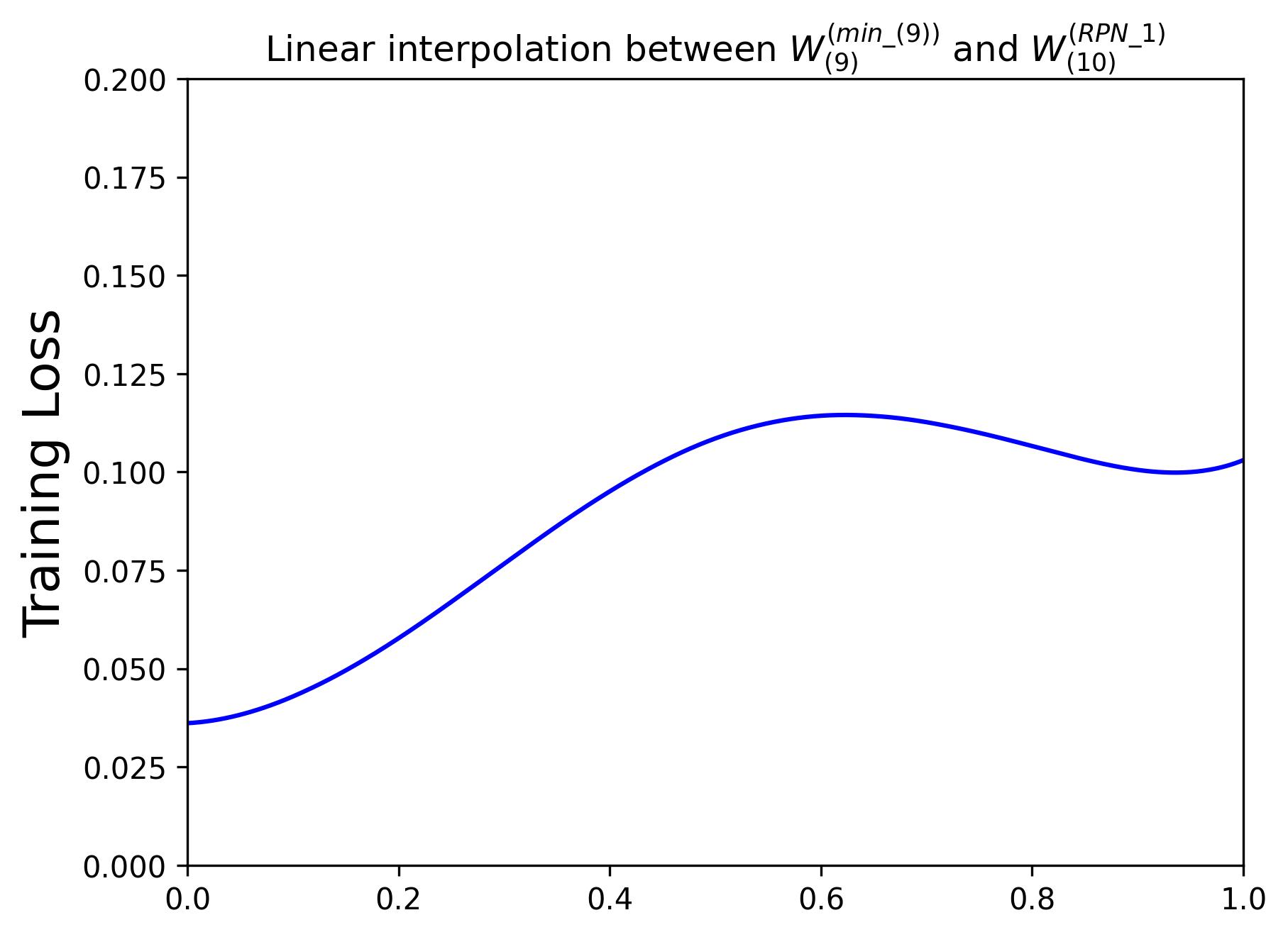}  
\caption{Training Loss along a straight line between $W^{(min\_(9))}_{(9)}$ and $W^{(RPN\_1)}_{(10)}$. } 
\end{figure}
\begin{figure}[h!]
\centering
\includegraphics[width=4cm, height=3.5cm]{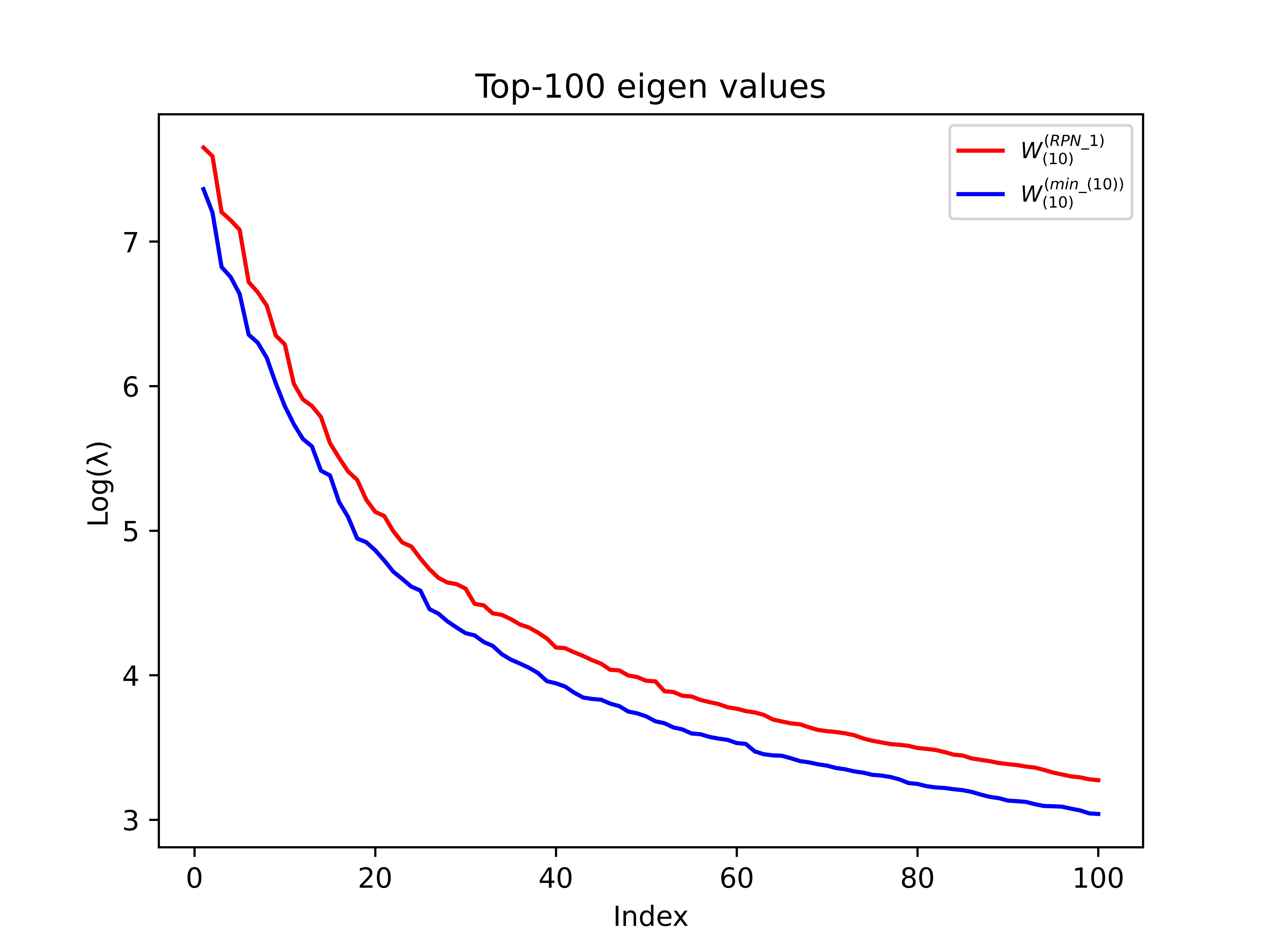}
\centering
\caption{Comparison of top-100 positive eigen values of the Hessian at $W^{(RPN\_1)}_{(10)}$ and $W^{(min\_(10))}_{(10)}$.} 
\end{figure} 

This showcases the importance of pruning strategy in neural network pruning. In general, given that at each level, IMP finds solutions in the neighborhood of the dense network minimum, an arbitrary perturbation in the pruning process can lose us this advantage. 
\subsection{Result 7: Why fine-tuning doesn't perform at par with rewinding?  }
Fine-tuning perturbs the pruned baseline by a small amount; hence, SGD more likely stays near the baseline minimum and does not explore the minima outside the baseline. However, rewinding takes the SGD out of the baseline minimum and is more likely to converge to a better minimum in the pruned space (which was undiscoverable in the original space).  \par
Fig. 21 presents the comparison of top-100 positive eigen values of the Hessian at $W^{(FT)}_{(10)}$ and $W^{(min\_(10))}_{(10)}$.
\begin{figure}[h!]
\centering
\includegraphics[width=4cm, height=3.5cm]{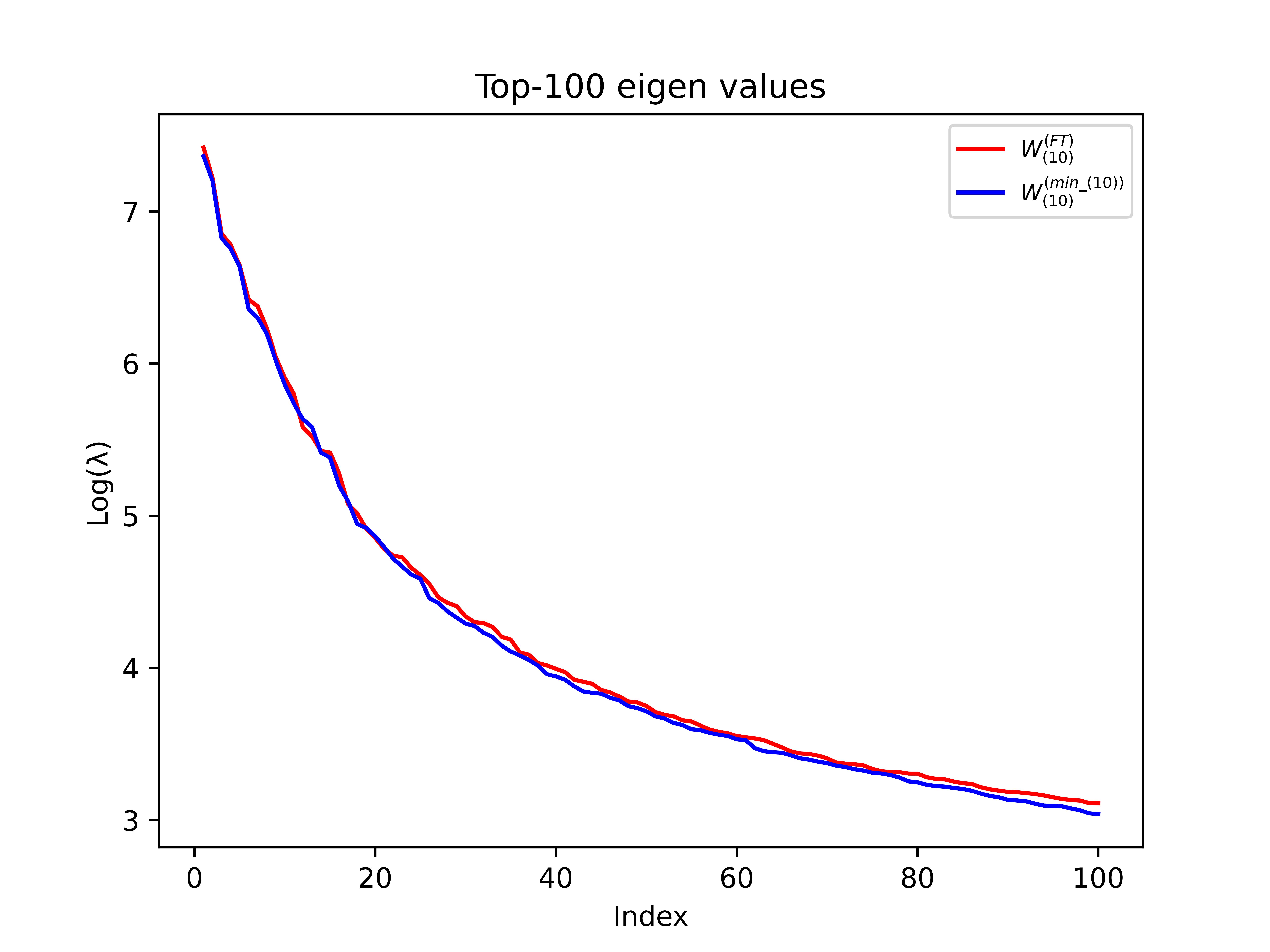}  
\caption{Comparison of top-100 positive eigen values of the Hessian at $W^{(FT)}_{(10)}$ and $W^{(min\_(10))}_{(10)}$.} 
\end{figure}
 It can be observed from the figure that $W^{(FT)}_{(10)}$ has larger eigen values of the Hessian than $W^{(min\_(10))}_{(10)}$, which implies a smaller volume for the basin around $W^{(FT)}_{(10)}$. This elucidates the importance of rewinding the learning rate schedule while re-training the pruned network.

\section{Conclusion and Scope for Further Work}
In this work, we have studied the loss landscape characteristics and volume/geometry of
the IMP solutions at different levels in order to answer some important questions about the IMP and the lottery ticket hypothesis. The study provided the following important insights among others: 1) there exist special type of solutions in the loss landscape, which perform well but have a very small volume in the original space, and the IMP procedure uncovers such solutions; 2) there exists a barrier between the IMP solutions at successive levels in the loss landscape; 3) IMP solutions obtained using rewinding lie within the same loss
sublevel set, and random pruning or random initialization of a pruned network take the SGD out of that sublevel set. These insights help better understand the underlying principles behind  IMP and the lottery-ticket hypothesis. \par
We have shown that solutions exist that have narrow profiles along certain dimensions and flatter profiles along others; however, the narrow profiles of these solutions are along sparser dimensions. There is a possibility of the existence of solutions that have narrow profiles along certain dimensions and flatter profiles along others, with narrow profiles along non-sparser dimensions. These solutions are clearly not important from the pruning perspective. If so, an interesting question is: can such solutions serve some other useful properties? This is a potential future direction that can be explored. \par Another future direction would be to design a computationally efficient algorithm that makes SGD directly converge to the good sparser solutions without going through the computationally expensive pruning and re-training cycles.
\bibliography{refs}
\bibliographystyle{IEEEtran}
\section*{\textbf{Appendix-I: Further Experimental Details}} \par
Some finer details of the experiments conducted in this study are mentioned here for completeness.\\
\textbf{ResNet-20 on CIFAR-10.} We train ResNet-20 on CIFAR-10 for $160$ epochs with SGD and a batchsize of $128$. We use learning rate = $0.1$,
momentum = $0.9$, and weight decay = $0.0001$. The learning rate is decayed by a factor of $10$ at $80$ and $120$ epochs. We use iterative magnitude pruning with weight rewinding (IMP-WR) and run $10$ iterations of IMP. We prune $20\%$ of the smallest magnitude weights in each iteration. The prunable parameters
are the weights of the convolutional layers and the fully-connected layers.\\
\textbf{VGG-16 on CIFAR-10.} We train VGG-16 on CIFAR-10 for $160$ epochs with SGD and a batchsize of $128$. We use learning rate = $0.1$,
momentum = $0.9$, and weight decay = $0.0001$. The learning rate is decayed by a factor of $10$ at $80$
and $120$ epochs. We use IMP-WR and run $12$ iterations of IMP. We prune $40\%$ of the smallest magnitude weights at each round. The prunable parameters
are the weights of the convolutional layers and the fully-connected layers.  \\
\textbf{Inverse volume of basin.} In our experiments, we calculate the logarithm of the product of top-$100$ positive eigen values of the Hessian of the loss function to approximate the inverse volume of the basin. And the Hessian of the loss function is calculated on a randomly chosen subset of the training dataset ($1/5^{th}$ of the size of the full training dataset in the case of ResNet-20 and $1/10^{th}$ of the size of the full training dataset in the case of VGG-16). Fig. 22 shows that the trend with top-$200$ is consistent with top-$100$.
\begin{figure}[h!]
\centering
\includegraphics[width=4cm, height=3.5cm]{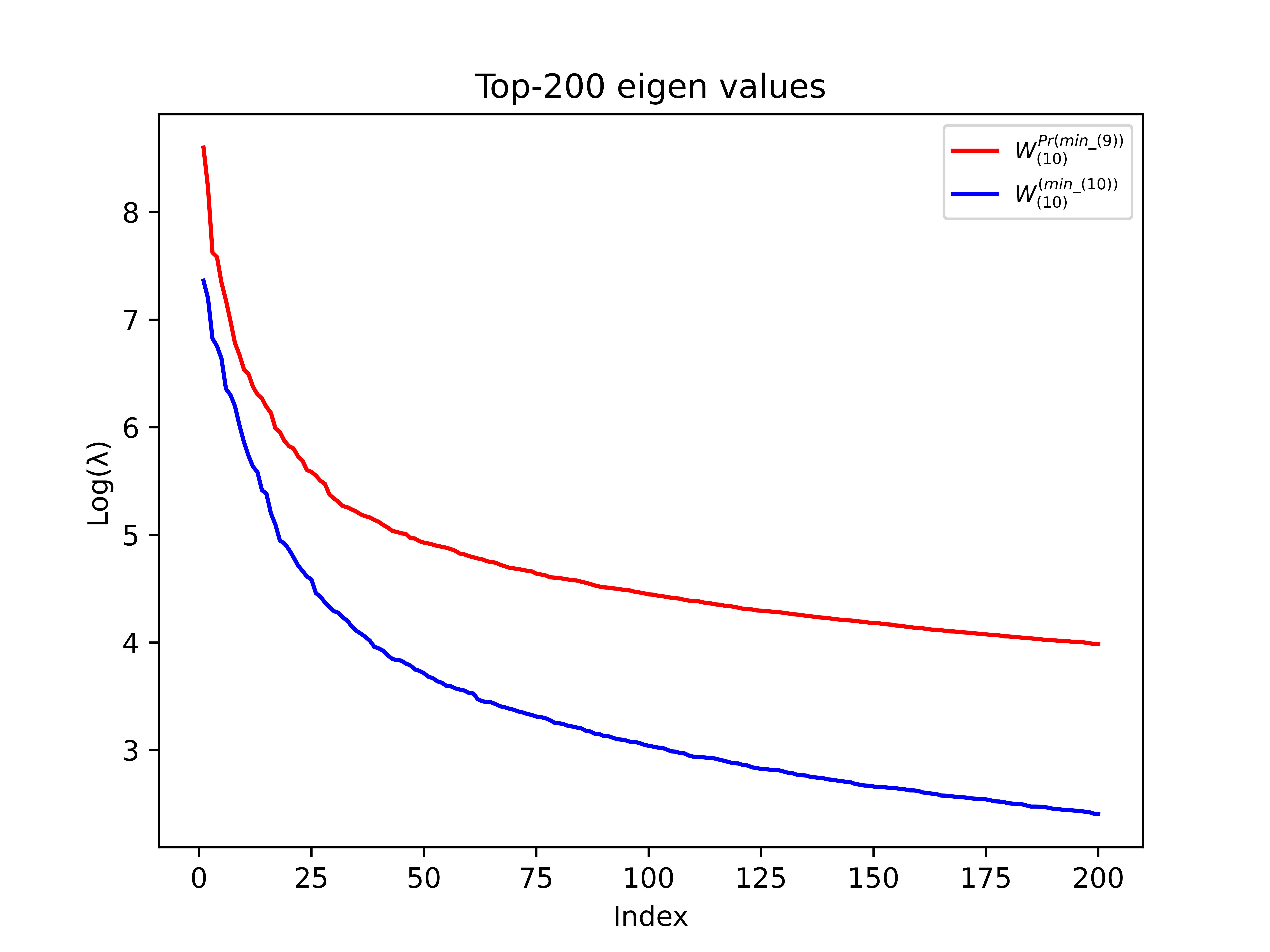}
\caption{Comparison of top-200 positive eigen values of the Hessian at $W_{(10)}^{(min\_(10))}$ and $W^{Pr{(min\_(9))}}_{(10)}$ in case of ResNet-20.}
\end{figure} \\
\textbf{Average radius of basin.} In our experiments, we have calculated the average radius of the basin around a solution by averaging the radii of the basin along $500$ random directions. These experiments have also been performed on a randomly chosen subset of the training dataset ($1/5^{th}$ of the size of the full training dataset).\\
\textbf{Training loss along a straight line connecting two solutions.} The training loss along a straight line connecting two solutions, $W^{(P)}$ and $W^{(Q)}$ is calculated as $Loss((1-\alpha)W^{(P)}+\alpha W^{(Q)})$, where $\alpha \in [0,1]$. In our linear interpolation experiments, we calculate training loss at $501$ points between the two solutions. \\
\textbf{Loss landscape plot.} Fig. 13 presents the training loss at and around the neighbourhood of the different points of interest. The loss is calculated using a randomly chosen subset of the training dataset ($1/5^{th}$ of the size of the full training dataset). 
The plot has been obtained by calculating training loss at $4200$ points in the high-dimensional space and then projecting the points on the two-dimensional plane spanned by $3$ points, $W_{(0)}^{(min\_(0))}$, $W_{(10)}^{(min\_(10))}$ and $W^{(RIPN)}$. However, the two axes formed by these points are not orthogonal. Hence, orthogonalization of these axes is carried out before projecting the points from the high-dimensional space. \\ Projection of a high-dimensional vector $r$ onto the plane spanned by orthogonal vectors $dx$ and $dy$ has been obtained as follows:
$x = (r.dx)/|dx|$,
$y = (r.dy)/|dy|$, where $x$ and $y$ are the projection coordinates along $dx$ and $dy$ respectively.\\
\section*{\textbf{Appendix-II: Experimental Results on VGG-16/CIFAR-10}}
Here, we present the evidence for our results from the experimentation conducted on VGG-16/CIFAR-10.
Fig. 23 presents the training loss and test accuracy at different levels of IMP-WR. 
\begin{figure}[h!]
\centering
\includegraphics[width=4cm, height=3.5cm]{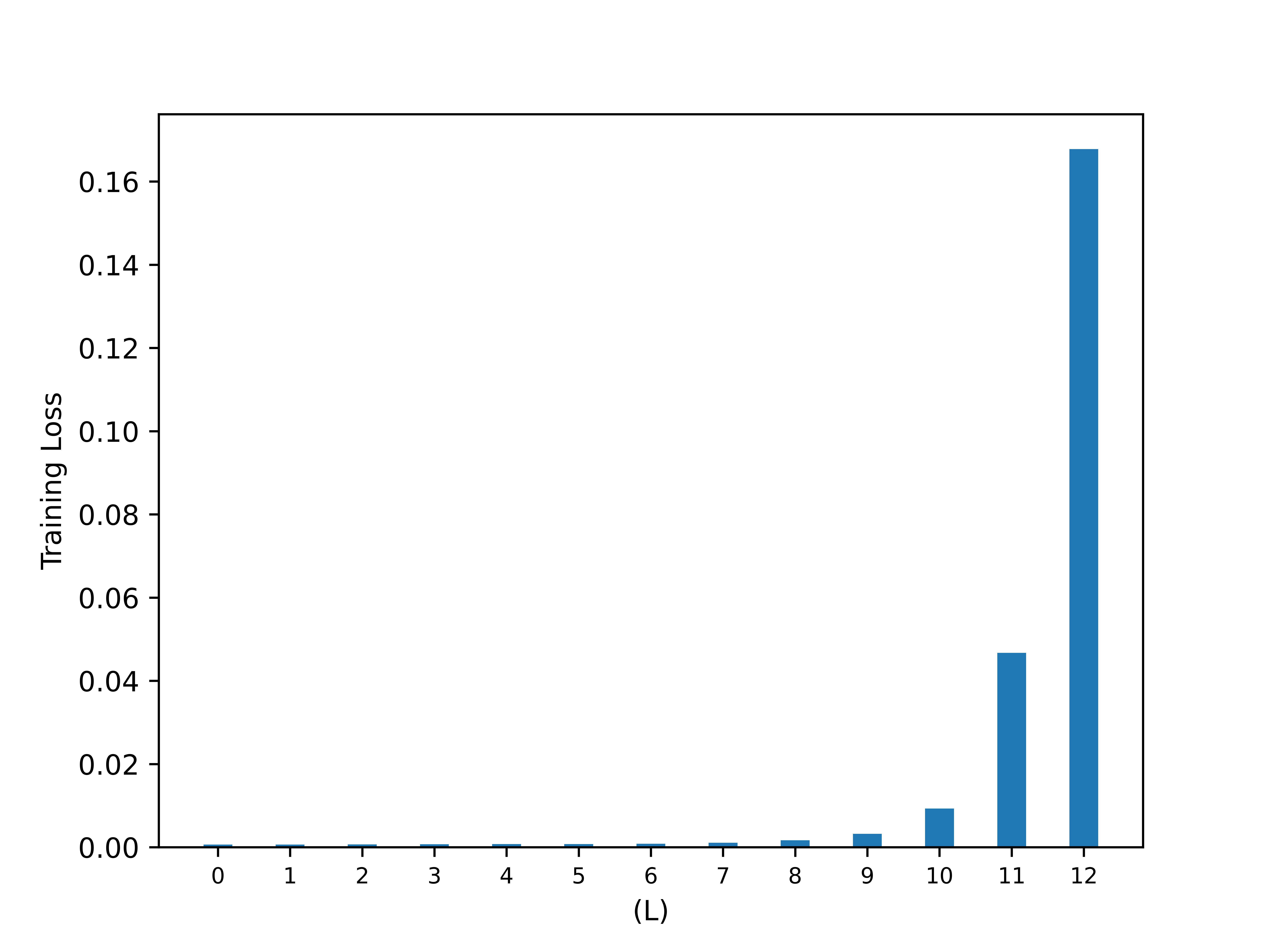}
\includegraphics[width=4cm, height=3.5cm]{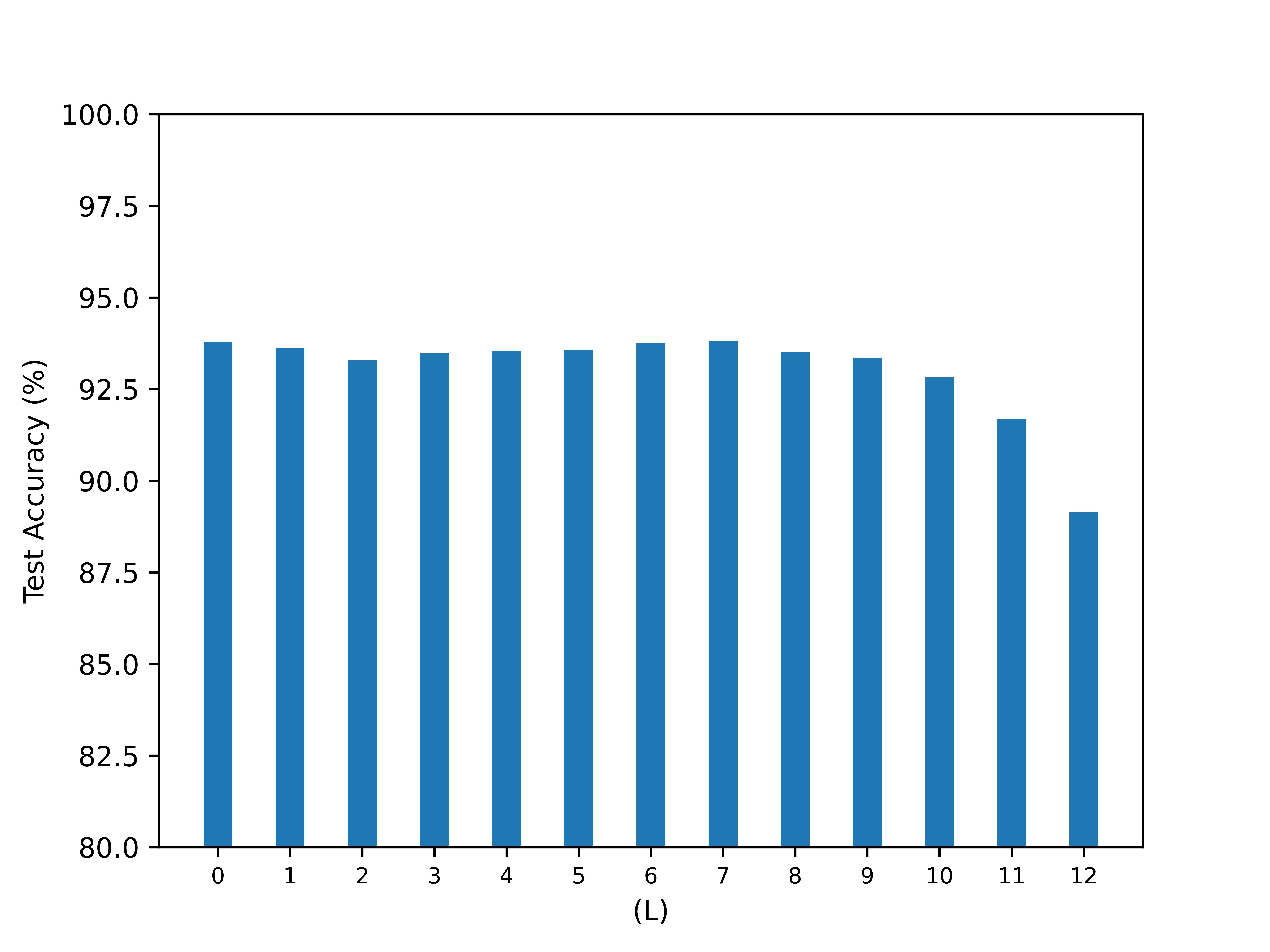}
\caption{Training loss and test accuracy at different levels of IMP-WR in case of VGG-16. \textbf{Left:} Training loss. \textbf{Right:} Test accuracy.}
\end{figure}
\subsection{Solution obtained with IMP-WR outperforms the solutions obtained with the other strategies.}
In the case of VGG-16, one-shot pruned network is obtained by pruning the weights of the trained dense network $W_{(0)}^{(min\_(0))}$ based on magnitude pruning in one go to attain the desired sparsity (same sparsity as that of $W_{(12)}^{(min\_(12))}$) and then rewinding the unpruned weights to their values at $W^{(rewind\_point)}$ and retraining. We represent the solution obtained using one-shot pruning by $W^{(one\_shot)}_{(12)}$. The fine-tuned network is obtained by pruning $40\%$ smallest magnitude weights from $W_{(11)}^{(min\_(11))}$ and then re-training the unpruned weights (without rewinding) with a learning rate of $0.001$ for $40$ epochs. We represent the solution obtained using fine-tuning by $W^{(FT)}_{(12)}$. Randomly initialized pruned network is obtained by pruning $40\%$ smallest magnitude weights from $W_{(11)}^{(min\_(11))}$ and then randomly initializing the unpruned weights and retraining. We represent the solution by $W^{(RIPN)}_{(12)}$. A randomly pruned network has been obtained by randomly pruning $40\%$ weights from $W_{(11)}^{(min\_(11))}$ and then rewinding and retraining. We represent the solution obtained in the aforementioned manner by $W^{(RPN)}_{(12)}$. 
Fig. 24 presents the comparison of training loss and test accuracy between $W_{(12)}^{(min\_(12))}$, $W^{(FT)}_{(12)}$, $W^{(RIPN)}_{(12)}$, $W^{(RPN)}_{(12)}$ and $W^{(one\_shot)}_{(12)}$. The plots clearly show that the solution obtained with IMP-WR ($W_{(12)}^{(min\_(12))}$) outperforms the solutions obtained with the other strategies like: fine-tuning ($W^{(FT)}_{(12)}$), random initialization of the pruned network ($W^{(RIPN)}_{(12)}$), random pruning ($W^{(RPN)}_{(12)}$) and one-shot pruning ($W^{(one\_shot)}_{(12)}$).
\begin{figure}[h!]
\centering
\includegraphics[width=4cm, height=3.5cm]{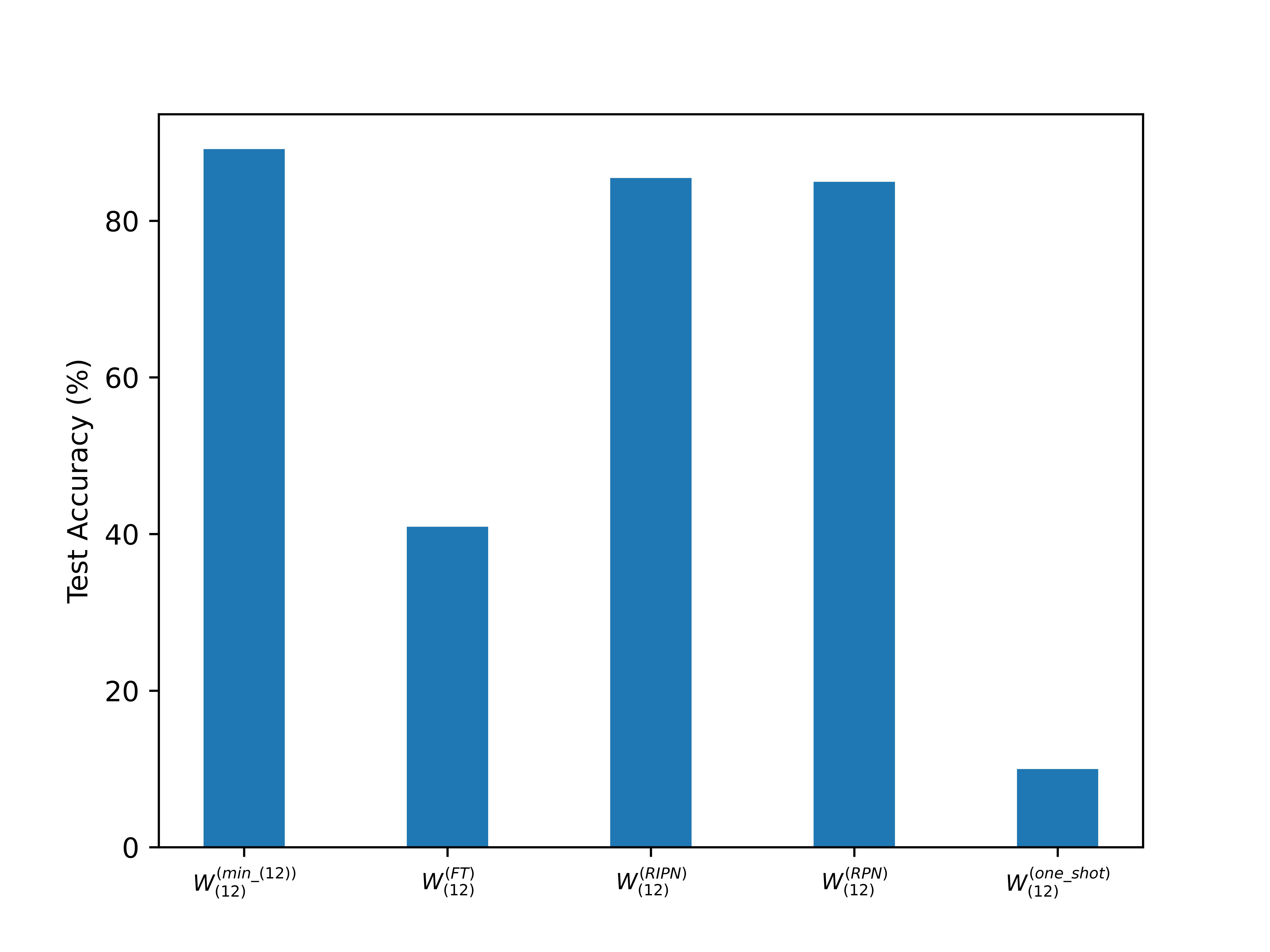}
\includegraphics[width=4cm, height=3.5cm]{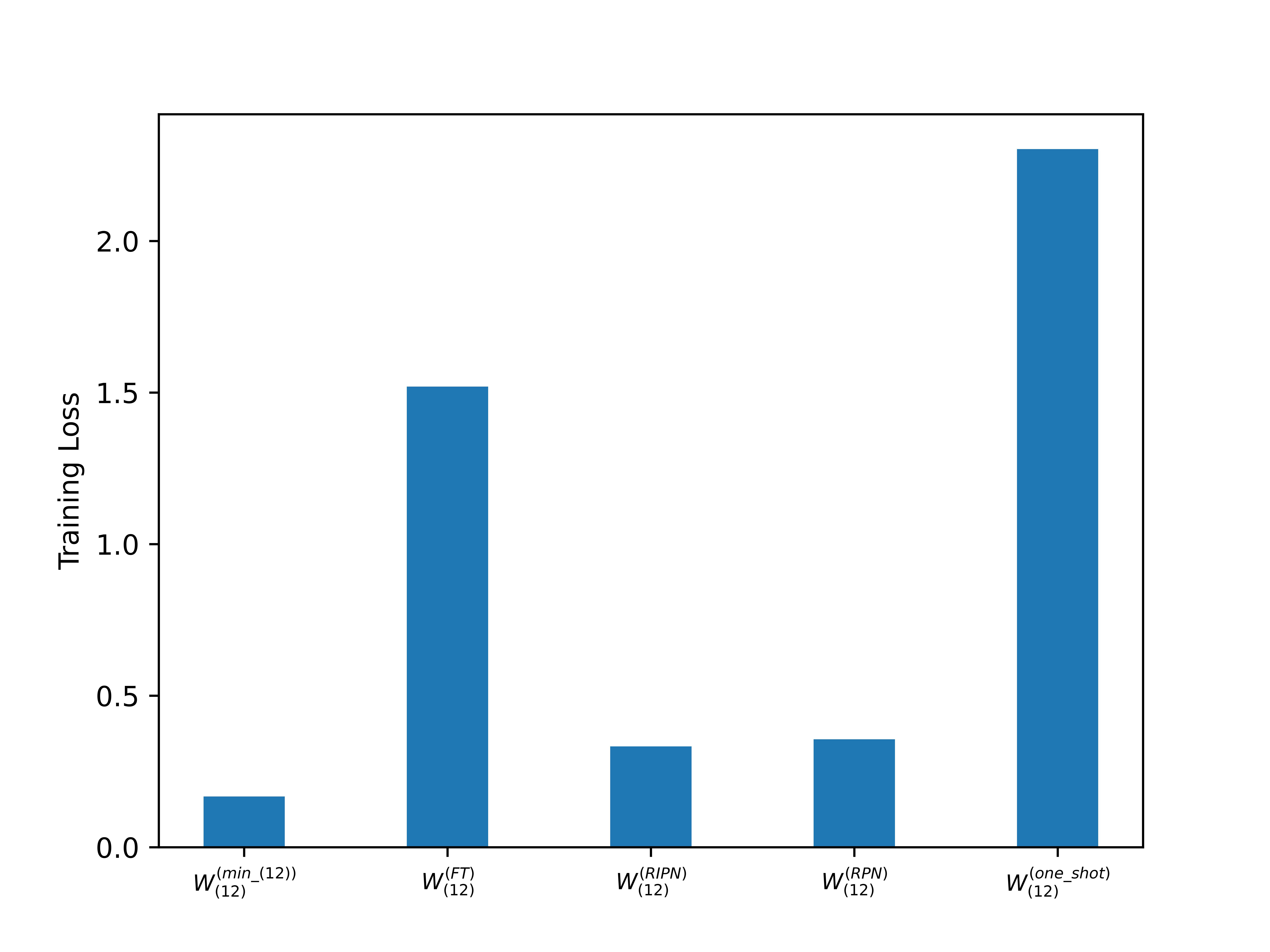}
\caption{Comparison of training loss and test accuracy between $W_{(12)}^{(min\_(12))}$, $W^{(FT)}_{(12)}$, $W^{(RIPN)}_{(12)}$, $W^{(RPN)}_{(12)}$ and $W^{(one\_shot)}_{(12)}$ in case of VGG-16. \textbf{Left:} Training loss. \textbf{Right:} Test accuracy.}
\end{figure}
\subsection{Path leading to $W_{(L)}^{(min\_(L))}$ is steeper than the path leading to $W^{Pr{(min\_(L-1))}}_{(L)}$.}
Fig. 25 presents the comparison of the logarithm of training loss versus epoch between level $(L)$ and level $(L-1)$ projected on level $(L)$. The plots clearly show that the path leading to $W_{(L)}^{(min\_(L))}$ is steeper than the path leading to $W^{Pr{(min\_(L-1))}}_{(L)}$.
\begin{figure*}
\centering
\includegraphics[width=4cm, height=3.5cm]{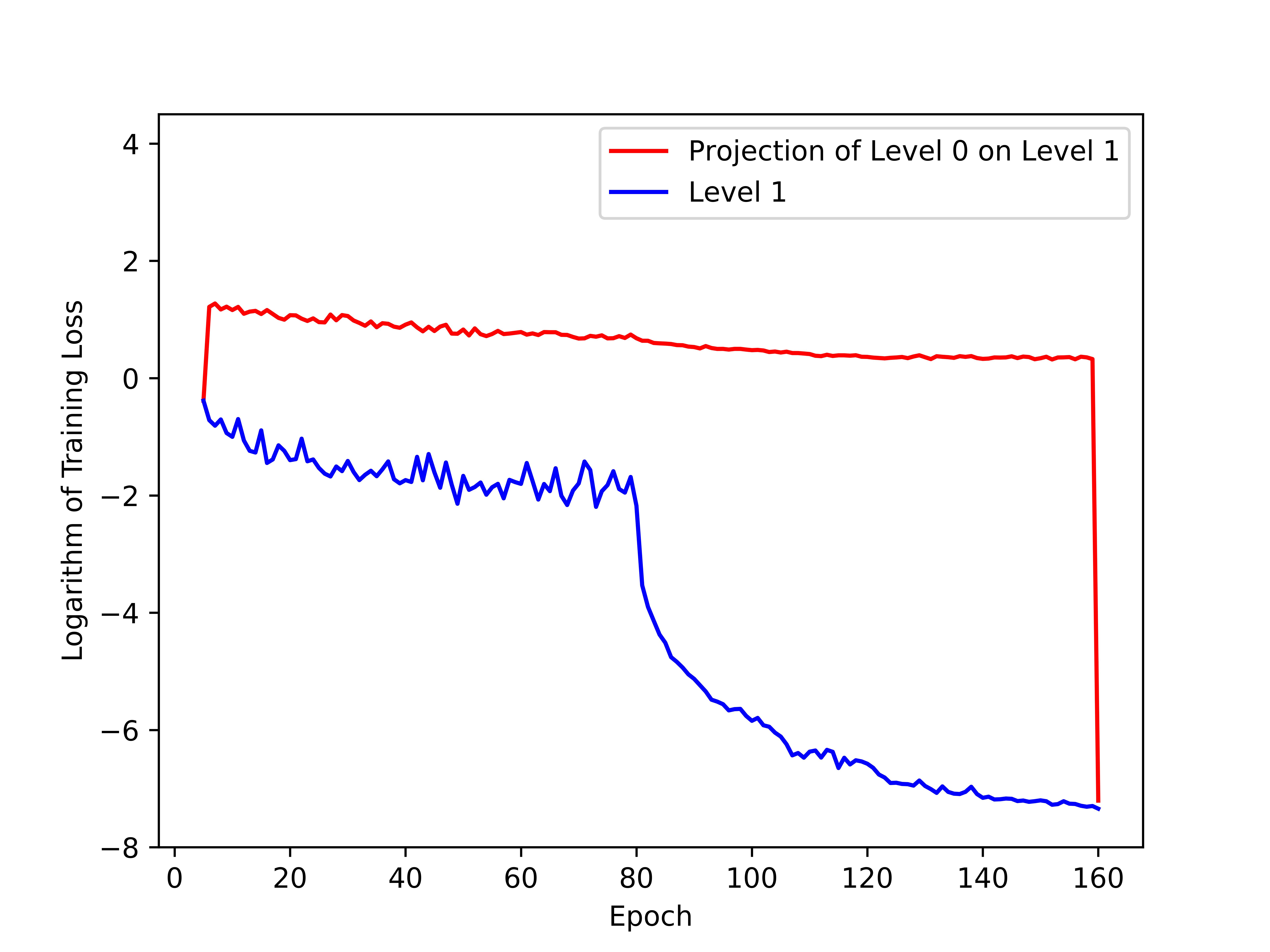}
\includegraphics[width=4cm, height=3.5cm]{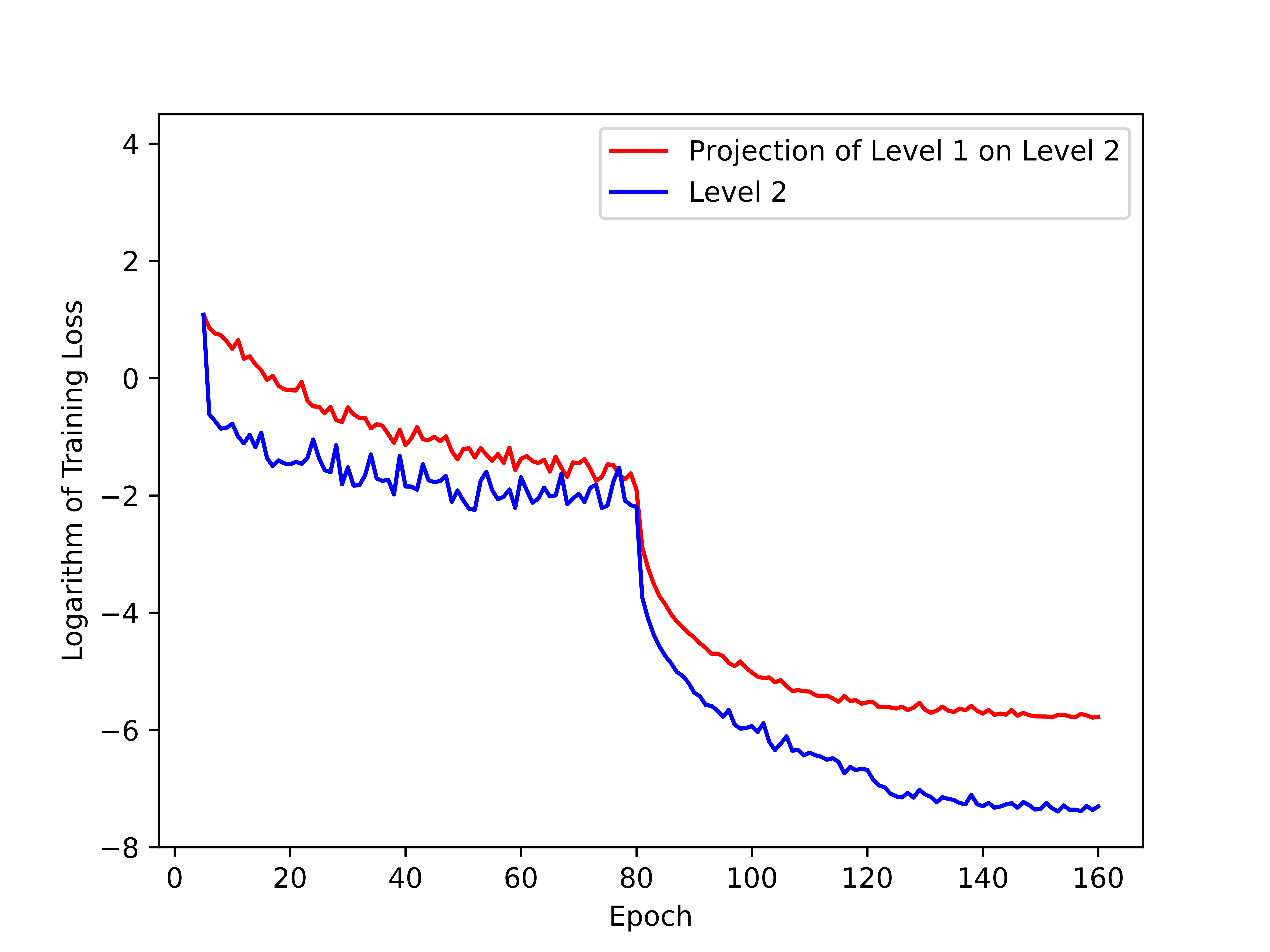} 
\includegraphics[width=4cm, height=3.5cm]{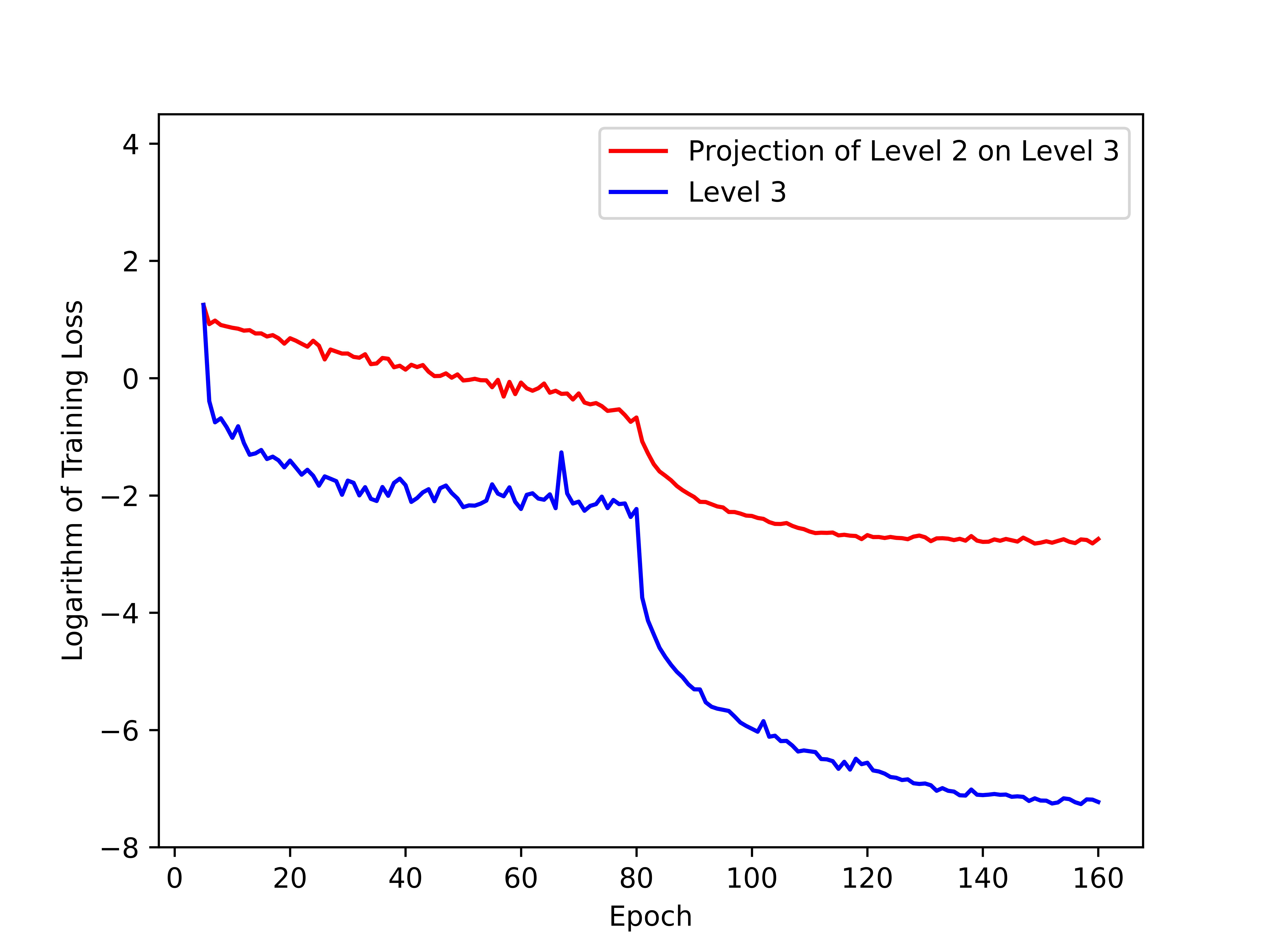} 
\includegraphics[width=4cm, height=3.5cm]{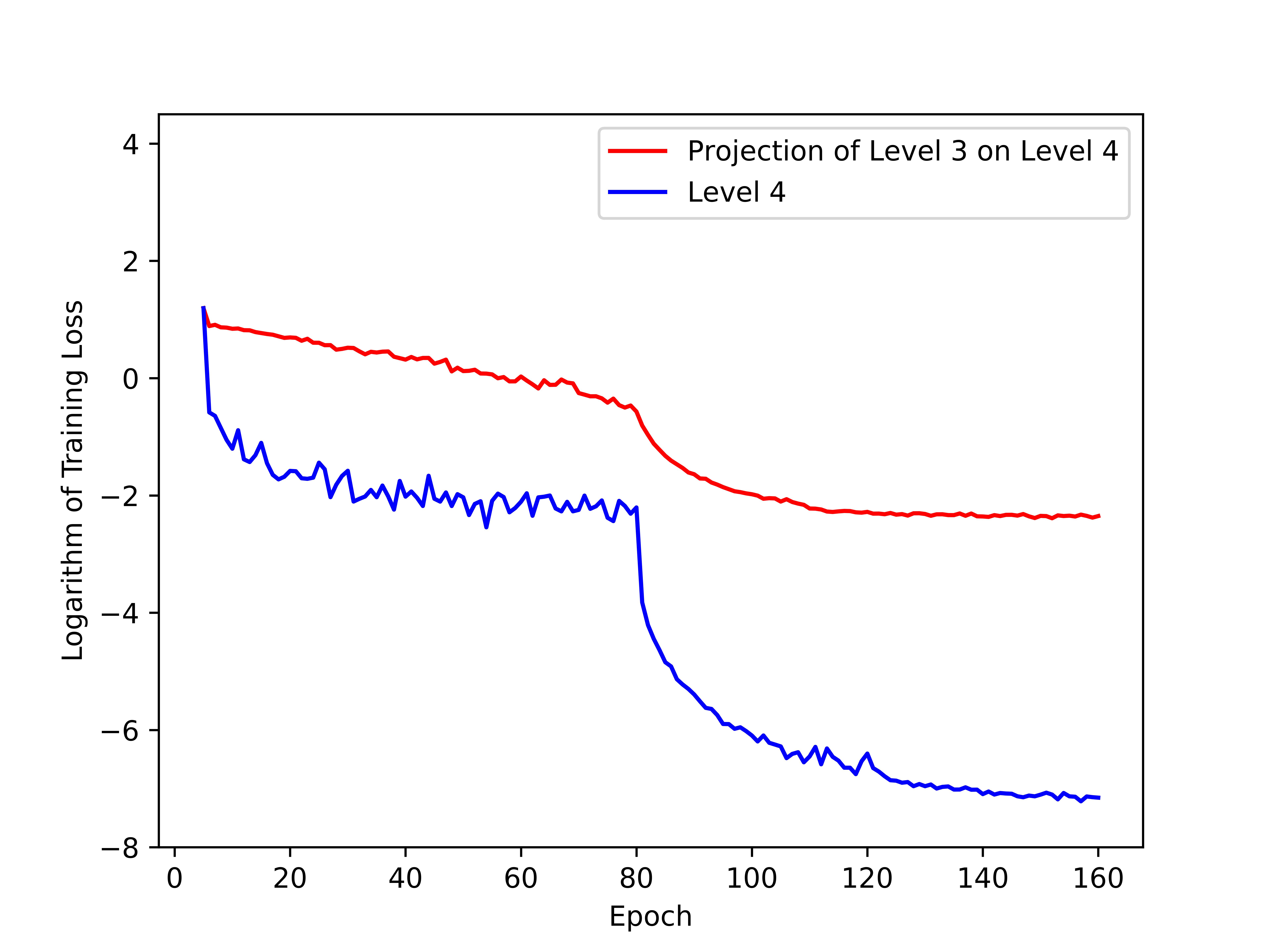} 
\includegraphics[width=4cm, height=3.5cm]{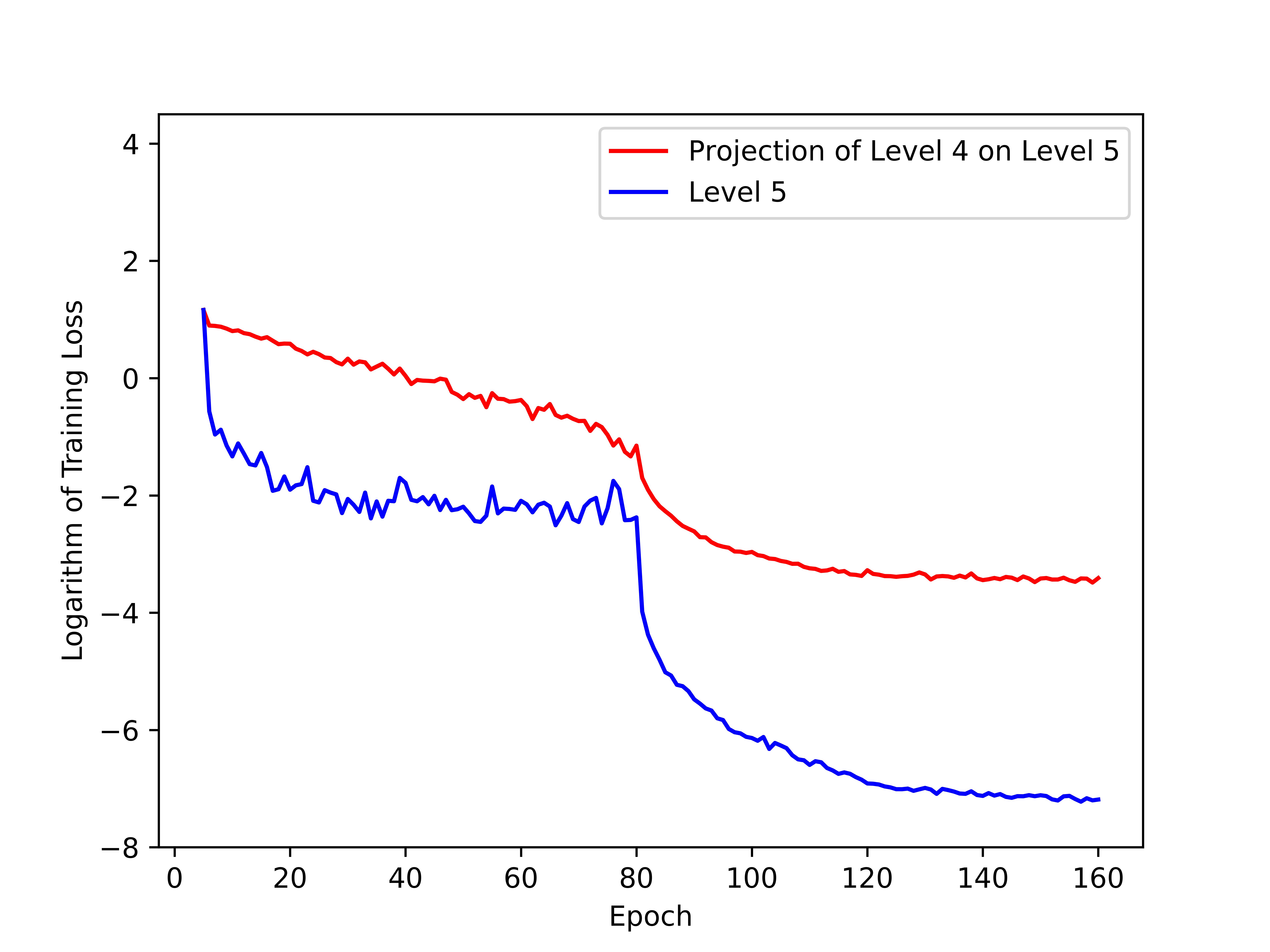} 
\includegraphics[width=4cm, height=3.5cm]{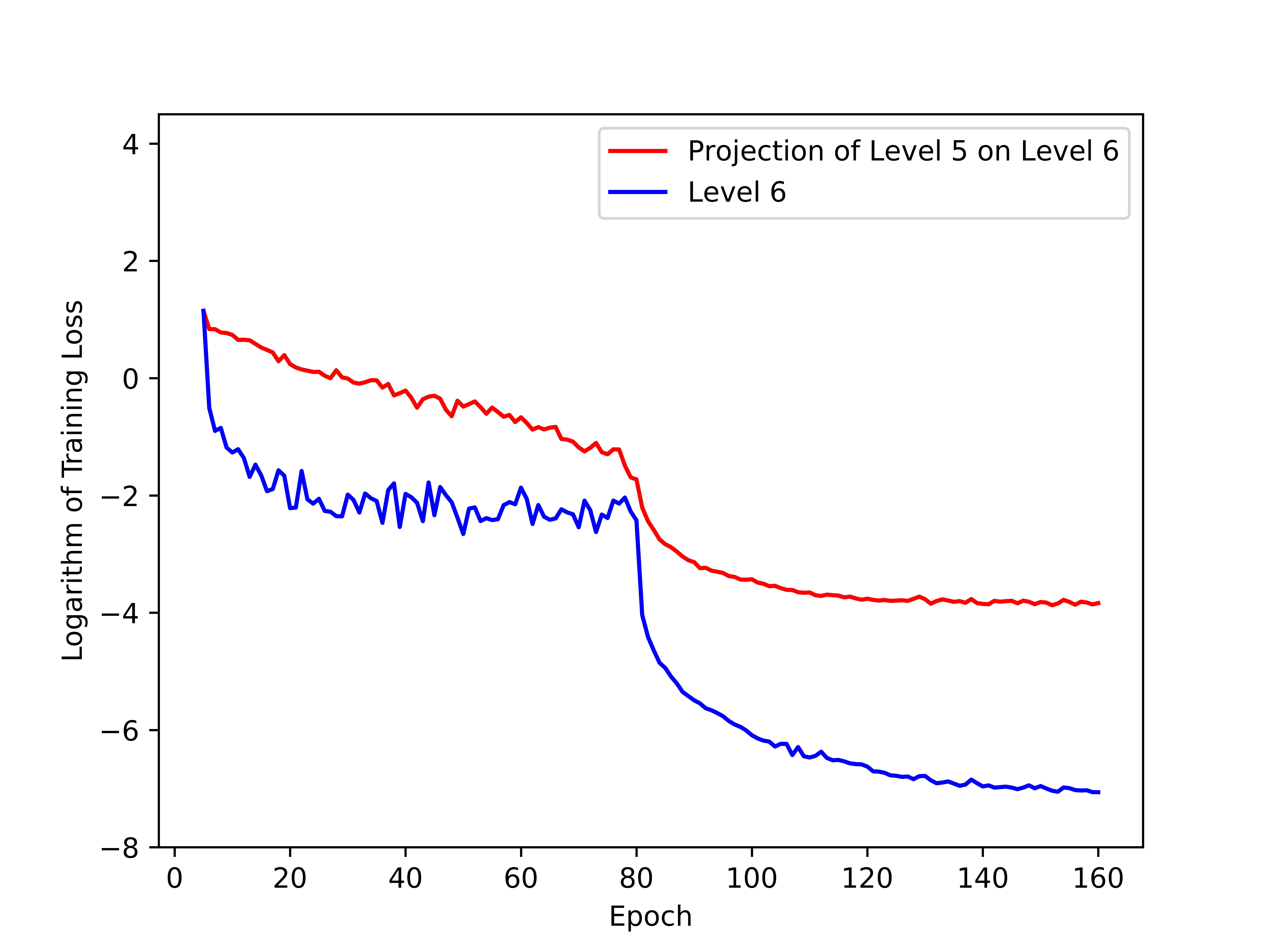} 
\includegraphics[width=4cm, height=3.5cm]{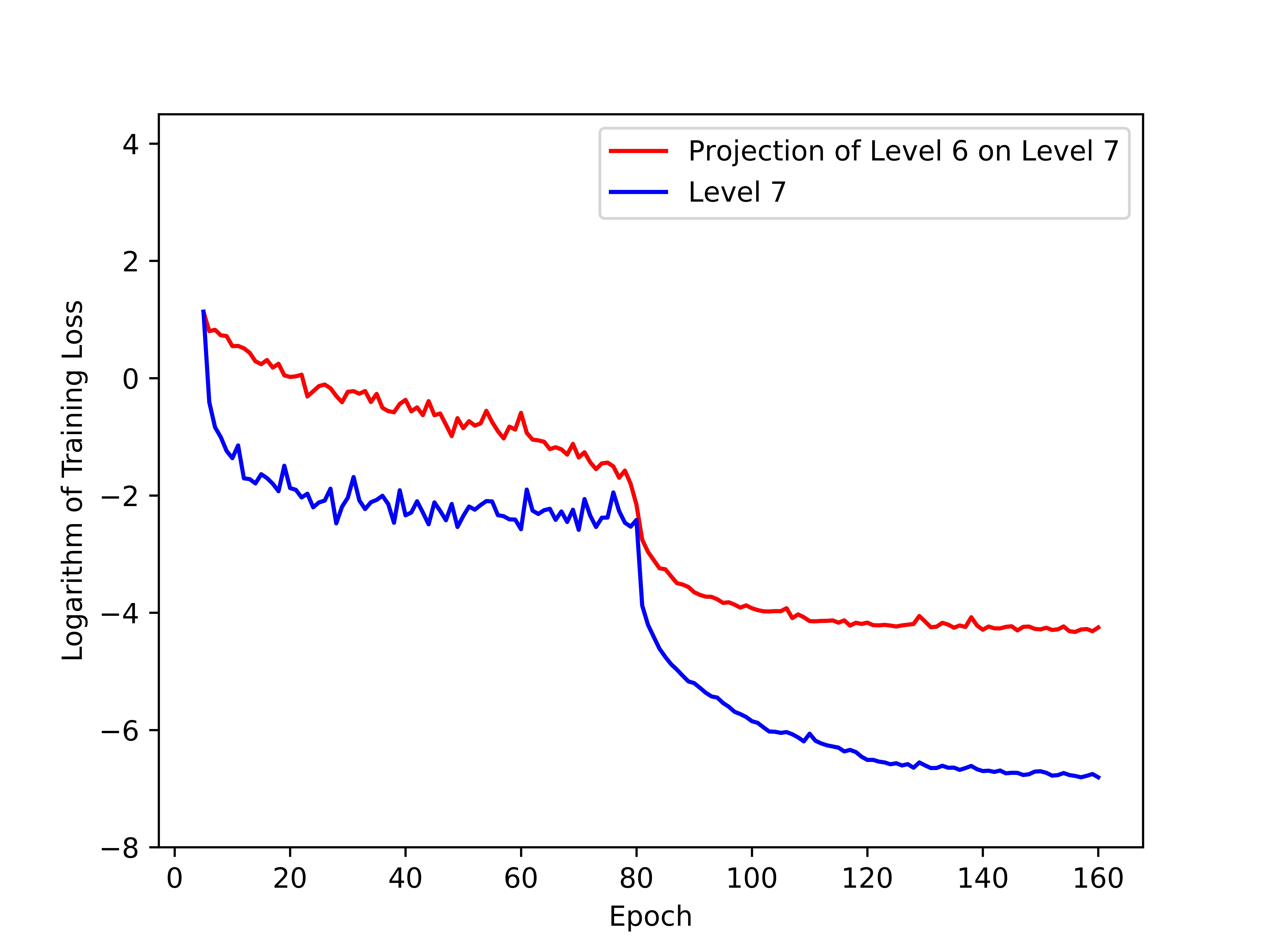} 
\includegraphics[width=4cm, height=3.5cm]{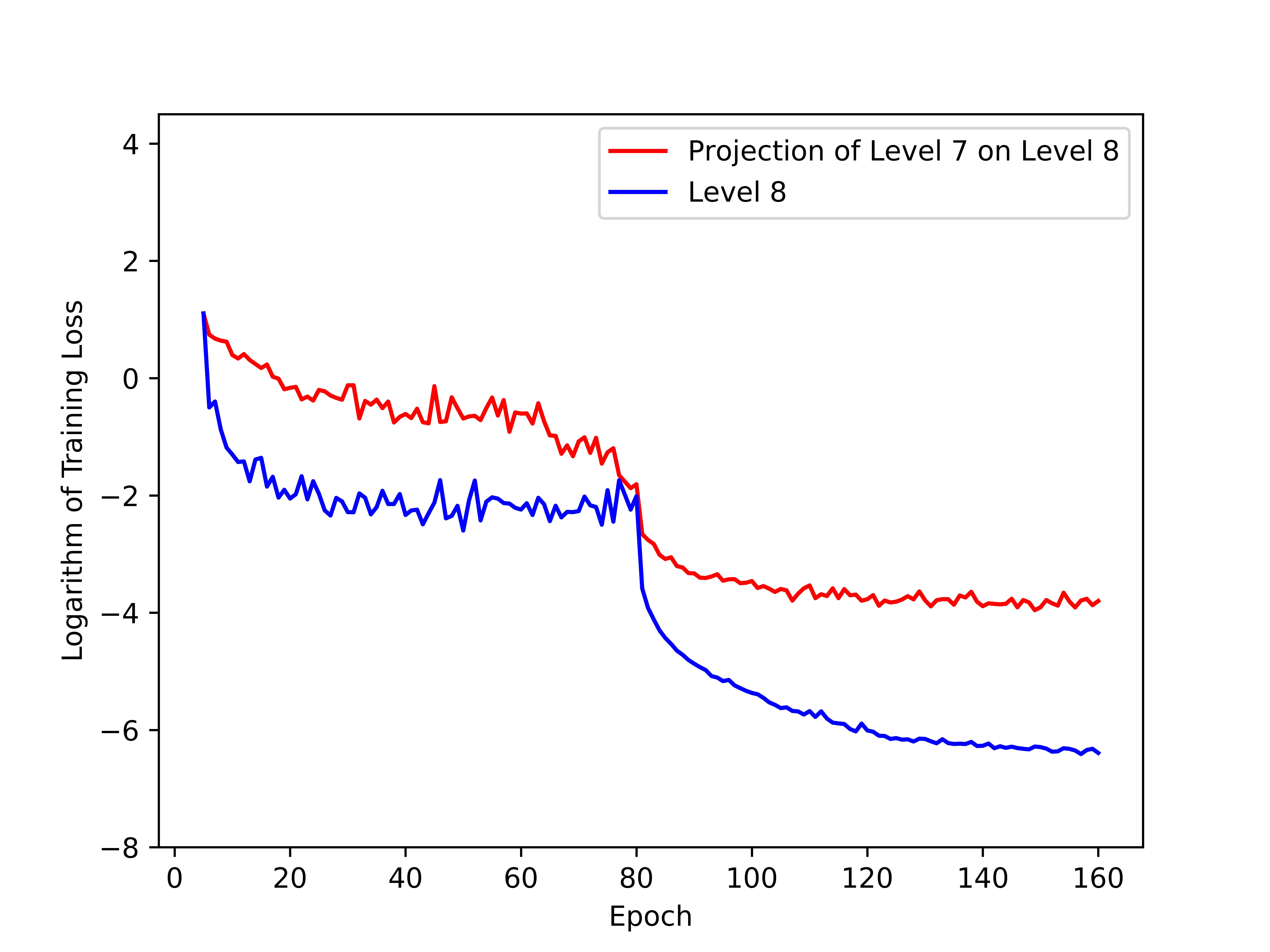} 
\includegraphics[width=4cm, height=3.5cm]{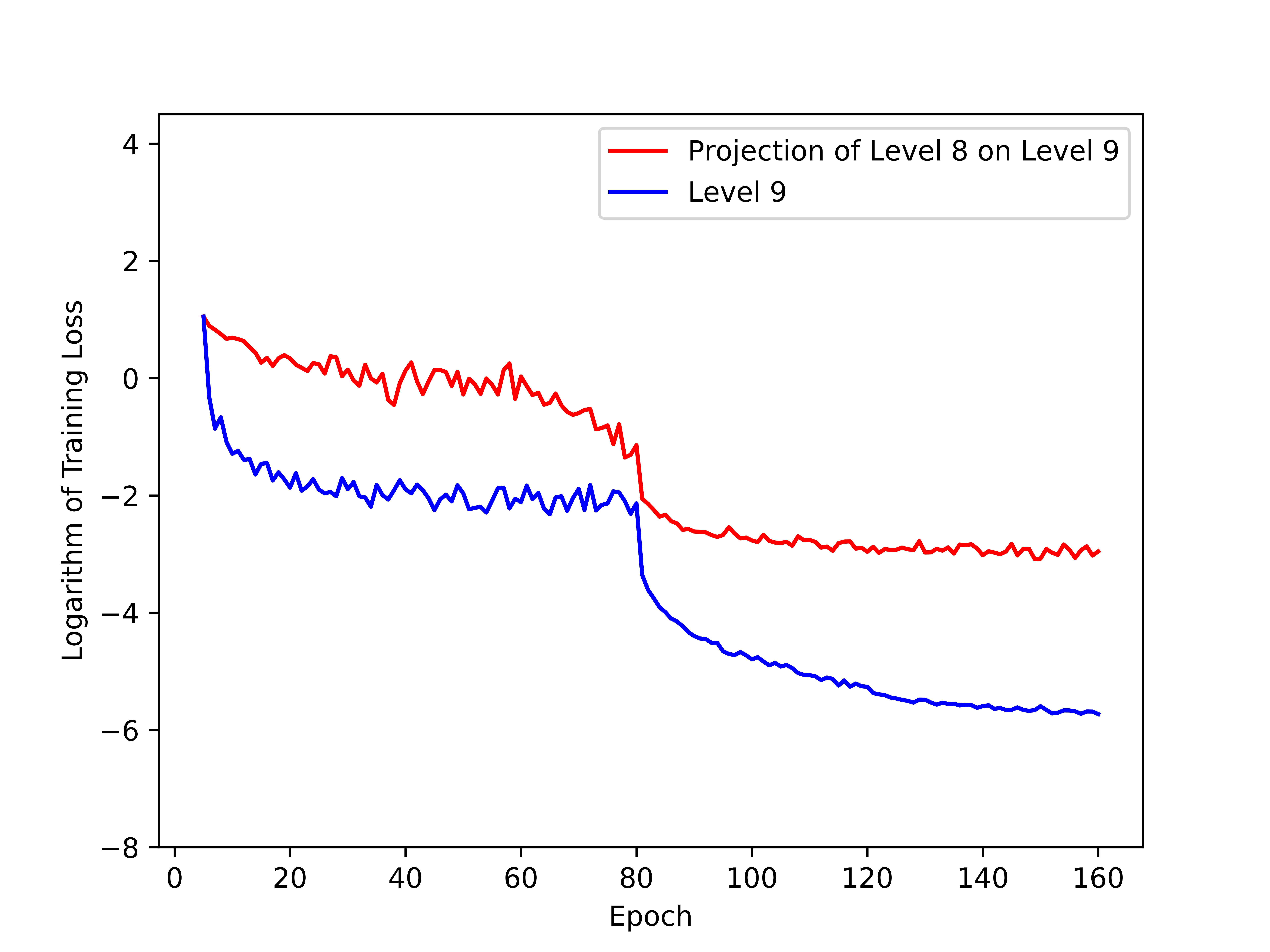}
\includegraphics[width=4cm, height=3.5cm]{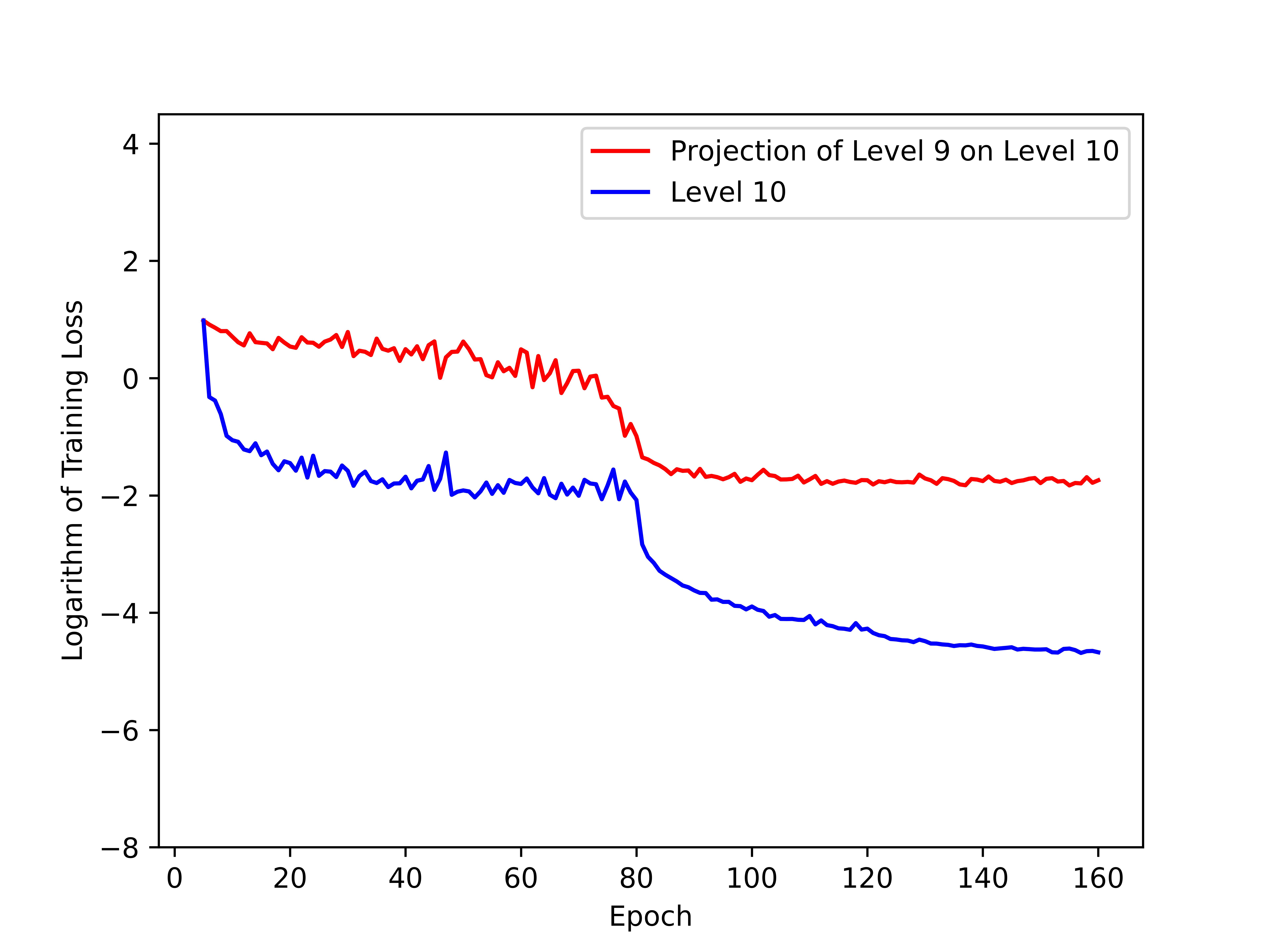} 
\includegraphics[width=4cm, height=3.5cm]{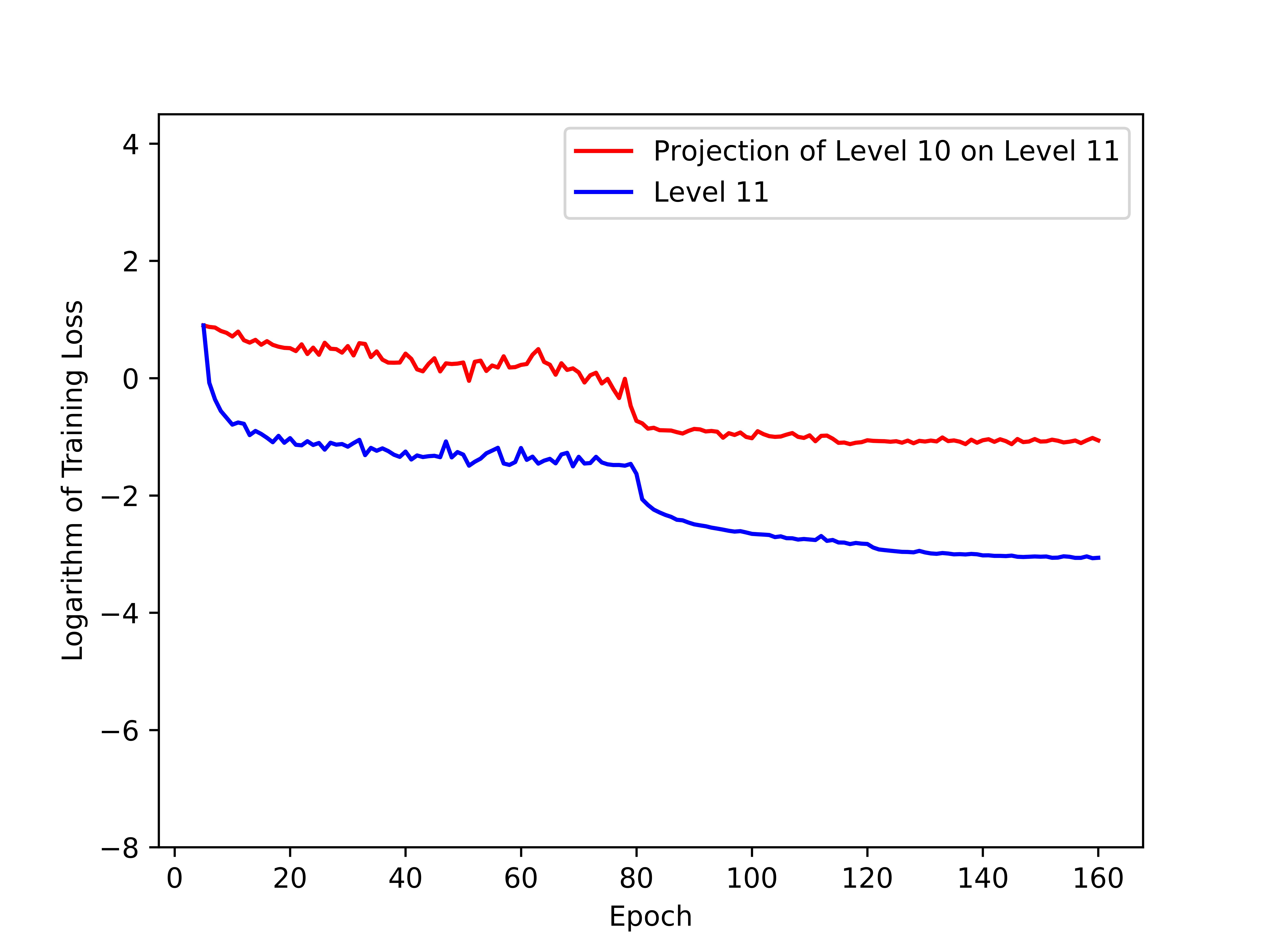}
\includegraphics[width=4cm, height=3.5cm]{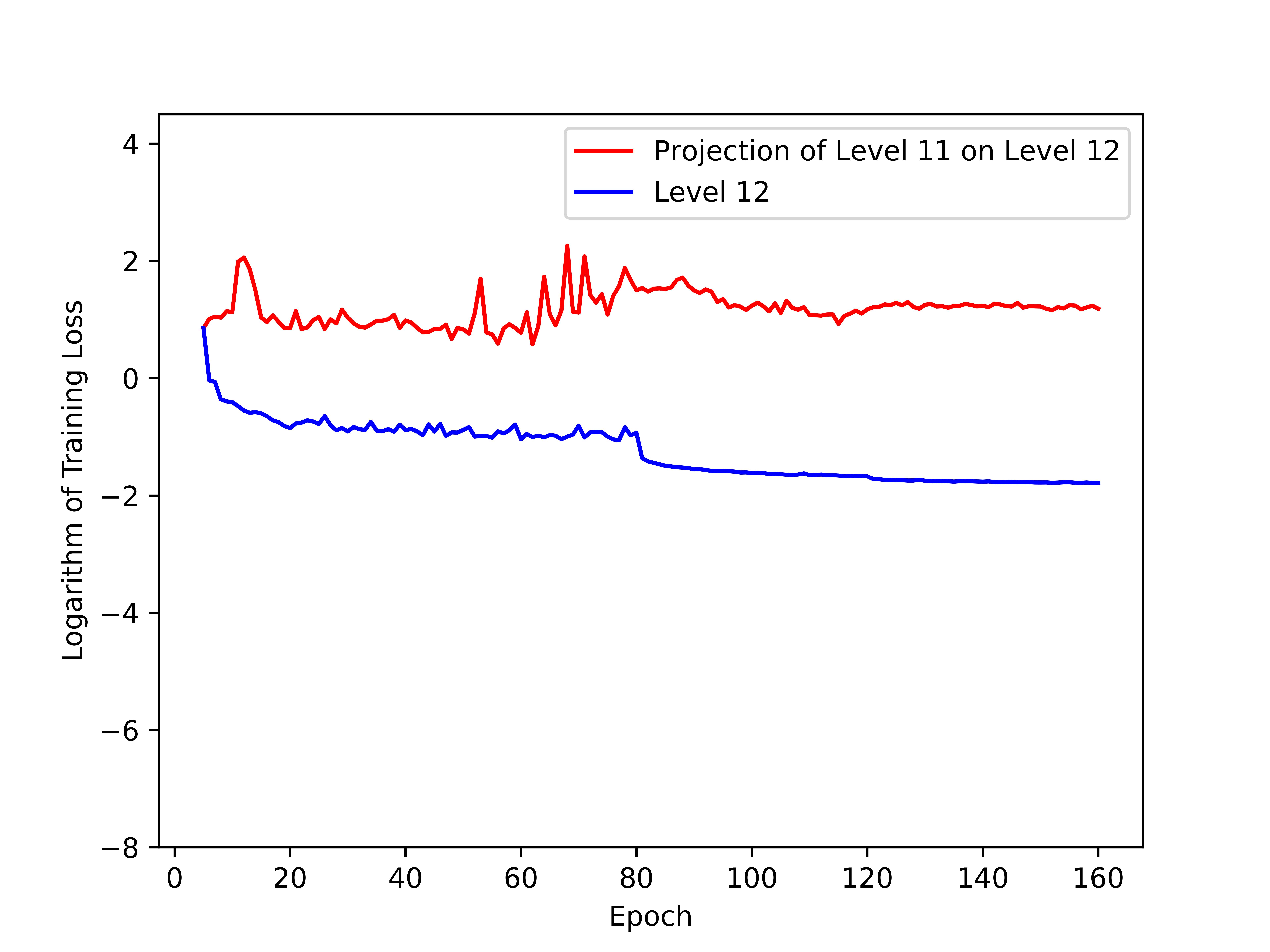} 
\caption{Comparison of the logarithm of training loss versus epoch between level ($L$) and level $(L-1)$ projected on level ($L$) in case of VGG-16. }
\end{figure*}
\subsection{Volume of the basin surrounding $W_{(L)}^{(min\_(L))}$ is larger than the volume of the basin surrounding $W^{Pr{(min\_(L-1))}}_{(L)}$.}
Fig. 26 presents the comparison of top-100 positive eigen values of the Hessian at $W_{(L)}^{(min\_(L))}$ and $W^{Pr{(min\_(L-1))}}_{(L)}$ for $L =$ \{$1,6,11$\}. As before, the figure shows that the eigen values of the Hessian at $W_{(L)}^{(min\_(L))}$ are smaller than that at $W^{Pr{(min\_(L-1))}}_{(L)}$, which means larger volume for the basin around $W_{(L)}^{(min\_(L))}$ than the volume of the basin around $W^{Pr{(min\_(L-1))}}_{(L)}$. This is also depicted in Table VI.
\begin{figure*}
\centering
\includegraphics[width=4cm, height=3.5cm]{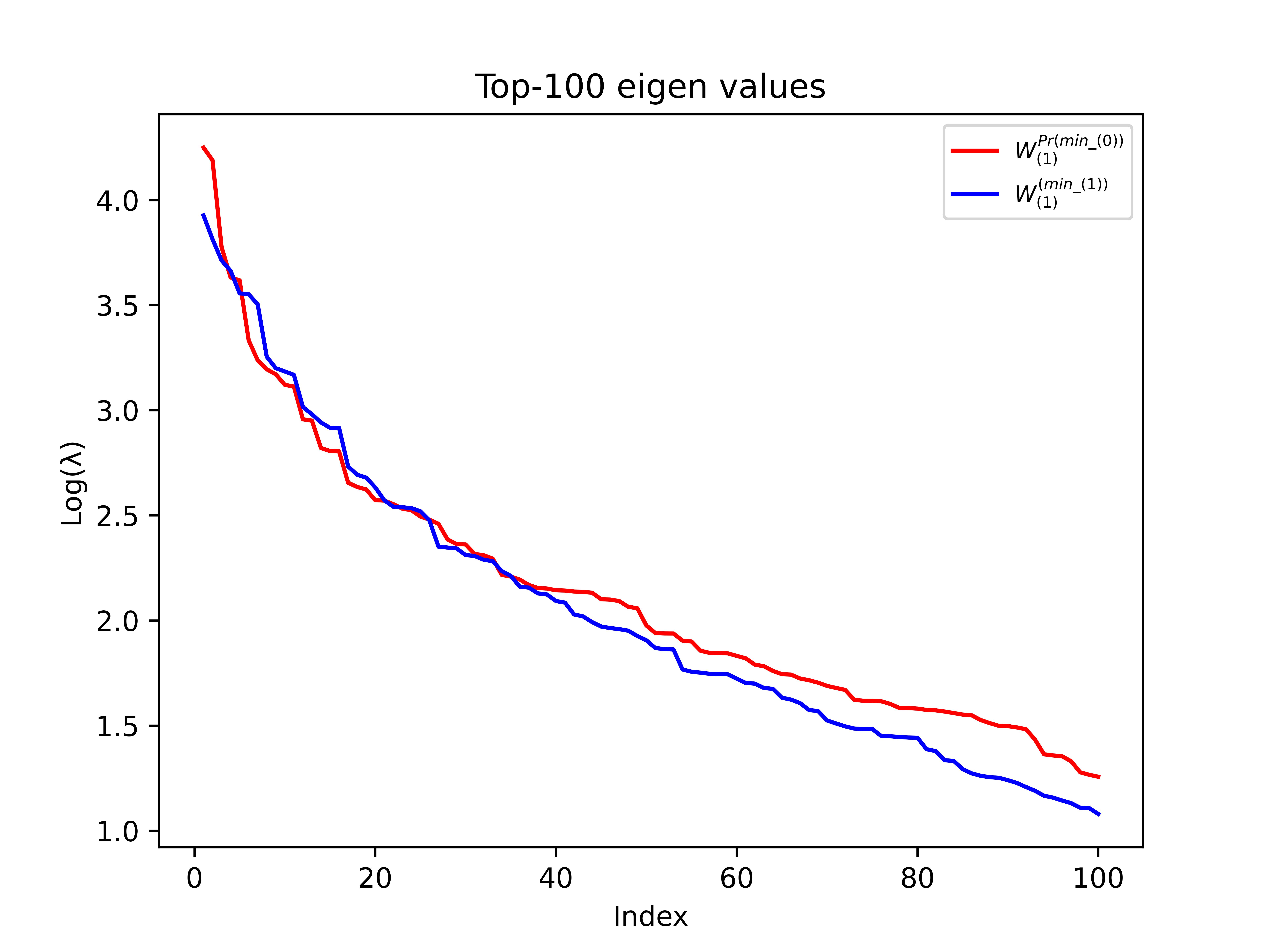} 
\includegraphics[width=4cm, height=3.5cm]{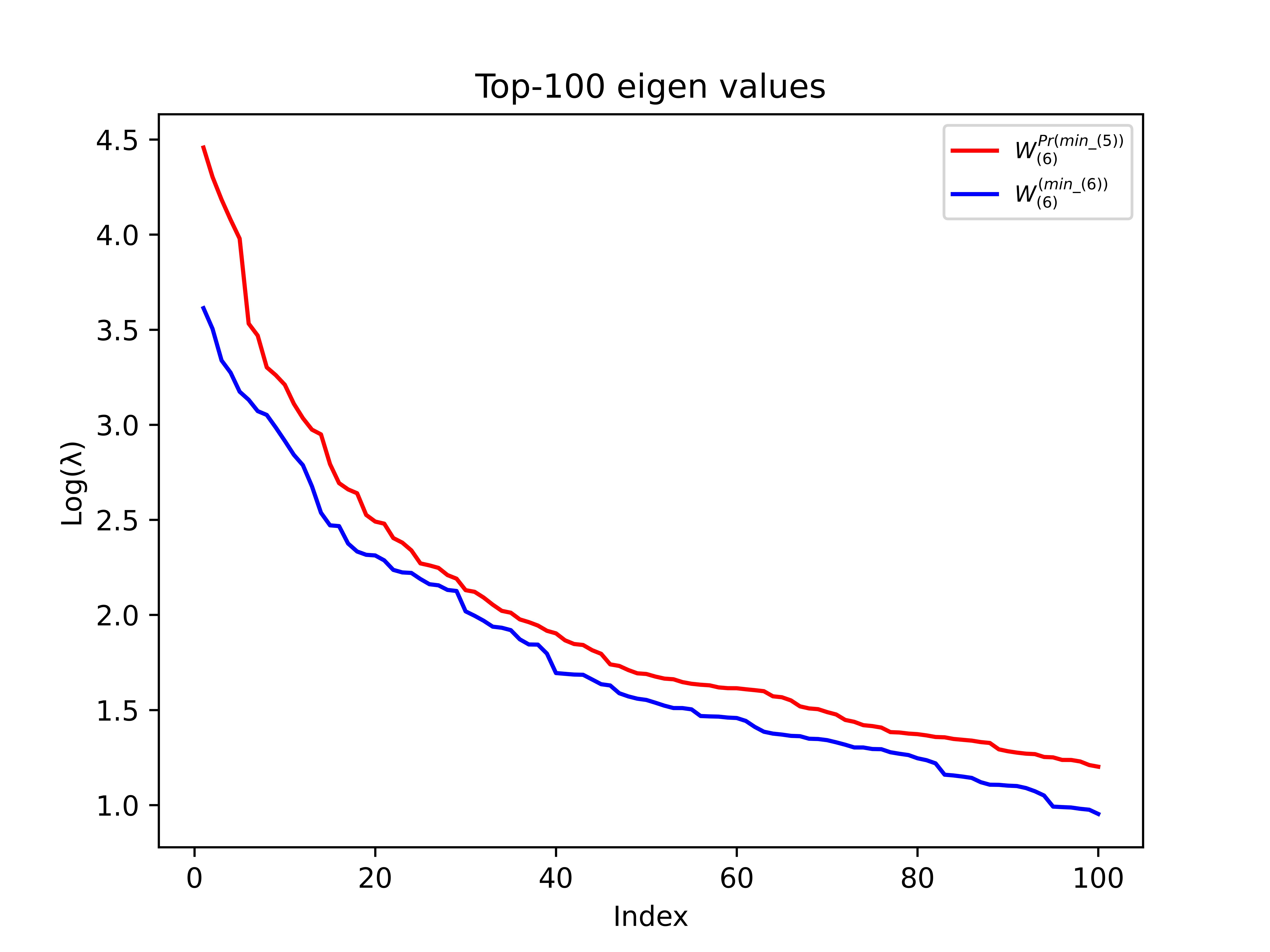} 
\includegraphics[width=4cm, height=3.5cm]{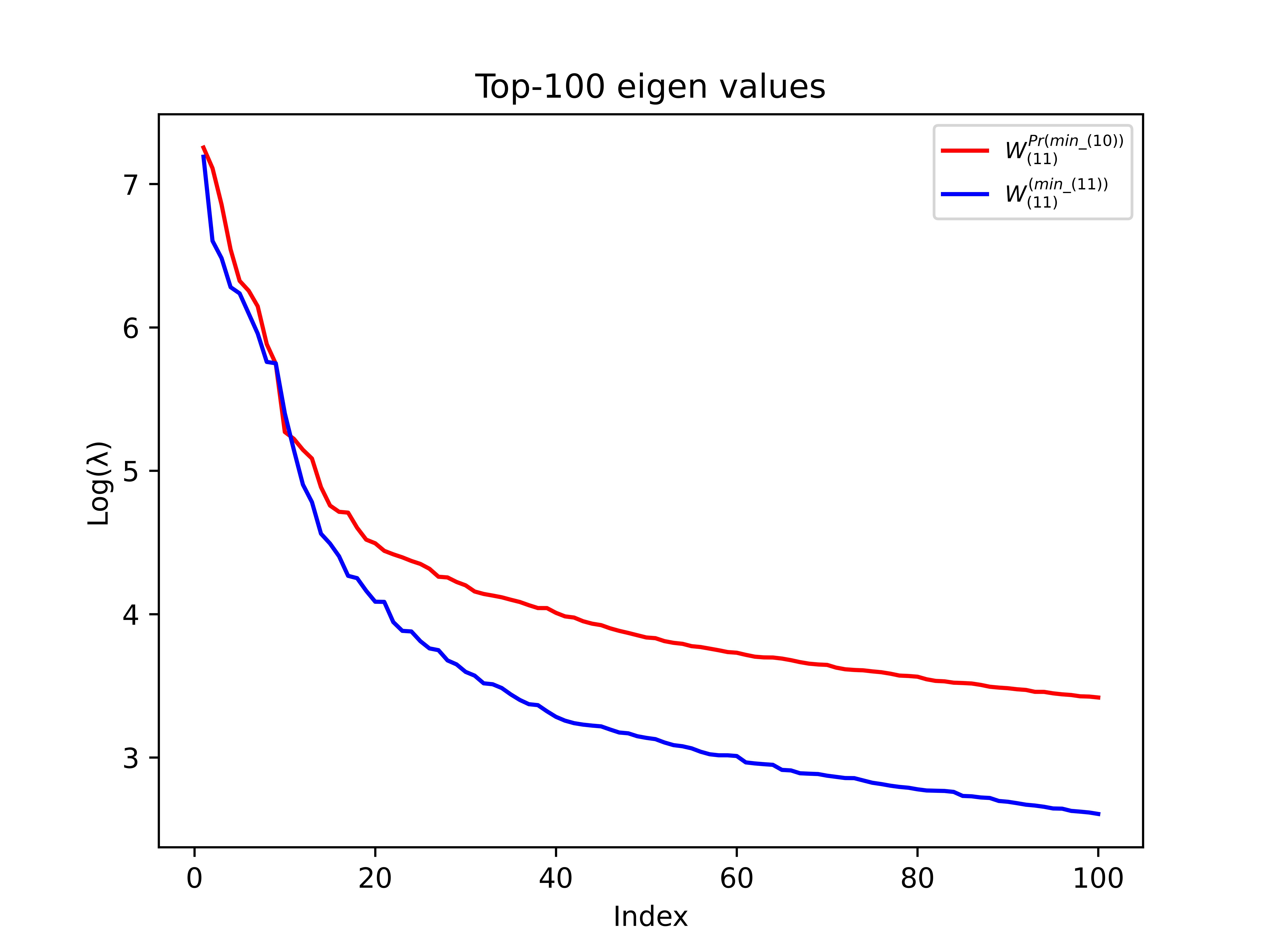} 
\caption{Comparison of top-100 positive eigen values of the Hessian at $W_{(L)}^{(min\_(L))}$ and $W^{Pr{(min\_(L-1))}}_{(L)}$ for $L =$ \{$1,6,11$\} in case of VGG-16.} 
\end{figure*}
\begin{table}[h!]
\caption{Comparison of inverse volume of basin, $V^{'} (100)$ at $W^{Pr{(min\_(L-1))}}_{(L)}$ and $W_{(L)}^{(min\_(L))}$ for $L =$ \{$1,6,11$\}.}
\begin{center}
\begin{tabular}{|p{0.03\textwidth} |  p{0.14\textwidth} | p{0.1\textwidth}|} 
  \hline
 $L$ &Solution & $V^{'} (100)$  \\ 
  \hline
  \multirow{2}{4em}{1} 
&$W^{Pr{(min\_(0))}}_{(1)}$ 
 &212.779
\\   \cline{2-3}
 &$W_{(1)}^{(min\_(1))}$
 &203.681\\
  \hline
  \multirow{2}{4em}{6}
  &$W^{Pr{(min\_(5))}}_{(6)}$ 
 & 198.106\\  \cline{2-3}
 &$W_{(6)}^{(min\_(6))}$
 & 177.233\\ 
  \hline 
 \multirow{2}{4em}{11}
  &$W^{Pr{(min\_(10))}}_{(11)}$
 &414.844\\  \cline{2-3}
  &$W_{(11)}^{(min\_(11))}$
 & 352.86\\ 
  \hline
\end{tabular}
\label{product}
\end{center}
\end{table}
\subsection{Volume of the basin surrounding $W_{(L-1)}^{(min\_(L-1))}$ is larger than the volume of the basin surrounding $W^{RPr{(min\_(L))}}_{(L-1)}$.}
Fig. 27 presents the comparison of top-100 positive eigen values of the Hessian at $W_{(L-1)}^{(min\_(L-1))}$ and $W^{RPr{(min\_(L))}}_{(L-1)}$ for $L =$ \{$1,6,11$\}. The figure shows that the eigen values of the Hessian at $W_{(L-1)}^{(min\_(L-1))}$ are smaller than that at $W^{RPr{(min\_(L))}}_{(L-1)}$, which means larger volume for the basin around $W_{(L-1)}^{(min\_(L-1))}$ than that of the basin around $W^{RPr{(min\_(L))}}_{(L-1)}$. This is also depicted in table VII.
\begin{figure*}
\centering
\includegraphics[width=4cm, height=3.5cm]{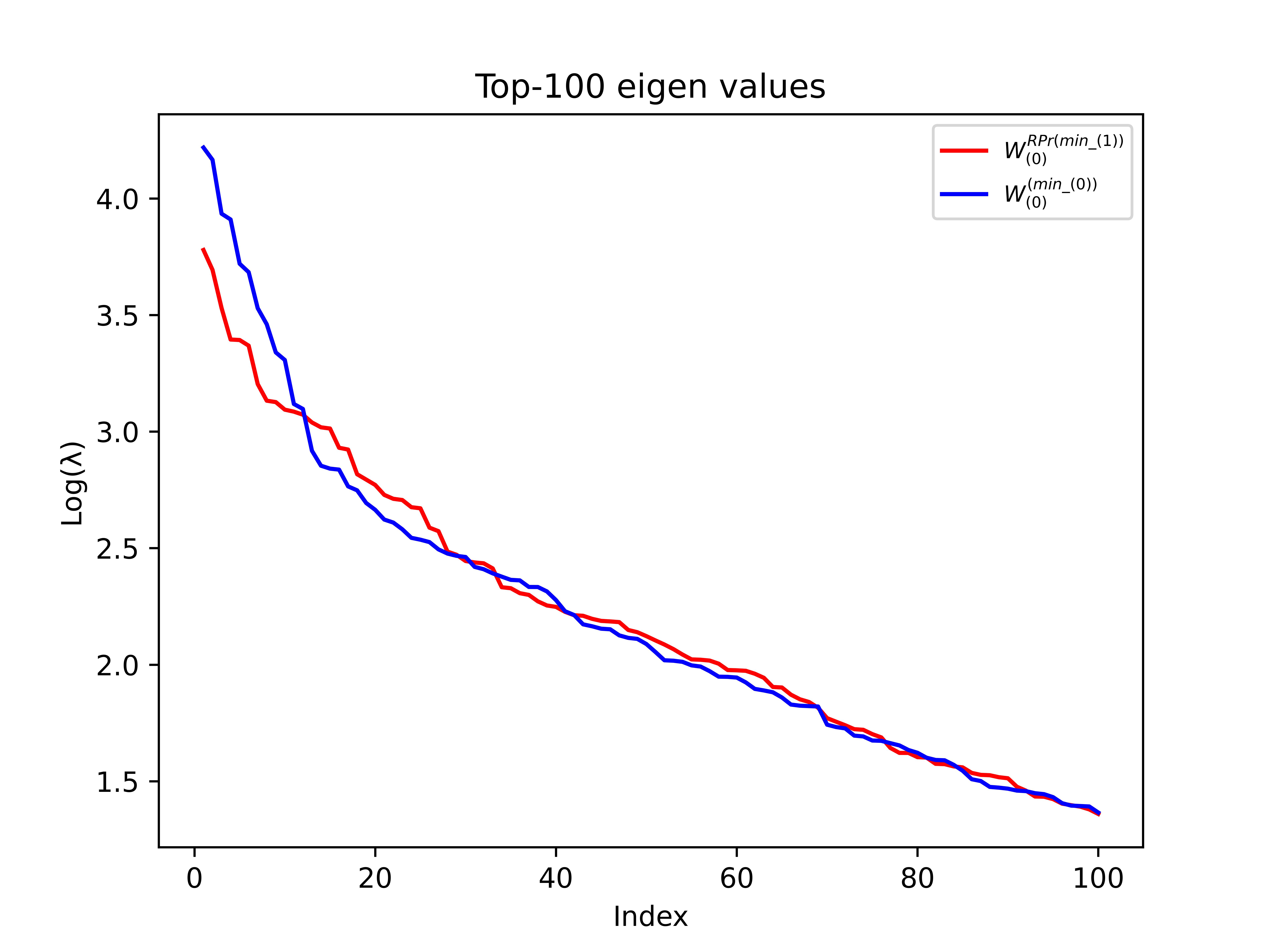} 
\includegraphics[width=4cm, height=3.5cm]{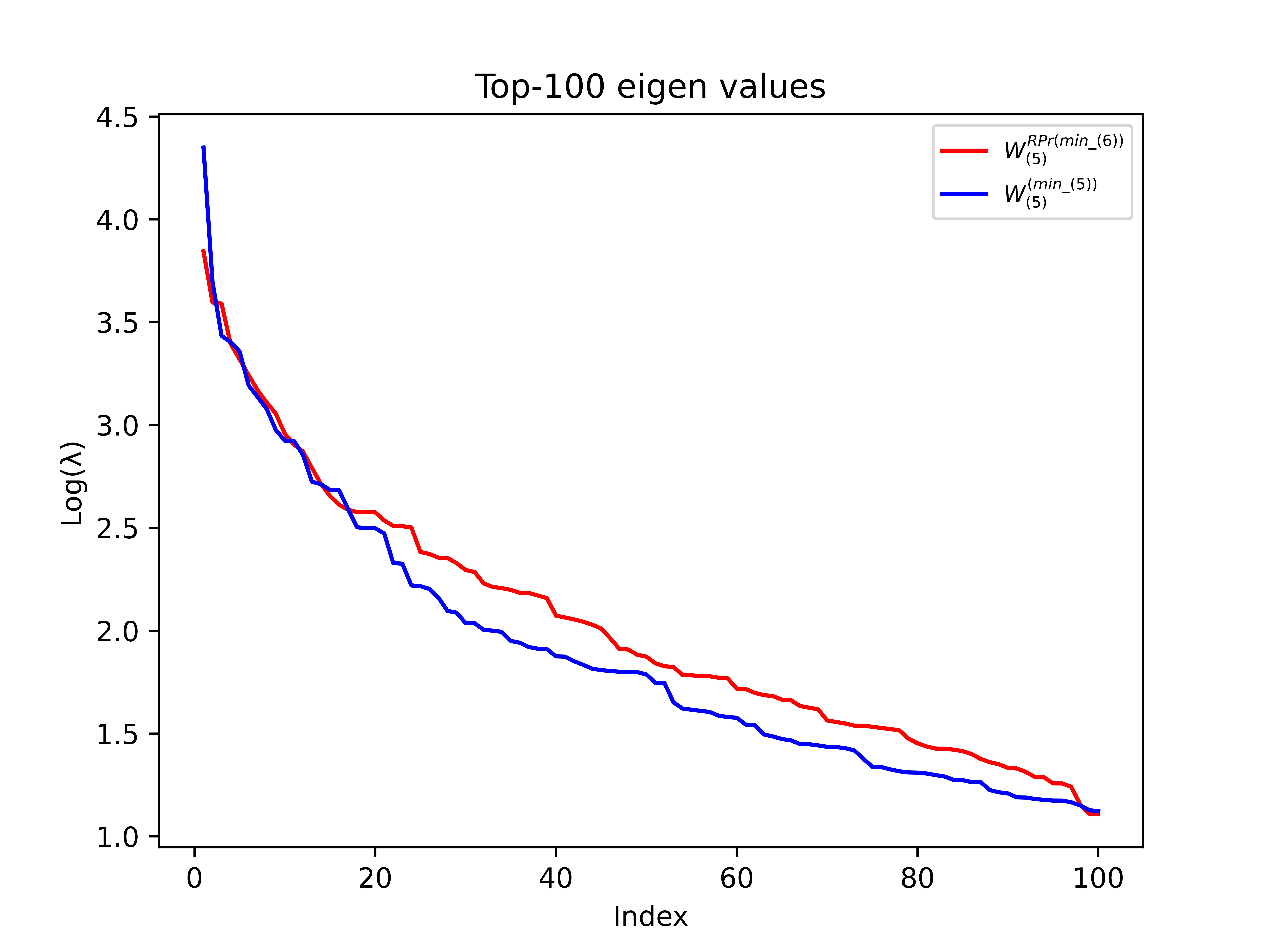} 
\includegraphics[width=4cm, height=3.5cm]{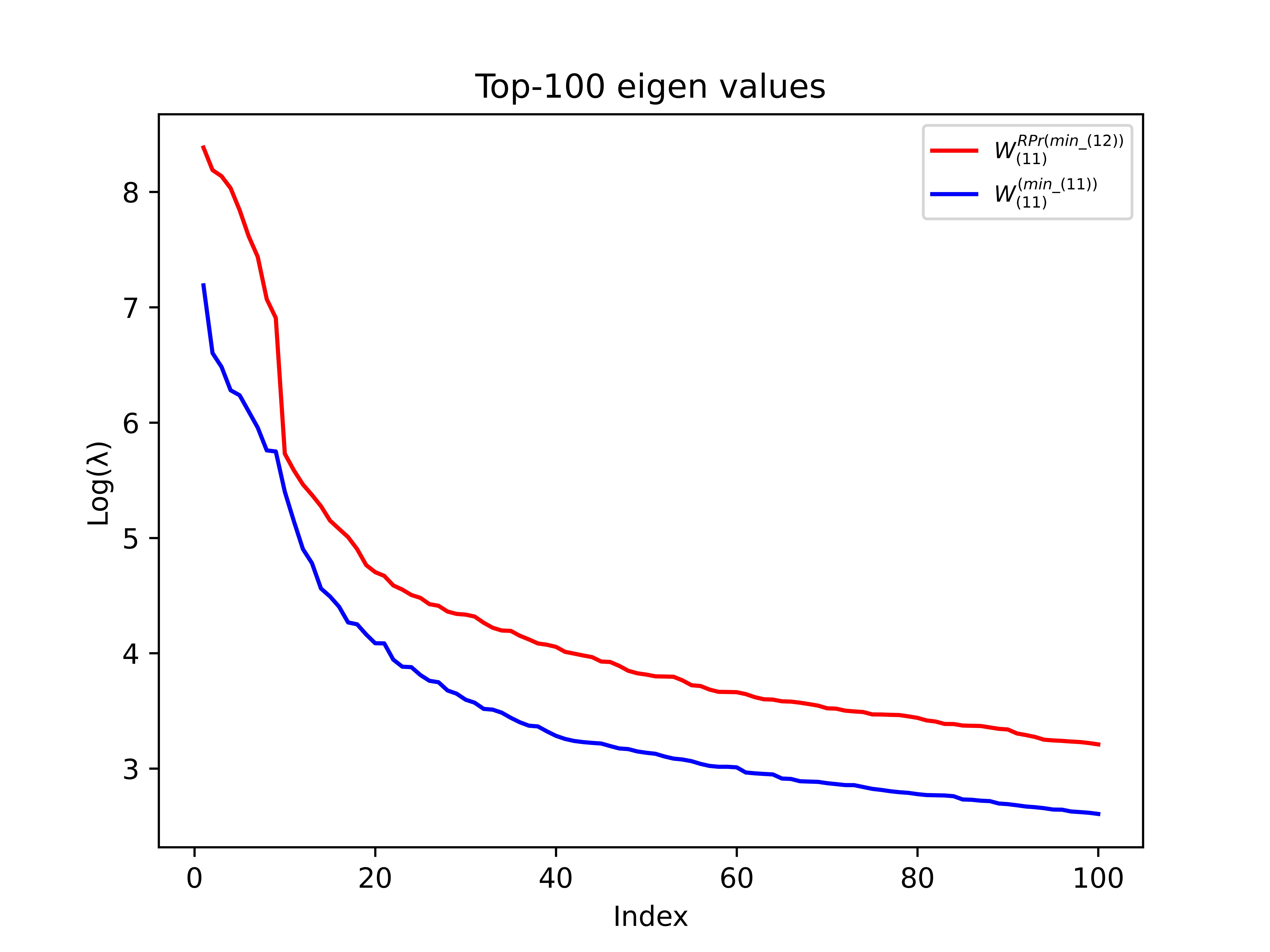} 
{\caption{Comparison of top-100 positive eigen values of the Hessian at $W_{(L-1)}^{(min\_(L-1))}$ and $W^{RPr{(min\_(L))}}_{(L-1)}$ for $L =$ \{$1,6,12$\} in case of VGG-16.} }
\end{figure*}
\begin{table}[h!]
\caption{Comparison of $V^{'} (100)$ at $W^{RPr{(min\_(L))}}_{(L-1)}$ and $W_{(L-1)}^{(min\_(L-1))}$ for $L =$ \{$1,6,12$\} in case of VGG-16.}
\begin{center}
\begin{tabular}{|p{0.03\textwidth} | p{0.14\textwidth} | p{0.1\textwidth}|} 
  \hline
 $L$ &Solution& $V^{'} (100)$  \\ 
  \hline
  \multirow{2}{4em}{1} 
&$W^{RPr{(min\_(1))}}_{(0)}$ 
 &219.996\\  \cline{2-3}
 &$W_{(0)}^{(min\_(0))}$
 & 220.915\\
  \hline
  \multirow{2}{4em}{6}
  &$W^{RPr{(min\_(6))}}_{(5)}$
 & 201.835\\  \cline{2-3}
 &$W_{(5)}^{(min\_(5))}$
 &189.044\\ 
  \hline 
 \multirow{2}{4em}{12}
  &$W^{RPr{(min\_(12))}}_{(11)}$
 &426.294\\  \cline{2-3}
  &$W_{(11)}^{(min\_(11))}$
 &352.86\\ 
  \hline
\end{tabular}
\label{product}
\end{center}
\end{table}
\subsection{There exists a barrier between the IMP solutions at successive levels in the loss landscape.}
Fig. 28 presents the training loss along a straight line connecting $W^{(min\_(L-1))}_{(L-1)}$ and $W^{(min\_(L))}_{(L)}$. The x-axis represents the interpolation co-efficient $\alpha$. Each plot depicts the training loss at $501$ points between $W^{(min\_(L-1))}_{(L-1)}$ and $W^{(min\_(L))}_{(L)}$. The barriers between successive minima are clearly visible.
\begin{figure*}
\centering
\includegraphics[width=4cm, height=3.5cm]{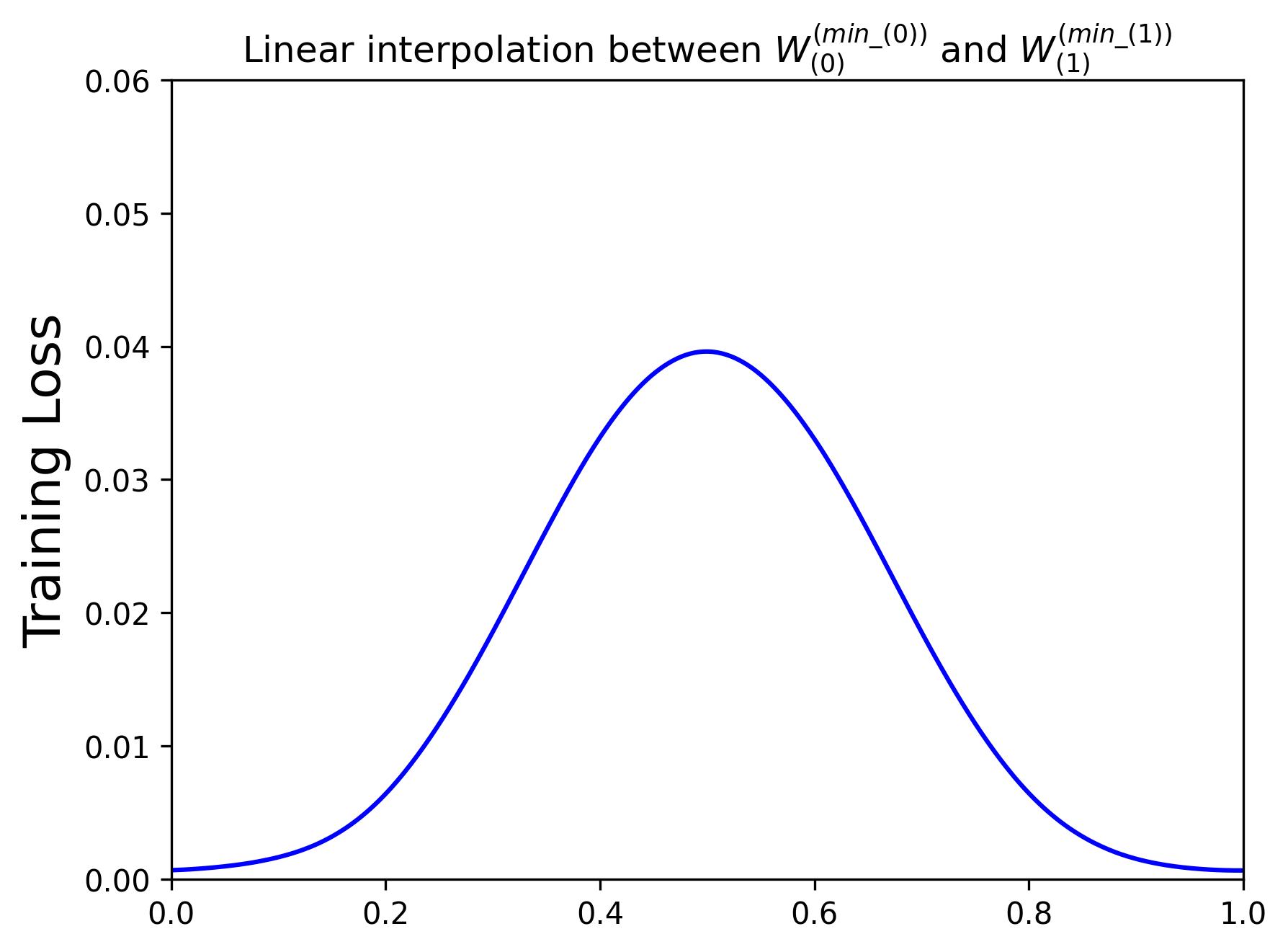}
\includegraphics[width=4cm, height=3.5cm]{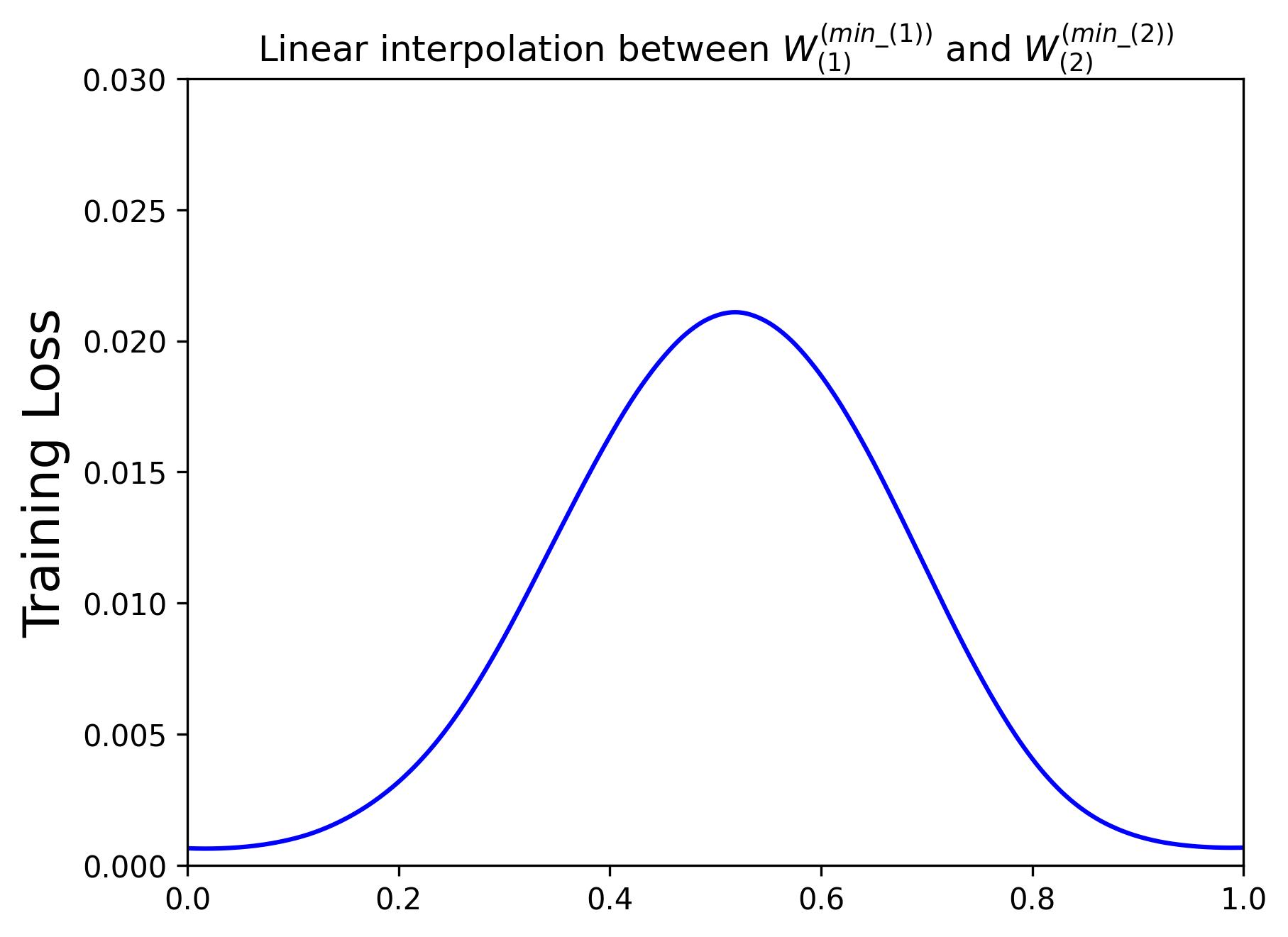} 
\includegraphics[width=4cm, height=3.5cm]{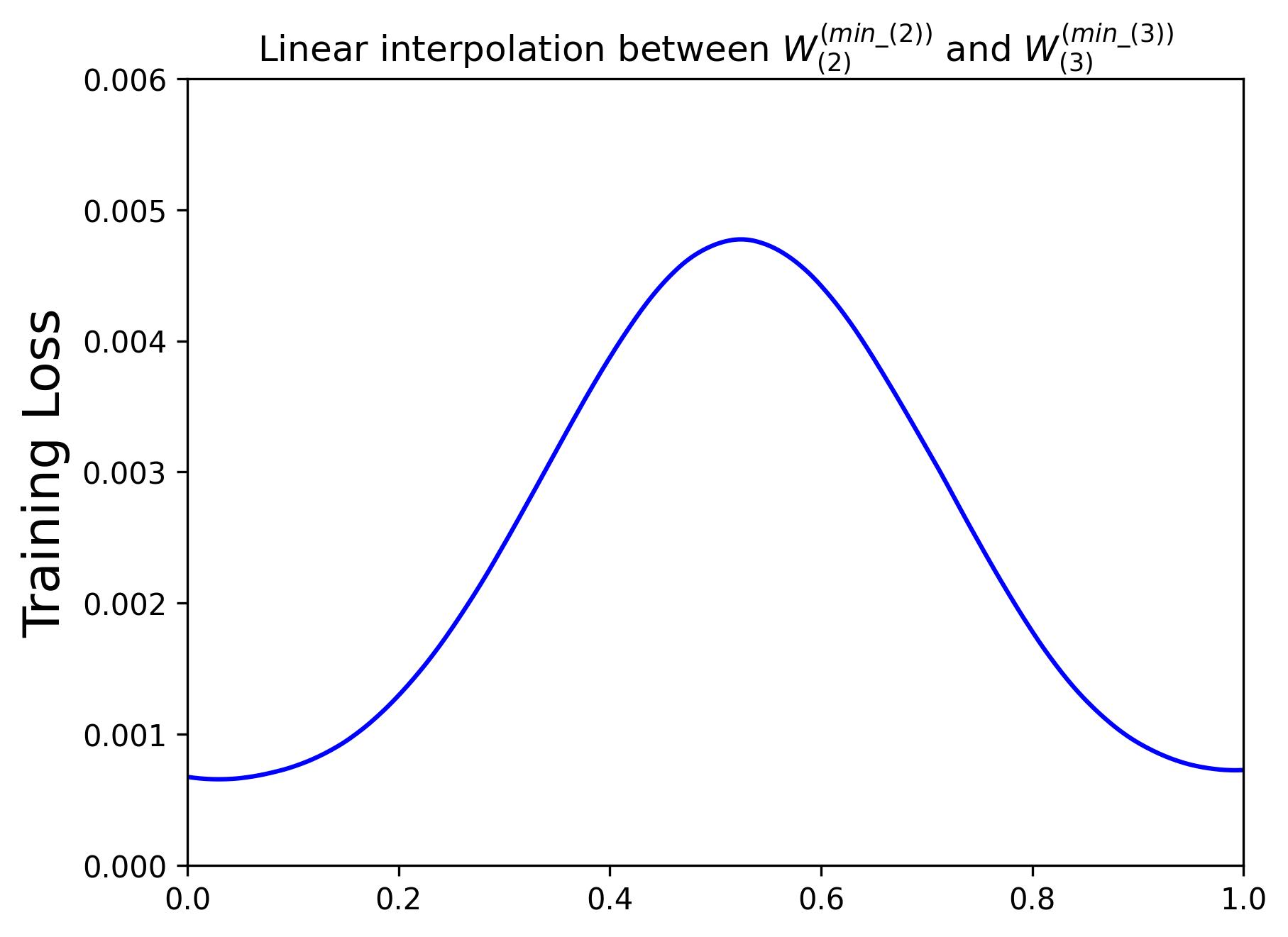} 
\includegraphics[width=4cm, height=3.5cm]{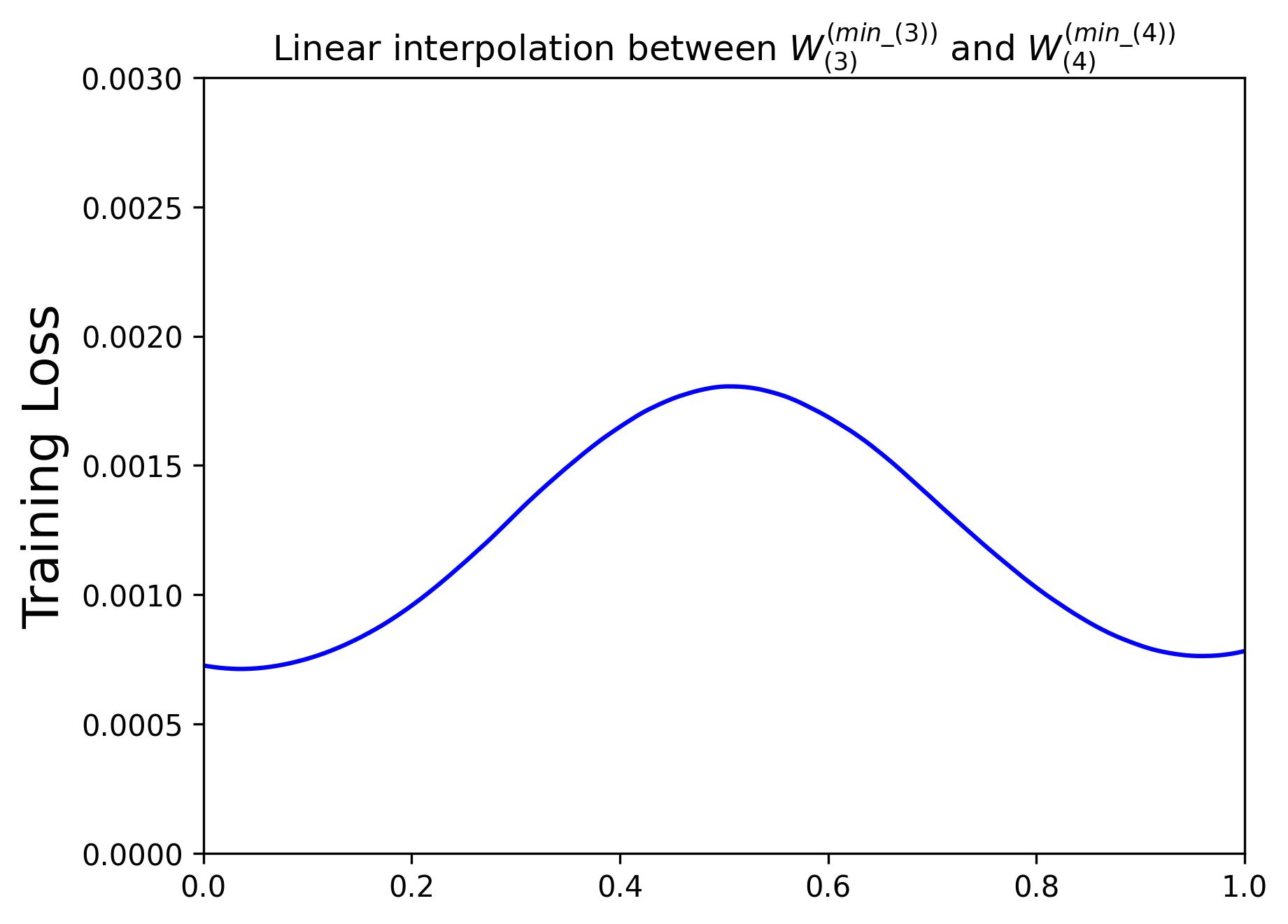} 
\includegraphics[width=4cm, height=3.5cm]{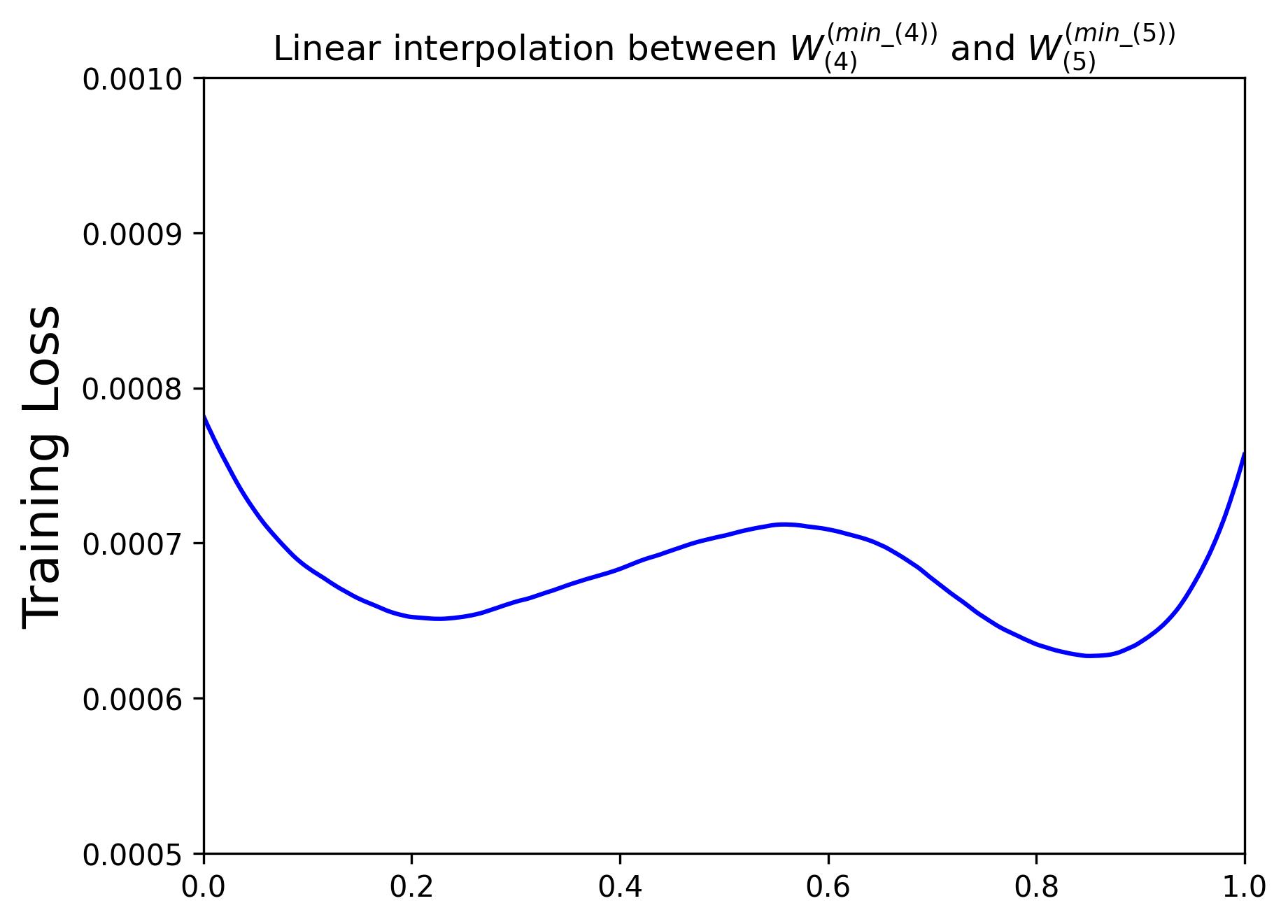} 
\includegraphics[width=4cm, height=3.5cm]{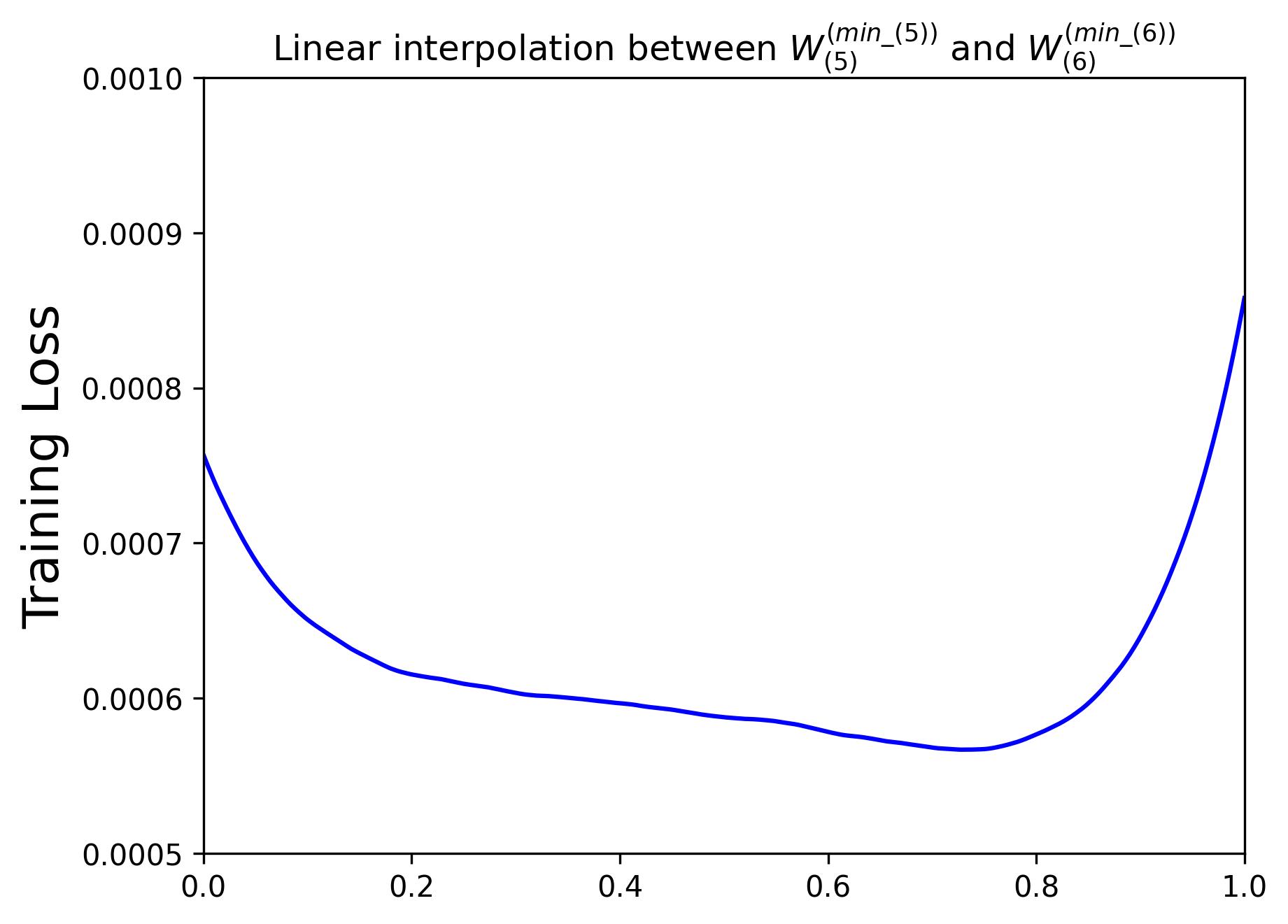} 
\includegraphics[width=4cm, height=3.5cm]{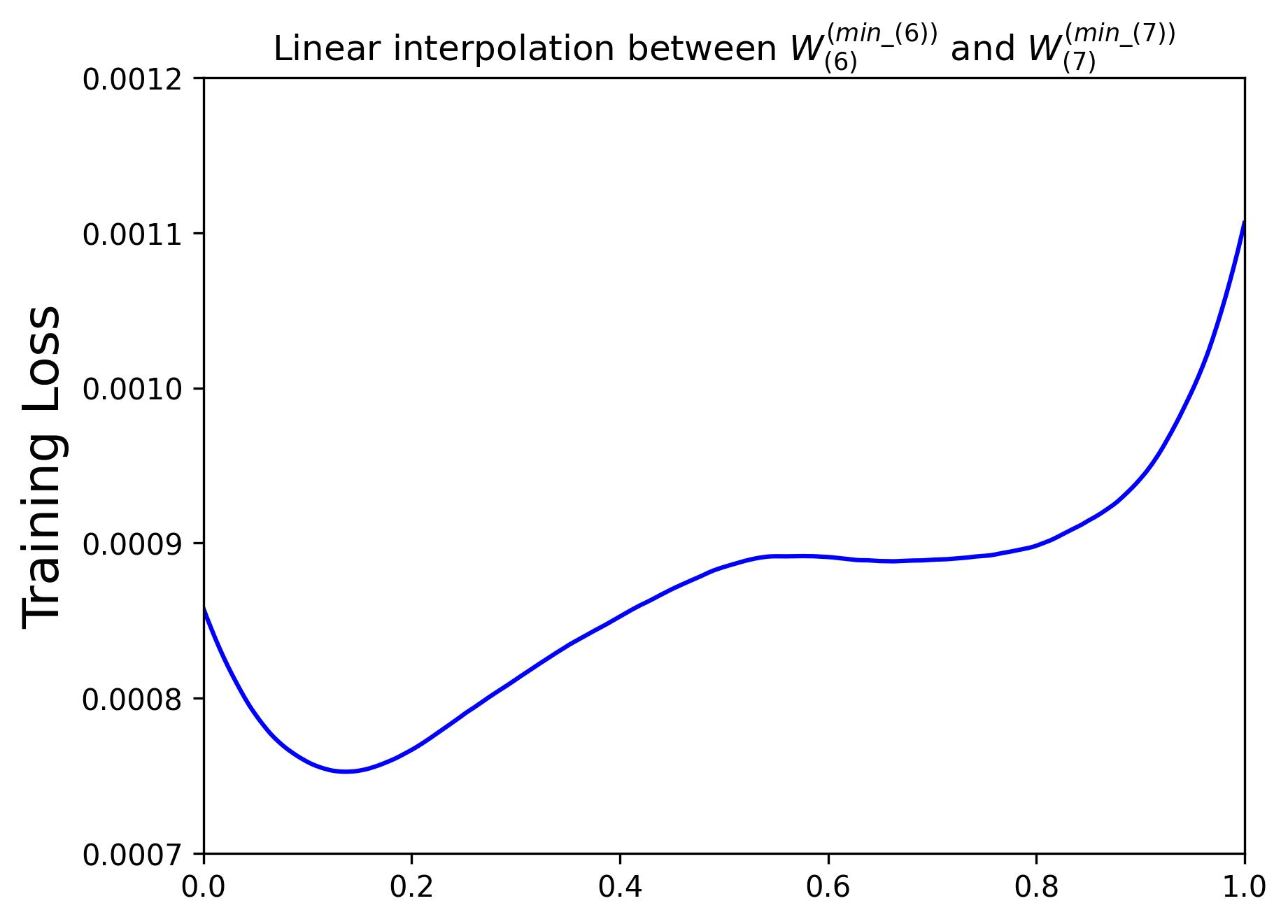} 
\includegraphics[width=4cm, height=3.5cm]{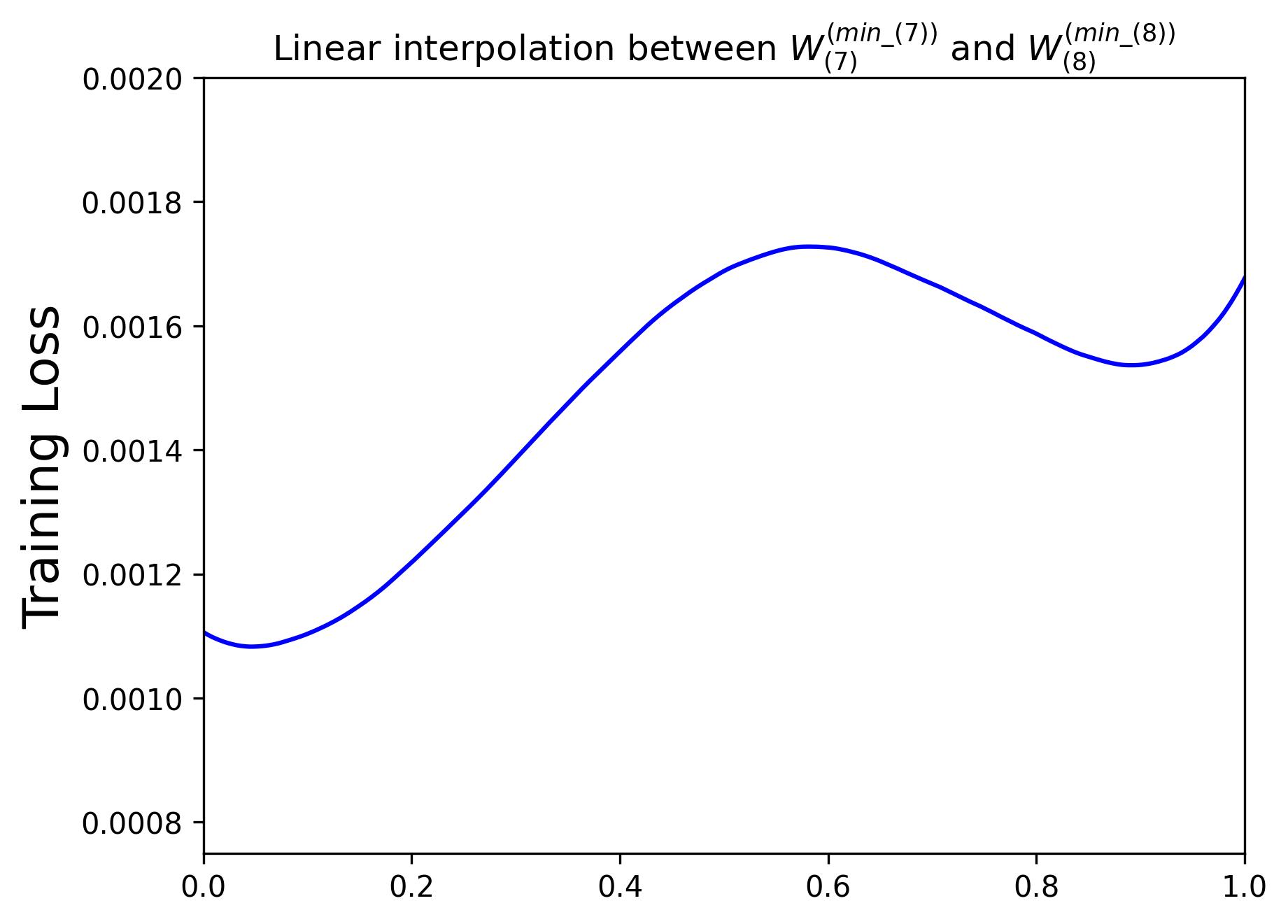} 
\includegraphics[width=4cm, height=3.5cm]{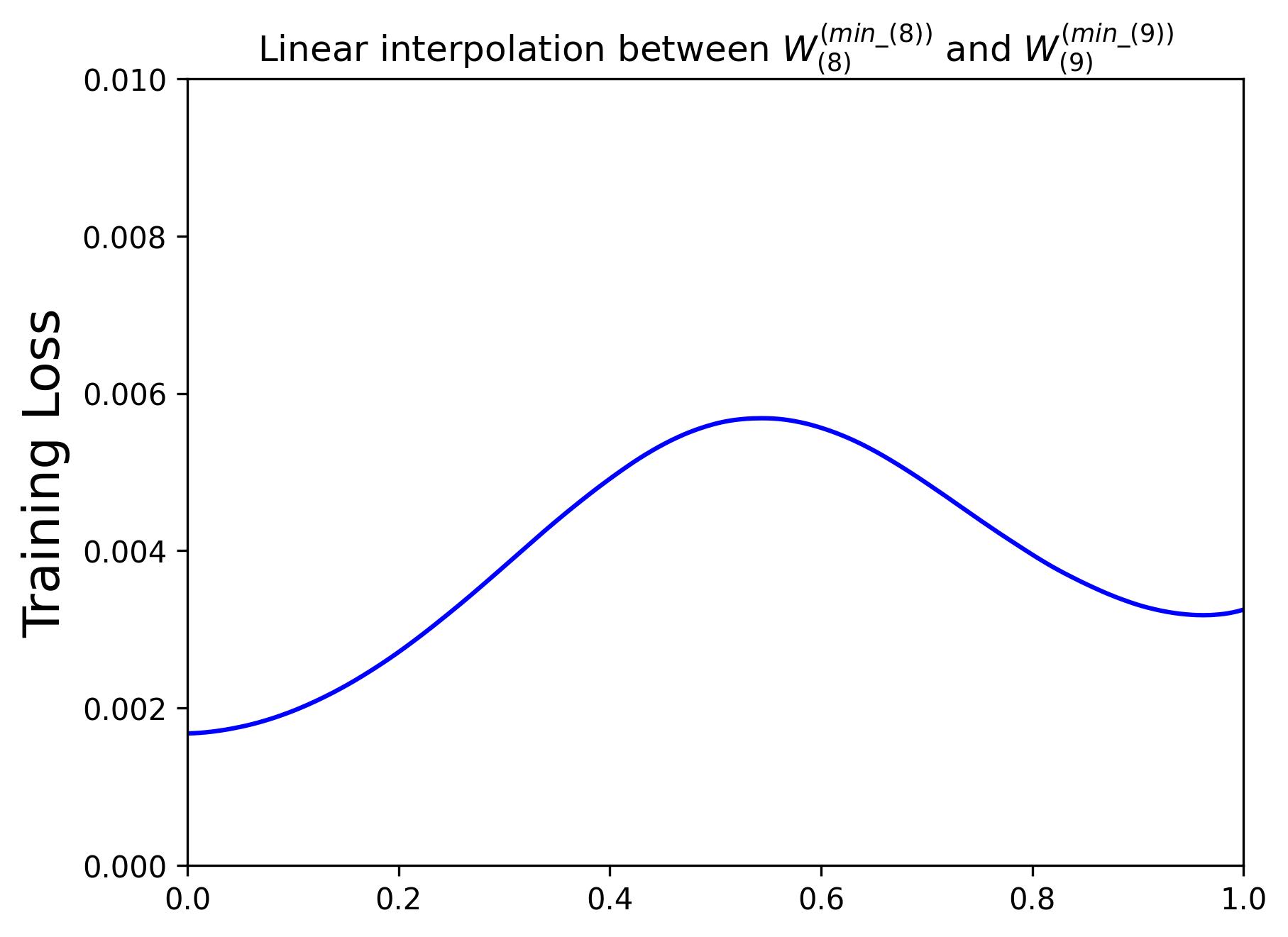}
\includegraphics[width=4cm, height=3.5cm]{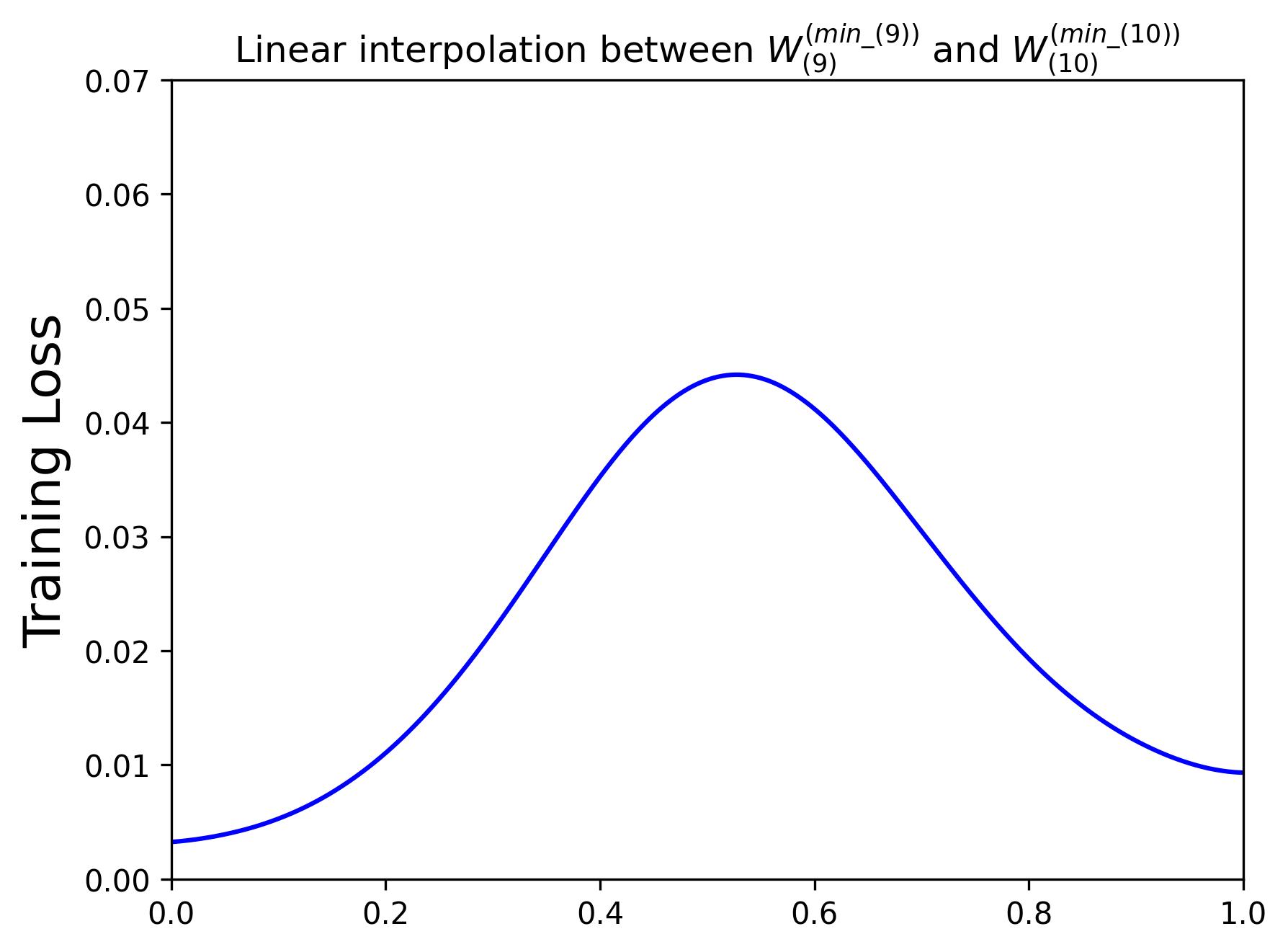} 
\includegraphics[width=4cm, height=3.5cm]{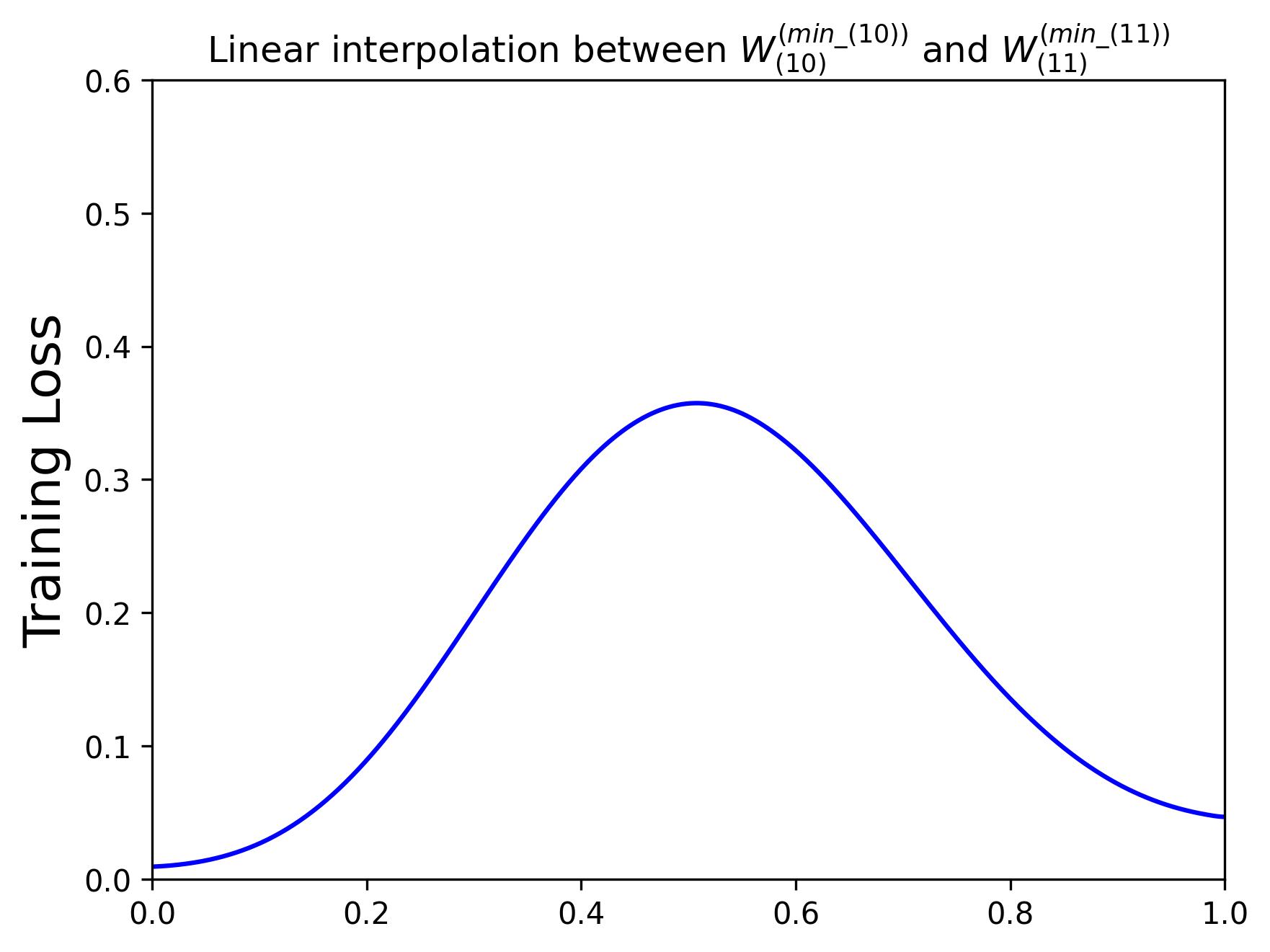}
\includegraphics[width=4cm, height=3.5cm]{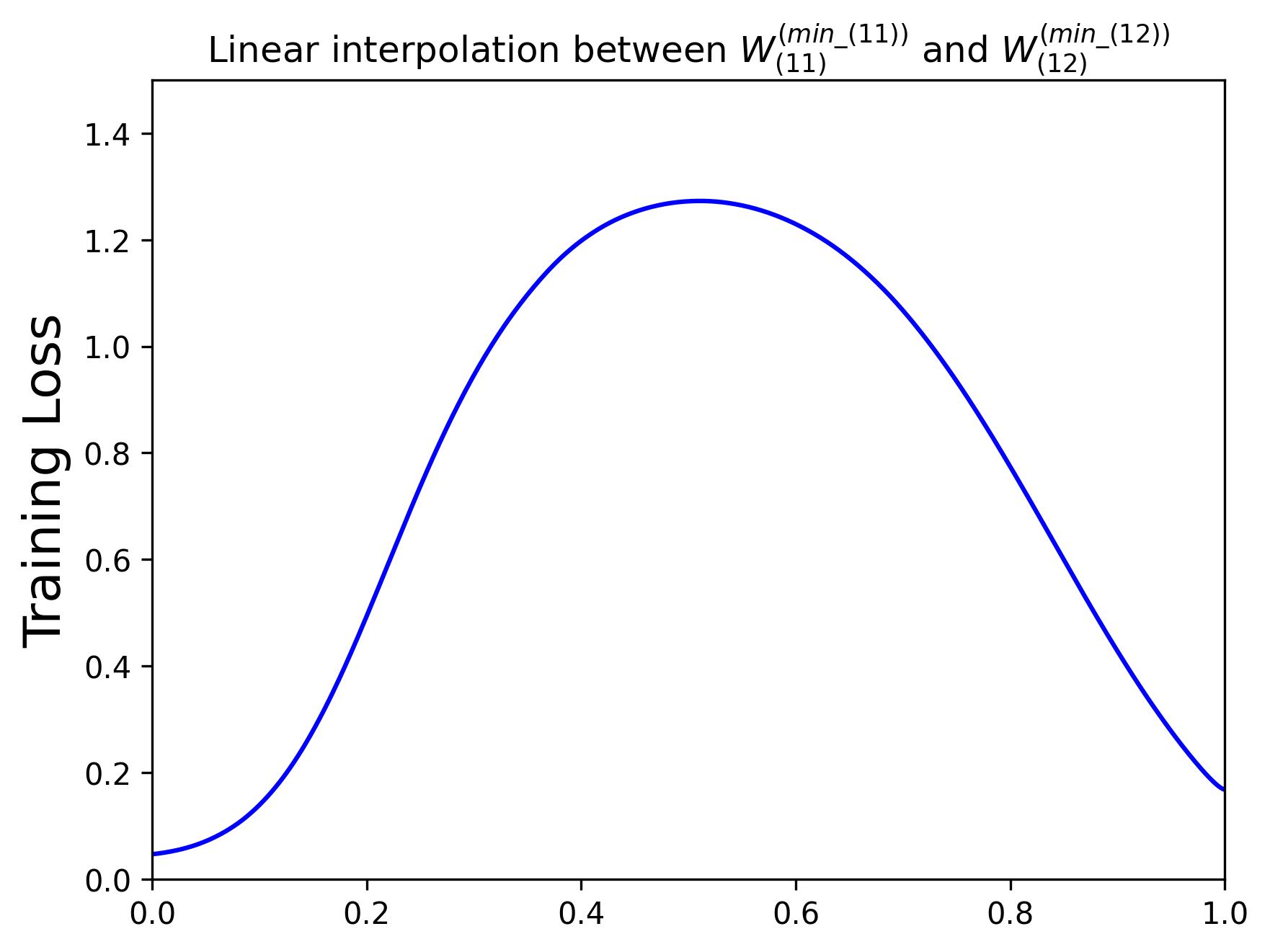} 
\caption{Training Loss along a straight line connecting $W^{(min\_(L-1))}_{(L-1)}$ and $W^{(min\_(L))}_{(L)}$ in case of VGG-16. } 
\end{figure*}
\subsection{Random initialization of a pruned network takes SGD out of the loss sublevel set and converges to a minimum with inferior performance (than that of the minimum obtained with rewinding).}
Fig. 29 presents the training loss along a straight line connecting $W^{(min\_(11))}_{(11)}$ (baseline for $W^{(RIPN)}_{(12)}$) and $W^{(RIPN)}_{(12)}$. The figure shows a huge barrier between the two points, which demonstrates that these two points lie in different loss sublevel sets.\par
\begin{figure}[h!]
\centering
\includegraphics[width=4cm, height=3.5cm]{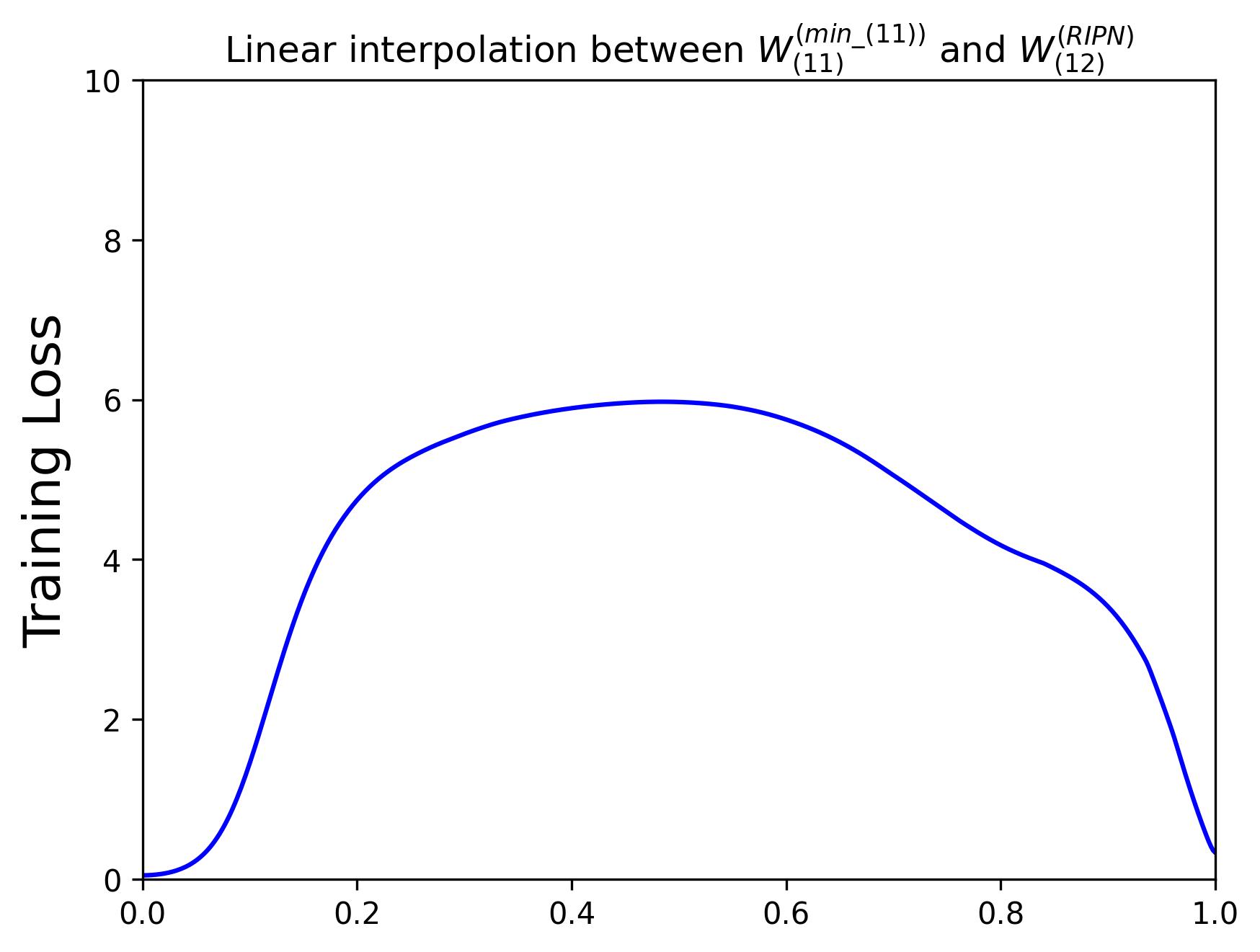}  
\caption{Training Loss along a straight line between $W^{(min\_(11))}_{(11)}$ and $W^{(RIPN)}_{(12)}$. } 
\end{figure}
A comparison of top-100 positive eigen values of the Hessian at $W^{(RIPN)}_{(12)}$ and $W^{(min\_(12))}_{(12)}$ is given in Fig. 30.
It can be observed from the figure that the Hessian at $W^{(RIPN)}_{(12)}$ has larger eigen values than that at $W^{(min\_(12))}_{(12)}$, which indicates a smaller volume for the basin around $W^{(RIPN)}_{(12)}$ (Table VIII). 
\begin{figure}[h!]
\centering
\includegraphics[width=4cm, height=3.5cm]{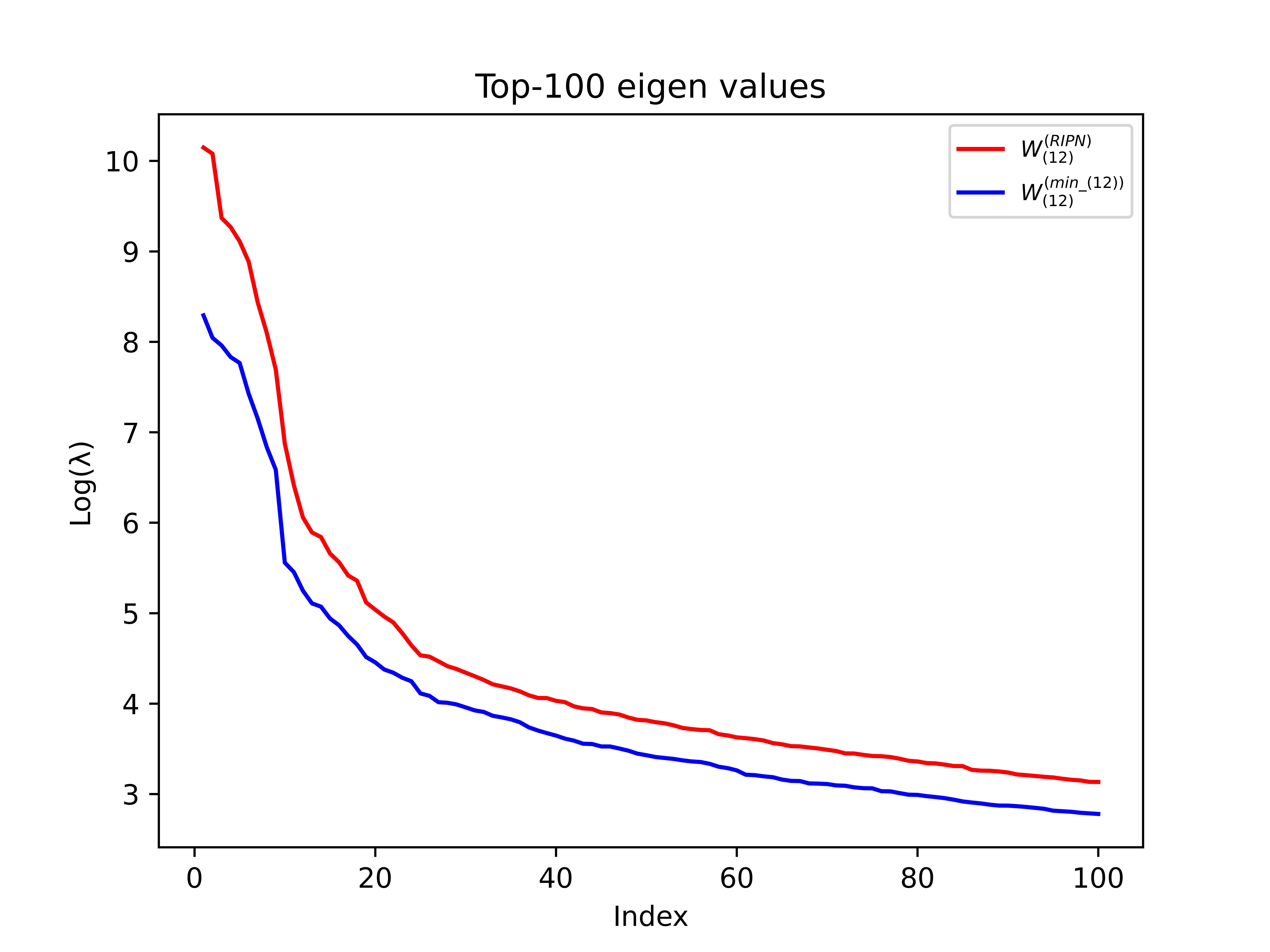}
\centering
\caption{Comparison of top-100 positive eigen values of the Hessian at $W^{(RIPN)}_{(12)}$ and $W^{(min\_(12))}_{(12)}$.} 
\end{figure} 
\begin{center}
\begin{table}[h!]
\caption{Comparison of $V^{'} (100)$ at $W^{(RIPN)}_{(12)}$ and $W^{(min\_(12))}_{(12)}$}.
\centering
\begin{tabular}{ | p{0.1\textwidth} | p{0.1\textwidth} | p{0.1\textwidth}|} 
  \hline
  Solution   & $V^{'}(100)$  \\ 
  \hline
$W^{(RIPN)}_{(12)}$
 &442.242\\ 
  \hline
   $W^{(min\_(12))}_{(12)}$   &389.957\\
  \hline
\end{tabular}
\label{product}
\end{table}
\end{center}
\subsection{Random pruning takes SGD out of the loss sublevel set and converges to a minimum with inferior performance (than that of the minimum obtained with magnitude based pruning).}
Fig. 31 presents the training loss along a straight line connecting $W^{(min\_(11))}_{(11)}$ (baseline for $W^{(RPN)}_{(12)}$) and $W^{(RPN)}_{(12)}$. The figure shows a huge barrier between the two points, which demonstrates that these two points lie in different loss sublevel sets.\par
\begin{figure}[h!]
\centering
\includegraphics[width=4cm, height=3.5cm]{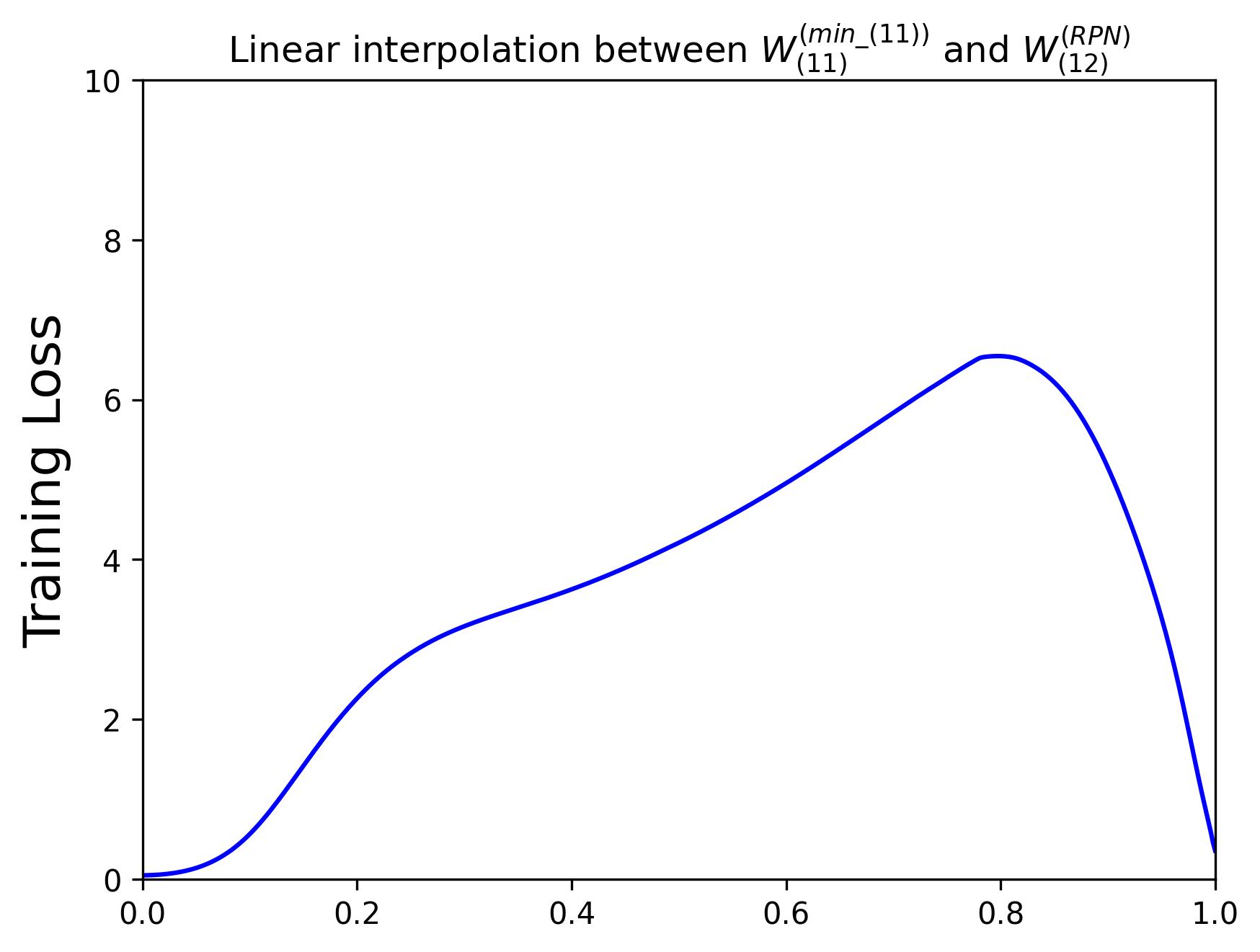}  
\caption{Training Loss along a straight line between $W^{(min\_(11))}_{(11)}$ and $W^{(RPN)}_{(12)}$. } 
\end{figure}
A comparison of top-100 positive eigen values of the Hessian at $W^{(RPN)}_{(12)}$ and $W^{(min\_(12))}_{(12)}$ is given in Fig. 32.
It can be observed from the figure that the Hessian at $W^{(RPN)}_{(12)}$ has larger eigen values than that at $W^{(min\_(12))}_{(12)}$, which indicates a smaller volume for the basin around $W^{(RPN)}_{(12)}$ (Table IX). 
\begin{figure}[h!]
\centering
\includegraphics[width=4cm, height=3.5cm]{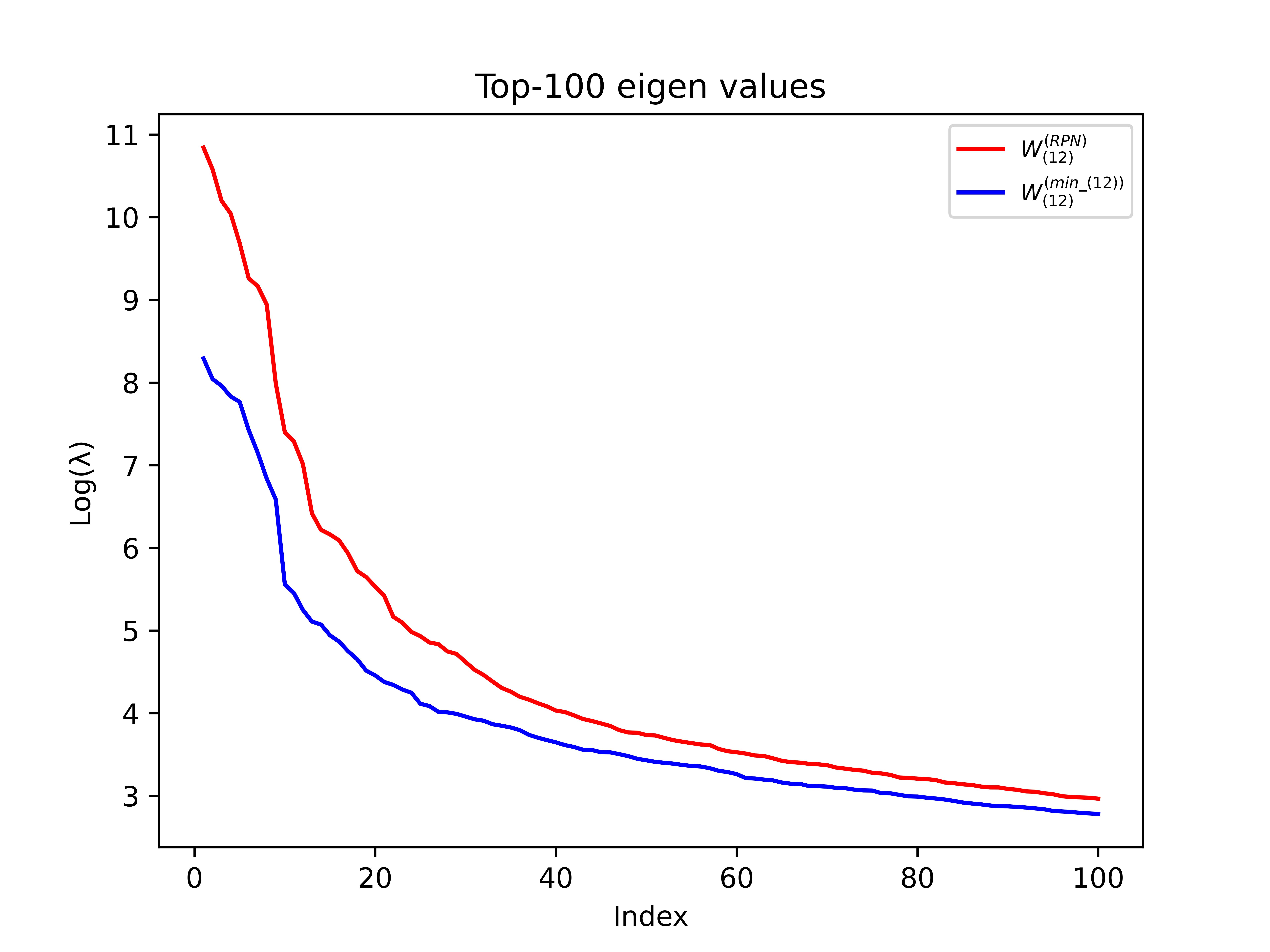}
\centering
\caption{Comparison of top-100 positive eigen values of the Hessian at $W^{(RPN)}_{(12)}$ and $W^{(min\_(12))}_{(12)}$.} 
\end{figure}
\begin{table}[h!]
\caption{Comparison of $V^{'} (100)$ at $W^{(RPN)}_{(12)}$ and $W^{(min\_(12))}_{(12)}$}.
\begin{center}
\begin{tabular}{ | p{0.1\textwidth} | p{0.1\textwidth} | p{0.1\textwidth}|} 
  \hline
  Solution   & $V^{'}(100)$  \\ 
  \hline
$W^{(RPN)}_{(12)}$
 &451.384\\ 
  \hline
   $W^{(min\_(12))}_{(12)}$
   &389.957\\
  \hline
\end{tabular}
\label{product}
\end{center}
\end{table}

\end{document}